\PassOptionsToPackage{dvipsnames}{xcolor}
\documentclass{article} %
\usepackage{arxiv_style,times}

\usepackage{amsmath,amsfonts,bm}

\def\eqref#1{equation~\ref{#1}}

\def\1{\bm{1}}

\DeclareMathAlphabet{\mathsfit}{\encodingdefault}{\sfdefault}{m}{sl}
\SetMathAlphabet{\mathsfit}{bold}{\encodingdefault}{\sfdefault}{bx}{n}

\usepackage{wrapfig}
\usepackage{booktabs}
\usepackage{graphicx}
\usepackage{caption}
\usepackage{subcaption}
\usepackage{array}
\usepackage{xcolor}
\usepackage{tikz}
\usepackage{enumitem}
\usepackage[most]{tcolorbox}
\usepackage{soul}

\definecolor{tzero}{HTML}{0072B2}
\definecolor{tfour}{HTML}{009E73}
\definecolor{tten}{HTML}{D55E00}

\newcommand{\answerbox}[2]{%
    \begingroup
    \sethlcolor{#1!15}%
    \hl{#2}%
    \endgroup
}

\newcommand{\answerframe}[2]{%
    {\setlength{\fboxrule}{0.25pt}%
    \fcolorbox{black}{white}{%
        \parbox[c][#1][c]{\linewidth}{%
            \centering
            #2
        }%
    }}%
}

\newcommand{\methodname}{CoAl}
\newcommand{\good}[2]{\textcolor{ForestGreen}{#1} {\tiny\textcolor{ForestGreen}{(#2)}}}
\newcommand{\bad}[2]{\textcolor{red}{#1} {\tiny\textcolor{red}{(#2)}}}
\newcommand{\tie}[2]{#1 {\scriptsize(#2)}}

\usepackage{hyperref}
\usepackage{url}

\title{
Learning to read the contextual tokens in Diffusion Transformers
}

\def\mystrut{\rule{0pt}{1.0\normalbaselineskip}}
\author{
\begin{tabular}{@{}l}
Omer Dahary$^{1}$\thanks{Equal contribution.}\quad Etai Sella$^{1*}$\quad Hadar Averbuch-Elor$^2$\quad Daniel Cohen-Or$^1$\quad Or Patashnik$^1$\mystrut \\
\end{tabular}\\
$^1$Tel Aviv University\mystrut\quad
$^2$Cornell University\\
\mystrut
}

\iclrfinalcopy %
\begin{document}

\maketitle
\vspace{-30pt}
\begin{center}
    \begin{minipage}[t]{0.158\linewidth}
        \centering
        \includegraphics[width=\linewidth]{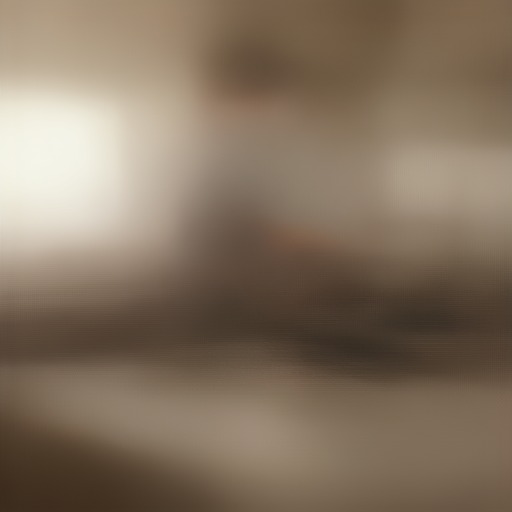}\\[2pt]
        \footnotesize cooking
    \end{minipage}
    \hfill
    \begin{minipage}[t]{0.158\linewidth}
        \centering
        \includegraphics[width=\linewidth]{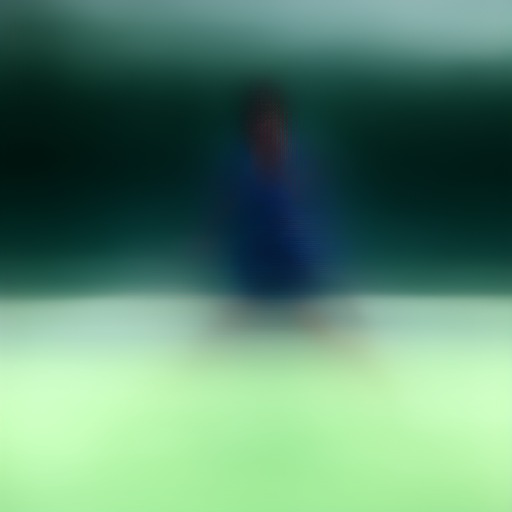}\\[2pt]
        \footnotesize playing tennis
    \end{minipage}
    \hfill
    \begin{minipage}[t]{0.158\linewidth}
        \centering
        \includegraphics[width=\linewidth]{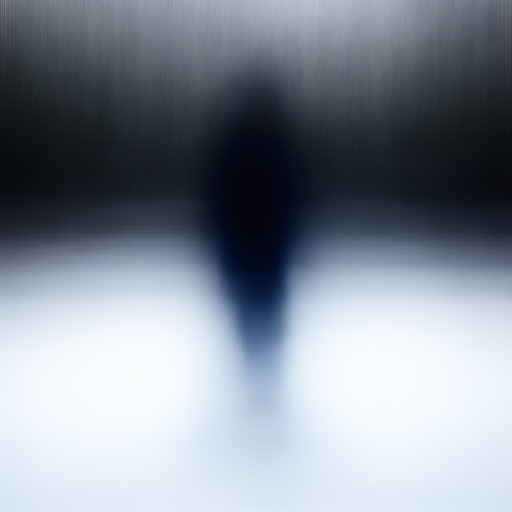}\\[2pt]
        \footnotesize skateboarding
    \end{minipage}
    \hfill
    \begin{minipage}[t]{0.158\linewidth}
        \centering
        \includegraphics[width=\linewidth]{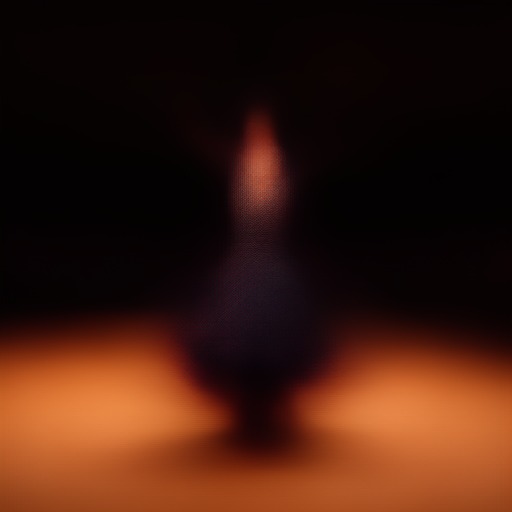}\\[2pt]
        \footnotesize dancing
    \end{minipage}
    \hfill
    \begin{minipage}[t]{0.158\linewidth}
        \centering
        \includegraphics[width=\linewidth]{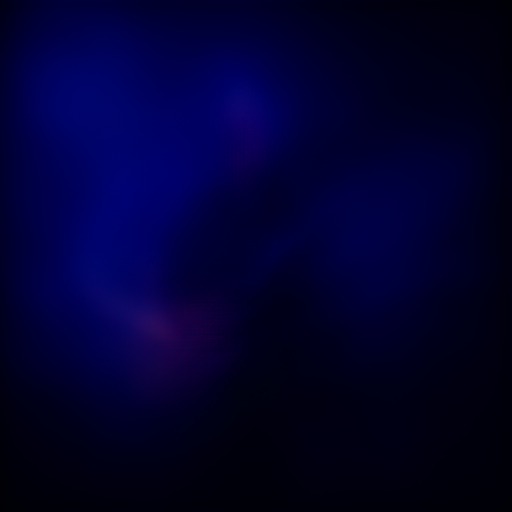}\\[2pt]
        \footnotesize playing guitar
    \end{minipage}
    \hfill
    \begin{minipage}[t]{0.158\linewidth}
        \centering
        \includegraphics[width=\linewidth]{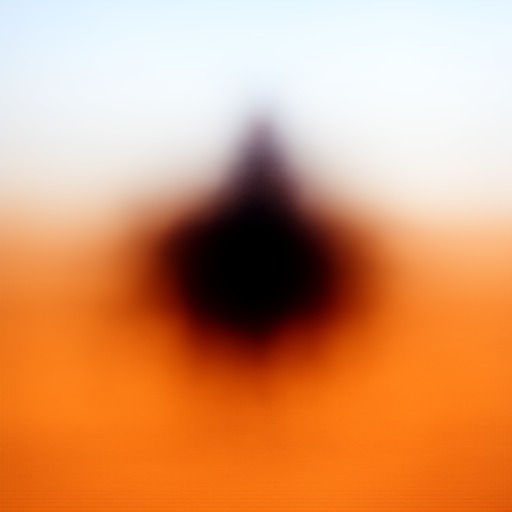}\\[2pt]
        \footnotesize riding a horse
    \end{minipage}

    \captionof{figure}{
        \textbf{Our \emph{Contextual Reader} answers the question ``What is this person doing?'' directly from the contextual tokens.} We show six generations conditioned on different prompts at an early denoising stage, where the visual predictions remain ambiguous. The text below each image is decoded from the contextual tokens when the MM-DiT is provided with an empty prompt. Despite this, our reader correctly identifies the activity, demonstrating that contextual tokens encode rich semantic information about the emerging image early in generation.
    }
    \label{fig:teaser}
\end{center}

\begin{abstract}

Multimodal Diffusion Transformers (MM-DiTs) jointly process visual and textual representations throughout generation. 
These models repeatedly update the text tokens through multimodal attention, forming dynamic \textit{contextual tokens} whose function is not well understood.
In this work, we introduce a framework for \textit{reading} this contextual space through natural-language interrogation. We train a lightweight bottleneck network that maps intermediate contextual tokens into the input space of a frozen Large Language Model (LLM), allowing the LLM to answer questions about the emerging image directly from these hidden representations. Our reader reveals that contextual tokens encode a rich, global representation of the emerging scene:
generation-specific semantics, including attributes left underspecified by the prompt, are accessible surprisingly early in denoising, while increasingly fine-grained details become readable over time.
Remarkably, this information remains decodable even when
the MM-DiT receives an empty prompt,
showing that contextual tokens accumulate substantial image-specific information from the evolving visual representation itself. We further find that
generations with more readable contextual representations tend to receive higher human-preference scores.
Building on
these observations, we introduce Contextual Alignment,
a training technique that explicitly reinforces the visual-semantic information encoded in the contextual tokens, improving generation quality and distributional coverage.
Together, our results establish contextual tokens as both an interpretable view into the internal dynamics of MM-DiTs and an effective target for improving generative models.

\end{abstract}

\section{Introduction}
\label{sec:intro}

In recent years, diffusion models have fundamentally transformed visual content creation by enabling high-fidelity generation directly from text prompts~\citep{rombach2022high,saharia2022photorealistic,ramesh2022hierarchical}. Building on this foundation, Multimodal Diffusion Transformers (MM-DiTs) have emerged as a dominant paradigm for state-of-the-art image and video synthesis~\citep{esser2024scaling,kong2024hunyuanvideo,chefer2026self}. Central to these models are multimodal attention layers, which 
enable bidirectional interaction between textual and visual representations throughout generation.

Within this shared multimodal space, visual tokens form a spatial representation of the evolving image and are explicitly supervised by the denoising objective.
Their intermediate features can therefore be directly visualized as images~\citep{patashnik2023localizing,tumanyan2023plug}, making them especially accessible to analysis and intervention. This has enabled methods that guide visual features for controlled generation~\citep{ho2022classifier,zhang2023adding,chefer2023attend}, manipulate them for image editing~\citep{meng2021sdedit,avrahami2022blended,hertz2022prompt}, and align them with pretrained representations to improve generation~\citep{yu2024representation,leng2025repa,jiang2025no}.

The text stream, however, is considerably less understood. In modern MM-DiTs, text tokens do not remain fixed conditioning vectors. Instead, they are repeatedly updated through multimodal attention, absorbing information from the emerging visual representation and evolving into what prior work termed \textit{contextual tokens}~\citep{dahary2026fly}. Although this cross-modal exchange has proven effective for generation~\citep{esser2024scaling}, the function of the resulting contextual representations remains unclear. Unlike visual tokens, contextual tokens have no explicit spatial correspondence and receive no dedicated supervision, making them inherently difficult to interpret. Consequently, it remains unknown what visual information they encode and how their dynamics ultimately steer the generation process.

In this work, we propose a framework to directly \textit{read} and interpret the contextual space of MM-DiTs by enabling natural language interrogation of their hidden representations. Rather than inspecting raw attention maps or feature vectors, our approach enables asking plain-text questions about visual elements in the emerging image and receiving direct answers decoded straight from the contextual tokens. To achieve this, we train a lightweight bottleneck network that projects intermediate contextual tokens into the input space of a Large Language Model (LLM). Because the LLM relies exclusively on the projected contextual tokens to answer visual questions, the bottleneck learns to extract the model’s internal visual semantics directly from the text-side representation. In doing so, we can read the model's evolving generative intent in plain text, revealing how it internally resolves underspecified prompt attributes over the course of generation. %

Through our reading mechanism, we find that contextual tokens encode generation-specific semantics surprisingly early in denoising. Remarkably, this remains true even when no prompt is fed into the model, showing that the tokens can acquire substantial information from the evolving visual representation alone. Figure~\ref{fig:teaser} demonstrates this setting: despite the MM-DiT receiving an empty prompt, the reader can identify the person's activity at an early denoising stage, before it can be readily inferred from the visual prediction.
Moreover, we find that readability continues to increase over the course of denoising, with the reader recovering progressively finer details about the emerging image. Importantly, this readability can also vary across different generations from the same prompt. Generations with clearer contextual representations tend to achieve higher human-preference scores, making contextual readability a meaningful signal of generation quality.

Building on these observations, we transition from passively reading the contextual tokens to actively supervising them during model training.
To this end, we introduce Contextual Alignment, a regularization objective that explicitly reinforces the visual-semantic information encoded in the contextual space.
We evaluate our supervised training framework through quantitative and qualitative experiments spanning both full model training and fine-tuning. Our experiments show that contextual supervision complements existing alignment methods, improving generation quality and distributional coverage. Furthermore, while existing alignment methods provide only limited additional gains in the fine-tuning regime, contextual supervision remains highly effective when adapting an already extensively trained model to a specific dataset.

Overall, our work establishes the contextual space as a semantically rich representation that can both expose the internal dynamics of multimodal generation and provide an effective target for improving generative models beyond the standard diffusion objective.

\section{Related Work}

\paragraph{Multimodal Diffusion Transformers.}

Diffusion models have traditionally relied on convolutional U-Net architectures as their denoising backbone~\citep{saharia2022photorealistic,ramesh2022hierarchical,rombach2022high,podell2023sdxl}. Transformer-based denoisers later offered a more scalable alternative~\citep{bao2023all,peebles2023scalable}, leading to increasingly large transformer-based text-to-image models~\citep{chen2024pixartalpha,chen2024pixartsigma}. Stable Diffusion 3~\citep{esser2024scaling} took this development further by introducing the Multimodal Diffusion Transformer (MM-DiT), in which textual and visual representations jointly participate in attention and continuously update one another. This bidirectional interaction has since become a prominent design in large-scale image generation and editing models, including FLUX~\citep{flux2024,labs2025flux}, Qwen-Image~\citep{wu2025qwen,zhao2026qwen}, and others~\citep{zhuo2024lumina,cai2025hidream,xiao2025omnigen}.

The same paradigm has become important in video generation, where CogVideoX~\citep{yang2025cogvideox} and HunyuanVideo~\citep{kong2024hunyuanvideo} jointly process textual and spatiotemporal representations. Other models extend joint attention to appearance and motion~\citep{chefer2025videojam} or audio and video~\citep{wang2026audiogen,wang2025kling,shan2025hunyuanvideo}. Together, these developments establish joint multimodal attention as a general mechanism for evolving representations across modalities.

\paragraph{Understanding Latent Representations in Diffusion Models}

A growing body of work has investigated the internal representations learned by diffusion models. Studies of visual features have shown that intermediate representations encode the semantics of individual patches and the correspondence between them, enabling a range of downstream applications~\citep{tumanyan2023plug,tang2023emergent,patashnik2023localizing,cao2023masactrl,alaluf2024cross,dahary2024yourself,dahary2025decisive,sella2025instancegen,luo2024readout}. Other studies examine textual representations in text-to-image models, studying how semantic information is encoded and distributed across textual tokens~\citep{rassin2022dalle,toker2024diffusion,kaplan2026follow,zhuang2024magnet}.

With the emergence of MM-DiTs, recent work has also begun to examine the evolving text-side representations formed through multimodal attention. ConceptAttention~\citep{helbling2025conceptattention} uses these representations to localize concepts in the generated image, while Contextual Repulsion~\citep{dahary2026fly} demonstrates their utility as an intervention space for increasing generation diversity. Several works use causal interventions to reveal functional roles of contextual tokens under specific conditioning settings. Padding Tone~\citep{toker2025padding} shows that, under textual conditioning, contextualized padding tokens can contribute additional visual information to the generated image; Vision-Language Binding~\citep{ge2026vision} shows that contextual tokens can acquire semantic information from a reference image in FLUX.2 editing; Li et al.~\citep{li2026text} show that LLM-encoded template tokens in Qwen-Image can act as implicit registers for maintaining object identity. Altogether, these interventions demonstrate that contextual tokens can carry information relevant to generation. However, they do not directly reveal the semantic content of the contextual space or how it evolves over denoising. We instead decode this space into natural language, revealing a surprisingly broad range of image-specific semantics encoded within it. Remarkably, this information can be decoded in the absence of textual conditioning and even before it becomes readily apparent in the visual prediction itself.

\paragraph{Representation Alignment for Generation.}

Recent work has shown that explicitly encouraging diffusion models to learn stronger internal representations can substantially improve both training efficiency and generation quality. REPA~\citep{yu2024representation} introduced this direction by aligning intermediate visual features with representations from a pretrained vision encoder. Subsequent works extend alignment to attention features~\citep{wang2026repa}, the latent space~\citep{leng2025repa,zheng2026diffusion}, representations derived from the diffusion model itself~\citep{jiang2025no,chefer2026self}, and text--image contrastive objectives~\citep{lee2026aligning}. Recent analyses have also examined the role of local and global information in representation alignment. iREPA~\citep{singh2025matters} shows that, for spatial visual features, alignment benefits primarily from preserving and enhancing spatial structure rather than global semantics. REG~\citep{wu2026representation} instead shows that global semantic information can benefit class-conditional generation when represented separately through a high-level semantic token alongside the image latents.

We extend these insights to text-conditioned generation and find that global visual semantics are implicitly encoded in the contextual tokens. We therefore align them with a global semantic representation, complementing the local, spatial representations reinforced by existing alignment methods.

\section{Reading the Contextual Space}
\label{sec:reading}

We introduce the \textit{Contextual Reader}, a framework for decoding the internal contextual representations of Multimodal Diffusion Transformers (MM-DiTs) into natural language. We formulate this as a visual question answering task.
At a selected timestep during generation, the reader receives the contextual representations produced by all MM-DiT layers together with a question about the emerging image.
The question can probe the image at different levels of detail, including properties that are not explicitly specified by the prompt. The reader then generates a textual answer using the contextual representation and the question. This allows us to examine which visual semantics are encoded in the contextual space as generation progresses.

\subsection{Reader Architecture}
\label{subsec:reader_overview}

\begin{figure}[t]
    \centering
    \includegraphics[width=\linewidth]{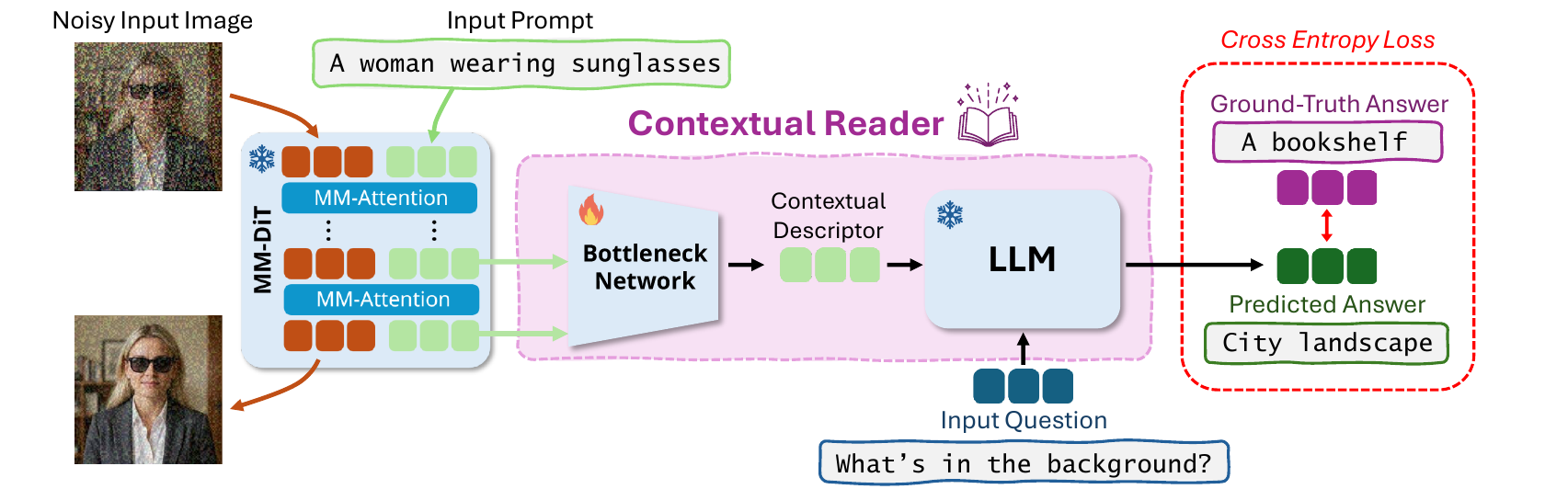}
            \caption{
        \textbf{Training the Contextual Reader.}
        At a selected denoising timestep, contextual tokens from multiple frozen MM-DiT layers are aggregated by a trainable Bottleneck Network into a compact Contextual Descriptor. A frozen LLM receives this descriptor together with a visual question and predicts an answer about the emerging image. We train only the Bottleneck Network using cross-entropy against the ground-truth answer.
    }
    \label{fig:reader}
\end{figure}

Figure~\ref{fig:reader} illustrates the architecture and training procedure of our Contextual Reader.
The reader has two main components.
First, a \textit{Bottleneck Network} aggregates contextual tokens from multiple MM-DiT layers into a small set of tokens, which we call the \textit{Contextual Descriptor}. Second, a frozen LLM receives this descriptor together with the visual question and generates the final answer. %

\paragraph{Multi-Layer Contextual Extraction.}
At diffusion timestep $t$, each block $l$ produces a sequence of contextual tokens
$c^{(l)}(t) \in \mathbb{R}^{M \times d}$,
where $M$ is the text sequence length, including padding tokens, and $d$ is the model hidden dimension.
We aggregate the contextual tokens from all $L$ transformer blocks.
To retain information about where and when each representation was extracted,
we augment each sequence with learnable embeddings of its layer and timestep.
We then concatenate these augmented tokens along the sequence dimension to form
$C(t) \in \mathbb{R}^{LM \times d}$.
For full details on the feature extraction, see
Appendix~\ref{subsec:reader_features}.

\paragraph{Bottleneck Network and Language Interface.}
Our language interface is inspired by LLaVA~\citep{Liu2023VisualIT}, which connects an external representation to a language model for visual question answering through a learned projection. We adopt a similar interface for the contextual space. We use a lightweight Bottleneck Network to compress $C(t)$ into a small set of tokens. The bottleneck consists of a 2-layer Q-Former with $Q$ learnable query tokens, where $Q \ll L \cdot M$, followed by a linear projection into the input embedding space of the frozen LLM.

The Q-Former queries attend to $C(t)$ through cross-attention, allowing them to aggregate information across layers and contextual-token positions. A linear projection maps the resulting query features to the embedding dimension of the frozen LLM, producing the \textit{Contextual Descriptor} $p \in \mathbb{R}^{Q \times d_{\text{LLM}}}$. The descriptor is provided to the LLM together with the tokenized question. The LLM then generates the answer autoregressively. For full architectural details, see Appendix~\ref{subsec:reader_interface}.

\subsection{Training the Contextual Reader}
\label{subsec:reader_training}

The Contextual Reader is trained to decode semantic properties of the corresponding final image from contextual representations at arbitrary denoising timesteps. Each training sample therefore consists of a contextual representation, a visual question, and a target answer for the corresponding generated image. We construct this supervision directly from the MM-DiT generation process.

\paragraph{Dataset Construction.}

To define a consistent set of questions that are meaningful across generations, we focus on images containing a single human subject. We curate 15 questions spanning broad scene properties, such as camera angle and background, as well as finer-grained properties, such as pose, appearance, and object interaction.

To generate the images, we use a subset of human-centric, single-subject prompts from MS-COCO~\citep{lin2014microsoft}. To encourage the reader to distinguish between different outcomes under the same textual conditioning, we generate five seeds for each prompt. Our dataset comprises 10,000 generations from 2,000 training prompts and 2,500 held-out generations from 500 prompts. For each generated image, an off-the-shelf Vision-Language Model provides answers to the 15 questions, which serve as the target supervision for the reader. For each generated image, we also store the prompt and initial noise used for generation, allowing us to retrieve its contextual representations at arbitrary denoising timesteps during training. The full list of questions and additional dataset construction details are provided in Appendix~\ref{subsec:reader_dataset}.

\paragraph{Training.}

During reader training, we sample a denoising timestep and a generated image from our dataset, together with its associated questions and answers. Using the stored prompt and initial noise, we reconstruct the noisy latent at the sampled timestep and run the frozen MM-DiT to obtain the contextual representations $C(t)$. The reader then produces answers to the corresponding visual questions, which are supervised against the ground-truth answers using an autoregressive cross-entropy loss.
We optimize only the Bottleneck Network, with all other components kept frozen. We train the Reader for 5,000 iterations with a batch size of 256; the validation loss decreases steeply during the first 2,000 iterations, with only modest improvements thereafter. Training details and validation curves are provided in Appendix~\ref{subsec:reader_training_details}. %

\subsection{Semantic Readability of the Contextual Space}
\label{subsec:contextual_analysis}
\begin{figure*}[t]
    \centering
    \scriptsize

    \setlength{\fboxrule}{1pt}
    \setlength{\fboxsep}{0pt}

    \begin{tabular}{@{}p{0.485\textwidth}@{\hspace{0.03\textwidth}}p{0.485\textwidth}@{}}

    \begin{minipage}[t]{\linewidth}
        \centering

        {\small\textbf{Prompt:} ``A person holding a scarf.''}

        \vspace{3pt}

        \begin{tabular}{@{}c@{\hspace{2pt}}c@{\hspace{2pt}}c@{\hspace{2pt}}c@{}}
            {\scriptsize\textbf{8\%}} &
            {\scriptsize\textbf{100\%}} &
            {\scriptsize\textbf{8\%}} &
            {\scriptsize\textbf{100\%}} \\[2pt]

            \fcolorbox{black}{white}{%
                \includegraphics[width=0.23\linewidth]
                {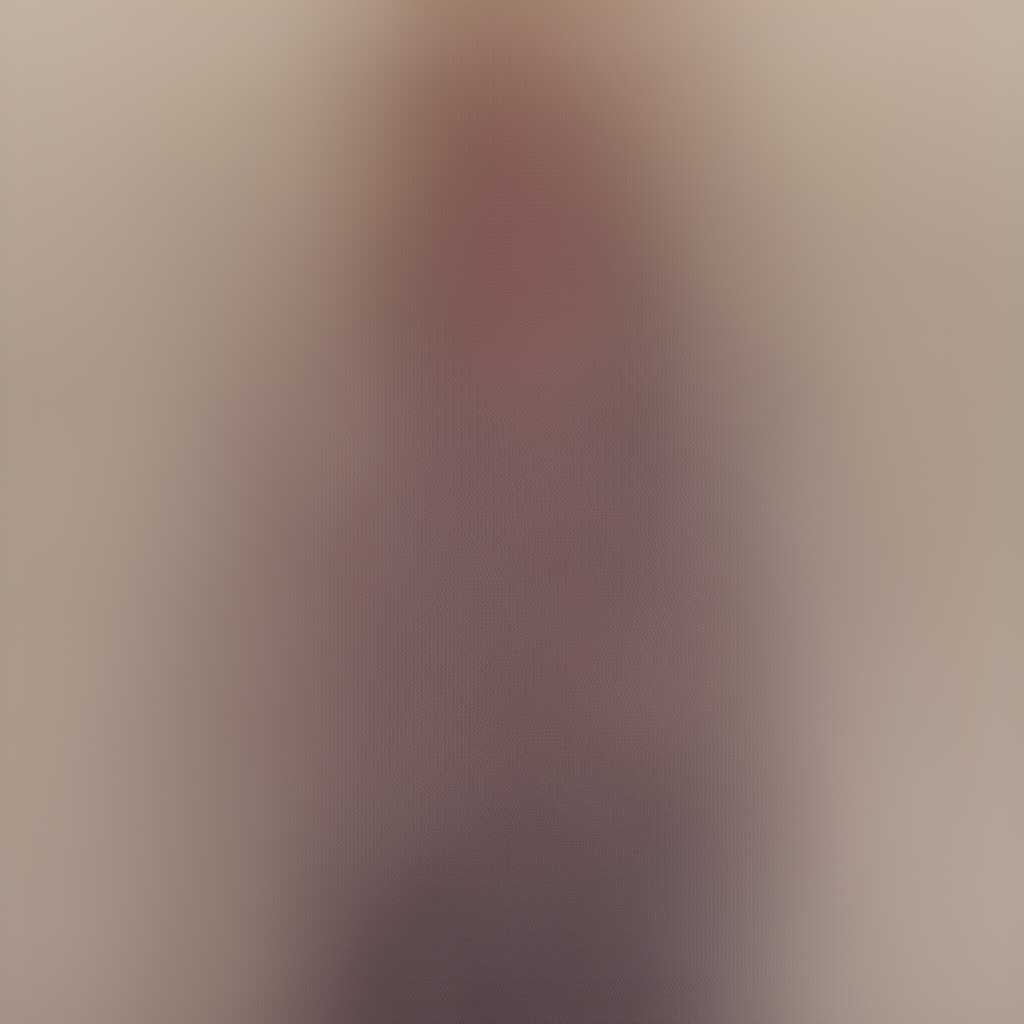}} &
            \fcolorbox{black}{white}{%
                \includegraphics[width=0.23\linewidth]
                {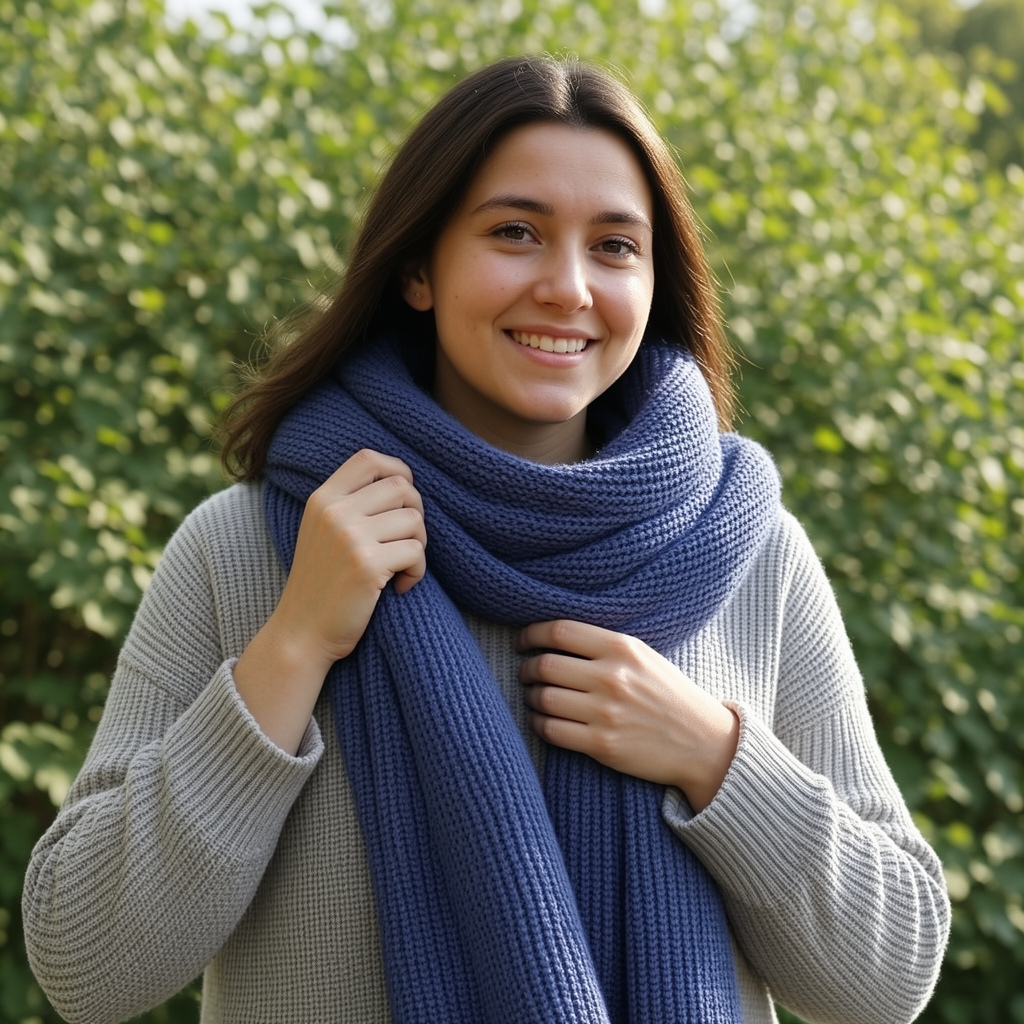}} &
            \fcolorbox{black}{white}{%
                \includegraphics[width=0.23\linewidth]
                {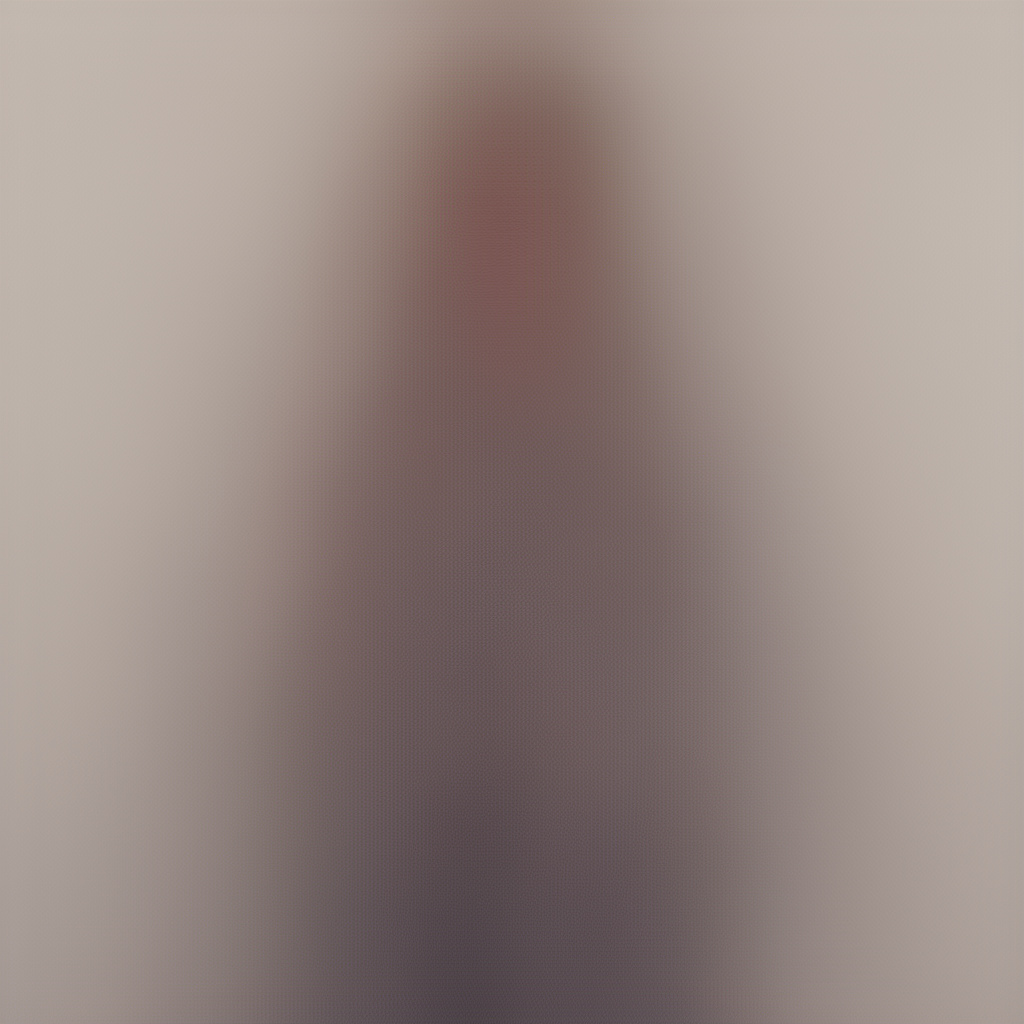}} &
            \fcolorbox{black}{white}{%
                \includegraphics[width=0.23\linewidth]
                {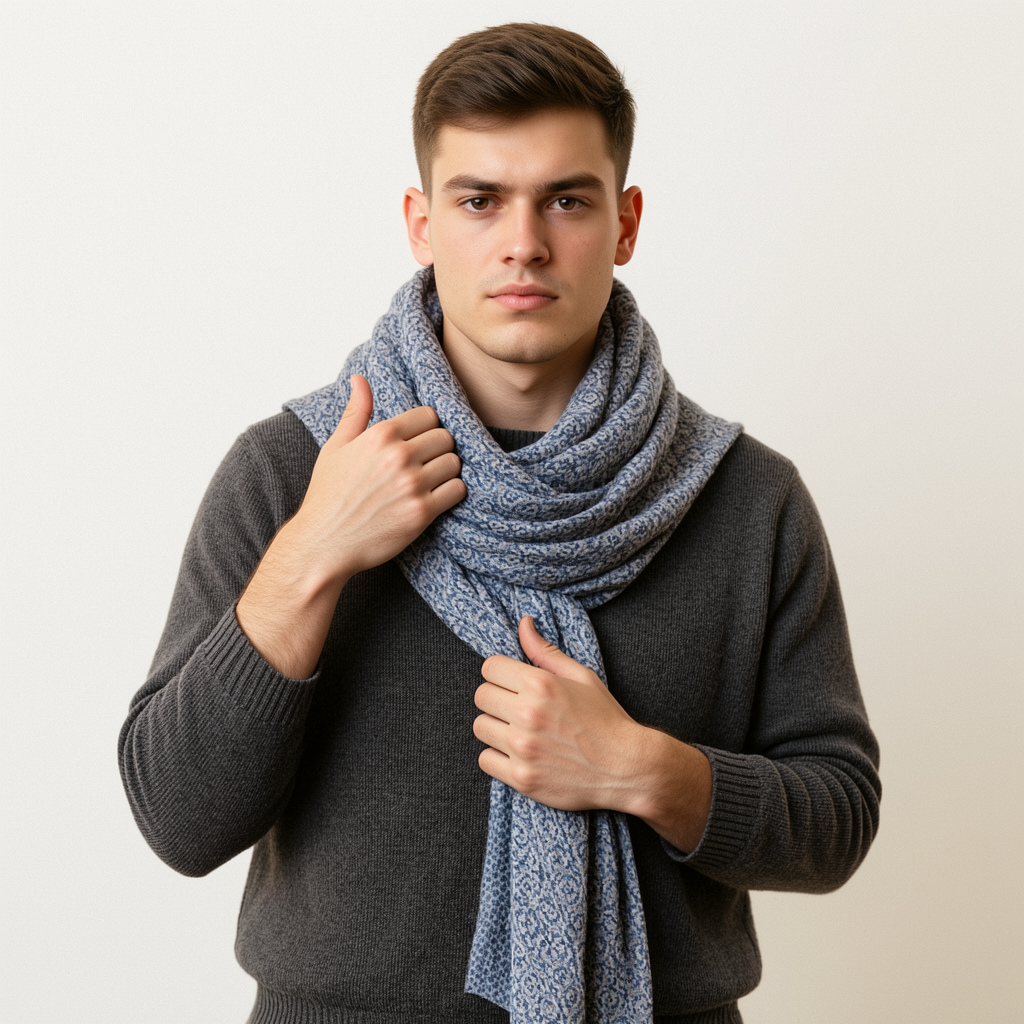}}
        \end{tabular}

        \vspace{4pt}

        \begin{tabular}{@{}p{0.47\linewidth}@{\hspace{0.04\linewidth}}p{0.47\linewidth}@{}}
            \centering\textbf{Seed A} &
            \centering\textbf{Seed B}
        \end{tabular}

        \vspace{3pt}

        \textit{What is in the background?}

        \vspace{1pt}

        \begin{tabular}{@{}p{0.47\linewidth}@{\hspace{0.04\linewidth}}p{0.47\linewidth}@{}}
            \centering
            \answerframe{0.35cm}{Green bushes and trees.}
            &
            \centering
            \answerframe{0.35cm}{A plain white wall.}
        \end{tabular}

        \vspace{4pt}

        \textit{What is the person's hair like?}

        \vspace{1pt}

        \begin{tabular}{@{}p{0.47\linewidth}@{\hspace{0.04\linewidth}}p{0.47\linewidth}@{}}
            \centering
            \answerframe{0.55cm}{Long, dark brown.}
            &
            \centering
            \answerframe{0.55cm}{Short, brown, neatly styled.}
        \end{tabular}

    \end{minipage}

    &

    \begin{minipage}[t]{\linewidth}
        \centering

        {\small\textbf{Prompt:} ``A person looking into a mirror.''}

        \vspace{3pt}

        \begin{tabular}{@{}c@{\hspace{2pt}}c@{\hspace{2pt}}c@{\hspace{2pt}}c@{}}
            {\scriptsize\textbf{8\%}} &
            {\scriptsize\textbf{100\%}} &
            {\scriptsize\textbf{8\%}} &
            {\scriptsize\textbf{100\%}} \\[2pt]

            \fcolorbox{black}{white}{%
                \includegraphics[width=0.23\linewidth]
                {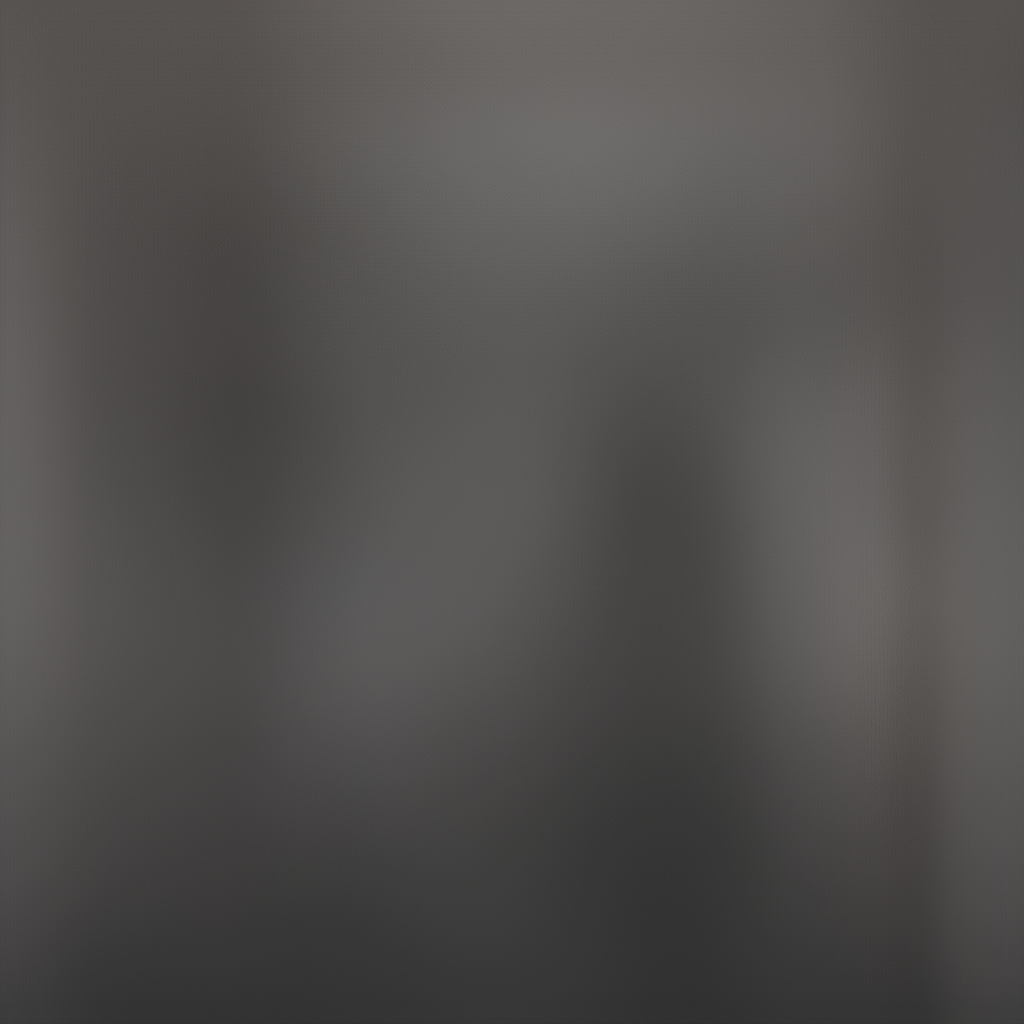}} &
            \fcolorbox{black}{white}{%
                \includegraphics[width=0.23\linewidth]
                {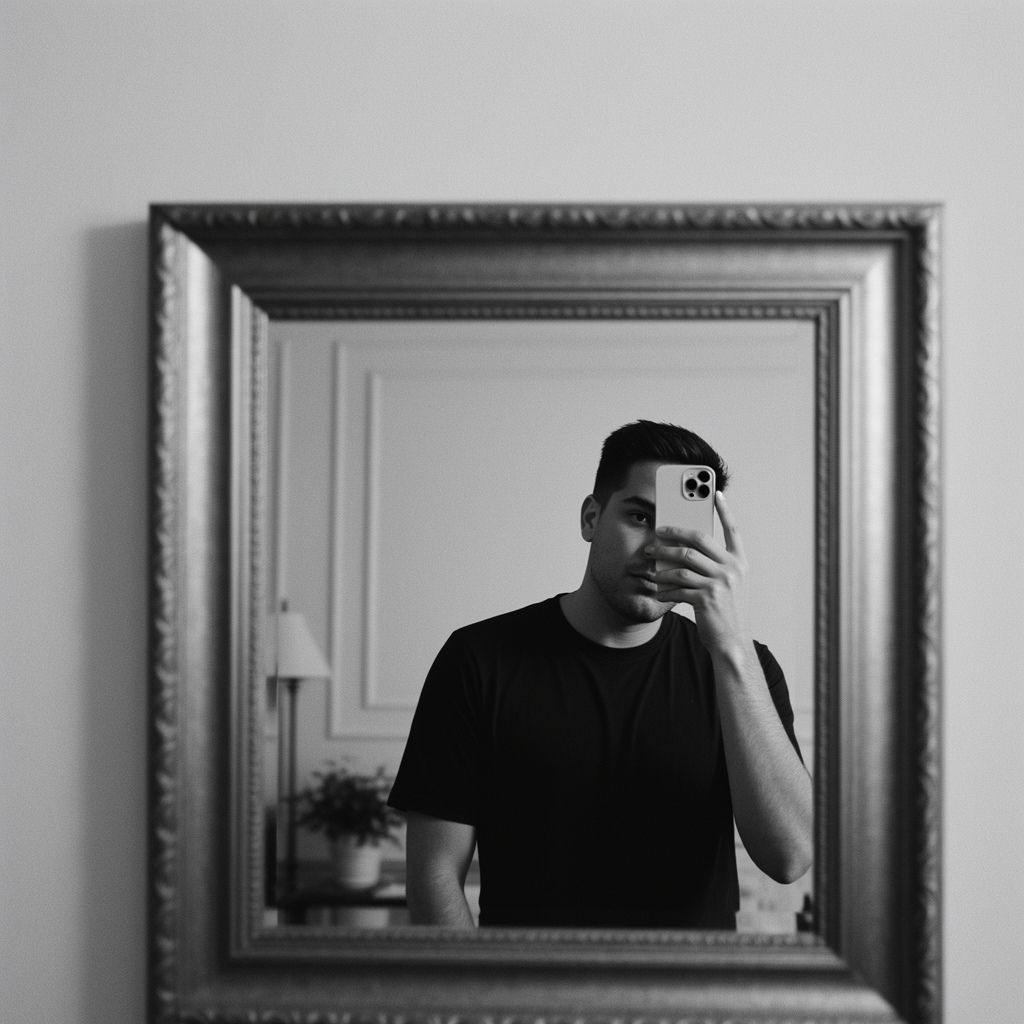}} &
            \fcolorbox{black}{white}{%
                \includegraphics[width=0.23\linewidth]
                {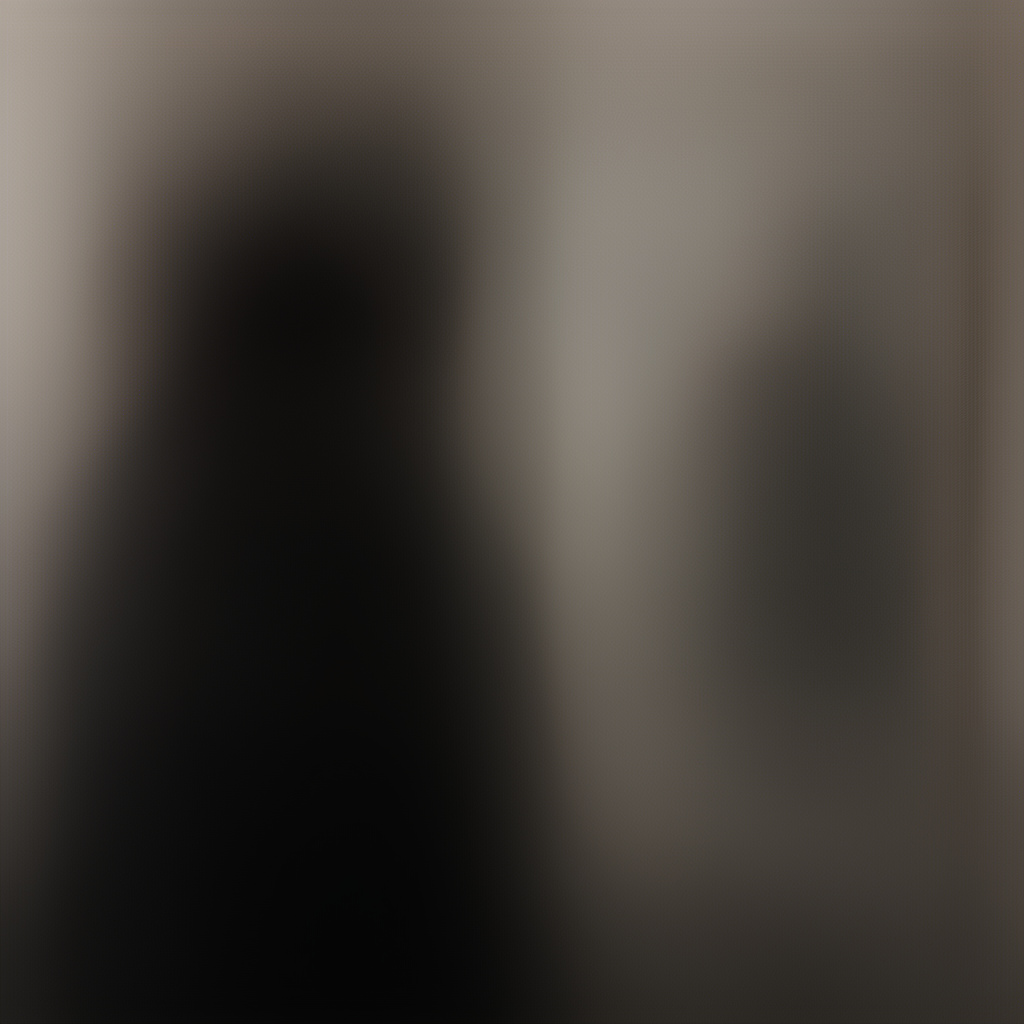}} &
            \fcolorbox{black}{white}{%
                \includegraphics[width=0.23\linewidth]
                {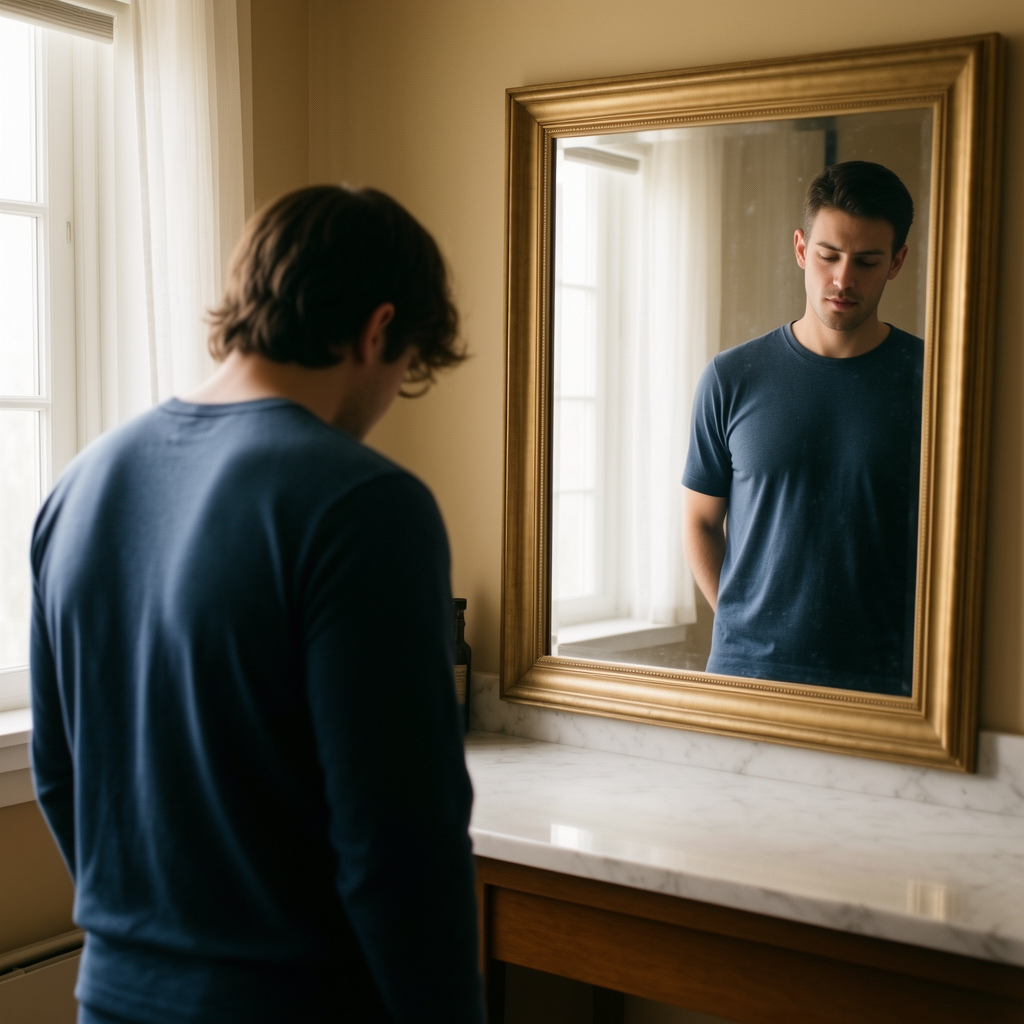}}
        \end{tabular}

        \vspace{4pt}

        \begin{tabular}{@{}p{0.47\linewidth}@{\hspace{0.04\linewidth}}p{0.47\linewidth}@{}}
            \centering\textbf{Seed A} &
            \centering\textbf{Seed B}
        \end{tabular}

        \vspace{3pt}

        \textit{What is the person holding?}

        \vspace{1pt}

        \begin{tabular}{@{}p{0.47\linewidth}@{\hspace{0.04\linewidth}}p{0.47\linewidth}@{}}
            \centering
            \answerframe{0.35cm}{A smartphone.}
            &
            \centering
            \answerframe{0.35cm}{Unknown.}
        \end{tabular}

        \vspace{4pt}

        \textit{What is the color palette?}

        \vspace{1pt}

        \begin{tabular}{@{}p{0.47\linewidth}@{\hspace{0.04\linewidth}}p{0.47\linewidth}@{}}
            \centering
            \answerframe{0.55cm}{Black and white.}
            &
            \centering
            \answerframe{0.55cm}{Neutral tones, with beige, blue, brown and white.}
        \end{tabular}

    \end{minipage}

    \end{tabular}

    \caption{
        \textbf{Exemplary plain-text predictions from our conditional Contextual Reader on FLUX.2.}
For each prompt, we compare two random seeds that resolve underspecified
attributes differently. We show the predicted clean images at $8\%$ of the
denoising trajectory for both seeds, followed by their final images, and
apply the reader to the contextual tokens at the same $8\%$ timestep to
obtain the corresponding answers. Despite the visual predictions remaining
ambiguous at the $8\%$ timestep, the reader can recover seed-specific
information that is not specified by the prompt.
    }

    \label{fig:seed_comparison}

\end{figure*}

Having established a mechanism for reading the contextual tokens, we now examine what this representation reveals about the emerging generation. We study this across two pretrained MM-DiTs with different architectures: SD3.5~\citep{esser2024scaling}, which maintains separate image and text streams while allowing them to exchange information through multimodal attention, and FLUX.2~\citep{chefer2026self}, which additionally includes single-stream blocks that jointly process the concatenated image and text tokens. 

To study how access to the prompt affects what can be decoded from the contextual space, we consider two conditioning settings. In one, the MM-DiT receives the original prompt (\emph{conditional setting}), allowing us to study how the contextual tokens become enriched with image-specific information beyond what the prompt specifies. In the other, the MM-DiT receives an empty prompt (\emph{unconditional setting}), allowing us to examine how much image-specific information can be recovered without access to the prompt.

\paragraph{Qualitative Analysis.}

Figure~\ref{fig:seed_comparison} shows examples from the conditional setting,
where two random seeds are generated from the same prompt. Although the textual conditioning is identical, the resulting images resolve underspecified properties differently, and the reader recovers these seed-specific differences from the contextual tokens. For the prompt
``A person holding a scarf,'' the reader identifies that the background is ``green bushes and trees'' in one seed, and ``a plain white wall'' in the other.
Turning to the prompt “A person looking into a mirror,” the reader identifies that the person is holding a smartphone in one seed. In the other seed, the person's hands fall outside the image, and the reader correspondingly answers ``Unknown.''
These distinctions can often be decoded at an early denoising step, before they are clearly visible in the corresponding visual prediction.

\begin{figure*}[t]
    \centering
    \scriptsize

    \setlength{\fboxrule}{1pt}
    \setlength{\fboxsep}{0pt}

    \begin{minipage}[t]{0.49\textwidth}
        \centering

        \newlength{\skiwidth}
        \setlength{\skiwidth}{0.23\linewidth}

        \begin{tabular}{@{}c@{\hspace{2pt}}c@{\hspace{2pt}}c@{\hspace{2pt}}c@{}}
            \textcolor{tzero}{\textbf{0\%}} &
            \textcolor{tfour}{\textbf{8\%}} &
            \textcolor{tten}{\textbf{20\%}} &
            \textbf{100\%}
            \\[2pt]

            \fcolorbox{tzero}{white}{%
                \includegraphics[width=\skiwidth]
                {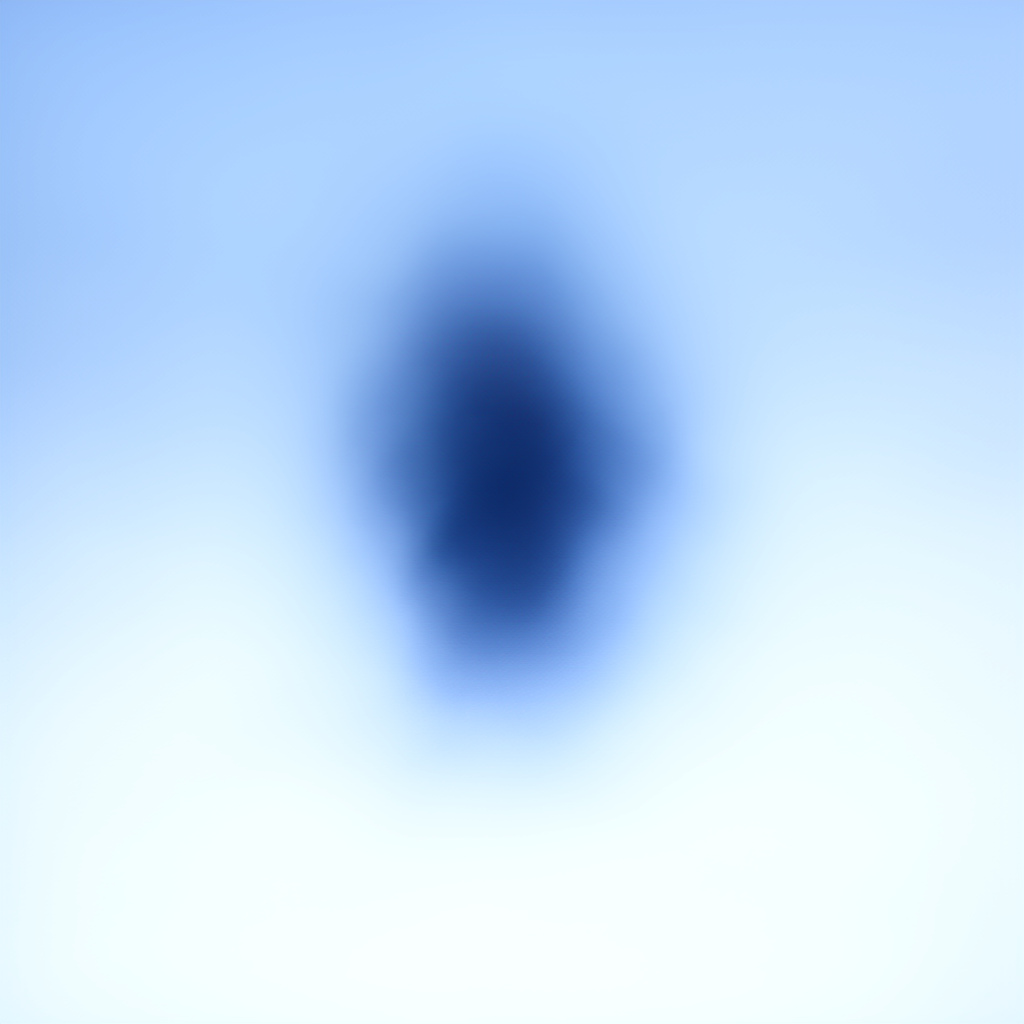}%
            }
            &
            \fcolorbox{tfour}{white}{%
                \includegraphics[width=\skiwidth]
                {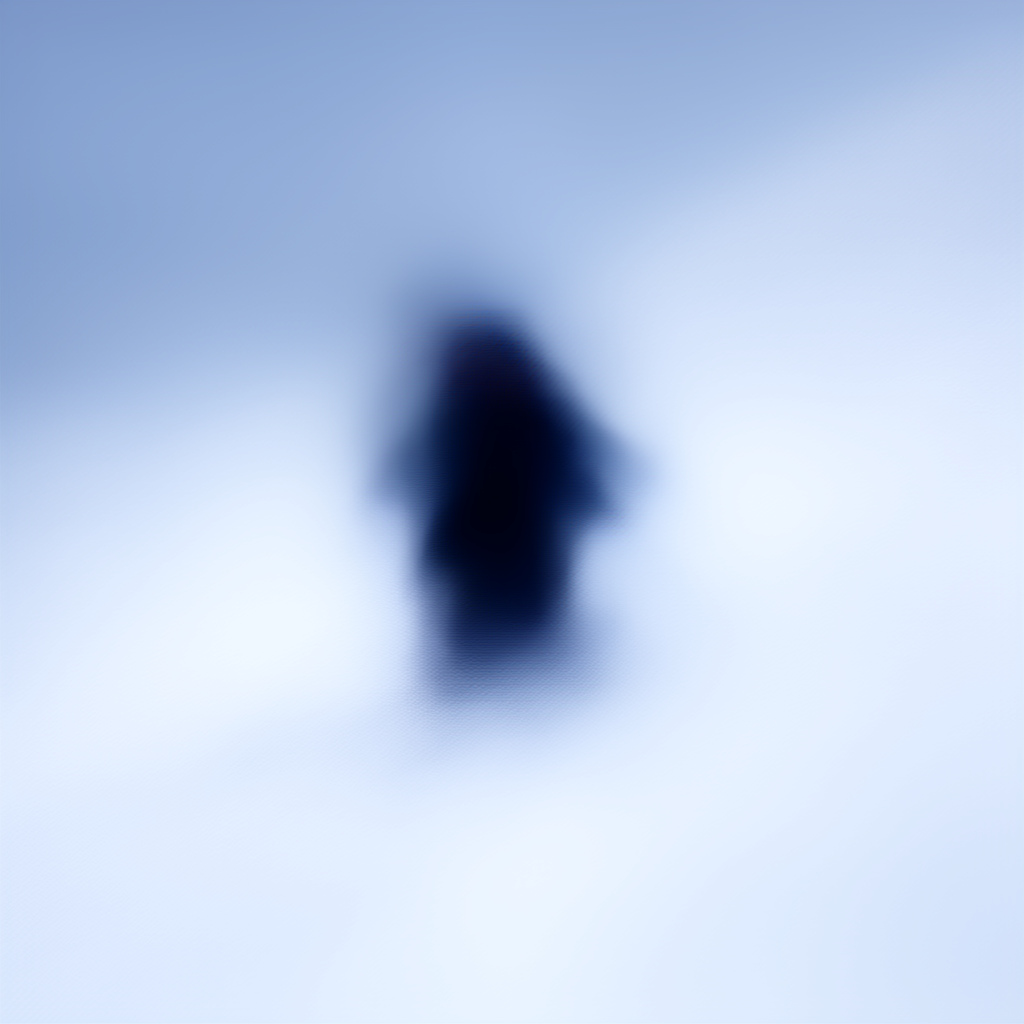}%
            }
            &
            \fcolorbox{tten}{white}{%
                \includegraphics[width=\skiwidth]
                {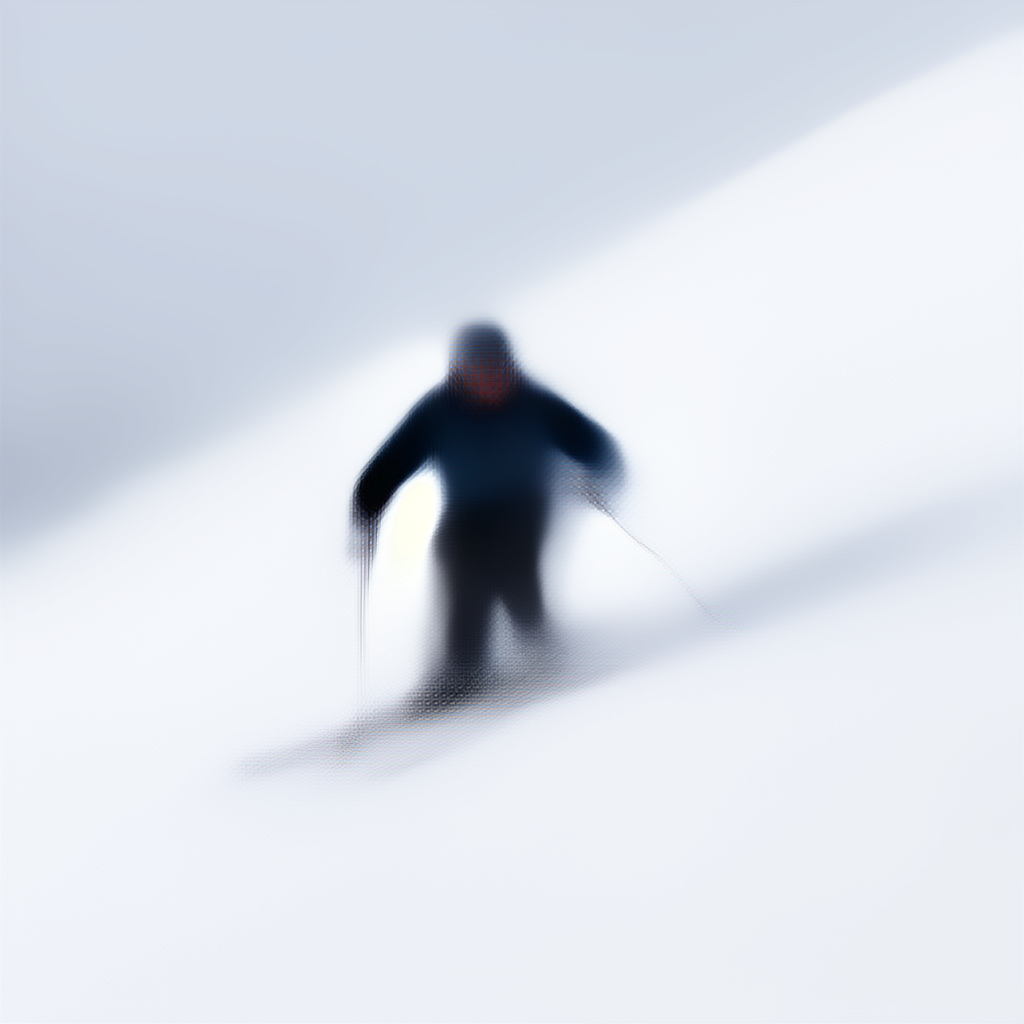}%
            }
            &
            \fcolorbox{black}{white}{%
                \includegraphics[width=\skiwidth]
                {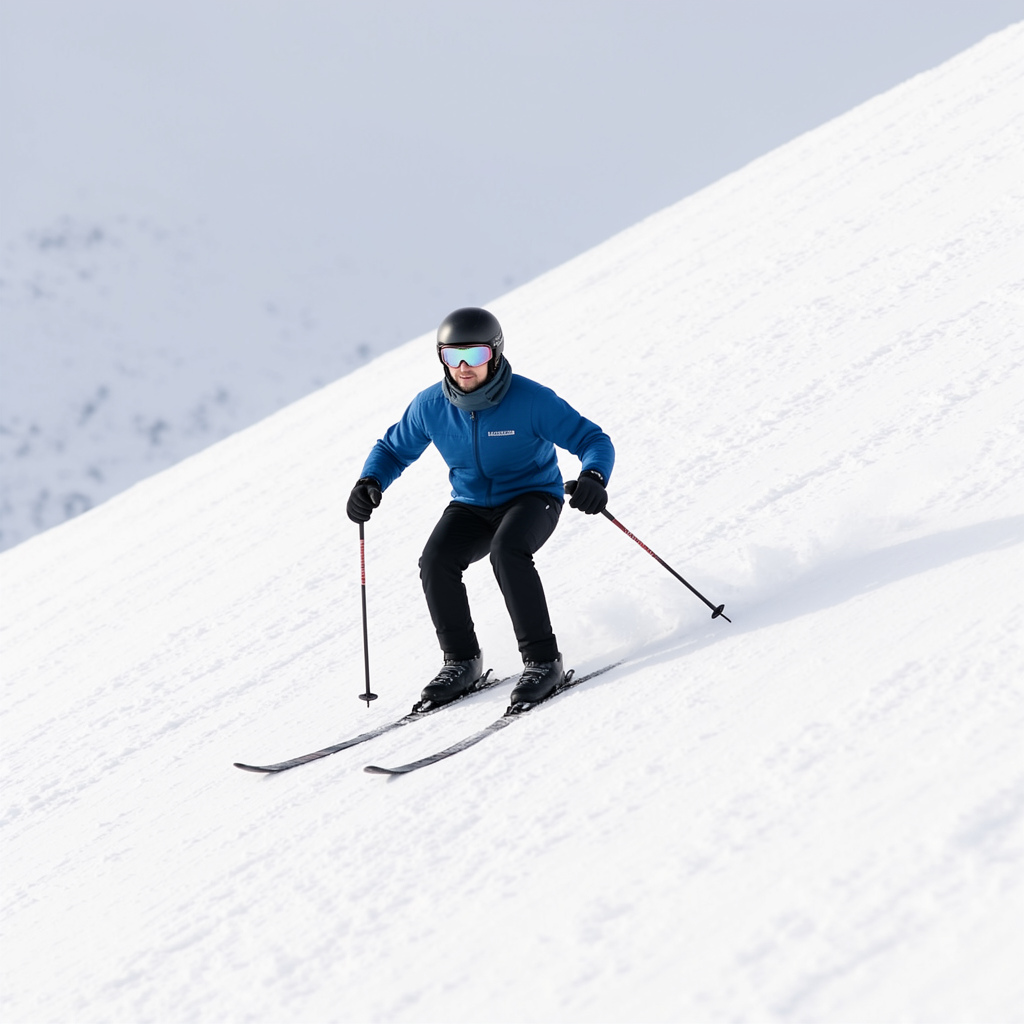}%
            }
        \end{tabular}

    \end{minipage}
    \hfill
    \begin{minipage}[t]{0.49\textwidth}
        \centering

        \newlength{\podiumwidth}
        \setlength{\podiumwidth}{0.23\linewidth}

        \begin{tabular}{@{}c@{\hspace{2pt}}c@{\hspace{2pt}}c@{\hspace{2pt}}c@{}}
            \textcolor{tzero}{\textbf{0\%}} &
            \textcolor{tfour}{\textbf{8\%}} &
            \textcolor{tten}{\textbf{20\%}} &
            \textbf{100\%}
            \\[2pt]

            \fcolorbox{tzero}{white}{%
                \includegraphics[width=\podiumwidth]
                {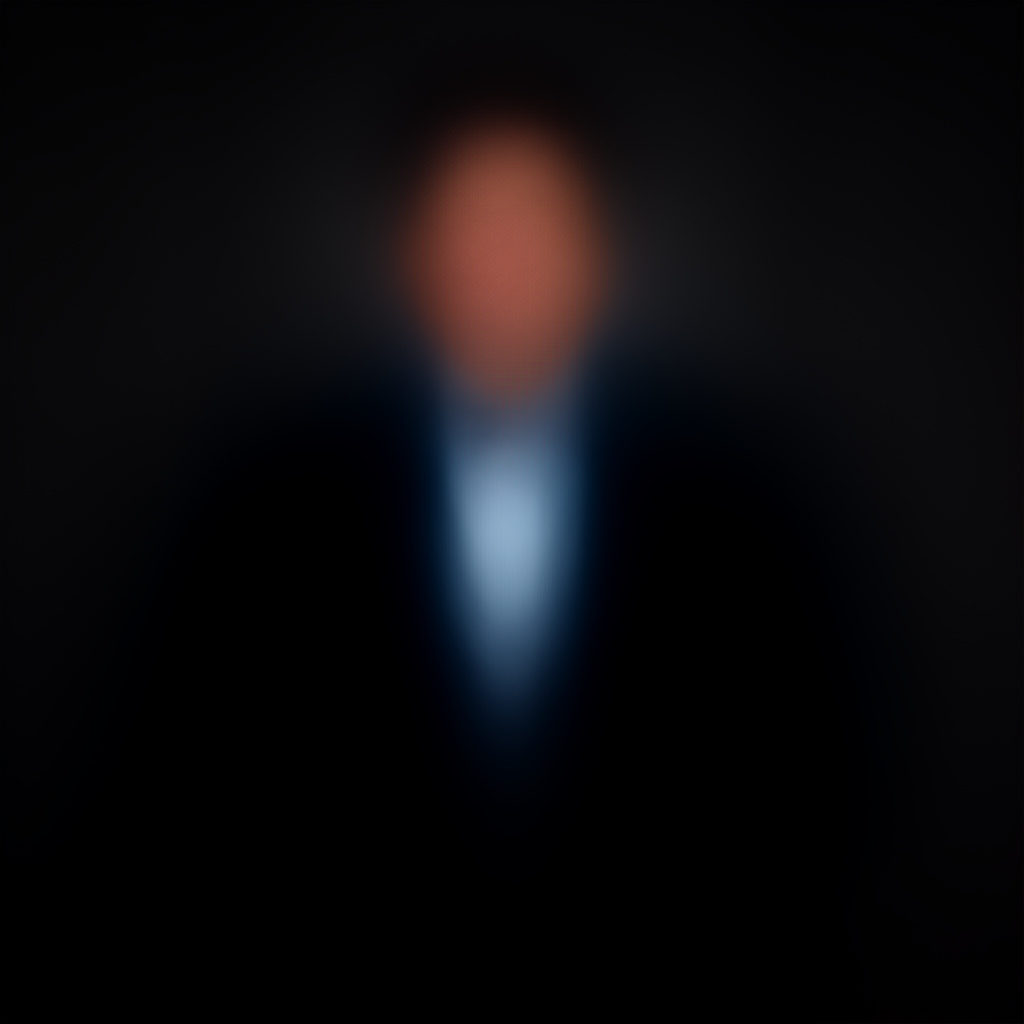}%
            }
            &
            \fcolorbox{tfour}{white}{%
                \includegraphics[width=\podiumwidth]
                {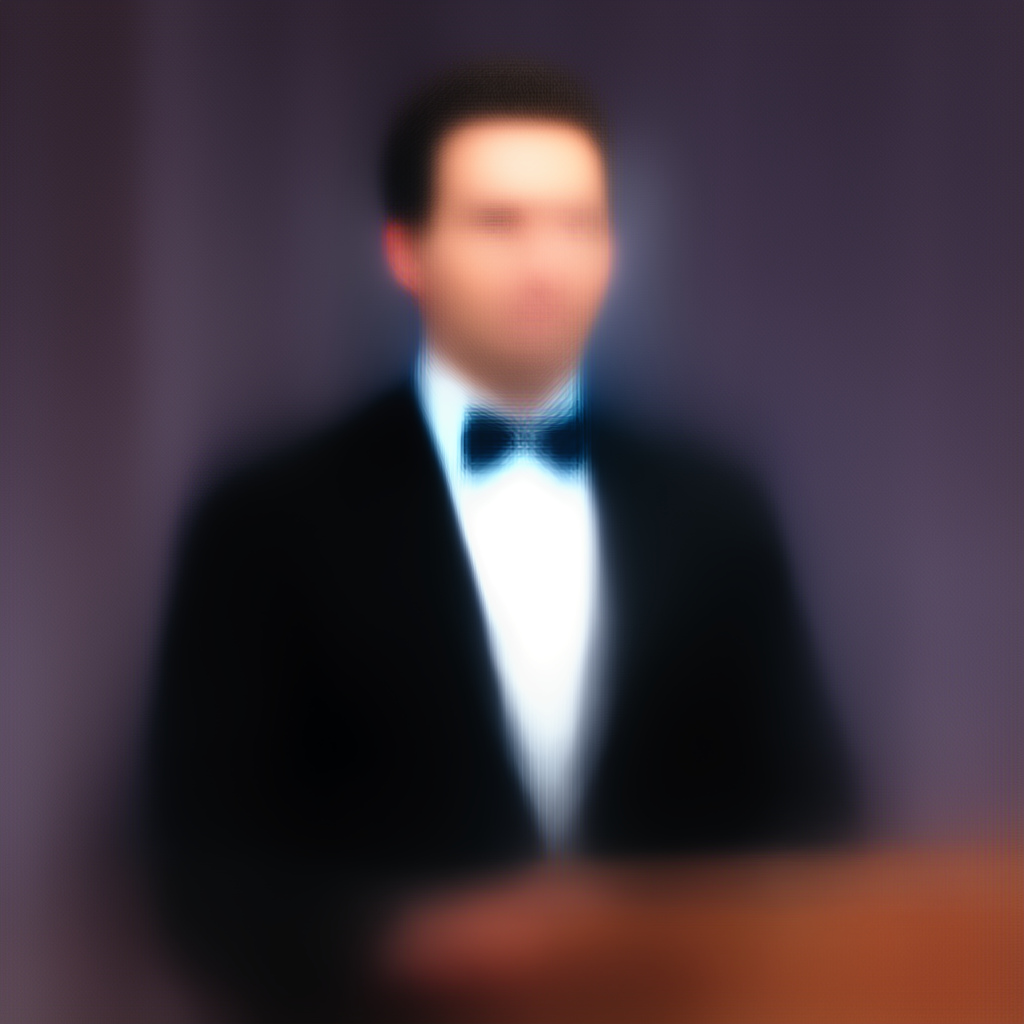}%
            }
            &
            \fcolorbox{tten}{white}{%
                \includegraphics[width=\podiumwidth]
                {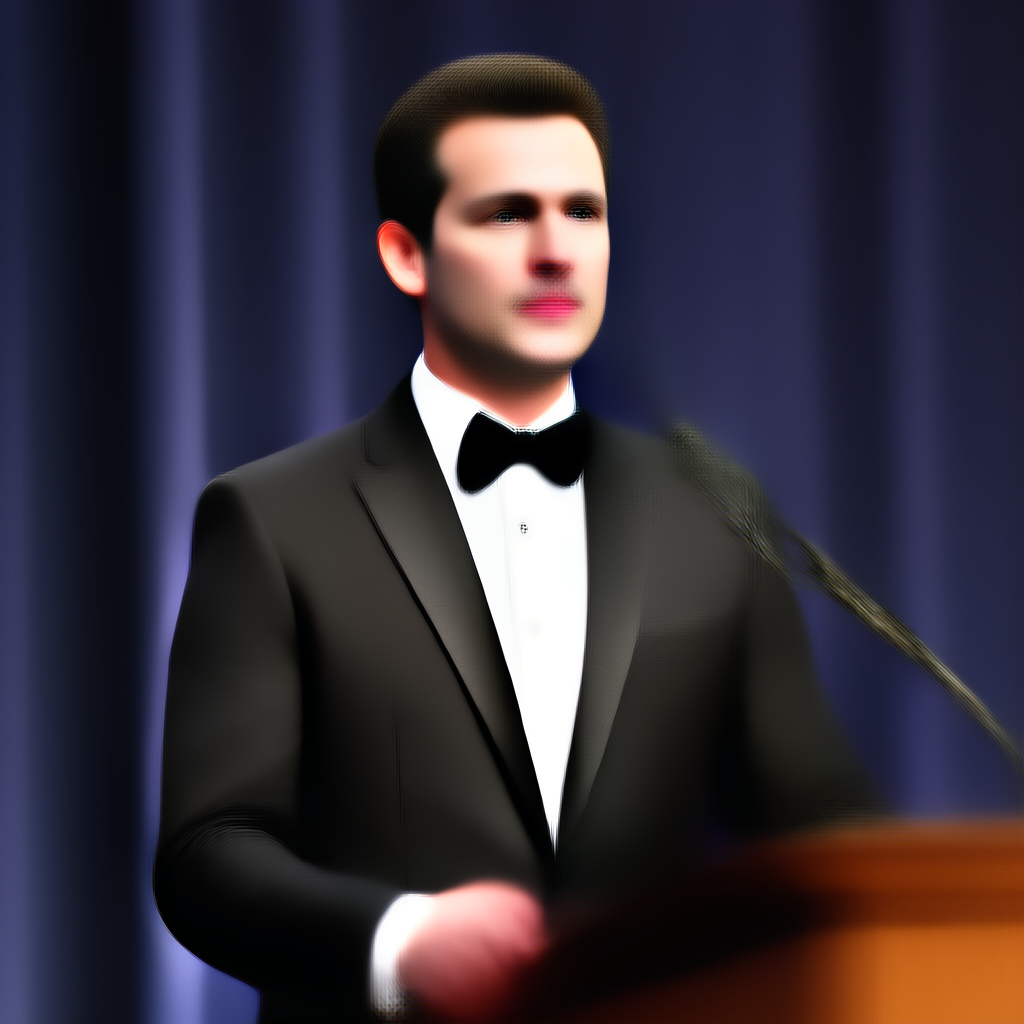}%
            }
            &
            \fcolorbox{black}{white}{%
                \includegraphics[width=\podiumwidth]
                {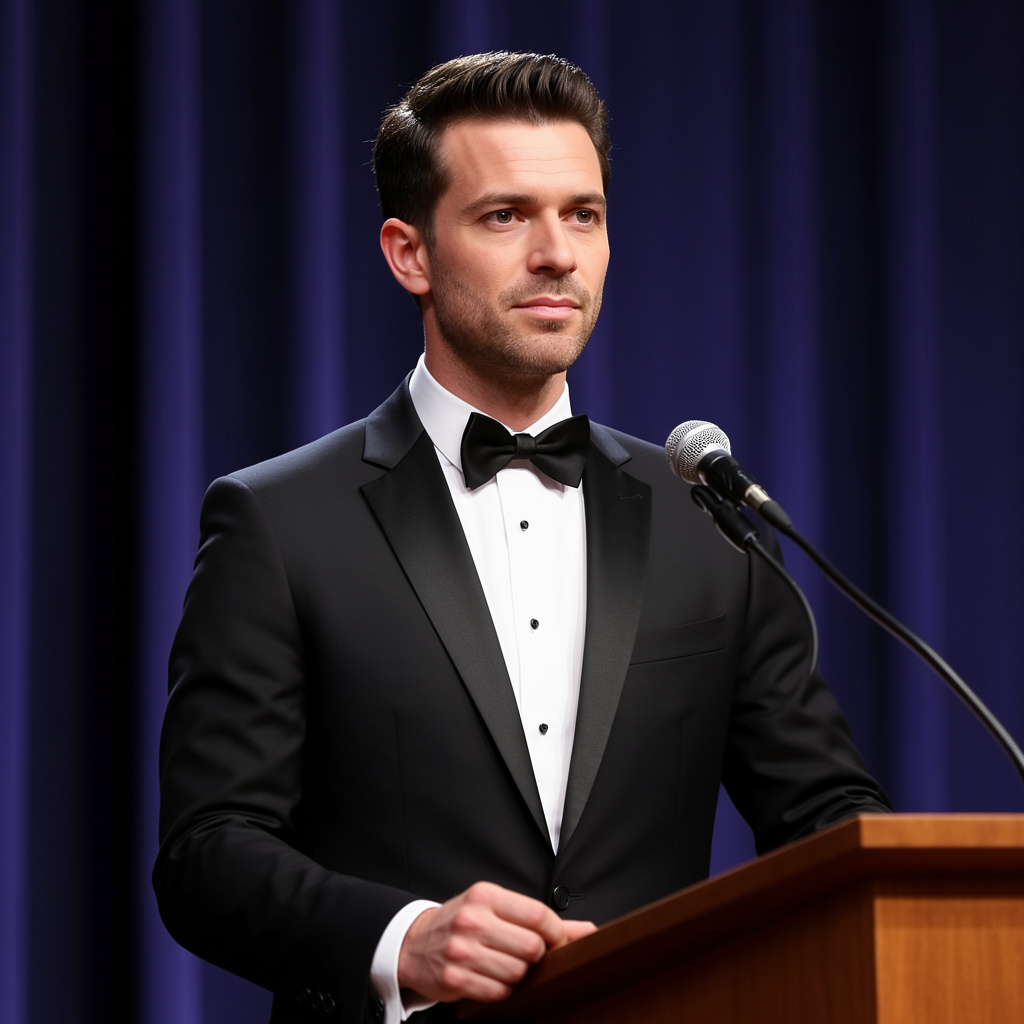}%
            }
        \end{tabular}

    \end{minipage}

    \vspace{3pt}

    \begin{minipage}[t]{0.49\textwidth}
        \centering

        \parbox[t]{\linewidth}{%
            \centering
            \textit{What is the person wearing?}

            \vspace{0.5pt}

            \answerframe{0.55cm}{%
                \answerbox{tzero}{Black shirt and jeans.}
                \hspace{3pt}
                \answerbox{tfour}{Black jacket, gloves, and helmet.}
                \hspace{3pt}
                \answerbox{tten}{Blue jacket, black pants, helmet, gloves, goggles.}
            }%
        }

        \vspace{4pt}

        \parbox[t]{\linewidth}{%
            \centering
            \textit{What is the person's pose?}

            \vspace{0.5pt}

            \answerframe{0.55cm}{%
                \answerbox{tzero}{Sitting on a chair.}
                \hspace{3pt}
                \answerbox{tfour}{Standing on skis with arms outstretched, holding poles.}
                \hspace{3pt}
                \answerbox{tten}{Skiing downhill with knees bent, poles angled forward.}
            }%
        }

        \vspace{4pt}

        \parbox[t]{\linewidth}{%
            \centering
            \textit{What is the person's hair like?}

            \vspace{0.5pt}

            \answerframe{0.35cm}{%
                \answerbox{tzero}{Short and dark.}
                \hspace{3pt}
                \answerbox{tfour}{Unknown.}
                \hspace{3pt}
                \answerbox{tten}{Unknown.}
            }%
        }

    \end{minipage}
    \hfill
    \begin{minipage}[t]{0.49\textwidth}
        \centering

        \parbox[t]{\linewidth}{%
            \centering
            \textit{What objects are visible?}

            \vspace{0.5pt}

            \answerframe{0.55cm}{%
                \answerbox{tzero}{Man, white shirt.}
                \hspace{3pt}
                \answerbox{tfour}{Man, suit, tie, table.}
                \hspace{3pt}
                \answerbox{tten}{Man, suit, tie, microphone, podium, background curtains.}
            }%
        }

        \vspace{4pt}

        \parbox[t]{\linewidth}{%
            \centering
            \textit{What is the person doing?}

            \vspace{0.5pt}

            \answerframe{0.55cm}{%
                \answerbox{tzero}{Sitting at a desk.}
                \hspace{3pt}
                \answerbox{tfour}{Holding a microphone.}
                \hspace{3pt}
                \answerbox{tten}{Giving a speech with a microphone.}
            }%
        }

        \vspace{4pt}

        \parbox[t]{\linewidth}{%
            \centering
            \textit{What is in the background?}

            \vspace{0.5pt}

            \answerframe{0.35cm}{%
                \answerbox{tzero}{Wall.}
                \hspace{3pt}
                \answerbox{tfour}{Plain wall with no decorations.}
                \hspace{3pt}
                \answerbox{tten}{Curtains.}
            }%
        }

    \end{minipage}

    \caption{
        \textbf{Exemplary plain-text predictions from our unconditional
        Contextual Reader on FLUX.2.}
        We show predicted clean images at $0\%$, $8\%$, and
        $20\%$ of the denoising trajectory, together with the final
        generation. Colored highlights associate each intermediate
        timestep with the corresponding reader answer obtained from the contextual tokens at that timestep, shown below. Despite
        receiving no text prompt, the reader recovers increasingly
        specific semantic information about the eventual image as
        denoising progresses.
    }

    \label{fig:unconditional_results}

\end{figure*}

Figure~\ref{fig:unconditional_results} removes the prompt altogether and follows the reader over the denoising trajectory. Even without textual conditioning, highly specific semantics can be decoded early in generation: the reader recognizes that the person is skiing by $8\%$ of the trajectory, and by $20\%$ identifies that they are wearing goggles, before these details are clearly observable in the corresponding visual predictions. Since the model receives no prompt in this setting, these semantics can only arise from the interaction between the contextual tokens and the evolving visual representation. %

Together, these examples show that the contextual tokens can expose surprisingly specific, generation-dependent semantics at early stages of denoising, often before those semantics are clearly expressed in the model's current visual prediction. Additional results are in Appendix~\ref{subsec:additional_qual}.

\paragraph{Quantifying Contextual Readability.}
\begin{figure}[t]
    \centering
    \includegraphics[width=\linewidth]{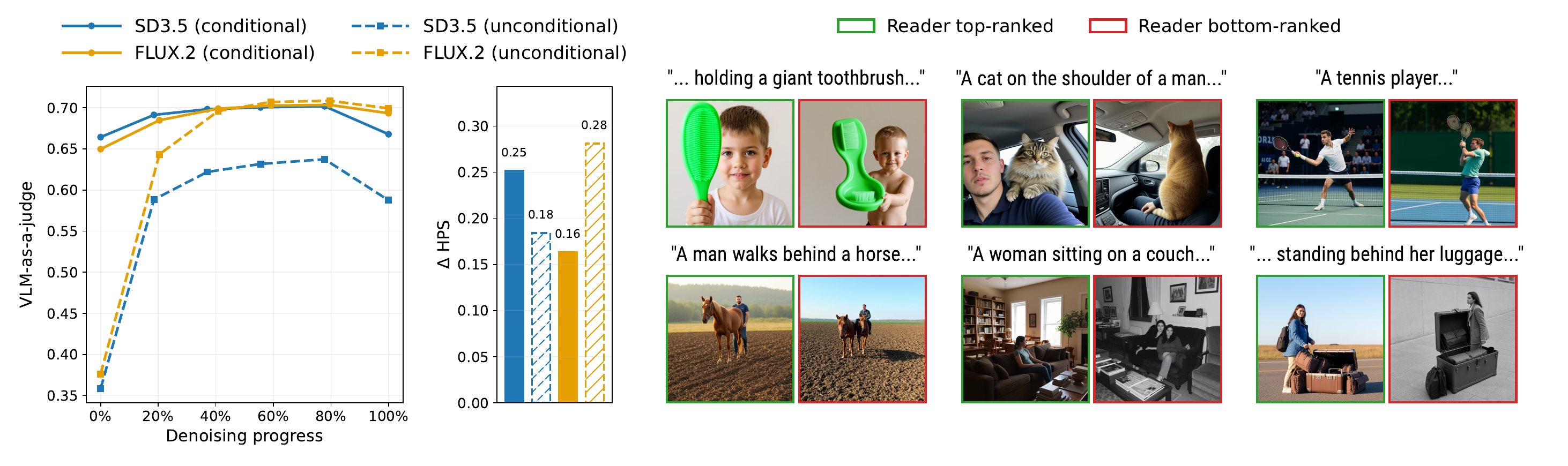}
    \vspace{-16pt}
    \caption{
    \textbf{Contextual readability emerges early and reflects final generation quality.}
    \textbf{Left:} VLM-as-a-judge evaluation across the denoising trajectory shows that contextual scores increase as generation progresses; with the input prompt they are already high from the start. With an empty prompt they rise rapidly as image-specific information enters the contextual space.
    \textbf{Middle:} Ranking seeds by reader scores separates generations with different HPS scores; $\Delta$HPS denotes the HPS difference between the top- and bottom-ranked seeds.
    \textbf{Right:} Representative top- and bottom-ranked generations illustrate that higher contextual readability is associated with stronger final generations. %
    }
    \label{fig:reader_metrics}
\end{figure}

We next quantify the behavior observed qualitatively in Figures~\ref{fig:seed_comparison} and~\ref{fig:unconditional_results}. At each timestep, we use VLM-as-a-judge~\citep{lee2024prometheus} to compare the reader's free-form answer with the ground-truth answer and assess its semantic correctness and specificity (see Appendix~\ref{subsec:vlm_judge} for details and Appendices~\ref{subsec:reader_comparison} and~\ref{subsec:unseen_questions} for additional quantitative analyses). The left panel of Figure~\ref{fig:reader_metrics} shows a consistent trend across both architectures. With the input prompt, the contextual representation is informative from the beginning of generation, but its readability continues to increase throughout denoising, reaching its highest value at $80\%$. Since some questions can already be partially answered from the prompt, this continued improvement indicates that the contextual tokens progressively capture attributes left underspecified by the textual conditioning.
Readability then shows a small decrease at the final $100\%$ timestep, as generation shifts toward refining high-frequency details rather than the semantic properties probed by our questions.

With an empty prompt, performance starts substantially lower and rises rapidly as denoising progresses.
Notably, it eventually surpasses the conditional reader for FLUX.2 and approaches the conditional performance for
SD3.5,
showing that the contextual tokens can become highly semantically informative even without direct access to the prompt.
These semantics emerge despite the contextual tokens never being explicitly supervised to represent visual content. Because they arise solely under the denoising objective, this suggests that such semantic encoding is useful to the generation process itself. We therefore examine whether the clarity of this implicit representation is also related to the quality of the resulting image.

The middle and right panels of Figure~\ref{fig:reader_metrics} address this relationship.
For each prompt, we first rank the five generations according to their reader score averaged across denoising timesteps. We then compare the HPS scores~\citep{ma2025hpsv3} of the top- and bottom-ranked generations.
The middle panel reports $\Delta$HPS, defined as the HPS score of the reader top-ranked generation minus that of the bottom-ranked generation. Across architectures and conditioning settings, $\Delta$HPS is consistently positive, indicating that generations with higher contextual readability tend to receive higher HPS scores.
The right panel provides qualitative examples of this relationship: for each prompt, we show the top-ranked generation in green and the bottom-ranked generation in red. The top- and bottom-ranked generations often differ in both visual fidelity and prompt alignment. For example, the bottom-ranked samples contain visible artifacts, such as a malformed toothbrush or tennis racket, and can also misinterpret the prompt, showing a person riding rather than walking behind a horse, or two women sitting on a couch rather than one.
This indicates that contextual readability provides a meaningful signal for identifying higher-quality final generations. We therefore turn, in the next section, to explicitly supervising the visual-semantic information encoded in the contextual tokens.

\begin{figure}[t]
    \centering
    \includegraphics[width=\linewidth]{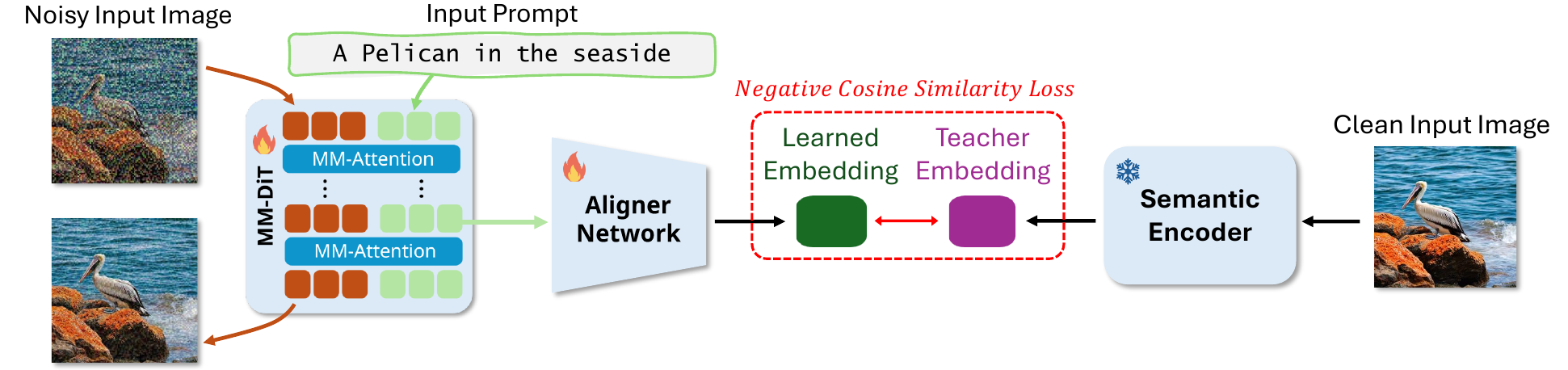}
            \caption{\textbf{Contextual Alignment.} A lightweight Aligner Network maps contextual tokens from the MM-DiT into a learned embedding, optimized via a negative cosine similarity loss to match the teacher embedding produced by a frozen semantic encoder from the clean input image.
}
    \label{fig:coal}
\end{figure}

\section{Supervising the Contextual Space}
\label{sec:contextual_supervision}

The previous section reveals that contextual tokens encode rich, global visual-semantic information about the emerging image, and that the clarity of this information is related to the quality of the resulting generation. We therefore seek to make this implicit semantic structure an explicit training signal. Representation alignment provides a natural mechanism for doing so: a pretrained vision encoder provides a target representation, and the diffusion model is trained to align its intermediate visual features with this target~\citep{yu2024representation,chefer2026self,wang2026repa}. These methods primarily align spatial visual representations, and
have been shown to benefit from spatially structured teacher representations rather than global semantic ones~\citep{singh2025matters}. Since our analysis reveals that contextual tokens encode global visual semantics, we instead use a global semantic teacher to supervise them, providing a complementary form of representation alignment.
We therefore introduce \textbf{Contextual Alignment} (\methodname{}), which aligns contextual tokens with a global semantic representation of the corresponding clean image.

\subsection{Contextual Alignment}
\label{subsec:contextual_alignment}

Our training process is illustrated in Figure~\ref{fig:coal}. During training, a frozen semantic encoder maps the clean image to a global teacher embedding, and a lightweight network, the \emph{Aligner Network}, maps the contextual tokens from a single layer to a predicted embedding. The embedding is then aligned with the teacher via a negative cosine similarity loss. We incorporate this loss into the diffusion objective and optimize the MM-DiT and the Aligner Network jointly; the Aligner Network is used exclusively during training, leaving the MM-DiT's inference cost unchanged.

Seeking a global, semantic, and language-aligned teacher, we use the pooled SigLIP image embedding~\citep{zhai2023sigmoid}, which SigLIP trains to align with caption embeddings through a contrastive image-text loss, making it a compact summary of the image. The Aligner Network itself is a small Q-Former, as in the Contextual Reader, but kept even more minimal: a single layer with one learned query. Its limited capacity makes closing the gap to the teacher depend on adaptation of the MM-DiT, rather than on the Aligner Network absorbing the alignment objective. Likewise, by aligning a single, early layer, we encourage the MM-DiT to encode the teacher's semantics in the contextual tokens early on, leaving the remaining layers free to refine and use them as needed. Both our choice of teacher embedding and our layer selection are confirmed empirically (Appendix~\ref{subsec:coal_ablations}).

\paragraph{Null-Condition Weighting} 

When training text-to-image diffusion models, the text prompt is occasionally dropped to support classifier-free guidance (CFG) at inference~\citep{ho2022classifier}.
When the prompt is provided to the model, the alignment objective has an easy shortcut: the contextual tokens can reduce the alignment loss by leveraging information already provided by the prompt, rather than capturing image-specific semantics that the prompt leaves unspecified.
On null-condition iterations, however, the tokens are initialized from an empty prompt, closing off this shortcut, meaning any alignment must instead be drawn through interaction between the contextual and image tokens. We therefore upweight the alignment loss specifically on null-condition iterations, encouraging a flow of semantic information from the image tokens into the contextual tokens. In practice, we apply this upweighting only in the later stages of full model training, or throughout fine-tuning.

\subsection{Experimental Setup and Baselines}
\label{subsec:coal_setup}

Training a generator from random initialization is the standard protocol for evaluating alignment methods, but for many practitioners the realistic use case is to adapt a pretrained generator to a dataset that reflects what they want to generate. We therefore evaluate \methodname{} in two complementary regimes. First, we train a dual-stream MM-DiT from random initialization on MS-COCO 2014~\citep{lin2014microsoft}. Second, we fine-tune SD3~\citep{esser2024scaling}, a pretrained dual-stream MM-DiT, on the curated subset of Fine-T2I~\citep{ma2026fine}, a dataset specifically designed for T2I fine-tuning that pairs professional photographs with long, detailed prompts. This tests whether contextual supervision remains effective after extensive pretraining, when adapting to a new data distribution.

We compare against REPA~\citep{yu2024representation} and its later variants, HASTE~\citep{wang2026repa} and SRA~\citep{jiang2025no}. For full model training, we test whether \methodname{} complements these visual alignment methods, comparing each objective with and without \methodname{}. For fine-tuning, however, REPA and its variants offer no meaningful gains over the vanilla denoising objective, so we instead combine \methodname{} with vanilla flow matching, including REPA, HASTE, and SRA for completeness. In the main manuscript, we report FID~\citep{heusel2017gans}, CLIP~\citep{radford2021learning}, HPS~\citep{wu2023better}, and DINOv2~\citep{oquab2023dinov2} Precision and Recall~\citep{kynkaanniemi2019improved}. For fine-tuning, we omit CLIP and HPS. These metrics are trained on broad, general-purpose distributions of images, captions, and preferences, so they capture what users prefer on average, which need not coincide with the characteristics of a particular fine-tuning dataset. Indeed, the ground-truth photographs score below the base model on CLIP and HPS, and therefore comparing along these axes is ill-posed. We instead evaluate how well the generated images fit and cover the target distribution, using FID, KID, and DINOv2 Precision and Recall. Our full result tables, including additional metrics and baselines, are provided in Appendix~\ref{subsec:detailed_quantitative}.

\subsection{Main Results}
\label{subsec:coal_results}
\begin{figure}[t]
\centering
\newlength{\qualvimgw}
\setlength{\qualvimgw}{0.15\linewidth}
\setlength{\tabcolsep}{1.5pt}
\renewcommand{\arraystretch}{0.9}
\begin{tabular}{cccccc}

\multicolumn{6}{p{0.95\linewidth}}{\centering\scriptsize\itshape ``A close-up photograph of a single, delicate pink flower with light purple hues, featuring a textured, soft appearance...''} \\[2pt]
\includegraphics[width=\qualvimgw]{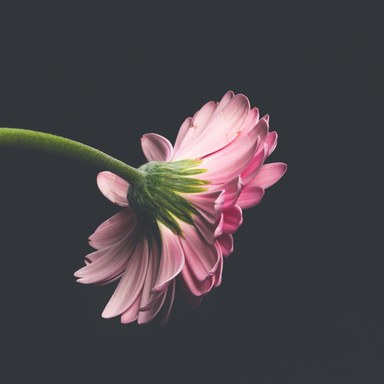} &
\includegraphics[width=\qualvimgw]{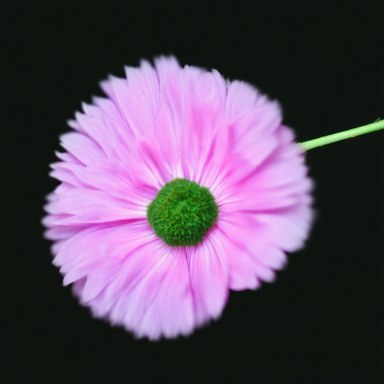} &
\includegraphics[width=\qualvimgw]{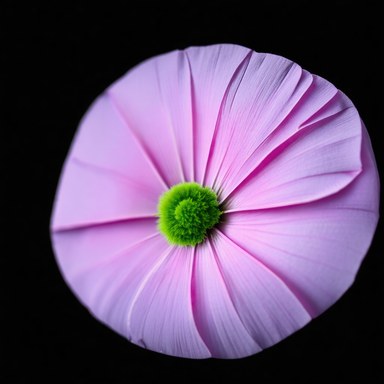} &
\includegraphics[width=\qualvimgw]{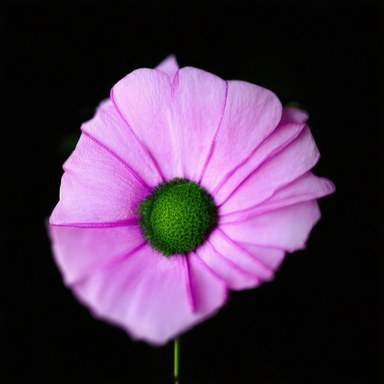} &
\includegraphics[width=\qualvimgw]{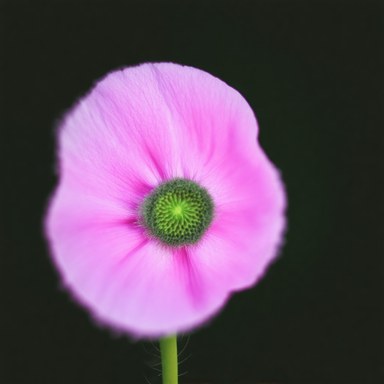} &
\includegraphics[width=\qualvimgw]{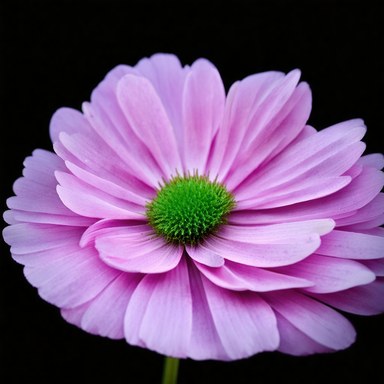} \\[3pt]

\multicolumn{6}{p{0.95\linewidth}}{\centering\scriptsize\itshape ``A bottle of red wine pours into a large wine glass on a round wooden table, set against a smooth, neutral-colored wall.''} \\[2pt]
\includegraphics[width=\qualvimgw]{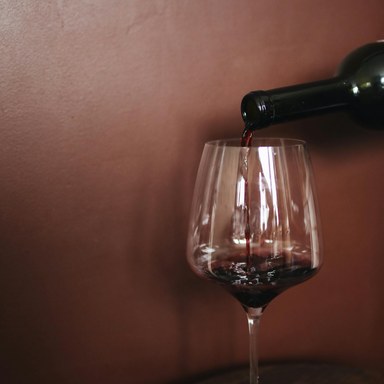} &
\includegraphics[width=\qualvimgw]{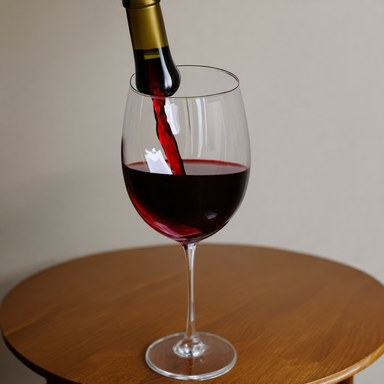} &
\includegraphics[width=\qualvimgw]{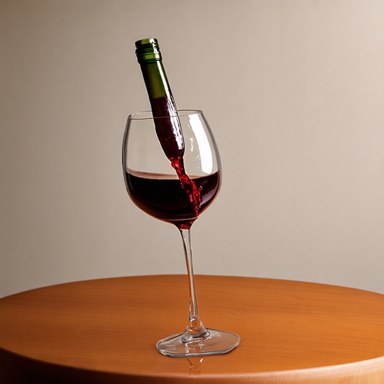} &
\includegraphics[width=\qualvimgw]{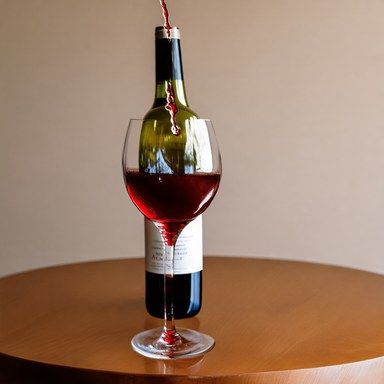} &
\includegraphics[width=\qualvimgw]{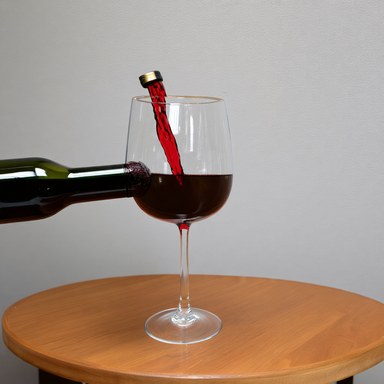} &
\includegraphics[width=\qualvimgw]{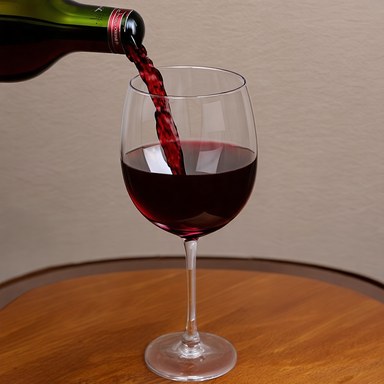} \\[3pt]

\multicolumn{6}{p{0.95\linewidth}}{\centering\scriptsize\itshape ``A woman with curly, shoulder-length brown hair and a {\color{red}relaxed expression}, partially submerged in a bubble bath...''} \\[2pt]
\includegraphics[width=\qualvimgw]{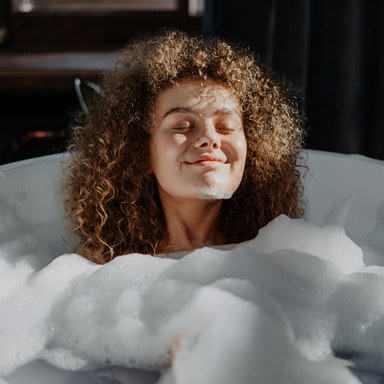} &
\includegraphics[width=\qualvimgw]{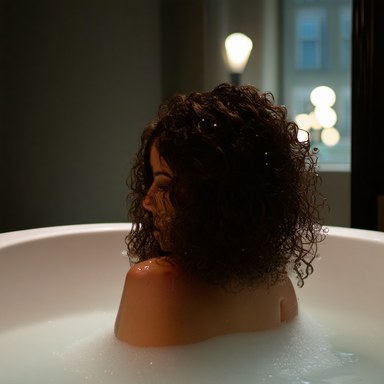} &
\includegraphics[width=\qualvimgw]{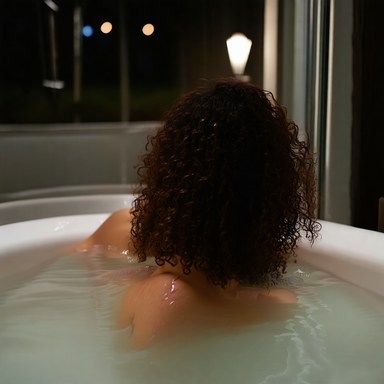} &
\includegraphics[width=\qualvimgw]{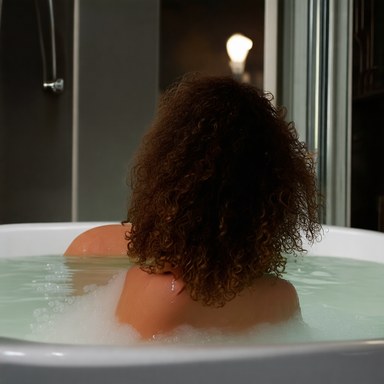} &
\includegraphics[width=\qualvimgw]{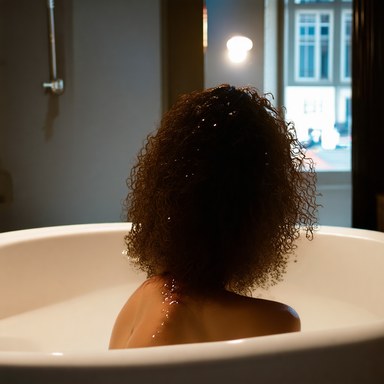} &
\includegraphics[width=\qualvimgw]{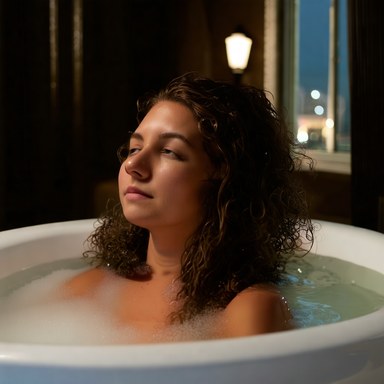} \\[3pt]

\multicolumn{6}{p{0.95\linewidth}}{\centering\scriptsize\itshape ``A {\color{red}Canon} EOS camera with a 24-105mm lens, resting on a wooden surface...''} \\[2pt]
\includegraphics[width=\qualvimgw]{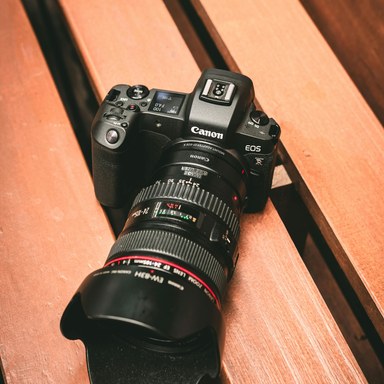} &
\includegraphics[width=\qualvimgw]{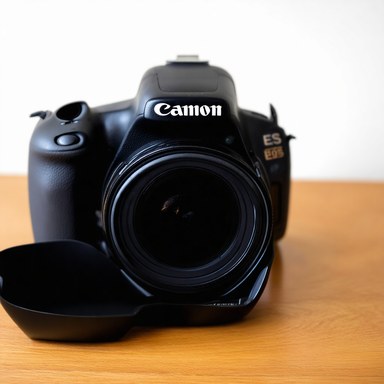} &
\includegraphics[width=\qualvimgw]{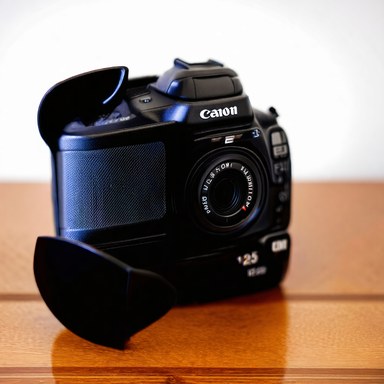} &
\includegraphics[width=\qualvimgw]{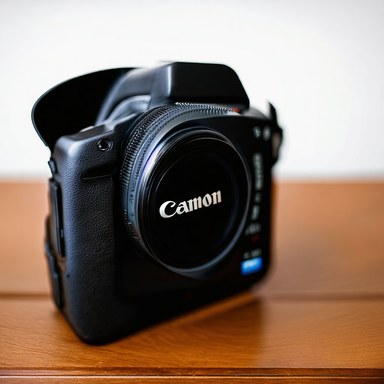} &
\includegraphics[width=\qualvimgw]{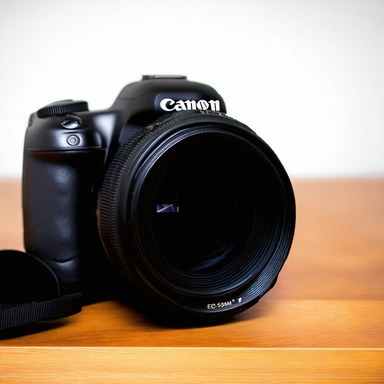} &
\includegraphics[width=\qualvimgw]{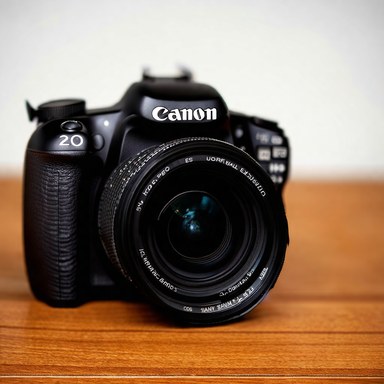} \\[3pt]

\multicolumn{6}{p{0.95\linewidth}}{\centering\scriptsize\itshape ``Two white swans with yellow beaks and orange bills, floating on calm, blue water...''} \\[2pt]
\includegraphics[width=\qualvimgw]{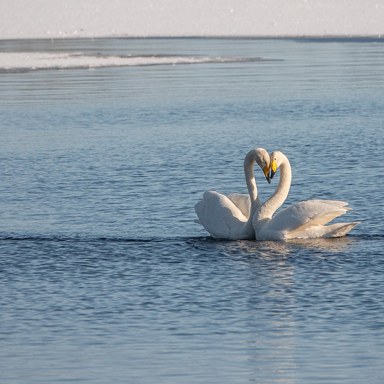} &
\includegraphics[width=\qualvimgw]{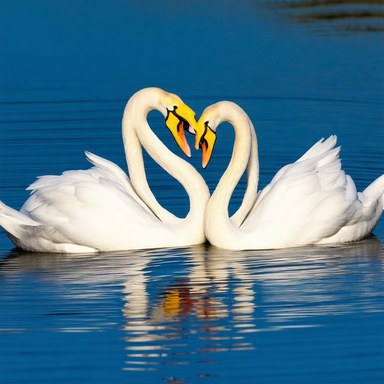} &
\includegraphics[width=\qualvimgw]{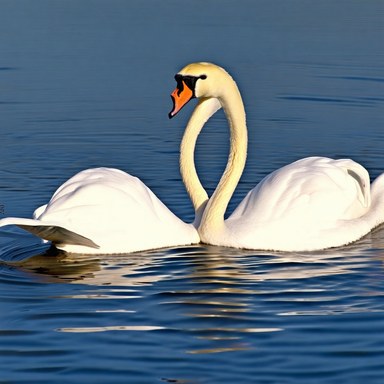} &
\includegraphics[width=\qualvimgw]{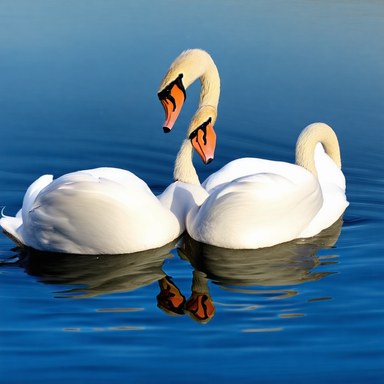} &
\includegraphics[width=\qualvimgw]{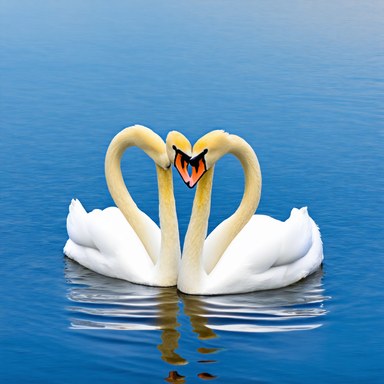} &
\includegraphics[width=\qualvimgw]{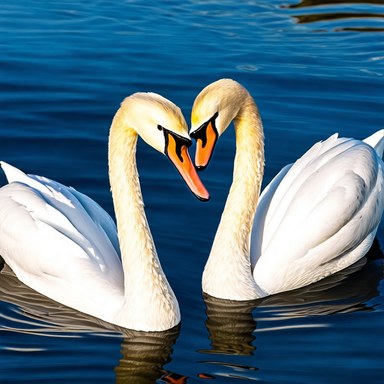} \\[3pt]

\multicolumn{6}{p{0.95\linewidth}}{\centering\scriptsize\itshape ``A close-up, {\color{red}high-resolution photograph} featuring three diverse young women...''} \\[2pt]
\includegraphics[width=\qualvimgw]{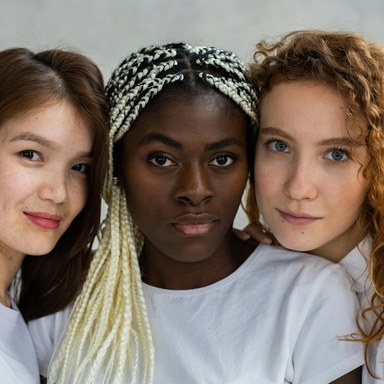} &
\includegraphics[width=\qualvimgw]{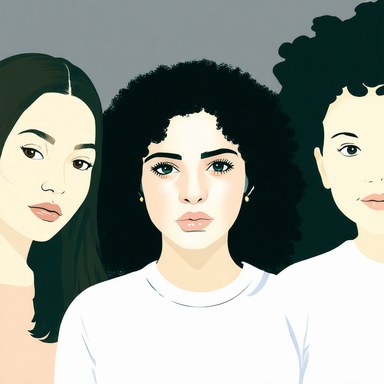} &
\includegraphics[width=\qualvimgw]{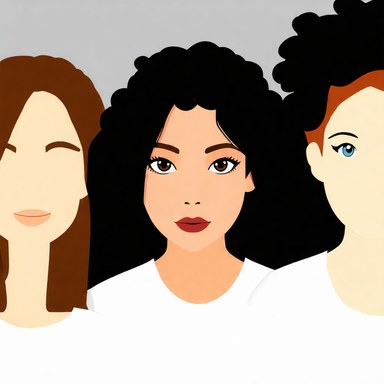} &
\includegraphics[width=\qualvimgw]{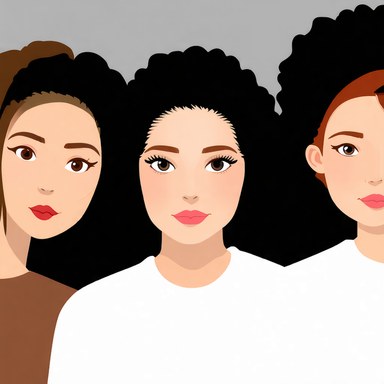} &
\includegraphics[width=\qualvimgw]{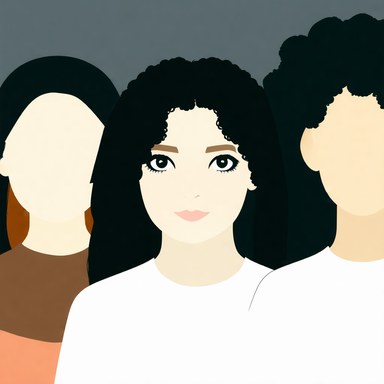} &
\includegraphics[width=\qualvimgw]{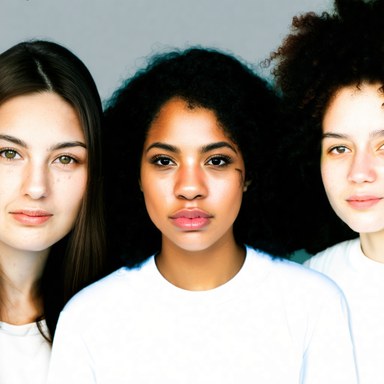} \\[3pt]

\footnotesize Reference & \footnotesize Vanilla & \footnotesize REPA & \footnotesize HASTE & \footnotesize SRA & \footnotesize \textbf{Ours} \\

\end{tabular}
\caption{
\textbf{Qualitative comparison on SD3-Medium fine-tuned on Fine-T2I.} The ``Reference'' column shows the held-out images from Fine-T2I.}
\label{fig:repa_qualitative}
\end{figure}

\begin{table}[t]
    \centering

    \caption{Quantitative comparisons with representation-alignment baselines.}
    \label{tab:side_by_side_comparison}

    \begin{subtable}[t]{0.57\linewidth}
        \centering
        \small
        \setlength{\tabcolsep}{5pt}
        \begin{tabular}{lccccc}
        \toprule
        Method & FID$\downarrow$ & CLIP{\scriptsize$\times10^{2}$}$\uparrow$ &
        HPS{\scriptsize$\times10^{2}$}$\uparrow$ & Prec.$\uparrow$ & Rec.$\uparrow$ \\
        \midrule
        REPA                 & 4.95 & 24.20 & 21.38 & 0.591 & 0.338 \\
        \hspace{0.5em}+ Ours & \textbf{4.29} & \textbf{24.40} & \textbf{21.60} &
        \textbf{0.628} & \textbf{0.407} \\
        \midrule
        HASTE                 & 4.87 & 24.21 & 21.42 & 0.609 & 0.361 \\
        \hspace{0.5em}+ Ours & \textbf{4.44} & \textbf{24.32} & \textbf{21.56} &
        \textbf{0.623} & \textbf{0.400} \\
        \midrule
        SRA                   & 5.89 & 23.49 & 20.37 & 0.515 & 0.261 \\
        \hspace{0.5em}+ Ours & \textbf{4.36} & \textbf{24.28} & \textbf{21.42} &
        \textbf{0.616} & \textbf{0.398} \\
        \bottomrule
        \end{tabular}
        \caption{Full model training.}
        \label{tab:mscoco_scratch_comparison_side}
    \end{subtable}%
    \hfill
    \begin{subtable}[t]{0.37\linewidth}
        \centering
        \small
        \setlength{\tabcolsep}{3pt}
        \begin{tabular}{lccc}
        \toprule
        Method & FID$\downarrow$ & Prec.$\uparrow$ & Rec.$\uparrow$ \\
        \midrule
        Base Model      & 19.30 & 0.993 & 0.939 \\
        Vanilla Flow  & 17.18 & 0.993 & 0.966 \\
        REPA            & 17.45 & 0.993 & 0.966 \\
        HASTE           & 17.53 & \textbf{0.994} & 0.963 \\
        SRA             & 17.48 & 0.991 & 0.964 \\
        \midrule
        Ours            & \textbf{16.75} & 0.993 & \textbf{0.969} \\
        \bottomrule
        \end{tabular}
        \caption{Model fine-tuning.}
        \label{tab:finet2i_comparison_side}
    \end{subtable}

\end{table}

\paragraph{Quantitative results.}
Table~\ref{tab:mscoco_scratch_comparison_side} reports results for training a new model from random weights. Contextual Alignment improves all three representation-alignment baselines across all evaluated metrics, further supporting the complementary role of global semantic alignment in the contextual space. Table~\ref{tab:finet2i_comparison_side} reports results for our fine-tuning experiment. In this setting, REPA, HASTE, and SRA provide no meaningful gains over vanilla fine-tuning, with all three achieving worse FID than this baseline. Contextual Alignment continues to improve generation quality, achieving the best FID and Recall, while remaining competitive in Precision. This demonstrates the effectiveness of contextual semantic supervision for fine-tuning pretrained generators.

\paragraph{Qualitative results.}
We present qualitative comparisons with the baseline in Figure~\ref{fig:repa_qualitative}, where we observe improvements across several aspects of generation, including overall image quality and sharpness, subject anatomy, compositional coherence, and prompt following.
We present additional results in Appendix~\ref{subsec:coal_additional_qual}.

\section{Conclusions}

We have presented a framework for reading the internal perceptual representations of generative models through natural language. More broadly, our work suggests that a generative model knows surprisingly early what it is going to generate. Semantic details of the final image are already present internally before they become perceptually apparent in the emerging image. Even more strikingly, this remains true without the text prompt: the model develops a rich semantic picture of the emerging image through the generation process itself. This observation also has a useful consequence: generations in which this internal representation is clearer tend to result in better images. This suggests that the representation is not only something we can read, but also something worth strengthening. Our Contextual Alignment does exactly this, leading to improved generations.

Our findings also highlight how much remains to be understood about the internal process of generative models. We can now read some of the semantic information that emerges during generation, but how these internal representations form, evolve, and ultimately shape the generated image remains largely unexplored. The fact that meaningful information appears so early, even before it becomes perceptually apparent, raises many new questions about the generation process itself. We believe that further studying these internal representations can deepen our understanding of generative models and reveal new opportunities for controlling and improving them.

\section*{Acknowledgements}
We thank Maya Vishnevsky, Daniel Garibi, Nir Goren, Wenxuan Peng, and Hao Phung for their early feedback and insightful discussions.

\bibliography{main}

@article{saharia2022photorealistic,
  title={Photorealistic text-to-image diffusion models with deep language understanding},
  author={Saharia, Chitwan and Chan, William and Saxena, Saurabh and Li, Lala and Whang, Jay and Denton, Emily L and Ghasemipour, Kamyar and Gontijo Lopes, Raphael and Karagol Ayan, Burcu and Salimans, Tim and others},
  journal={Advances in Neural Information Processing Systems},
  volume={35},
  pages={36479--36494},
  year={2022}
}

@article{ramesh2022hierarchical,
  title={Hierarchical text-conditional image generation with clip latents},
  author={Ramesh, Aditya and Dhariwal, Prafulla and Nichol, Alex and Chu, Casey and Chen, Mark},
  journal={arXiv preprint arXiv:2204.06125},
  volume={1},
  number={2},
  pages={3},
  year={2022}
}

@inproceedings{rombach2022high,
  title={High-resolution image synthesis with latent diffusion models},
  author={Rombach, Robin and Blattmann, Andreas and Lorenz, Dominik and Esser, Patrick and Ommer, Bj{\"o}rn},
  booktitle={Proceedings of the IEEE/CVF conference on computer vision and pattern recognition},
  pages={10684--10695},
  year={2022}
}

@article{podell2023sdxl,
  title={Sdxl: Improving latent diffusion models for high-resolution image synthesis},
  author={Podell, Dustin and English, Zion and Lacey, Kyle and Blattmann, Andreas and Dockhorn, Tim and M{\"u}ller, Jonas and Penna, Joe and Rombach, Robin},
  journal={arXiv preprint arXiv:2307.01952},
  year={2023}
}

@inproceedings{peebles2023scalable,
  title={Scalable diffusion models with transformers},
  author={Peebles, William and Xie, Saining},
  booktitle={2023 IEEE/CVF International Conference on Computer Vision (ICCV)},
  pages={4172--4182},
  year={2023},
  organization={IEEE}
}

@inproceedings{esser2024scaling,
  title={Scaling rectified flow transformers for high-resolution image synthesis},
  author={Esser, Patrick and Kulal, Sumith and Blattmann, Andreas and Entezari, Rahim and M{\"u}ller, Jonas and Saini, Harry and Levi, Yam and Lorenz, Dominik and Sauer, Axel and Boesel, Frederic and others},
  booktitle={Forty-first international conference on machine learning},
  year={2024}
}

@misc{flux2024,
    author={Black Forest Labs},
    title={FLUX},
    year={2024},
    howpublished={\url{https://github.com/black-forest-labs/flux}},
}

@article{labs2025flux,
  title={Flux. 1 kontext: Flow matching for in-context image generation and editing in latent space},
  author={Labs, Black Forest and Batifol, Stephen and Blattmann, Andreas and Boesel, Frederic and Consul, Saksham and Diagne, Cyril and Dockhorn, Tim and English, Jack and English, Zion and Esser, Patrick and others},
  journal={arXiv preprint arXiv:2506.15742},
  year={2025}
}

@article{kong2024hunyuanvideo,
  title={Hunyuanvideo: A systematic framework for large video generative models},
  author={Kong, Weijie and Tian, Qi and Zhang, Zijian and Min, Rox and Dai, Zuozhuo and Zhou, Jin and Xiong, Jiangfeng and Li, Xin and Wu, Bo and Zhang, Jianwei and others},
  journal={arXiv preprint arXiv:2412.03603},
  year={2024}
}

@inproceedings{yang2025cogvideox,
  title={Cogvideox: Text-to-video diffusion models with an expert transformer},
  author={Yang, Zhuoyi and Teng, Jiayan and Zheng, Wendi and Ding, Ming and Huang, Shiyu and Xu, Jiazheng and Yang, Yuanming and Hong, Wenyi and Zhang, Xiaohan and Feng, Guanyu and others},
  booktitle={International Conference on Learning Representations},
  volume={2025},
  pages={83048--83077},
  year={2025}
}

@inproceedings{bao2023all,
  title={All are worth words: A vit backbone for diffusion models},
  author={Bao, Fan and Nie, Shen and Xue, Kaiwen and Cao, Yue and Li, Chongxuan and Su, Hang and Zhu, Jun},
  booktitle={2023 IEEE/CVF Conference on Computer Vision and Pattern Recognition (CVPR)},
  pages={22669--22679},
  year={2023},
  organization={IEEE}
}

@inproceedings{chen2024pixartalpha,
  title={Pixart-$\alpha$: Fast training of diffusion transformer for photorealistic text-to-image synthesis},
  author={Chen, Junsong and Yu, Jincheng and Ge, Chongjian and Yao, Lewei and Xie, Enze and Wang, Zhongdao and Kwok, James and Luo, Ping and Lu, Huchuan and Li, Zhenguo},
  booktitle={International conference on learning representations},
  volume={2024},
  pages={57611--57640},
  year={2024}
}

@inproceedings{chen2024pixartsigma,
  title={Pixart-$\sigma$: Weak-to-strong training of diffusion transformer for 4k text-to-image generation},
  author={Chen, Junsong and Ge, Chongjian and Xie, Enze and Wu, Yue and Yao, Lewei and Ren, Xiaozhe and Wang, Zhongdao and Luo, Ping and Lu, Huchuan and Li, Zhenguo},
  booktitle={European Conference on Computer Vision},
  pages={74--91},
  year={2024},
  organization={Springer}
}

@article{zhuo2024lumina,
  title={Lumina-next: Making lumina-t2x stronger and faster with next-dit},
  author={Zhuo, Le and Du, Ruoyi and Xiao, Han and Li, Yangguang and Liu, Dongyang and Huang, Rongjie and Liu, Wenze and Zhu, Xiangyang and Wang, Fu-Yun and Ma, Zhanyu and others},
  journal={Advances in Neural Information Processing Systems},
  volume={37},
  pages={131278--131315},
  year={2024}
}

@inproceedings{xiao2025omnigen,
  title={Omnigen: Unified image generation},
  author={Xiao, Shitao and Wang, Yueze and Zhou, Junjie and Yuan, Huaying and Xing, Xingrun and Yan, Ruiran and Li, Chaofan and Wang, Shuting and Huang, Tiejun and Liu, Zheng},
  booktitle={2025 IEEE/CVF Conference on Computer Vision and Pattern Recognition (CVPR)},
  pages={13294--13304},
  year={2025},
  organization={IEEE}
}

@article{cai2025hidream,
  title={Hidream-i1: A high-efficient image generative foundation model with sparse diffusion transformer},
  author={Cai, Qi and Chen, Jingwen and Chen, Yang and Li, Yehao and Long, Fuchen and Pan, Yingwei and Qiu, Zhaofan and Zhang, Yiheng and Gao, Fengbin and Xu, Peihan and others},
  journal={arXiv preprint arXiv:2505.22705},
  year={2025}
}

@article{wu2025qwen,
  title={Qwen-image technical report},
  author={Wu, Chenfei and Li, Jiahao and Zhou, Jingren and Lin, Junyang and Gao, Kaiyuan and Yan, Kun and Yin, Sheng-ming and Bai, Shuai and Xu, Xiao and Chen, Yilei and others},
  journal={arXiv preprint arXiv:2508.02324},
  year={2025}
}

@article{zhao2026qwen,
  title={Qwen-image-2.0 technical report},
  author={Zhao, Bing and Wu, Chenfei and Li, Deqing and Meng, Hao and Li, Jiahao and Zhang, Jie and Zhou, Jingren and Lin, Junyang and Gao, Kaiyuan and Cao, Kuan and others},
  journal={arXiv preprint arXiv:2605.10730},
  year={2026}
}

@article{chefer2025videojam,
  title={Videojam: Joint appearance-motion representations for enhanced motion generation in video models},
  author={Chefer, Hila and Singer, Uriel and Zohar, Amit and Kirstain, Yuval and Polyak, Adam and Taigman, Yaniv and Wolf, Lior and Sheynin, Shelly},
  journal={arXiv preprint arXiv:2502.02492},
  year={2025}
}

@article{wang2025kling,
  title={Kling-foley: Multimodal diffusion transformer for high-quality video-to-audio generation},
  author={Wang, Jun and Zeng, Xijuan and Qiang, Chunyu and Chen, Ruilong and Wang, Shiyao and Wang, Le and Zhou, Wangjing and Cai, Pengfei and Zhao, Jiahui and Li, Nan and others},
  journal={arXiv preprint arXiv:2506.19774},
  year={2025}
}

@article{shan2025hunyuanvideo,
  title={Hunyuanvideo-foley: Multimodal diffusion with representation alignment for high-fidelity foley audio generation},
  author={Shan, Sizhe and Li, Qiulin and Cui, Yutao and Yang, Miles and Wang, Yuehai and Yang, Qun and Zhou, Jin and Zhong, Zhao},
  journal={arXiv preprint arXiv:2508.16930},
  year={2025}
}

@inproceedings{wang2026audiogen,
  title={Audiogen-omni: A unified multimodal diffusion transformer for video-synchronized audio, speech, and song generation},
  author={Wang, Le and Wang, Jun and Qiang, Chunyu and Deng, Feng and Zhang, Chen and Gai, Kun},
  booktitle={ICASSP 2026-2026 IEEE International Conference on Acoustics, Speech and Signal Processing (ICASSP)},
  pages={16047--16051},
  year={2026},
  organization={IEEE}
}

@article{ho2022classifier,
  title={Classifier-free diffusion guidance},
  author={Ho, Jonathan and Salimans, Tim},
  journal={arXiv preprint arXiv:2207.12598},
  year={2022}
}

@inproceedings{zhang2023adding,
  title={Adding conditional control to text-to-image diffusion models},
  author={Zhang, Lvmin and Rao, Anyi and Agrawala, Maneesh},
  booktitle={2023 IEEE/CVF International Conference on Computer Vision (ICCV)},
  pages={3813--3824},
  year={2023},
  organization={IEEE}
}

@article{meng2021sdedit,
  title={Sdedit: Guided image synthesis and editing with stochastic differential equations},
  author={Meng, Chenlin and He, Yutong and Song, Yang and Song, Jiaming and Wu, Jiajun and Zhu, Jun-Yan and Ermon, Stefano},
  journal={arXiv preprint arXiv:2108.01073},
  year={2021}
}

@inproceedings{avrahami2022blended,
  title={Blended diffusion for text-driven editing of natural images},
  author={Avrahami, Omri and Lischinski, Dani and Fried, Ohad},
  booktitle={2022 IEEE/CVF Conference on Computer Vision and Pattern Recognition (CVPR)},
  pages={18187--18197},
  year={2022},
  organization={IEEE}
}

@article{hertz2022prompt,
  title={Prompt-to-prompt image editing with cross attention control},
  author={Hertz, Amir and Mokady, Ron and Tenenbaum, Jay and Aberman, Kfir and Pritch, Yael and Cohen-Or, Daniel},
  journal={arXiv preprint arXiv:2208.01626},
  year={2022}
}

@article{chefer2023attend,
  title={Attend-and-excite: Attention-based semantic guidance for text-to-image diffusion models},
  author={Chefer, Hila and Alaluf, Yuval and Vinker, Yael and Wolf, Lior and Cohen-Or, Daniel},
  journal={ACM transactions on Graphics (TOG)},
  volume={42},
  number={4},
  pages={1--10},
  year={2023},
  publisher={ACM New York, NY, USA}
}

@article{yu2024representation,
  title={Representation alignment for generation: Training diffusion transformers is easier than you think},
  author={Yu, Sihyun and Kwak, Sangkyung and Jang, Huiwon and Jeong, Jongheon and Huang, Jonathan and Shin, Jinwoo and Xie, Saining},
  journal={arXiv preprint arXiv:2410.06940},
  year={2024}
}

@inproceedings{leng2025repa,
  title={Repa-e: Unlocking vae for end-to-end tuning with latent diffusion transformers},
  author={Leng, Xingjian and Singh, Jaskirat and Hou, Yunzhong and Xing, Zhenchang and Xie, Saining and Zheng, Liang},
  booktitle={2025 IEEE/CVF International Conference on Computer Vision (ICCV)},
  pages={18262--18272},
  year={2025},
  organization={IEEE}
}

@article{jiang2025no,
  title={No other representation component is needed: Diffusion transformers can provide representation guidance by themselves},
  author={Jiang, Dengyang and Wang, Mengmeng and Li, Liuzhuozheng and Zhang, Lei and Wang, Haoyu and Wei, Wei and Dai, Guang and Zhang, Yanning and Wang, Jingdong},
  journal={arXiv preprint arXiv:2505.02831},
  year={2025}
}

@inproceedings{singh2025matters,
  title={What matters for representation alignment: Global information or spatial structure?},
  author={Singh, Jaskirat and Leng, Xingjian and Wu, Zongze and Zheng, Liang and Zhang, Richard and Shechtman, Eli and Xie, Saining},
  booktitle={The Fourteenth International Conference on Learning Representations},
  year={2025}
}

@article{chefer2026self,
  title={Self-supervised flow matching for scalable multi-modal synthesis},
  author={Chefer, Hila and Esser, Patrick and Lorenz, Dominik and Podell, Dustin and Raja, Vikash and Tong, Vinh and Torralba, Antonio and Rombach, Robin},
  journal={arXiv preprint arXiv:2603.06507},
  year={2026}
}

@article{wu2026representation,
  title={Representation entanglement for generation: Training diffusion transformers is much easier than you think},
  author={Wu, Ge and Zhang, Shen and Shi, Ruijing and Gao, Shanghua and Chen, Zhenyuan and Wang, Lei and Chen, Zhaowei and Gao, Hongcheng and Tang, Yao and Cheng, Ming-Ming and others},
  journal={Advances in Neural Information Processing Systems},
  volume={38},
  pages={7714--7743},
  year={2026}
}

@inproceedings{zheng2026diffusion,
  title={Diffusion transformers with representation autoencoders},
  author={Zheng, Boyang and Ma, Nanye and Tong, Shengbang and Xie, Saining},
  booktitle={International Conference on Learning Representations},
  volume={2026},
  pages={35791--35820},
  year={2026}
}

@article{lee2026aligning,
  title={Aligning text to image in diffusion models is easier than you think},
  author={Lee, Jaa-Yeon and Cha, Byunghee and Kim, Jeongsol and Ye, Jong Chul},
  journal={Advances in Neural Information Processing Systems},
  volume={38},
  pages={157106--157136},
  year={2026}
}

@inproceedings{dahary2026fly,
  title={On-the-fly Repulsion in the Contextual Space for Rich Diversity in Diffusion Transformers},
  author={Dahary, Omer and Koren, Benaya and Garibi, Daniel and Cohen-Or, Daniel},
  booktitle={Proceedings of the Special Interest Group on Computer Graphics and Interactive Techniques Conference Conference Papers},
  pages={1--11},
  year={2026}
}

@inproceedings{toker2025padding,
  title={Padding tone: A mechanistic analysis of padding tokens in t2i models},
  author={Toker, Michael and Galil, Ido and Orgad, Hadas and Gal, Rinon and Tewel, Yoad and Chechik, Gal and Belinkov, Yonatan},
  booktitle={Proceedings of the 2025 Conference of the Nations of the Americas Chapter of the Association for Computational Linguistics: Human Language Technologies (Volume 1: Long Papers)},
  pages={7618--7632},
  year={2025}
}

@article{helbling2025conceptattention,
  title={Conceptattention: Diffusion transformers learn highly interpretable features},
  author={Helbling, Alec and Meral, Tuna Han Salih and Hoover, Ben and Yanardag, Pinar and Chau, Duen Horng},
  journal={arXiv preprint arXiv:2502.04320},
  year={2025}
}

@inproceedings{ge2026vision,
  title={Vision-Language Binding in In-Context Image Generation},
  author={Ge, Chris and Gandikota, Rohit and Torralba, Antonio and Shaham, Tamar Rott},
  booktitle={Mechanistic Interpretability Workshop at ICML 2026},
  year={2026}
}

@article{li2026text,
  title={Text Template Tokens Are Implicit Semantic Registers in Diffusion Transformers},
  author={Li, Maohua and Li, Qirui and Zhou, Yanke and Li, Yiduo and Chi, Zhaosheng and Xu, Chao and Shen, Cuifeng and Xu, Yixuan and Tang, Hanlin and Liu, Kan and others},
  journal={arXiv preprint arXiv:2607.19139},
  year={2026}
}

@inproceedings{tumanyan2023plug,
  title={Plug-and-play diffusion features for text-driven image-to-image translation},
  author={Tumanyan, Narek and Geyer, Michal and Bagon, Shai and Dekel, Tali},
  booktitle={2023 IEEE/CVF Conference on Computer Vision and Pattern Recognition (CVPR)},
  pages={1921--1930},
  year={2023},
  organization={IEEE}
}

@inproceedings{dahary2024yourself,
  title={Be yourself: Bounded attention for multi-subject text-to-image generation},
  author={Dahary, Omer and Patashnik, Or and Aberman, Kfir and Cohen-Or, Daniel},
  booktitle={European Conference on Computer Vision},
  pages={432--448},
  year={2024},
  organization={Springer}
}

@inproceedings{dahary2025decisive,
  title={Be decisive: Noise-induced layouts for multi-subject generation},
  author={Dahary, Omer and Cohen, Yehonathan and Patashnik, Or and Aberman, Kfir and Cohen-Or, Daniel},
  booktitle={Proceedings of the Special Interest Group on Computer Graphics and Interactive Techniques Conference Conference Papers},
  pages={1--12},
  year={2025}
}

@inproceedings{sella2025instancegen,
  title={Instancegen: Image generation with instance-level instructions},
  author={Sella, Etai and Kleiman, Yanir and Averbuch-Elor, Hadar},
  booktitle={Proceedings of the Special Interest Group on Computer Graphics and Interactive Techniques Conference Conference Papers},
  pages={1--10},
  year={2025}
}

@inproceedings{patashnik2023localizing,
  title={Localizing object-level shape variations with text-to-image diffusion models},
  author={Patashnik, Or and Garibi, Daniel and Azuri, Idan and Averbuch-Elor, Hadar and Cohen-Or, Daniel},
  booktitle={2023 IEEE/CVF International Conference on Computer Vision (ICCV)},
  pages={22994--23004},
  year={2023},
  organization={IEEE}
}

@inproceedings{alaluf2024cross,
  title={Cross-image attention for zero-shot appearance transfer},
  author={Alaluf, Yuval and Garibi, Daniel and Patashnik, Or and Averbuch-Elor, Hadar and Cohen-Or, Daniel},
  booktitle={ACM SIGGRAPH 2024 conference papers},
  pages={1--12},
  year={2024}
}

@inproceedings{cao2023masactrl,
  title={Masactrl: Tuning-free mutual self-attention control for consistent image synthesis and editing},
  author={Cao, Mingdeng and Wang, Xintao and Qi, Zhongang and Shan, Ying and Qie, Xiaohu and Zheng, Yinqiang},
  booktitle={2023 IEEE/CVF International Conference on Computer Vision (ICCV)},
  pages={22503--22513},
  year={2023},
  organization={IEEE}
}

@article{tang2023emergent,
  title={Emergent correspondence from image diffusion},
  author={Tang, Luming and Jia, Menglin and Wang, Qianqian and Phoo, Cheng Perng and Hariharan, Bharath},
  journal={Advances in neural information processing systems},
  volume={36},
  pages={1363--1389},
  year={2023}
}

@inproceedings{luo2024readout,
  title={Readout guidance: Learning control from diffusion features},
  author={Luo, Grace and Darrell, Trevor and Wang, Oliver and Goldman, Dan B and Holynski, Aleksander},
  booktitle={2024 IEEE/CVF Conference on Computer Vision and Pattern Recognition (CVPR)},
  pages={8217--8227},
  year={2024},
  organization={IEEE}
}

@inproceedings{rassin2022dalle,
  title={DALLE-2 is seeing double: Flaws in word-to-concept mapping in Text2Image models},
  author={Rassin, Royi and Ravfogel, Shauli and Goldberg, Yoav},
  booktitle={Proceedings of the Fifth BlackboxNLP workshop on analyzing and interpreting neural networks for NLP},
  pages={335--345},
  year={2022}
}

@article{toker2024diffusion,
  title={Diffusion lens: Interpreting text encoders in text-to-image pipelines},
  author={Toker, Michael and Orgad, Hadas and Ventura, Mor and Arad, Dana and Belinkov, Yonatan},
  journal={arXiv preprint arXiv:2403.05846},
  year={2024}
}

@inproceedings{kaplan2026follow,
  title={Follow the Flow: On Information Flow Across Textual Tokens in Text-to-Image Models},
  author={Kaplan, Guy and Toker, Michael and Reif, Yuval and Belinkov, Yonatan and Schwartz, Roy},
  booktitle={Proceedings of the 64th Annual Meeting of the Association for Computational Linguistics (Volume 1: Long Papers)},
  pages={34139--34157},
  year={2026}
}

@article{zhuang2024magnet,
  title={Magnet: We never know how text-to-image diffusion models work, until we learn how vision-language models function},
  author={Zhuang, Chenyi and Hu, Ying and Gao, Pan},
  journal={Advances in Neural Information Processing Systems},
  volume={37},
  pages={57115--57149},
  year={2024}
}

@article{Liu2023VisualIT,
  title={Visual Instruction Tuning},
  author={Haotian Liu and Chunyuan Li and Qingyang Wu and Yong Jae Lee},
  journal={ArXiv},
  year={2023},
  volume={abs/2304.08485},
}

@article{wang2026repa,
  title={Repa works until it doesn’t: Early-stopped, holistic alignment supercharges diffusion training},
  author={Wang, Ziqiao and Zhao, Wangbo and Zhou, Yuhao and Li, Zekai and Liang, Zhiyuan and Shi, Mingjia and Zhao, Xuanlei and Zhou, Pengfei and Zhang, Kaipeng and Wang, Zhangyang and others},
  journal={Advances in Neural Information Processing Systems},
  volume={38},
  pages={136854--136887},
  year={2026}
}

@inproceedings{zhai2023sigmoid,
  title={Sigmoid loss for language image pre-training},
  author={Zhai, Xiaohua and Mustafa, Basil and Kolesnikov, Alexander and Beyer, Lucas},
  booktitle={2023 IEEE/CVF International Conference on Computer Vision (ICCV)},
  pages={11941--11952},
  year={2023},
  organization={IEEE}
}

@inproceedings{lee2024prometheus,
  title={Prometheus-vision: Vision-language model as a judge for fine-grained evaluation},
  author={Lee, Seongyun and Kim, Seungone and Park, Sue and Kim, Geewook and Seo, Minjoon},
  booktitle={Findings of the Association for Computational Linguistics: ACL 2024},
  pages={11286--11315},
  year={2024}
}

@inproceedings{ma2025hpsv3,
  title={Hpsv3: Towards wide-spectrum human preference score},
  author={Ma, Yuhang and Wu, Xiaoshi and Sun, Keqiang and Li, Hongsheng},
  booktitle={2025 IEEE/CVF International Conference on Computer Vision (ICCV)},
  pages={15086--15095},
  year={2025},
  organization={IEEE}
}

@inproceedings{yehezkel2025navigating,
  title={Navigating with annealing guidance scale in diffusion space},
  author={Yehezkel, Shai and Dahary, Omer and Voynov, Andrey and Cohen-Or, Daniel},
  booktitle={Proceedings of the SIGGRAPH Asia 2025 Conference Papers},
  pages={1--11},
  year={2025}
}

@article{wang2026towards,
  title={Towards a golden classifier-free guidance path via foresight fixed point iterations},
  author={Wang, Kaibo and Mao, Jianda and Wu, Tong and Xiang, Yang},
  journal={Advances in Neural Information Processing Systems},
  volume={38},
  pages={124994--125030},
  year={2026}
}

@inproceedings{lin2014microsoft,
  title={Microsoft coco: Common objects in context},
  author={Lin, Tsung-Yi and Maire, Michael and Belongie, Serge and Hays, James and Perona, Pietro and Ramanan, Deva and Doll{\'a}r, Piotr and Zitnick, C Lawrence},
  booktitle={European conference on computer vision},
  pages={740--755},
  year={2014},
  organization={Springer}
}

@article{ma2026fine,
  title={Fine-t2i: An open, large-scale, and diverse dataset for high-quality t2i fine-tuning},
  author={Ma, Xu and Zhang, Yitian and Dong, Qihua and Fu, Yun},
  journal={arXiv preprint arXiv:2602.09439},
  year={2026}
}

@misc{qwen2025qwen25technicalreport,
      title={Qwen2.5 Technical Report}, 
      author={Qwen and : and An Yang and Baosong Yang and Beichen Zhang and Binyuan Hui and Bo Zheng and Bowen Yu and Chengyuan Li and Dayiheng Liu and Fei Huang and Haoran Wei and Huan Lin and Jian Yang and Jianhong Tu and Jianwei Zhang and Jianxin Yang and Jiaxi Yang and Jingren Zhou and Junyang Lin and Kai Dang and Keming Lu and Keqin Bao and Kexin Yang and Le Yu and Mei Li and Mingfeng Xue and Pei Zhang and Qin Zhu and Rui Men and Runji Lin and Tianhao Li and Tianyi Tang and Tingyu Xia and Xingzhang Ren and Xuancheng Ren and Yang Fan and Yang Su and Yichang Zhang and Yu Wan and Yuqiong Liu and Zeyu Cui and Zhenru Zhang and Zihan Qiu},
      year={2025},
      eprint={2412.15115},
      archivePrefix={arXiv},
      primaryClass={cs.CL},
      url={https://arxiv.org/abs/2412.15115}, 
}

@article{zhu2025internvl3,
  title={Internvl3: Exploring advanced training and test-time recipes for open-source multimodal models},
  author={Zhu, Jinguo and Wang, Weiyun and Chen, Zhe and Liu, Zhaoyang and Ye, Shenglong and Gu, Lixin and Tian, Hao and Duan, Yuchen and Su, Weijie and Shao, Jie and others},
  journal={arXiv preprint arXiv:2504.10479},
  year={2025}
}

@article{heusel2017gans,
  title={Gans trained by a two time-scale update rule converge to a local nash equilibrium},
  author={Heusel, Martin and Ramsauer, Hubert and Unterthiner, Thomas and Nessler, Bernhard and Hochreiter, Sepp},
  journal={Advances in neural information processing systems},
  volume={30},
  year={2017}
}

@article{binkowski2018demystifying,
  title={Demystifying mmd gans},
  author={Bi{\'n}kowski, Miko{\l}aj and Sutherland, Danica J and Arbel, Michael and Gretton, Arthur},
  journal={arXiv preprint arXiv:1801.01401},
  year={2018}
}

@inproceedings{radford2021learning,
  title={Learning transferable visual models from natural language supervision},
  author={Radford, Alec and Kim, Jong Wook and Hallacy, Chris and Ramesh, Aditya and Goh, Gabriel and Agarwal, Sandhini and Sastry, Girish and Askell, Amanda and Mishkin, Pamela and Clark, Jack and others},
  booktitle={International conference on machine learning},
  pages={8748--8763},
  year={2021},
  organization={PMLR}
}

@inproceedings{lin2024evaluating,
  title={Evaluating text-to-visual generation with image-to-text generation},
  author={Lin, Zhiqiu and Pathak, Deepak and Li, Baiqi and Li, Jiayao and Xia, Xide and Neubig, Graham and Zhang, Pengchuan and Ramanan, Deva},
  booktitle={European Conference on Computer Vision},
  pages={366--384},
  year={2024},
  organization={Springer}
}

@article{wu2023better,
  title={Better aligning text-to-image models with human preference},
  author={Wu, Xiaoshi and Sun, Keqiang and Zhu, Feng and Zhao, Rui and Li, Hongsheng},
  journal={arXiv preprint arXiv:2303.14420},
  volume={1},
  number={3},
  year={2023}
}

@article{oquab2023dinov2,
  title={Dinov2: Learning robust visual features without supervision},
  author={Oquab, Maxime and Darcet, Timoth{\'e}e and Moutakanni, Th{\'e}o and Vo, Huy and Szafraniec, Marc and Khalidov, Vasil and Fernandez, Pierre and Haziza, Daniel and Massa, Francisco and El-Nouby, Alaaeldin and others},
  journal={arXiv preprint arXiv:2304.07193},
  year={2023}
}

@article{kynkaanniemi2019improved,
  title={Improved precision and recall metric for assessing generative models},
  author={Kynk{\"a}{\"a}nniemi, Tuomas and Karras, Tero and Laine, Samuli and Lehtinen, Jaakko and Aila, Timo},
  journal={Advances in neural information processing systems},
  volume={32},
  year={2019}
}
\bibliographystyle{iclr2027_conference}

\appendix
\section*{Appendix}

\section{Implementation Details}
\label{sec:reader_implementation}
\subsection{Contextual Reader}
\subsubsection{Contextual Feature Extraction and Encoding}
\label{subsec:reader_features}

We extract contextual representations from the attention operator of the frozen MM-DiT at the selected diffusion timestep. Specifically, we extract the output of the attention operator before the subsequent block-specific attention gate, residual addition, normalization, or MLP update. We retain only the contextual/text positions and exclude image-token positions.

For SD3.5, we capture the text-side output of joint attention from all transformer blocks. For FLUX.2, we capture the text-side output from all dual-stream blocks and the text positions from all single-stream blocks.

For the conditional reader, we use the contextual representations produced by the positive-prompt branch. For the unconditional reader, we use the model's empty-prompt setting. No image tokens are supplied to the reader.

Because the reader aggregates contextual representations from multiple MM-DiT layers, we add a learned layer embedding to the contextual tokens from each layer before concatenating them. We also provide the denoising timestep to the reader. The continuous noise level $\sigma$ is encoded using a 256-dimensional sinusoidal embedding followed by an MLP with dimensions $256\rightarrow256\rightarrow D$ and SiLU activation, where $D$ is the hidden dimension of the corresponding diffusion backbone. The resulting timestep embedding is added to every contextual token. We do not average or otherwise mix the representations across layers before the Q-Former; all layer-token pairs remain independently accessible.

\subsubsection{Bottleneck Network and Language-Model Interface}
\label{subsec:reader_interface}

The Bottleneck Network is a two-block, eight-head Q-Former with 16 learned queries. Its hidden dimension matches that of the corresponding diffusion backbone. Each Q-Former block uses a cross-first ordering. The learned queries first attend to all contextual tokens through pre-normalized cross-attention, followed by a residual connection. This is followed by pre-normalized self-attention among the queries and a residual connection, and then a pre-normalized feed-forward network with dimensions $D\rightarrow2048\rightarrow D$, GELU activation, and a residual connection. A final layer normalization is applied after the second block. No dropout is used.

The Q-Former outputs are mapped individually into the input embedding space of the frozen Qwen2.5-7B-Instruct~\citep{qwen2025qwen25technicalreport} language model using a learned affine projection. Each projected token is L2-normalized and rescaled to the mean norm of the LLM's vocabulary embeddings, excluding the padding token.

The resulting Contextual Descriptor is provided to the LLM as a sequence of 16 continuous prefix embeddings, followed by the textual question and its answer-format instruction. The original generation prompt is not provided to the LLM.

During teacher-forced training, the target consists of the ground-truth answer and the tokenizer EOS token. The cross-entropy loss is computed only over answer tokens.

\subsubsection{Dataset Construction}
\label{subsec:reader_dataset}

We construct the dataset for training and evaluating the Contextual Reader from the 2014 MS-COCO training split. We first restrict the source images to those containing exactly one non-crowd person instance according to the COCO instance annotations. We then filter the associated captions based on their textual content: each caption is classified by the text-only model Qwen2.5-7B-Instruct~\citep{qwen2025qwen25technicalreport} as describing either zero, one, multiple, or an ambiguous number of real people. We retain captions classified as describing exactly one real person without ambiguity. This procedure produces 62,069 eligible captions.

From these captions, we select 2,500 prompts from distinct COCO source images using deterministic stratified sampling. We first select 500 prompts for the held-out set and then select 2,000 prompts from the remaining images, ensuring no source-image overlap between the two sets. Each prompt is rendered with five randomly assigned seeds, such that no seed is shared between prompts. This results in 10,000 training generations and 2,500 held-out generations per diffusion backbone.

We generate images at $1024\times1024$ resolution. SD3.5 uses 28 FlowMatch Euler denoising steps and guidance scale 7.0, while FLUX.2 uses 50 FlowMatch Euler steps and guidance scale 4.0. The same prompt-to-seed mapping is used for both backbones.

We manually curate 15 visual questions designed to span diverse semantic aspects of human-centric images, including appearance, pose, activity, spatial configuration, and scene properties. The complete set of questions is shown in Figure~\ref{fig:questions}.

\begin{figure*}[t]
    \centering

    \begin{tcolorbox}[
        enhanced,
        width=0.96\textwidth,
        colback=blue!3,
        colframe=blue!45!black,
        boxrule=0.7pt,
        arc=0pt,
        left=10pt,
        right=10pt,
        top=8pt,
        bottom=6pt,
        borderline west={2pt}{0pt}{blue!65!black},
    ]

    \begin{minipage}{0.47\textwidth}
        \begin{enumerate}[
            leftmargin=*,
            labelsep=5pt,
            itemsep=2pt,
            topsep=0pt,
            parsep=0pt
        ]
            \item What is the person's apparent gender presentation?
            \item What age group does the person appear to belong to?
            \item What is the person wearing?
            \item What is the person's hair like?
            \item Where is the person located in the image?
            \item What is the person's detailed pose?
            \item Which direction is the person facing?
            \item What is the person doing?
        \end{enumerate}
    \end{minipage}
    \hfill
    \begin{minipage}{0.47\textwidth}
        \begin{enumerate}[
            leftmargin=*,
            labelsep=5pt,
            itemsep=2pt,
            topsep=0pt,
            parsep=0pt
        ]
            \setcounter{enumi}{8}
            \item What is in the background?
            \item What is the camera angle?
            \item What object is the person holding or using?
            \item Which parts of the person are visible in the image?
            \item How large is the person relative to the full image?
            \item What is the overall color scheme of the image?
            \item What objects are visible in the image?
        \end{enumerate}
    \end{minipage}

    \end{tcolorbox}

    \caption{\textbf{Visual questions used by the Contextual Reader.}
    We manually curate 15 questions spanning diverse semantic aspects of human-centric images.}
    \label{fig:questions}
\end{figure*}

For each generated image, we obtain an answer to each question using InternVL3-8B~\citep{zhu2025internvl3}. The teacher receives only the generated image and the visual question, without access to the original generation prompt, and is instructed to answer using visible image evidence. Answers are generated independently using deterministic decoding. We automatically validate the generated responses and retry invalid answers up to two times, replacing answers that remain invalid with ``unknown.''

\subsubsection{Reader Training}
\label{subsec:reader_training_details}

We train a separate reader for each diffusion backbone and conditioning setting. The diffusion backbone, text encoders, VAE, and LLM remain frozen throughout training. Only the Q-Former and the projection layer are optimized.

At each training iteration, we sample a continuous noise level
$\sigma \sim U[0,1]$,
allowing the reader to observe contextual representations throughout the denoising trajectory. For each generated image, we store its final latent $x_0$ and corresponding initial noise $\epsilon$, and construct the noisy latent as
$
z_t = (1-\sigma)x_0 + \sigma\epsilon.
$
We then run the frozen diffusion model on this latent and extract the contextual representations at the sampled noise level. Thus, contextual representations are extracted on the fly rather than stored for every timestep.

We train for 5,000 optimizer updates using AdamW with learning rate $10^{-4}$, weight decay $10^{-4}$, and gradient-norm clipping at 1.0. No learning-rate scheduler or warmup is used. Training uses bfloat16 mixed precision with an effective global batch size of 256. We use 8 NVIDIA A100 GPUs for training. Training takes approximately 4 hours for the SD3.5 reader and 8 hours for the FLUX.2 reader.

The resulting contextual representations are passed through the Bottleneck Network and LLM, and the reader is optimized using the autoregressive cross-entropy objective.

\begin{figure*}[!t]
    \centering
    \includegraphics[width=0.98\textwidth]{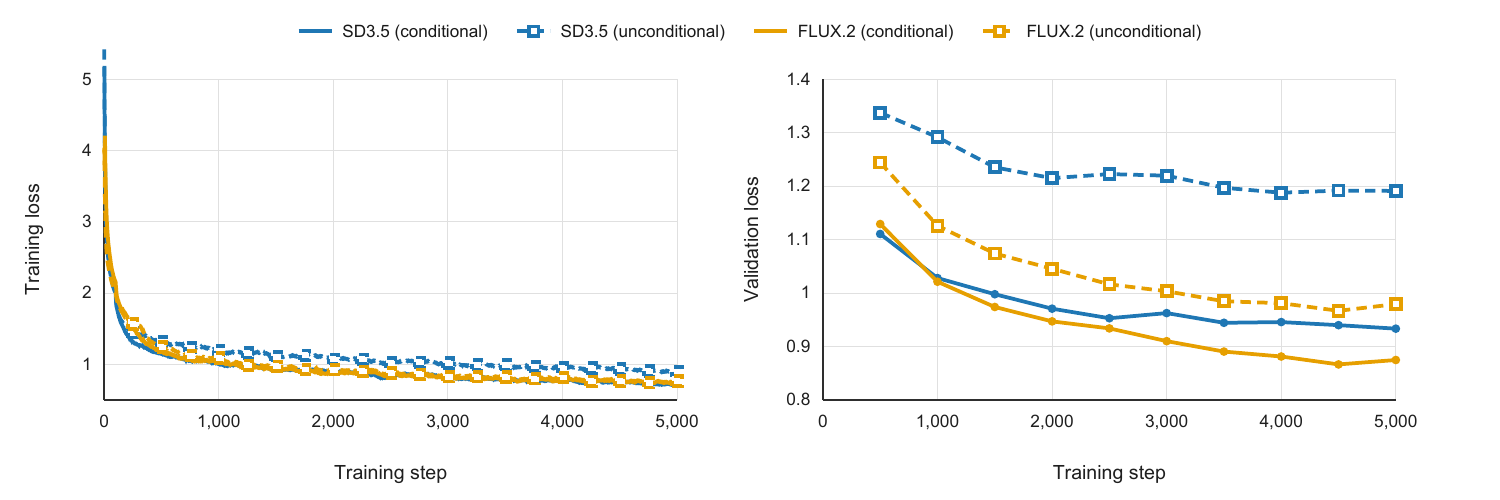}
    \caption{\textbf{Reader training and validation loss.}
    Training and validation loss over 5,000 optimization steps for the conditional and unconditional readers across both diffusion backbones. The validation set consists of $5\%$ of the training data and is separate from the held-out data used for reader evaluation.}
    \label{fig:reader_training_curves}
\end{figure*}

We additionally monitor training and validation loss during reader training, as shown in Figure~\ref{fig:reader_training_curves}. The validation loss decreases steeply until approximately 2,000 training steps, with further training yielding only modest additional reductions. The validation loss is computed on $5\%$ held-out prompts from the training data, separate from the data used for reader evaluation.

\subsubsection{VLM-as-a-Judge Details}
\label{subsec:vlm_judge}

We evaluate the reader's predictions using a VLM-as-a-judge. We use Prometheus-Vision 13B~\citep{lee2024prometheus} to compare each reader prediction with the corresponding reference answer produced by InternVL3-8B. The judge receives the visual question, the reader's prediction, the reference answer, and the final generated image. The generation prompt is not provided to the judge. The exact evaluation prompt and scoring rubric are shown in Figure~\ref{fig:vlm_judge}.

The judge assigns an integer score from 1 to 5 based on semantic correctness, completeness, and consistency with the image. A score of 5 indicates a semantically equivalent, fully correct, and image-consistent answer, while lower scores reflect increasing degrees of incorrectness, omission, ambiguity, or contradiction. Empty predictions receive a score of 1, and all other predictions are evaluated by Prometheus-Vision. The resulting scores are normalized to $[0,1]$.

For each denoising timestep, we compute the reader's readability as the mean normalized score across all validation images and questions.

\subsection{Contextual Alignment}
\subsubsection{Feature Extraction and Aligner Network Architecture}

\label{subsec:coal_architecture}
Across architectures, contextual features are extracted in the same manner as in the Contextual Reader (Appendix~\ref{subsec:reader_features}): tokens from the contextual indices are pooled immediately after the shared attention block. Unlike the Reader, however, in \methodname{} we extract only the contextual features from a single MM-DiT layer $l$ ($l=4$ by default) and no positional embedding is added to it.

The Aligner network is made up of a Q-Former followed by a small MLP. The Q-Former is $24$ head single-layer, single-query Q-Former which attends to the contextual features via pre-normalized cross-attention between the learnable query and the features. The MLP is made up of two linear layers and SiLU activation such that the dimension of the vector produced by the Q-Former is $D\rightarrow1024\rightarrow d_y$. With $D$ being the hidden dimension of the contextual tokens ($768$ in the MM-DiT used in the full model training setting, $1536$ in the SD3-Medium model used in the fine-tune setting). As explained in the main paper, the Aligner network, along with the MM-DiT, are optimized via a negative cosine similarity loss between the Aligner Network output and the teacher embedding.

\subsubsection{Null-Condition Weighting}
Null-Condition Weighting involves upweighting the \methodname{} loss for training samples in which the text condition has been dropped. In practice, this mechanism is activated after a set amount of training iterations $i_{start}$ has passed. After $i_{start}$ the \methodname{} loss is multiplied by a factor $w_{null} > 1.0$ for samples with dropped text conditioning and multiplied by $w_{text} < 1.0$ for text guided samples. These factors are applied on top of the overall \methodname{} loss weight $w_{CoAl}$. For full model training we use $w_{CoAl}=1.3$, $i_{start}=125{,}000$, $w_{null}=2.0$, and $w_{text}=1.0$; for fine-tuning we use $w_{CoAl}=0.75$, $i_{start}=0$, $w_{null}=6.0$, and $w_{text}=0.25$.

\subsubsection{Teacher Embeddings}
Our default teacher encoder is SigLIP, specifically the \href{https://huggingface.co/google/siglip-base-patch16-256}{\texttt{google/siglip-base-patch16-256}} model, producing $768$-dimensional embeddings. In our ablation study (Appendix \ref{subsec:coal_ablations}), we additionally consider CLIP (\href{https://huggingface.co/openai/clip-vit-large-patch14}{\texttt{openai/clip-vit-large-patch14}}), DINOv2 (\href{https://github.com/facebookresearch/dinov2}{\texttt{dinov2\_vitb14}}, the same encoder standard REPA itself uses as its own teacher), and Qwen3-VL (\href{https://huggingface.co/Qwen/Qwen3-VL-Embedding-2B}{\texttt{Qwen/Qwen3-VL-Embedding-2B}}).

\subsubsection{Training Configurations}
\paragraph{Full Model Training} In the full model training setting the MM-DiT architecture we train is based on the Stable Diffusion 3 model~\citep{esser2024scaling}, but is trained entirely from scratch (no pretrained weights) at a reduced scale: $24$ transformer layers, matching SD3's own depth, with a hidden dimension of $768$ and $24$ attention heads (32-dimensional per head), compared to SD3-Medium's $1536$-dimensional hidden size at the same head count (64-dimensional per head). Training data is MS-COCO 2014~\citep{lin2014microsoft}, downloaded from \href{https://cocodataset.org/#download}{this source}, using the official \texttt{train2014}/\texttt{val2014} split ($82{,}783 / 40{,}504$ images, each with $\sim$5 captions), at $256\times256$ resolution. We use the AdamW optimizer with a constant learning rate of $1\times10^{-4}$, no weight decay, and $\epsilon=1\times10^{-8}$, with a global batch size of $256$. We maintain an exponential moving average (EMA) of the model weights with decay $0.9999$, initialized as an exact copy of the model at step 0, and report all results using the EMA weights. We train for $150{,}000$ iterations using \texttt{fp16} mixed precision. Using 8 NVIDIA B200 GPUs training takes approximately $12.5$ hours.

\paragraph{Fine-Tuning} In the fine-tuning setting we fine-tune the full Stable Diffusion 3 Medium model~\citep{esser2024scaling} (all weights trainable, no LoRA or partial freezing), with $24$ transformer layers, a hidden dimension of $1536$, and $24$ attention heads (64-dimensional per head). Training data is the curated subset of the Fine-T2I dataset~\citep{ma2026fine}, available at \href{https://huggingface.co/datasets/ma-xu/fine-t2i}{huggingface.co/datasets/ma-xu/fine-t2i}, using $162{,}979$ training images and a held-out set of $5{,}000$ validation images, at $1024\times1024$ resolution. We use the AdamW optimizer with a constant learning rate of $1\times10^{-5}$,  weight decay $1\times10^{-4}$, and $\epsilon=1\times10^{-8}$, with a global batch size of $32$. As in the full model training setting, we maintain an EMA of the weights with decay $0.9999$ and report results using the EMA weights. Text conditioning is dropped with probability $0.15$ during training (for classifier-free guidance). We train for $30{,}000$ iterations using \texttt{bf16} mixed precision. Using 8 NVIDIA B200 GPUs training takes approximately $5.6$ hours.

\subsubsection{Evaluation}
We report Fr\'echet Inception Distance (FID) and Kernel Inception Distance (KID) using the \texttt{pytorch-fid} package's InceptionV3 network with 2048-dimensional pool3 features; the standard configuration for both metrics. KID uses a degree-3 polynomial kernel averaged over $100$ random subsets of size $1000$. CLIP similarity between generated images and their prompts is computed with \texttt{openai/clip-vit-large-patch14}. DINO precision and recall follow the Improved Precision and Recall formulation, using \texttt{facebook/dinov2-base} features with $k=5$ nearest neighbors. VQAScore is computed with the \texttt{t2v\_metrics} package using \texttt{clip-flant5-xxl}. For Human Preference Score we use HPS v2.1, via the \texttt{hpsv2} package (\texttt{xswu/HPSv2} checkpoint, the latest official release, downloaded through \texttt{huggingface\_hub}).

In the full model training setting, FID/KID/CLIP similarity/DINO precision-recall and HPS are computed over $40{,}192$ generated samples against the full MS-COCO 2014 validation set, while VQAScore is computed over a fixed random subset of $4{,}096$ samples. In the fine-tuning setting, all metrics are computed over the full $5{,}000$-image held-out validation set.

\begin{figure*}[!t]
    \centering

    \begin{tcolorbox}[
        enhanced,
        width=0.96\textwidth,
        colback=blue!3,
        colframe=blue!45!black,
        boxrule=0.7pt,
        arc=0pt,
        left=10pt,
        right=10pt,
        top=8pt,
        bottom=8pt,
        borderline west={1.5pt}{0pt}{blue!65!black},
    ]

    \begin{minipage}[t]{0.46\textwidth}
        \textbf{Evaluation Prompt}

        \vspace{3pt}

        \small
        \textbf{Instruction to evaluate:}\\
        \texttt{<question>}

        \vspace{4pt}

        \textbf{Response to evaluate:}\\
        \texttt{<reader prediction>}

        \vspace{4pt}

        \textbf{Reference Answer (Score 5):}\\
        \texttt{<ground-truth VLM answer>}

        \vspace{4pt}

        \textbf{Output format:}\\
        \texttt{Feedback: (feedback) [RESULT] (1--5)}

    \end{minipage}
    \hfill
    \begin{minipage}[t]{0.49\textwidth}
        \textbf{Scoring Rubric}

        \vspace{3pt}

        \small
        \begin{enumerate}[
            leftmargin=*,
            labelsep=5pt,
            itemsep=2pt,
            topsep=0pt,
            parsep=0pt
        ]
            \item The answer is empty, unrelated, or contradicts the reference answer or image on the requested attribute.
            \item The answer contains some relevant content, but an essential fact is wrong or missing.
            \item The answer is partially correct but has a material ambiguity, omission, or unsupported addition.
            \item All core facts are correct and image-supported; only a nonessential detail is omitted or added.
            \item The answer is semantically equivalent to the reference answer, fully correct, and image-consistent.
        \end{enumerate}
    \end{minipage}

    \end{tcolorbox}

    \caption{\textbf{VLM-as-a-judge prompt and scoring rubric.}
    Prometheus-Vision receives the visual question, the reader's prediction, the reference answer, and the generated image, and assigns a score from 1 to 5 according to the rubric.}
    \label{fig:vlm_judge}
\end{figure*}

\section{Additional Results and Experiments}
\label{sec:additional_results}

\subsection{Contextual Reader}

\subsubsection{Comparing Semantic Readability Across Representations}
\label{subsec:reader_comparison}

We compare the semantic readability of the contextual space with that of the latent and the prompt. We follow the evaluation protocol described in Appendix~\ref{subsec:vlm_judge} and consider two complementary questions:
whether contextual tokens provide semantic information beyond what is recoverable from the latent at early denoising stages,
and whether they encode information beyond what can be inferred from the textual conditioning alone.

\paragraph{Semantics Emerge Early in Contextual Tokens.}

\begin{figure}[t]
    \centering
    \includegraphics[width=\linewidth]{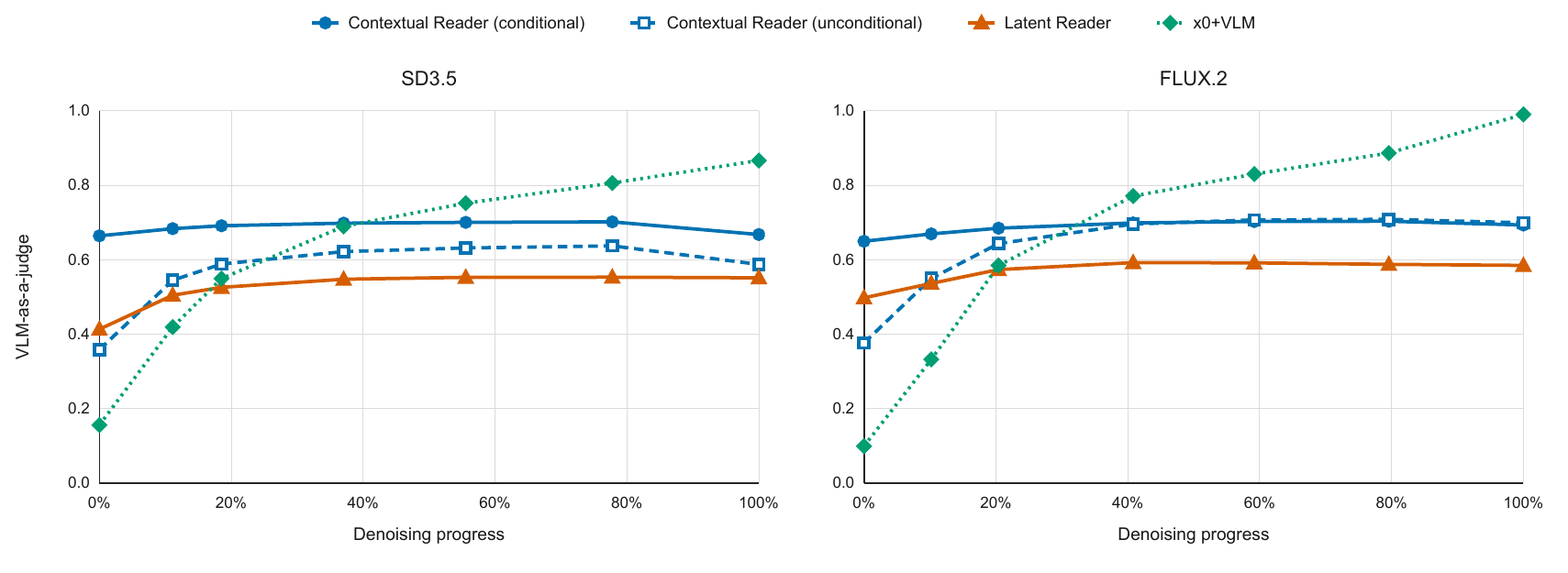}
    \caption{\textbf{Comparing semantic readability with the predicted image.}
    We compare the Contextual Reader with a Latent Reader operating on the output latent representation and an $x_0$+VLM baseline that answers questions directly from the predicted clean image. Results are shown for both conditional and unconditional readers across two diffusion backbones. The contextual representations provide higher readability during the early stages of denoising, while $x_0$+VLM achieves the highest readability from approximately $40\%$ onward.}
    \label{fig:reader_comparison}
\end{figure}

We first compare the contextual space with two alternative sources of information. First, we train a Latent Reader using the same LLM and reader architecture as the Contextual Reader, but provide it with the output latent representation at the corresponding denoising timestep instead of contextual tokens. Second, we use InternVL3-8B to answer the visual question from the predicted clean image at each timestep; we refer to this baseline as $x_0$+VLM. We note that the ground-truth answers used for evaluation are also generated by InternVL3-8B from the final images, making $x_0$+VLM more closely aligned with the evaluation target.

Figure~\ref{fig:reader_comparison} shows the resulting readability scores for both diffusion backbones. The conditional Contextual Reader consistently outperforms the Latent Reader throughout denoising, while the unconditional Contextual Reader outperforms the Latent Reader from $10\%$ onward despite having no access to the textual conditioning. In comparison, $x_0$+VLM achieves the highest readability from approximately $40\%$ onward. Remarkably, before this point, both the conditional and unconditional Contextual Readers achieve higher readability despite having no direct access to the image. These results show that contextual tokens provide a particularly informative representation of semantic information during the early stages of denoising.

\paragraph{Contextual Tokens Encode Generation-Specific Semantics.}

\begin{figure}[t]
    \centering
    \includegraphics[width=\linewidth]{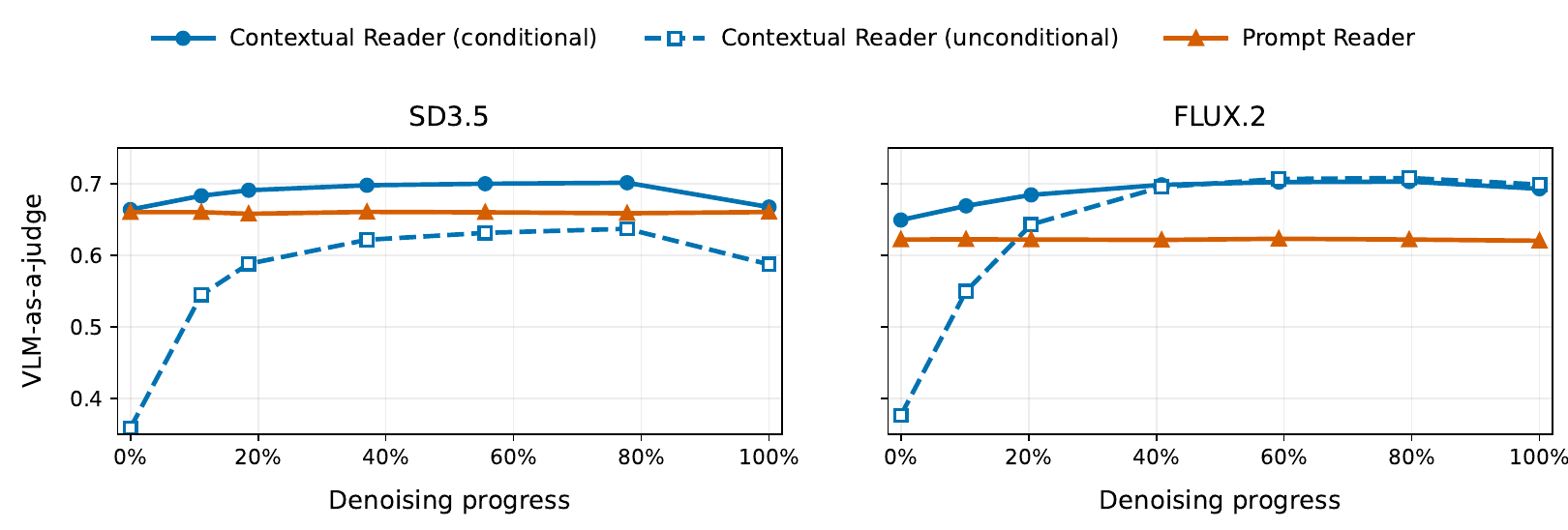}
    \caption{\textbf{Comparing semantic readability with the prompt.}
    We compare the Contextual Reader with a Prompt Reader using the same architecture but receiving only the input prompt. Results are shown for both conditional and unconditional readers across two diffusion backbones. The conditional Contextual Reader achieves higher readability than the Prompt Reader throughout denoising, while the unconditional Contextual Reader surpasses it for FLUX.2 and approaches it for SD3.5 as denoising progresses.}
    \label{fig:prompt_reader}
\end{figure}

We next examine how much of the semantic information decoded by the Contextual Reader can be attributed to the information available from the textual conditioning and the LLM's prior alone. To this end, we train a Prompt Reader using the same LLM and reader architecture as the Contextual Reader, but provide it only with the input prompt rather than contextual tokens. The Prompt Reader is trained with the same visual questions and target answers as the Contextual Reader.

Figure~\ref{fig:prompt_reader} compares the resulting readability scores for both diffusion backbones and conditioning settings. The Prompt Reader achieves constant readability throughout denoising, as expected since its input prompt remains unchanged. In contrast, the conditional Contextual Reader achieves higher readability than the Prompt Reader throughout denoising, with the largest gap between $20\%$ and $80\%$. The unconditional Contextual Reader, which receives no textual conditioning, surpasses the Prompt Reader as denoising progresses for FLUX.2 and approaches it for SD3.5. These results show that contextual tokens encode semantic information beyond what can be inferred from the textual conditioning and the LLM's prior.

\subsubsection{Generalization to Unseen Questions}
\label{subsec:unseen_questions}

\begin{figure}[t]
    \centering
    \includegraphics[width=\linewidth]{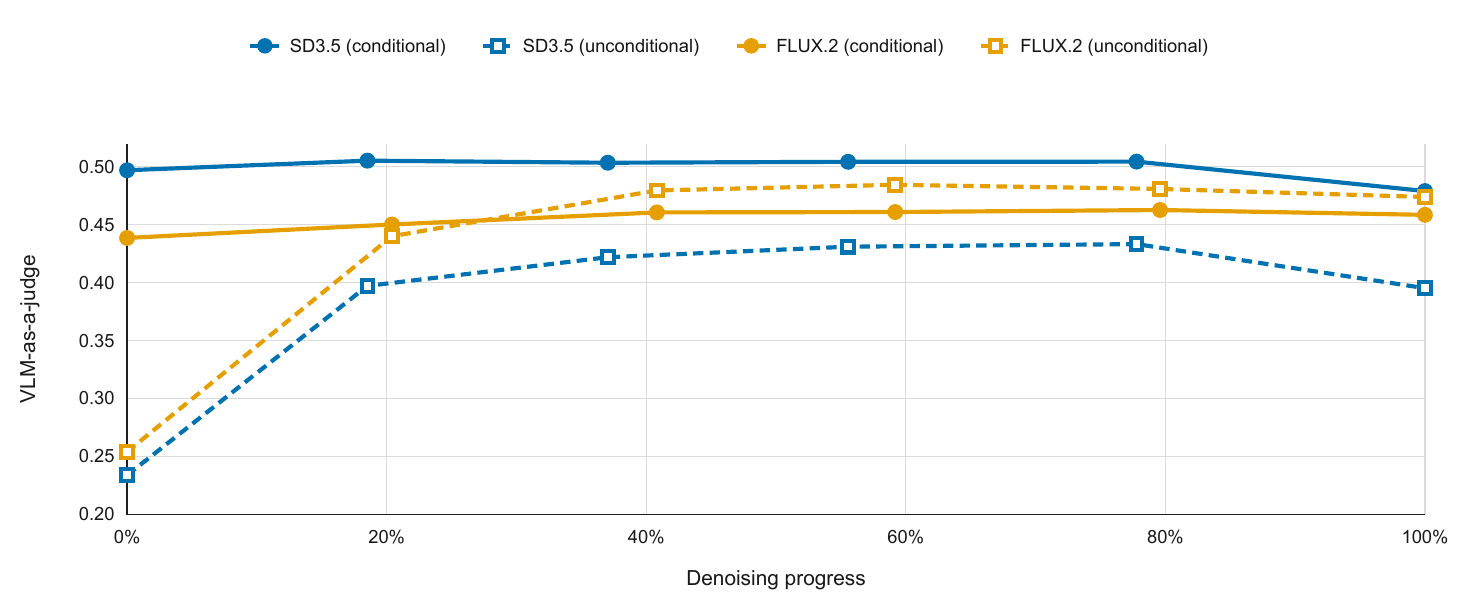}
    \caption{\textbf{Generalization to unseen questions.}
    For each question, we train a separate reader on the remaining 14 questions and evaluate it on the held-out question. We report the average semantic readability across all 15 held-out evaluations. Despite the reduced supervision, the reader maintains meaningful readability across denoising timesteps for both conditioning settings and diffusion backbones.}
    \label{fig:unseen_questions}
\end{figure}

To evaluate whether the reader generalizes beyond the questions used during training, we perform a held-out-question evaluation. For each of the 15 questions, we train a separate reader on the remaining 14 questions and evaluate it exclusively on the held-out question. We repeat this procedure for all questions and report the average readability across the 15 held-out evaluations.

Figure~\ref{fig:unseen_questions} shows the resulting readability scores for both diffusion backbones and conditioning settings. Compared to the evaluation on questions seen during training, readability decreases, as expected, but remains meaningful. The conditional readers reach scores of approximately $0.50$, corresponding to responses that are typically judged as partially correct, with some omissions. We observe the same trend for the unconditional readers as in our earlier experiments: their readability approaches that of the conditional readers as denoising progresses. SD3.5 reaches approximately $0.40$ (generally relevant but with some omissions), while FLUX.2 eventually surpasses the conditional reader later in denoising. These results show that the reader retains the ability to decode semantic information even for questions that were never observed during training.

\subsubsection{Additional Qualitative Results}
\label{subsec:additional_qual}

\paragraph{Final images corresponding to Figure~\ref{fig:teaser}.}

Figure~\ref{fig:teaser_final} shows the final images corresponding to the generations presented in Figure~\ref{fig:teaser}. While the early visual predictions are ambiguous, the corresponding denoising trajectories eventually resolve into the activities identified by the Contextual Reader.

\begin{center}
    \begin{minipage}[t]{0.158\linewidth}
        \centering
        \includegraphics[width=\linewidth]{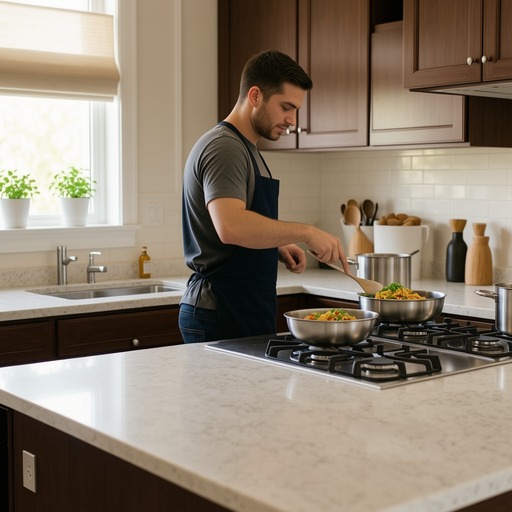}\\[2pt]
        \footnotesize cooking
    \end{minipage}
    \hfill
    \begin{minipage}[t]{0.158\linewidth}
        \centering
        \includegraphics[width=\linewidth]{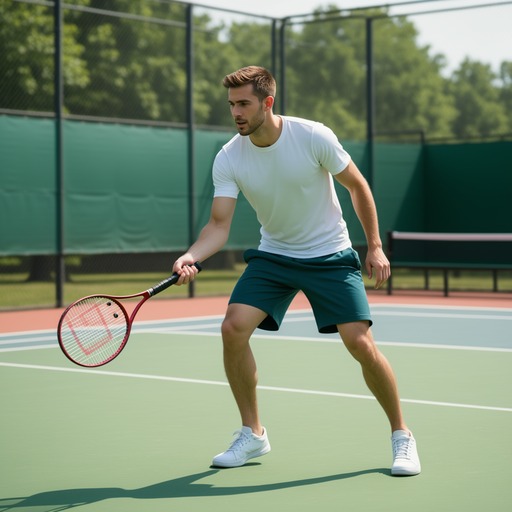}\\[2pt]
        \footnotesize playing tennis
    \end{minipage}
    \hfill
    \begin{minipage}[t]{0.158\linewidth}
        \centering
        \includegraphics[width=\linewidth]{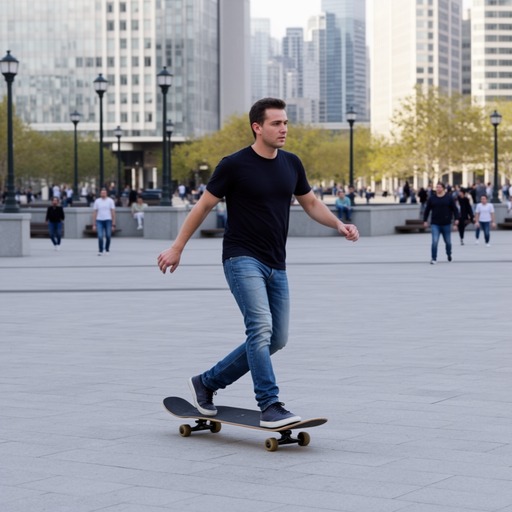}\\[2pt]
        \footnotesize skateboarding
    \end{minipage}
    \hfill
    \begin{minipage}[t]{0.158\linewidth}
        \centering
        \includegraphics[width=\linewidth]{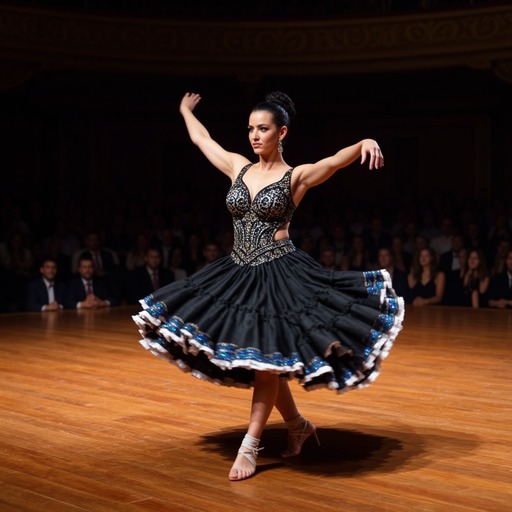}\\[2pt]
        \footnotesize dancing
    \end{minipage}
    \hfill
    \begin{minipage}[t]{0.158\linewidth}
        \centering
        \includegraphics[width=\linewidth]{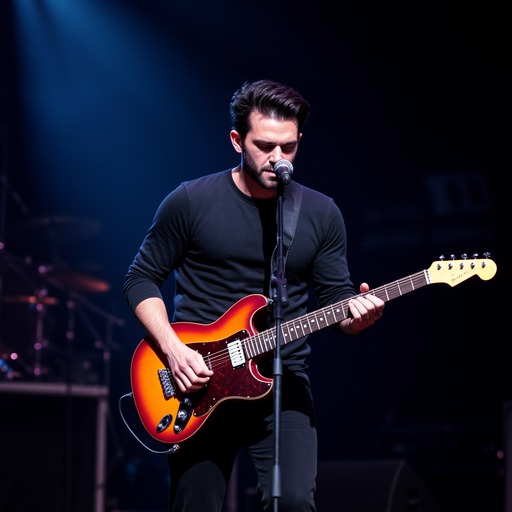}\\[2pt]
        \footnotesize playing guitar
    \end{minipage}
    \hfill
    \begin{minipage}[t]{0.158\linewidth}
        \centering
        \includegraphics[width=\linewidth]{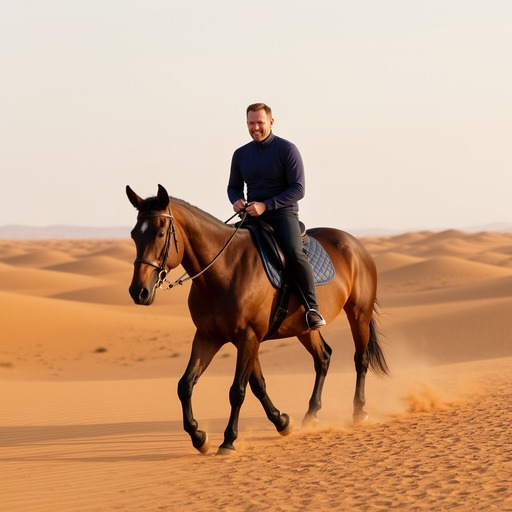}\\[2pt]
        \footnotesize riding a horse
    \end{minipage}

    \captionof{figure}{
    \textbf{Final images corresponding to the early visual predictions in Figure~\ref{fig:teaser}.}
    }
    \label{fig:teaser_final}
\end{center}

\begin{figure*}[t]
    \centering
    \footnotesize

    \setlength{\fboxrule}{1pt}
    \setlength{\fboxsep}{0pt}

    \begin{minipage}[t]{0.485\textwidth}
        \centering

        {\small\textbf{Prompt:} ``A youth counselor in the studio.''}

        \vspace{3pt}

        \newlength{\mirrorwidth}
        \setlength{\mirrorwidth}{0.235\linewidth}

        \begin{tabular}{@{}c@{\hspace{2pt}}c@{\hspace{2pt}}c@{\hspace{2pt}}c@{}}
            {\scriptsize\textbf{8\%}} &
            {\scriptsize\textbf{100\%}} &
            {\scriptsize\textbf{8\%}} &
            {\scriptsize\textbf{100\%}} \\[2pt]
            \fcolorbox{black}{white}{%
                \includegraphics[width=0.23\linewidth]{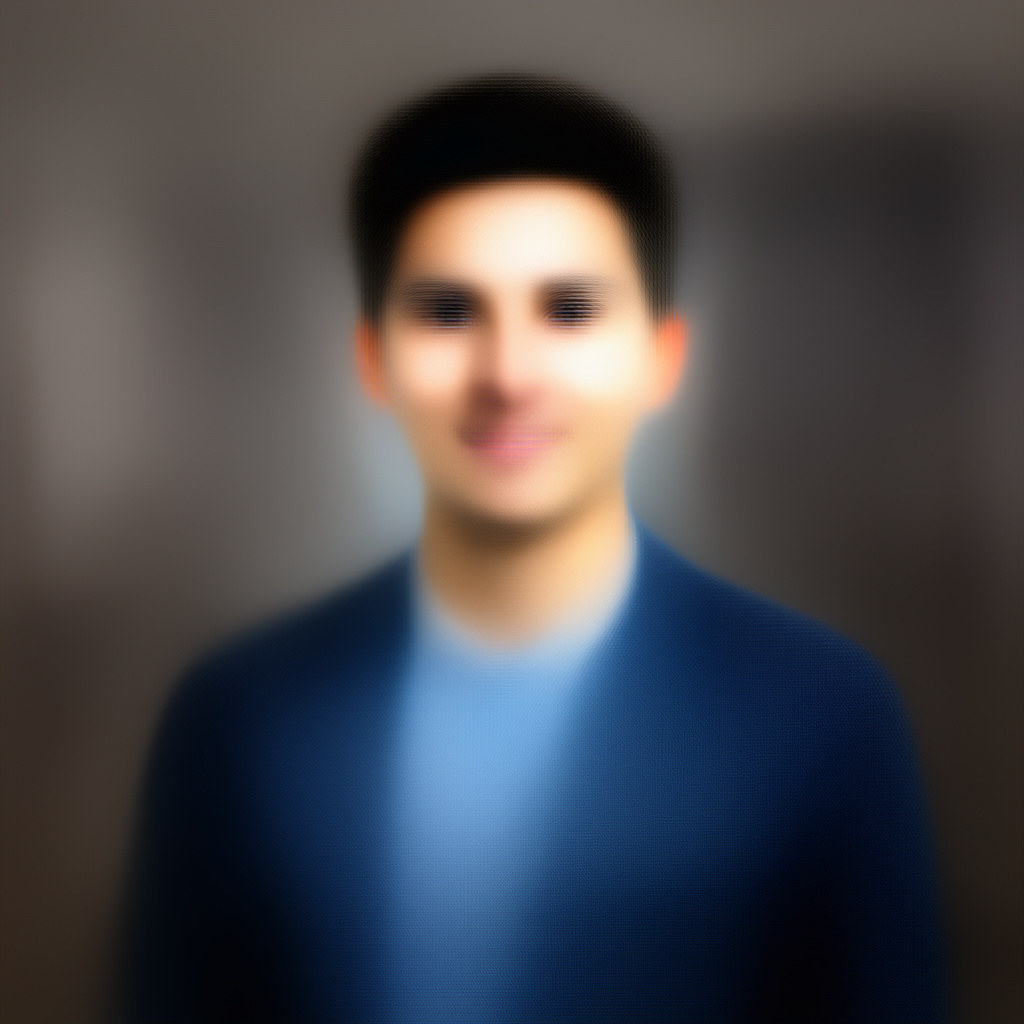}} &
            \fcolorbox{black}{white}{%
                \includegraphics[width=0.23\linewidth]{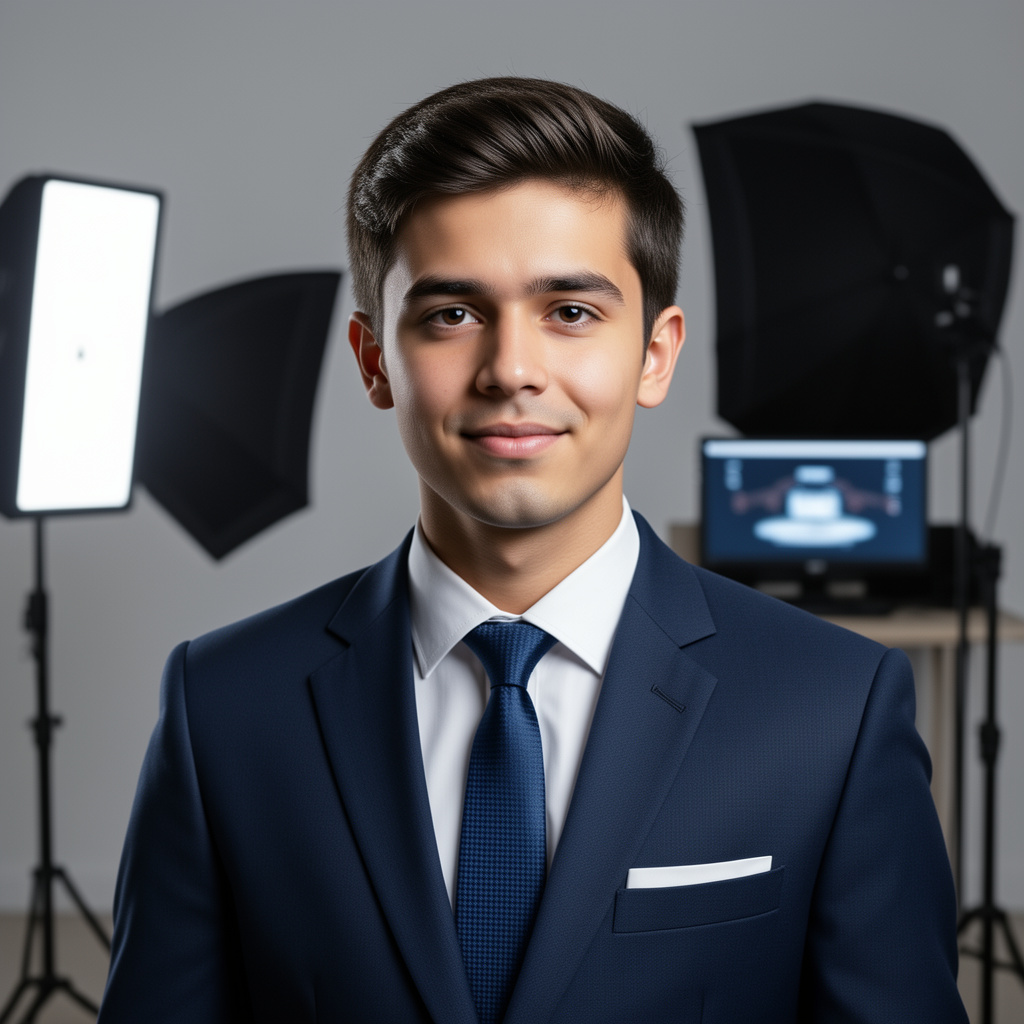}} &
            \fcolorbox{black}{white}{%
                \includegraphics[width=0.23\linewidth]{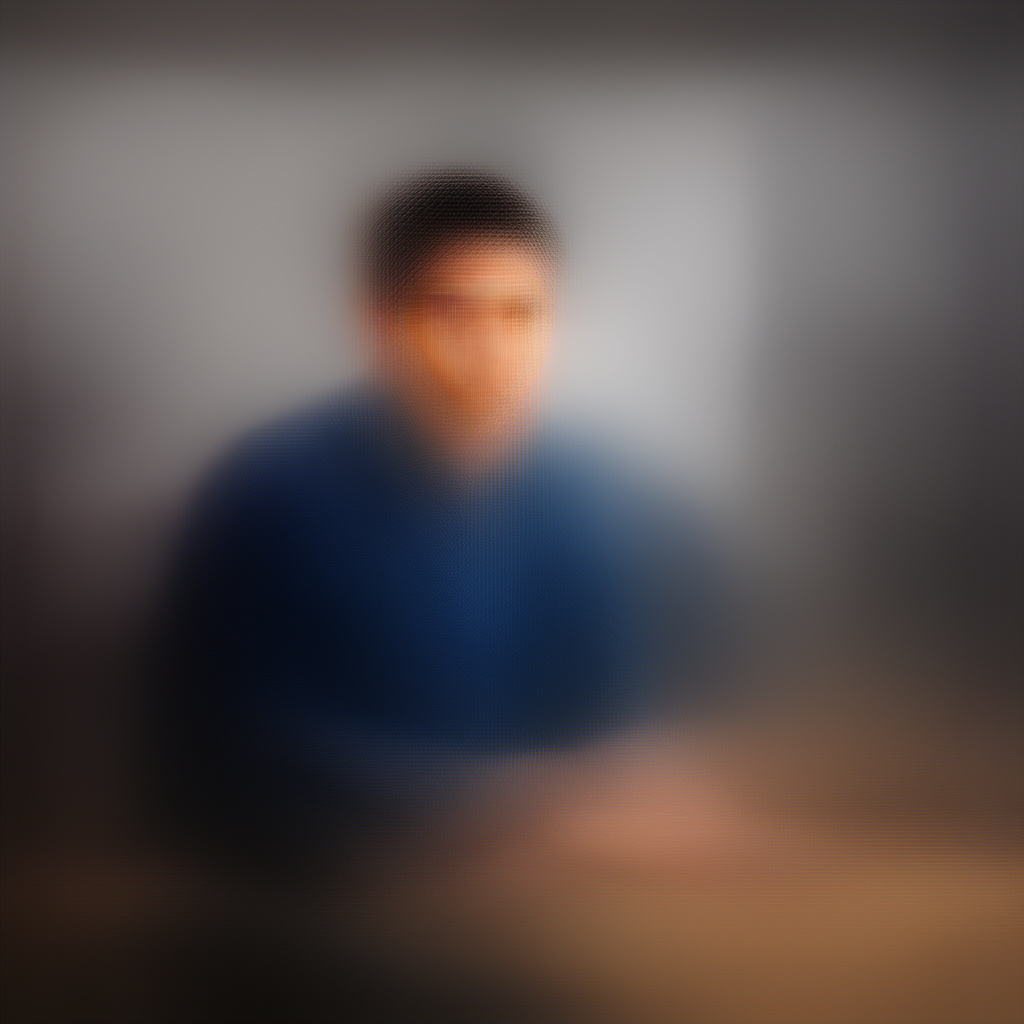}} &
            \fcolorbox{black}{white}{%
                \includegraphics[width=0.23\linewidth]{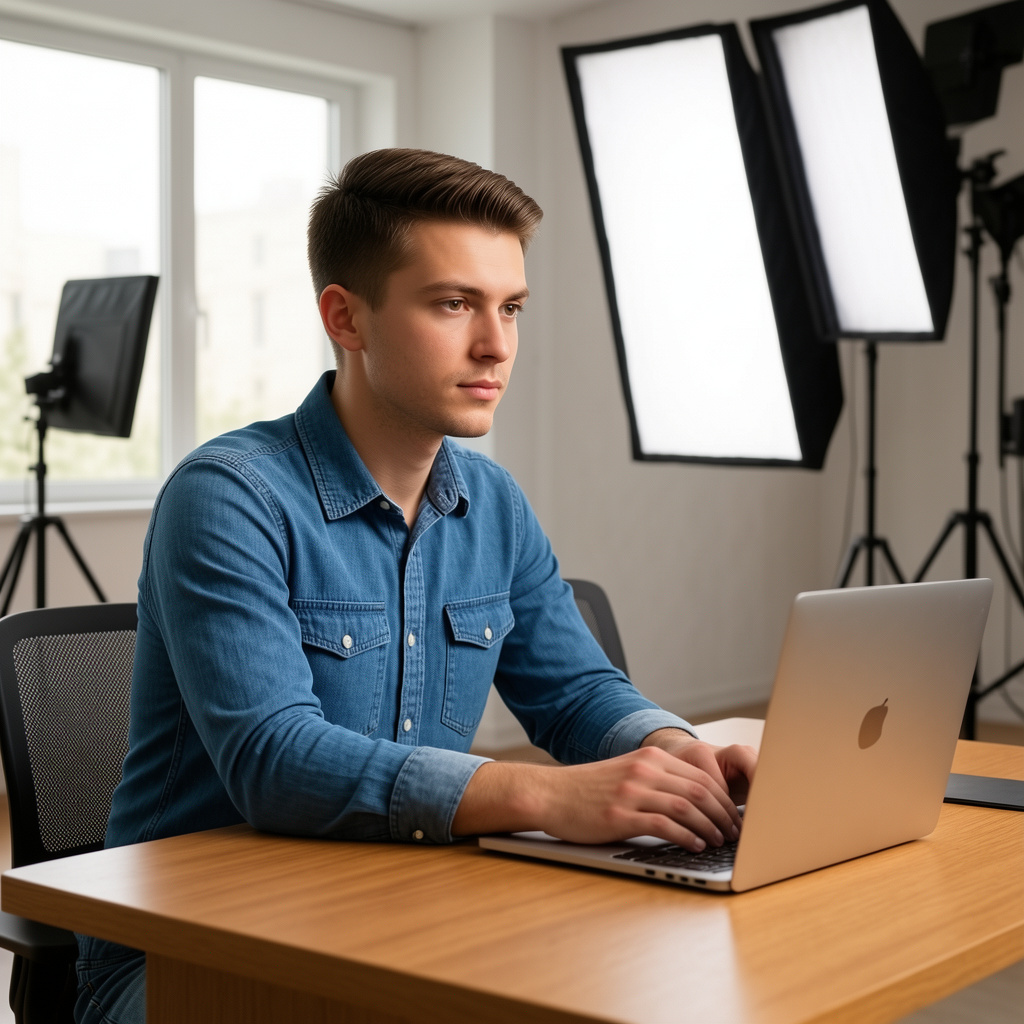}}
        \end{tabular}

        \vspace{4pt}

        \begin{tabular}{@{}p{0.47\linewidth}@{\hspace{0.04\linewidth}}p{0.47\linewidth}@{}}
            \centering\textbf{Seed A} & \centering\textbf{Seed B}
        \end{tabular}

        \vspace{3pt}

        \textit{What is the person's pose?}

        \vspace{1pt}

        \begin{tabular}{@{}p{0.47\linewidth}@{\hspace{0.04\linewidth}}p{0.47\linewidth}@{}}
            \centering
            \answerframe{0.65cm}{Standing, facing the camera.}
            &
            \centering
            \answerframe{0.65cm}{Sitting with hands on table.}
        \end{tabular}

        \vspace{4pt}

        \textit{What is the person wearing?}

        \vspace{1pt}

        \begin{tabular}{@{}p{0.47\linewidth}@{\hspace{0.04\linewidth}}p{0.47\linewidth}@{}}
            \centering
            \answerframe{0.65cm}{Suit, white shirt and blue tie.}
            &
            \centering
            \answerframe{0.65cm}{Blue button-down shirt.}
        \end{tabular}

    \end{minipage}
    \hfill
    \begin{minipage}[t]{0.485\textwidth}
        \centering

        {\small\textbf{Prompt:} ``A person holding a hat.''}

        \vspace{3pt}

        \setlength{\mirrorwidth}{0.235\linewidth}

        \begin{tabular}{@{}c@{\hspace{2pt}}c@{\hspace{2pt}}c@{\hspace{2pt}}c@{}}
            {\scriptsize\textbf{8\%}} &
            {\scriptsize\textbf{100\%}} &
            {\scriptsize\textbf{8\%}} &
            {\scriptsize\textbf{100\%}} \\[2pt]
            \fcolorbox{black}{white}{%
                \includegraphics[width=0.23\linewidth]{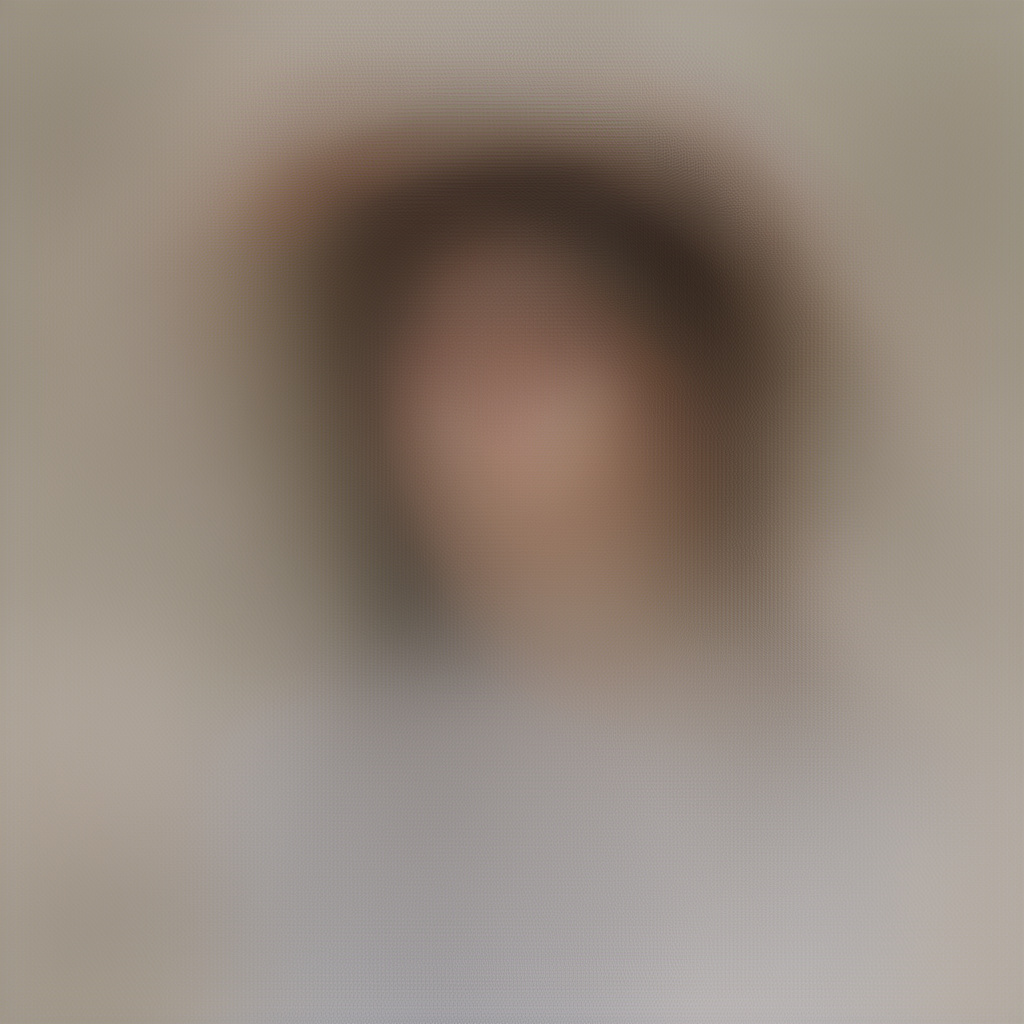}} &
            \fcolorbox{black}{white}{%
                \includegraphics[width=0.23\linewidth]{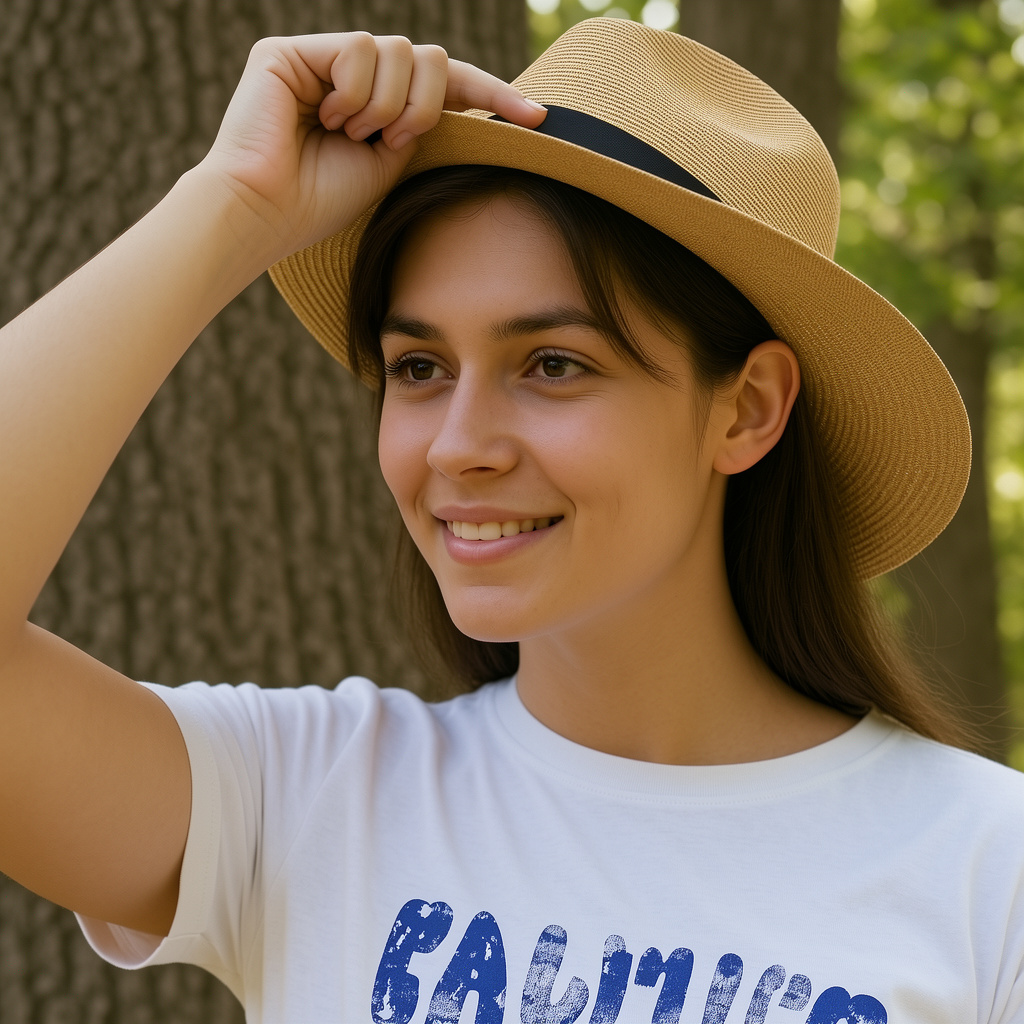}} &
            \fcolorbox{black}{white}{%
                \includegraphics[width=0.23\linewidth]{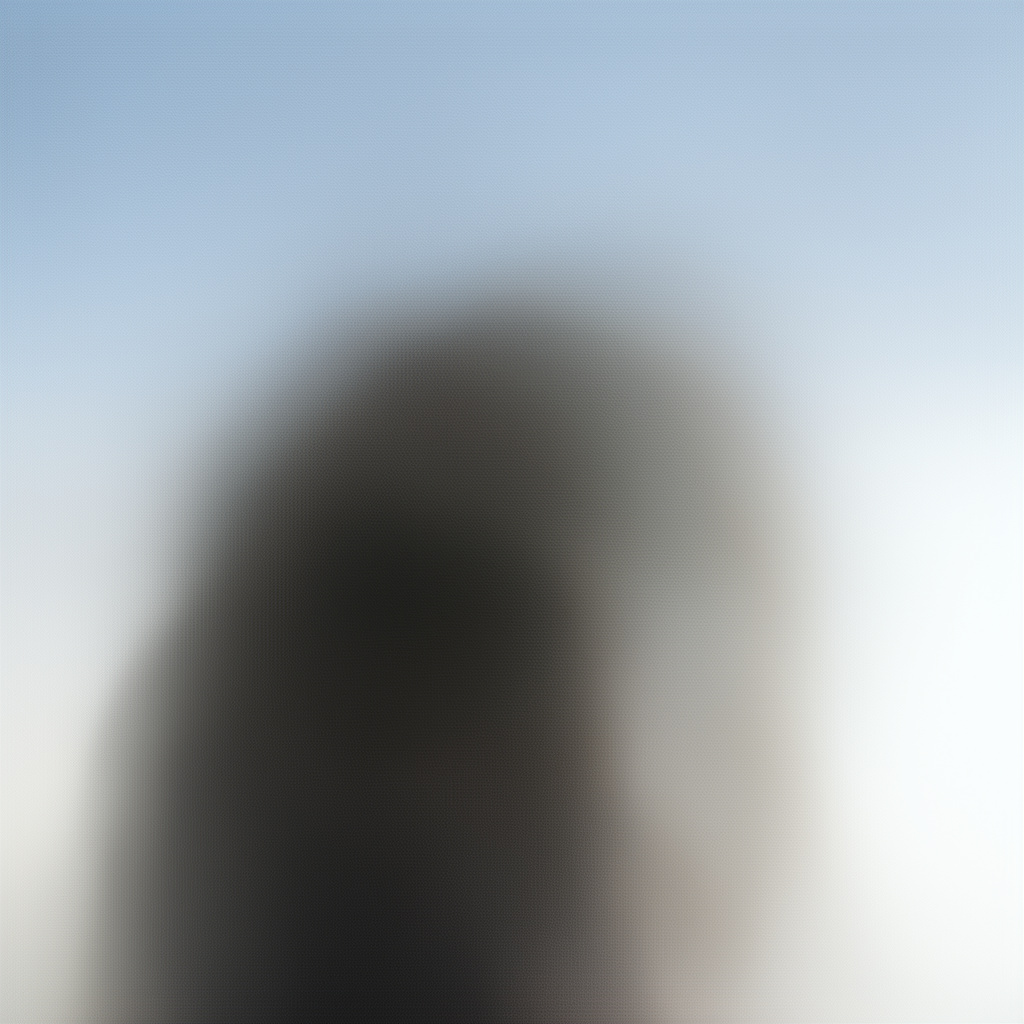}} &
            \fcolorbox{black}{white}{%
                \includegraphics[width=0.23\linewidth]{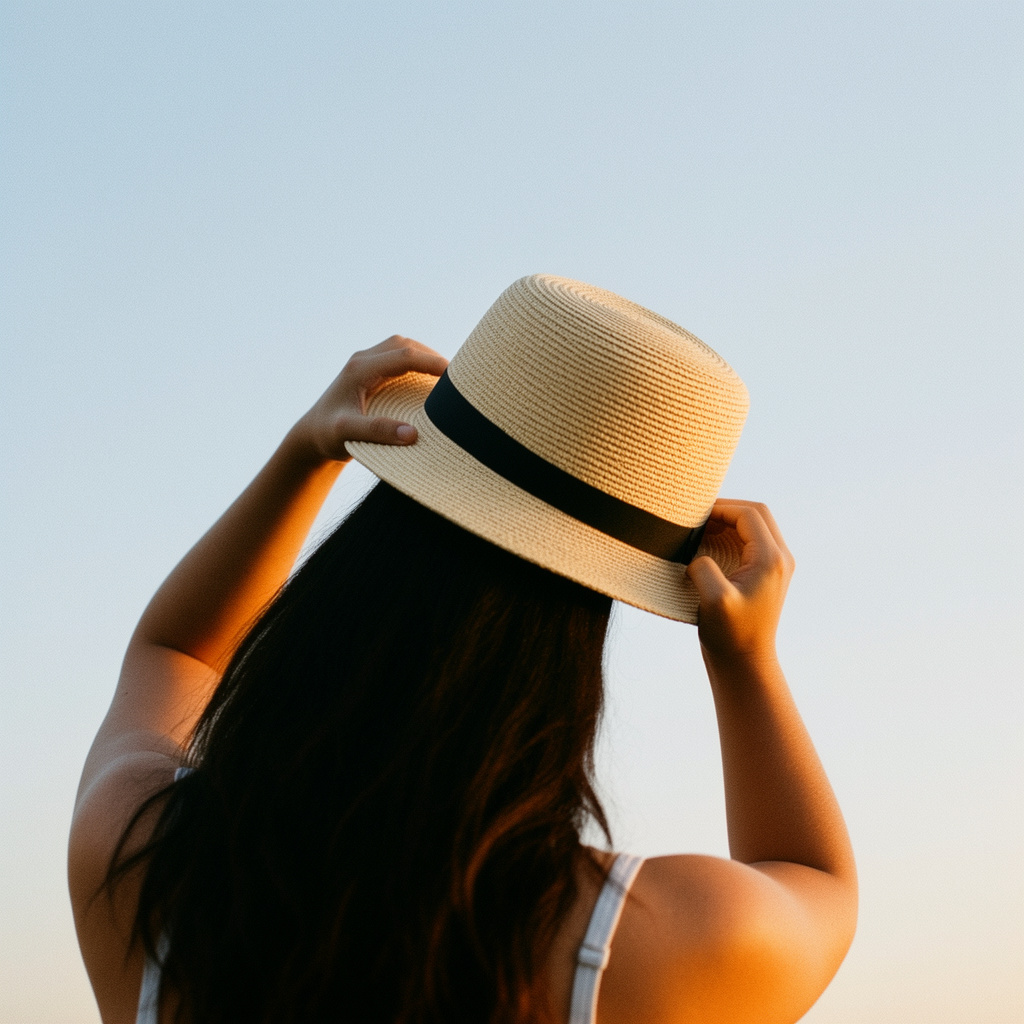}}
        \end{tabular}

        \vspace{4pt}

        \begin{tabular}{@{}p{0.47\linewidth}@{\hspace{0.04\linewidth}}p{0.47\linewidth}@{}}
            \centering\textbf{Seed A} & \centering\textbf{Seed B}
        \end{tabular}

        \vspace{3pt}

        \textit{What is the person's pose?}

        \vspace{1pt}

        \begin{tabular}{@{}p{0.47\linewidth}@{\hspace{0.04\linewidth}}p{0.47\linewidth}@{}}
            \centering
            \answerframe{0.65cm}{Holding a hat with one hand.}
            &
            \centering
            \answerframe{0.65cm}{Holding a hat with both hands.}
        \end{tabular}

        \vspace{4pt}

        \textit{What is in the background?}

        \vspace{1pt}

        \begin{tabular}{@{}p{0.47\linewidth}@{\hspace{0.04\linewidth}}p{0.47\linewidth}@{}}
            \centering
            \answerframe{0.65cm}{Trees.}
            &
            \centering
            \answerframe{0.65cm}{Sky.}
        \end{tabular}

    \end{minipage}

    \vspace{10pt}

    \begin{minipage}[t]{0.485\textwidth}
        \centering

        {\small\textbf{Prompt:} ``A person standing next to a brick wall.''}

        \vspace{3pt}

        \setlength{\mirrorwidth}{0.235\linewidth}

        \begin{tabular}{@{}c@{\hspace{2pt}}c@{\hspace{2pt}}c@{\hspace{2pt}}c@{}}
            {\scriptsize\textbf{8\%}} &
            {\scriptsize\textbf{100\%}} &
            {\scriptsize\textbf{8\%}} &
            {\scriptsize\textbf{100\%}} \\[2pt]
            \fcolorbox{black}{white}{%
                \includegraphics[width=0.23\linewidth]{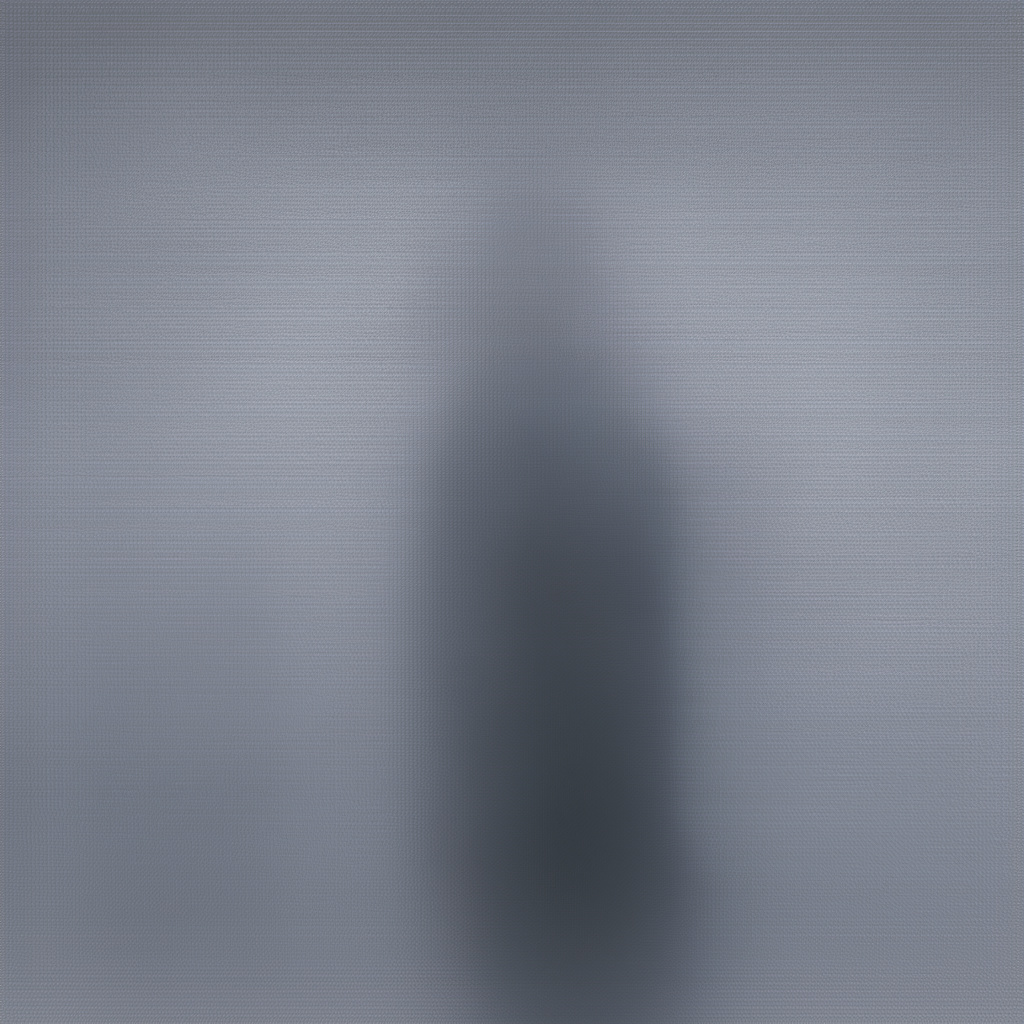}} &
            \fcolorbox{black}{white}{%
                \includegraphics[width=0.23\linewidth]{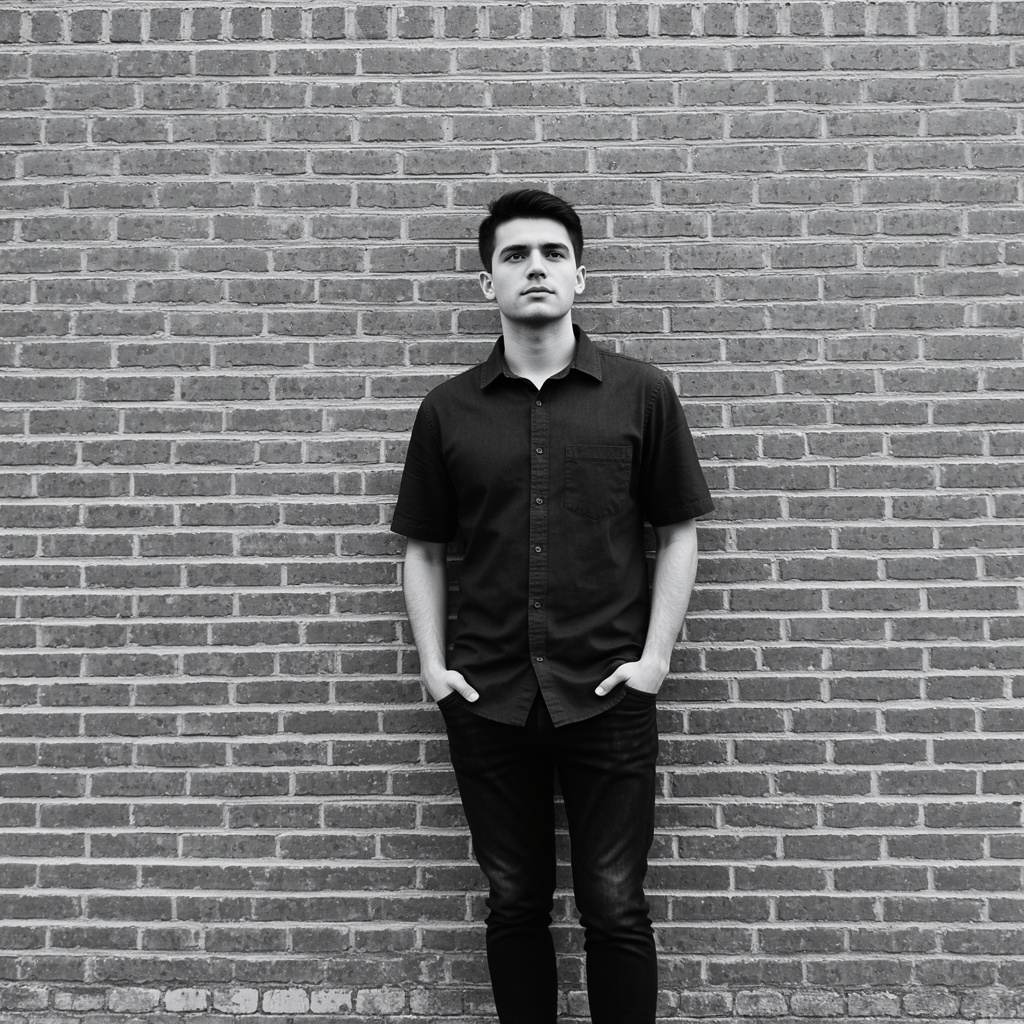}} &
            \fcolorbox{black}{white}{%
                \includegraphics[width=0.23\linewidth]{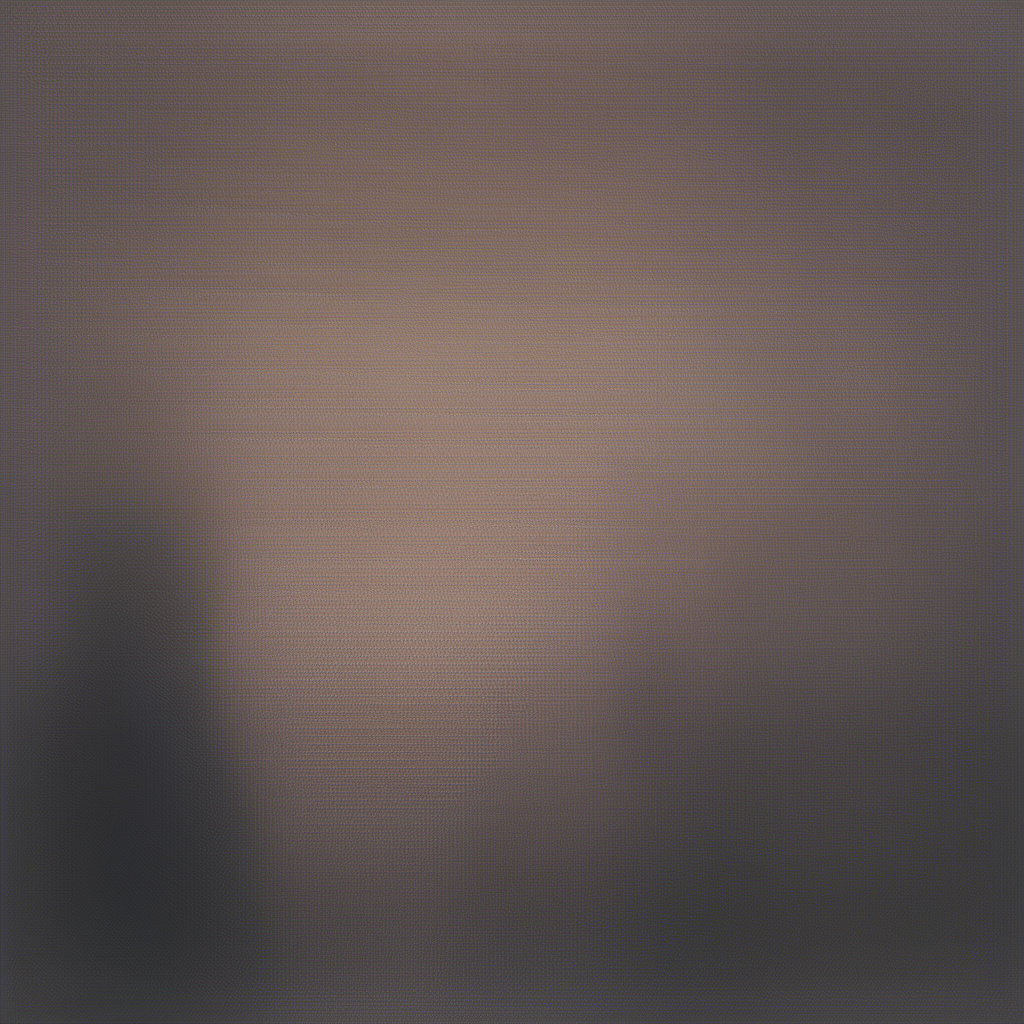}} &
            \fcolorbox{black}{white}{%
                \includegraphics[width=0.23\linewidth]{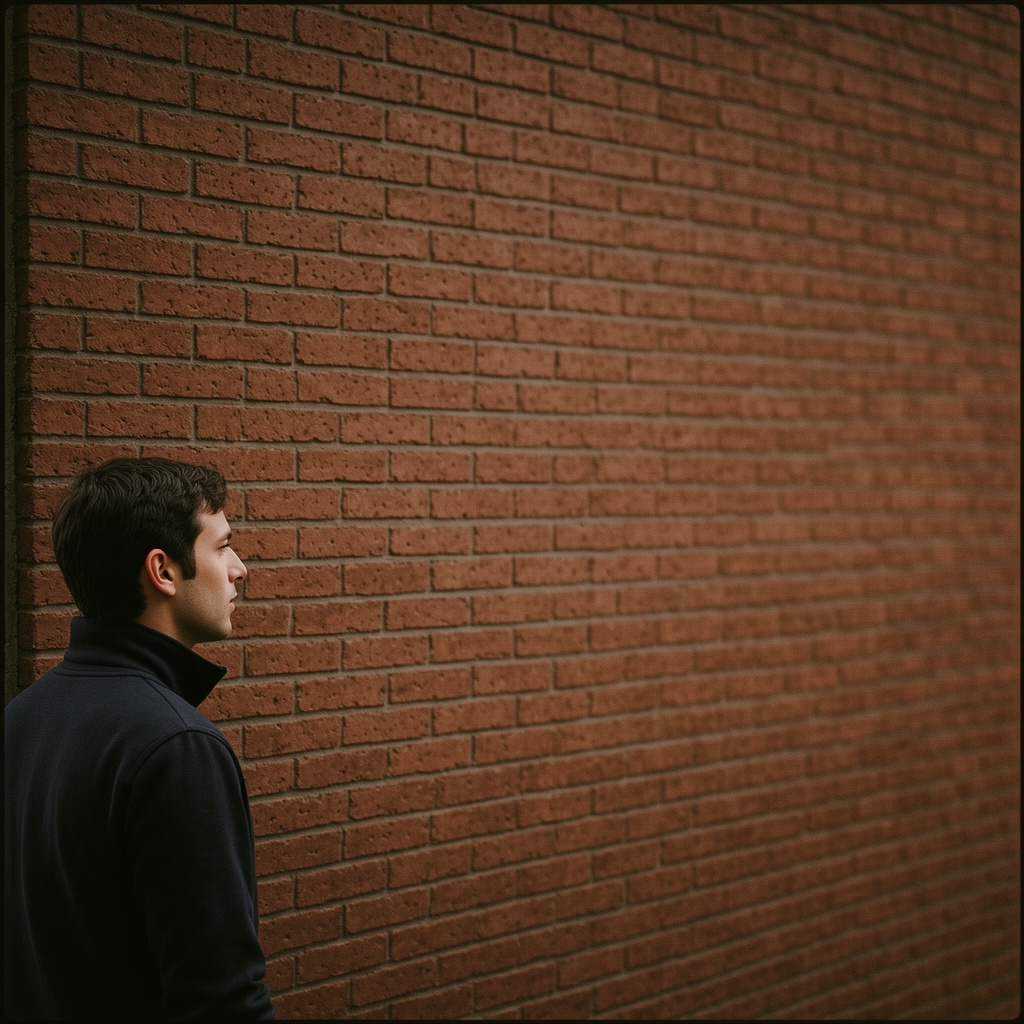}}
        \end{tabular}

        \vspace{4pt}

        \begin{tabular}{@{}p{0.47\linewidth}@{\hspace{0.04\linewidth}}p{0.47\linewidth}@{}}
            \centering\textbf{Seed A} & \centering\textbf{Seed B}
        \end{tabular}

        \vspace{3pt}

        \textit{Where is the person located in the image?}

        \vspace{1pt}

        \begin{tabular}{@{}p{0.47\linewidth}@{\hspace{0.04\linewidth}}p{0.47\linewidth}@{}}
            \centering
            \answerframe{0.65cm}{Middle.}
            &
            \centering
            \answerframe{0.65cm}{Left.}
        \end{tabular}

        \vspace{4pt}

        \textit{What is the camera angle?}

        \vspace{1pt}

        \begin{tabular}{@{}p{0.47\linewidth}@{\hspace{0.04\linewidth}}p{0.47\linewidth}@{}}
            \centering
            \answerframe{0.35cm}{Eye level.}
            &
            \centering
            \answerframe{0.35cm}{Side view.}
        \end{tabular}

    \end{minipage}
    \hfill
    \begin{minipage}[t]{0.485\textwidth}
        \centering

        {\small\textbf{Prompt:} ``A person walking near a river.''}

        \vspace{3pt}

        \setlength{\mirrorwidth}{0.235\linewidth}

        \begin{tabular}{@{}c@{\hspace{2pt}}c@{\hspace{2pt}}c@{\hspace{2pt}}c@{}}
            {\scriptsize\textbf{8\%}} &
            {\scriptsize\textbf{100\%}} &
            {\scriptsize\textbf{8\%}} &
            {\scriptsize\textbf{100\%}} \\[2pt]
            \fcolorbox{black}{white}{%
                \includegraphics[width=0.23\linewidth]{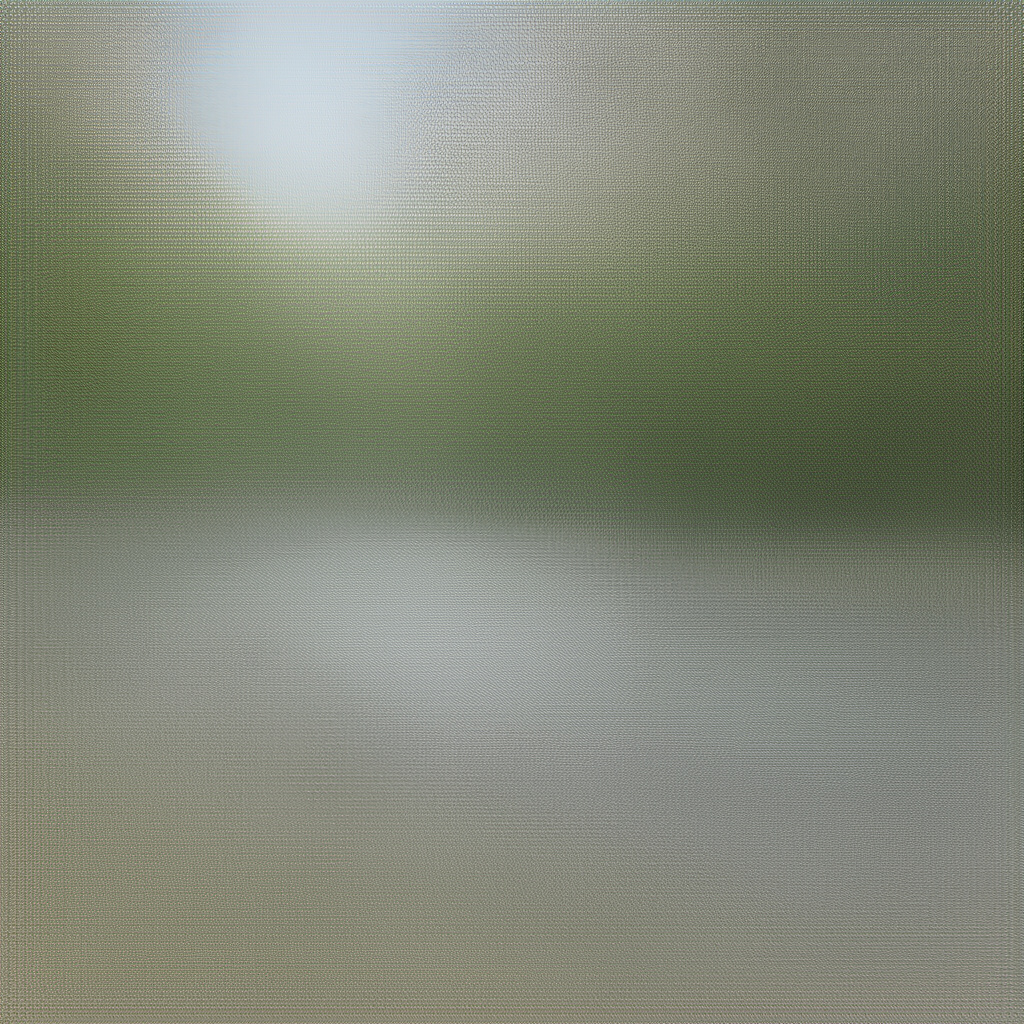}} &
            \fcolorbox{black}{white}{%
                \includegraphics[width=0.23\linewidth]{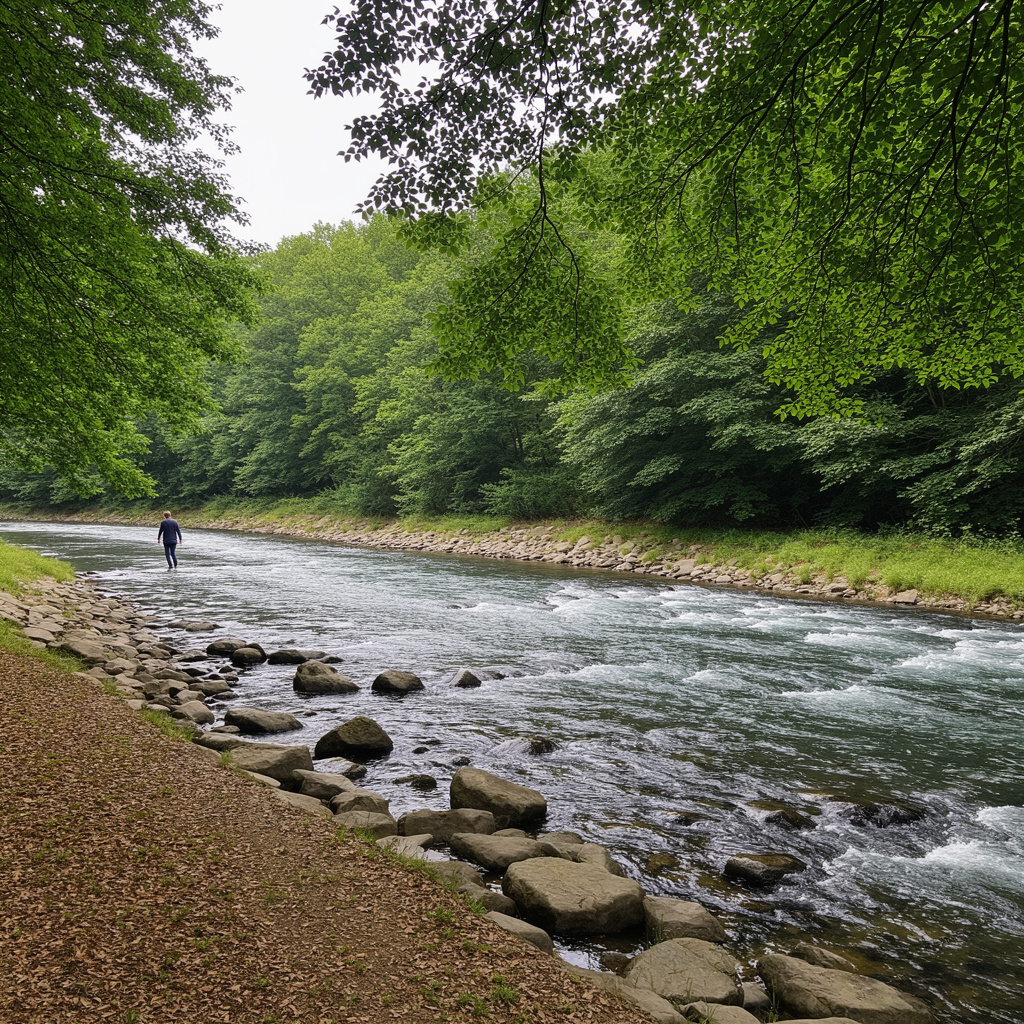}} &
            \fcolorbox{black}{white}{%
                \includegraphics[width=0.23\linewidth]{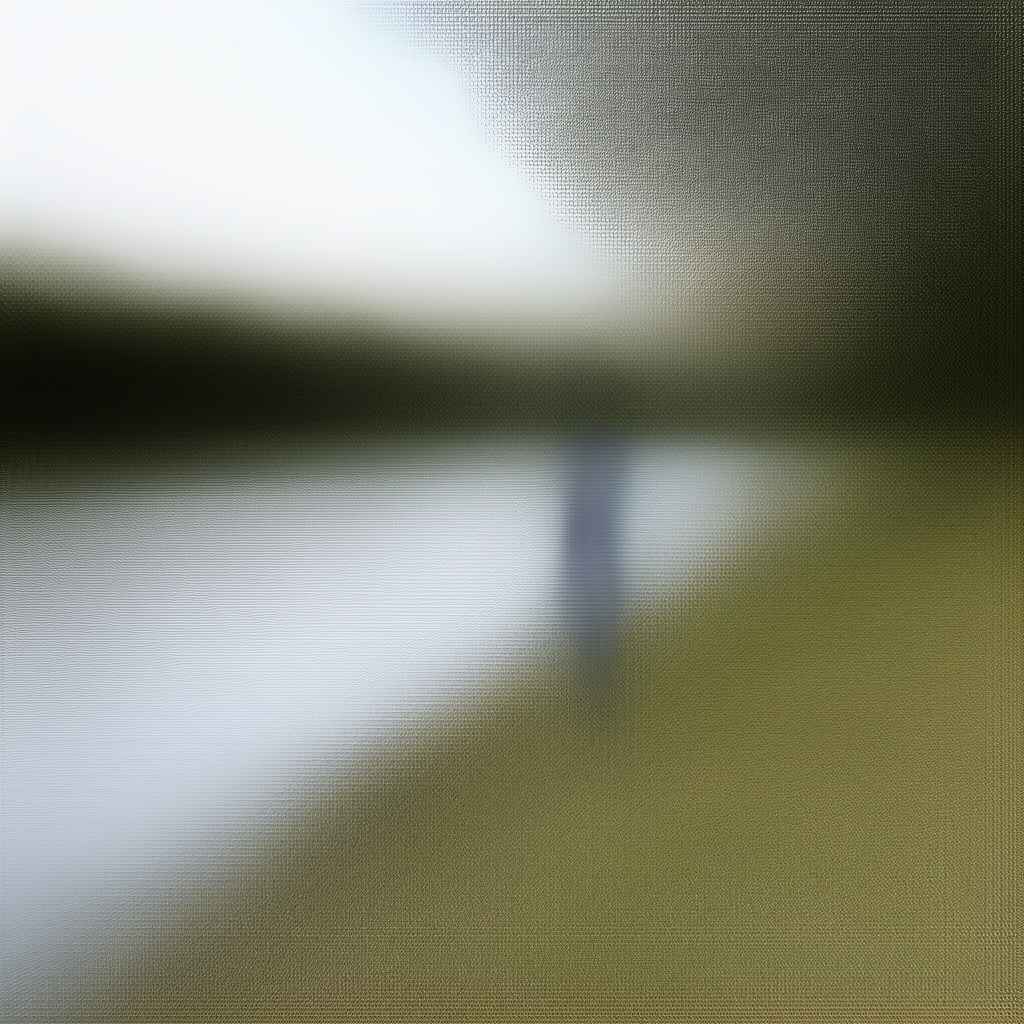}} &
            \fcolorbox{black}{white}{%
                \includegraphics[width=0.23\linewidth]{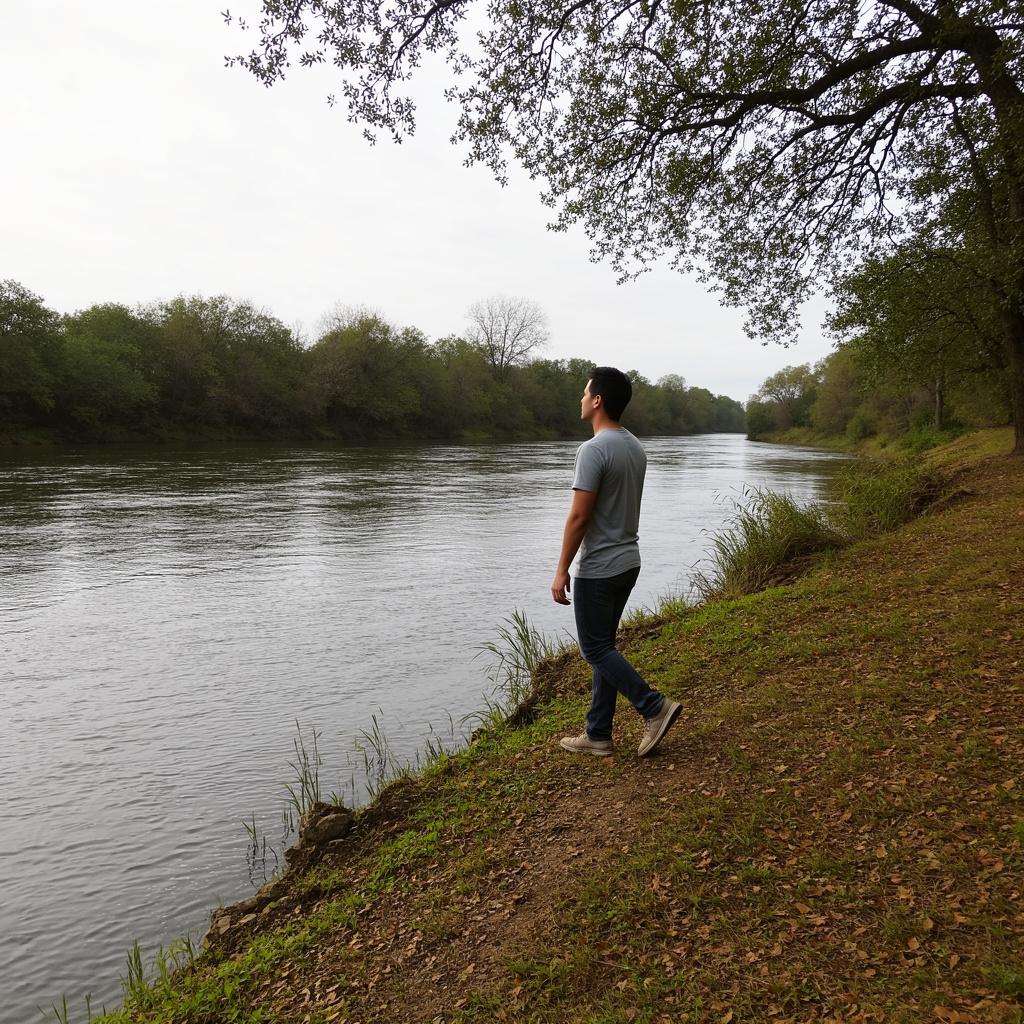}}
        \end{tabular}

        \vspace{4pt}

        \begin{tabular}{@{}p{0.47\linewidth}@{\hspace{0.04\linewidth}}p{0.47\linewidth}@{}}
            \centering\textbf{Seed A} & \centering\textbf{Seed B}
        \end{tabular}

        \vspace{3pt}

        \textit{What objects are visible?}

        \vspace{1pt}

        \begin{tabular}{@{}p{0.47\linewidth}@{\hspace{0.04\linewidth}}p{0.47\linewidth}@{}}
            \centering
            \answerframe{0.65cm}{Person, river, rocks, trees.}
            &
            \centering
            \answerframe{0.65cm}{Man, river, trees.}
        \end{tabular}

        \vspace{4pt}

        \textit{How large is the person relative to the full image?}

        \vspace{1pt}

        \begin{tabular}{@{}p{0.47\linewidth}@{\hspace{0.04\linewidth}}p{0.47\linewidth}@{}}
            \centering
            \answerframe{0.35cm}{Small.}
            &
            \centering
            \answerframe{0.35cm}{Medium.}
        \end{tabular}

    \end{minipage}

    \vspace{10pt}

    \begin{minipage}[t]{0.485\textwidth}
        \centering

        {\small\textbf{Prompt:} ``A person sitting alone on a bench.''}

        \vspace{3pt}

        \setlength{\mirrorwidth}{0.235\linewidth}

        \begin{tabular}{@{}c@{\hspace{2pt}}c@{\hspace{2pt}}c@{\hspace{2pt}}c@{}}
            {\scriptsize\textbf{8\%}} &
            {\scriptsize\textbf{100\%}} &
            {\scriptsize\textbf{8\%}} &
            {\scriptsize\textbf{100\%}} \\[2pt]
            \fcolorbox{black}{white}{%
                \includegraphics[width=0.23\linewidth]{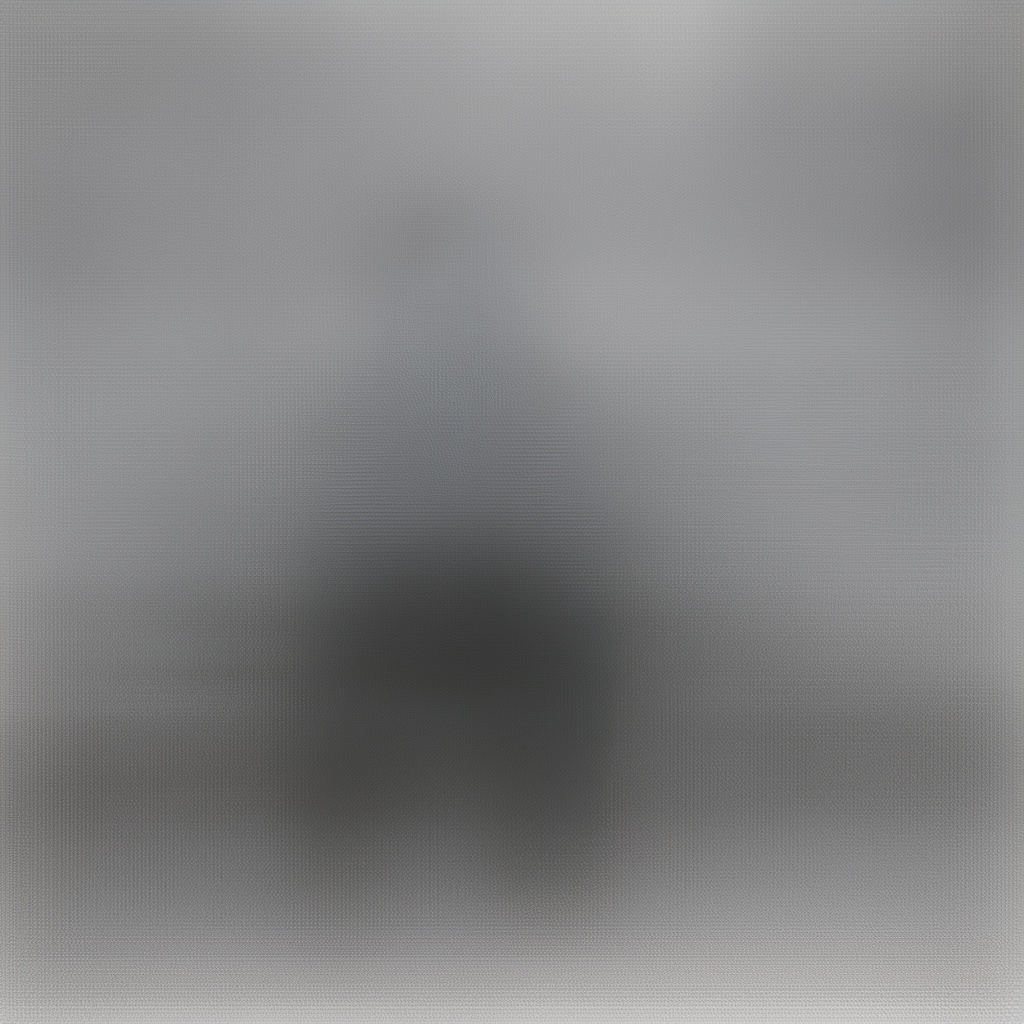}} &
            \fcolorbox{black}{white}{%
                \includegraphics[width=0.23\linewidth]{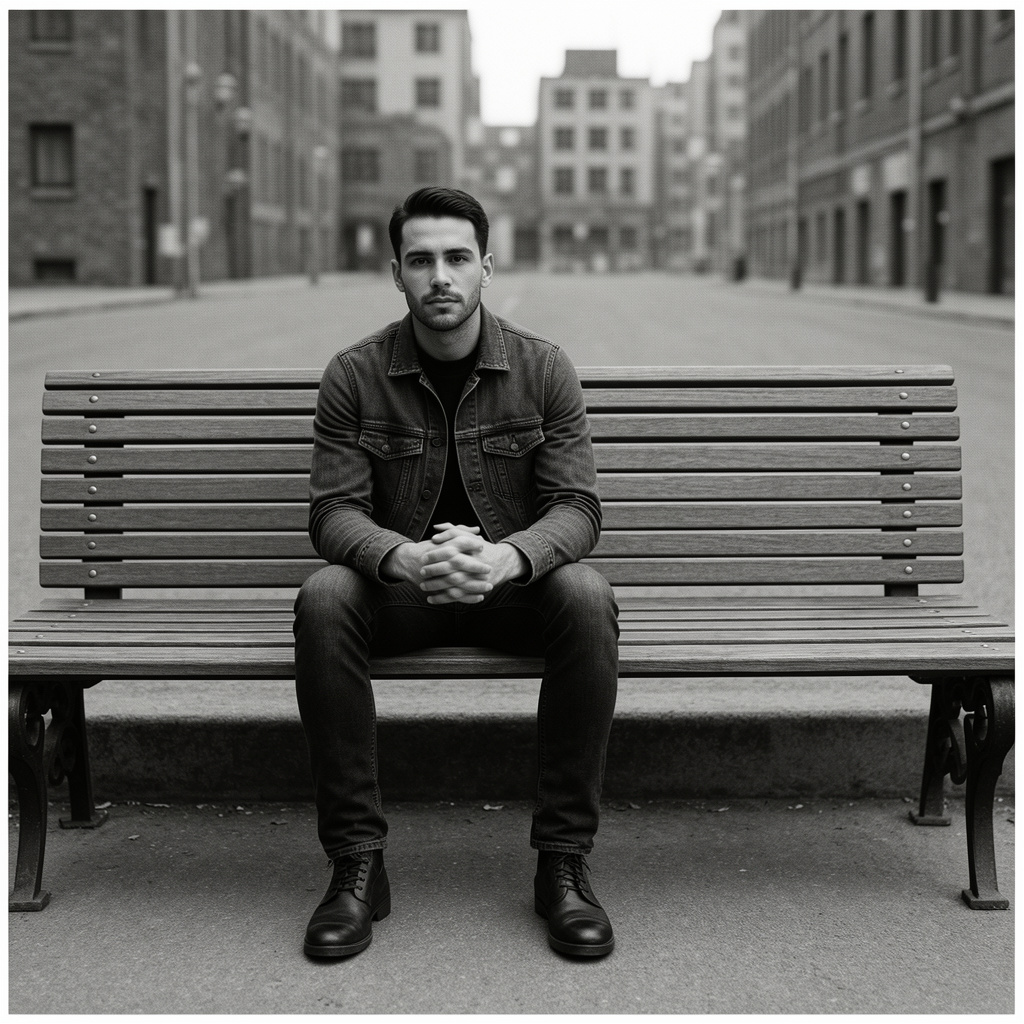}} &
            \fcolorbox{black}{white}{%
                \includegraphics[width=0.23\linewidth]{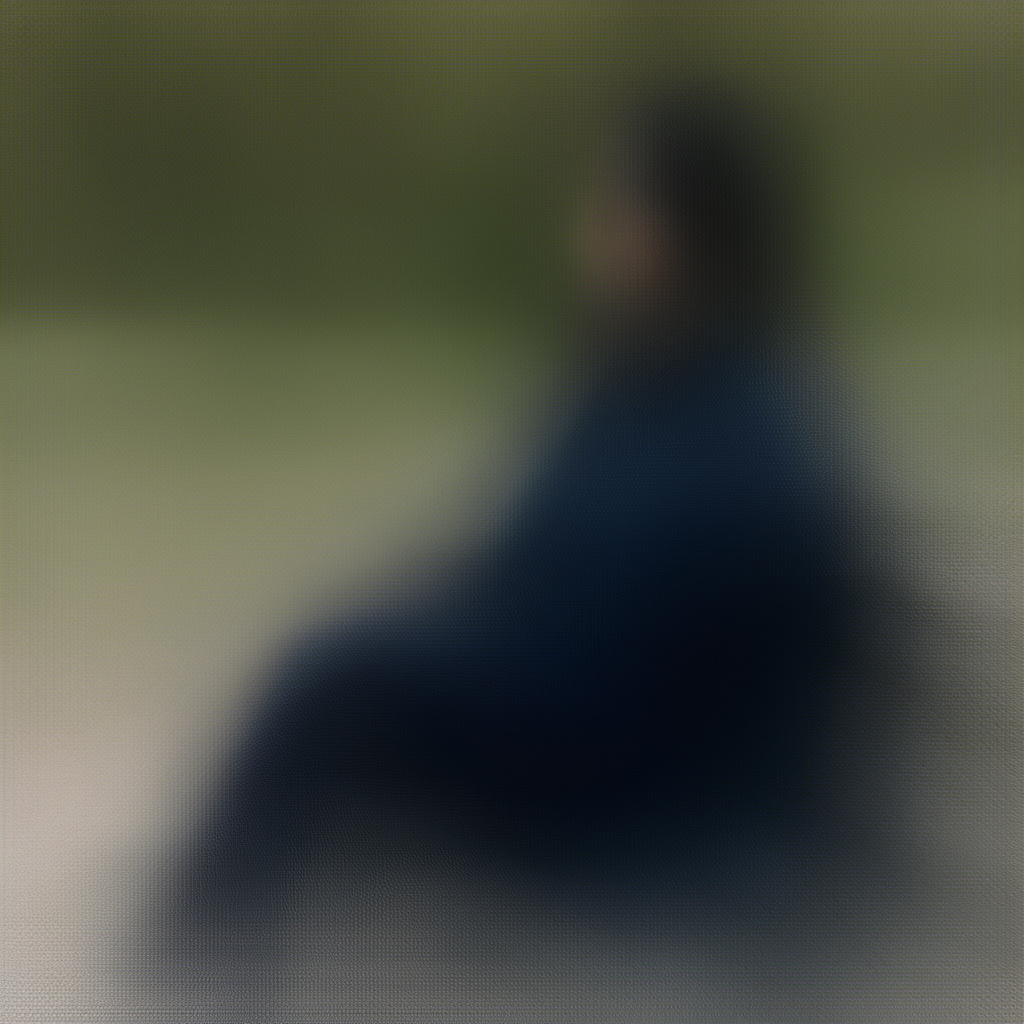}} &
            \fcolorbox{black}{white}{%
                \includegraphics[width=0.23\linewidth]{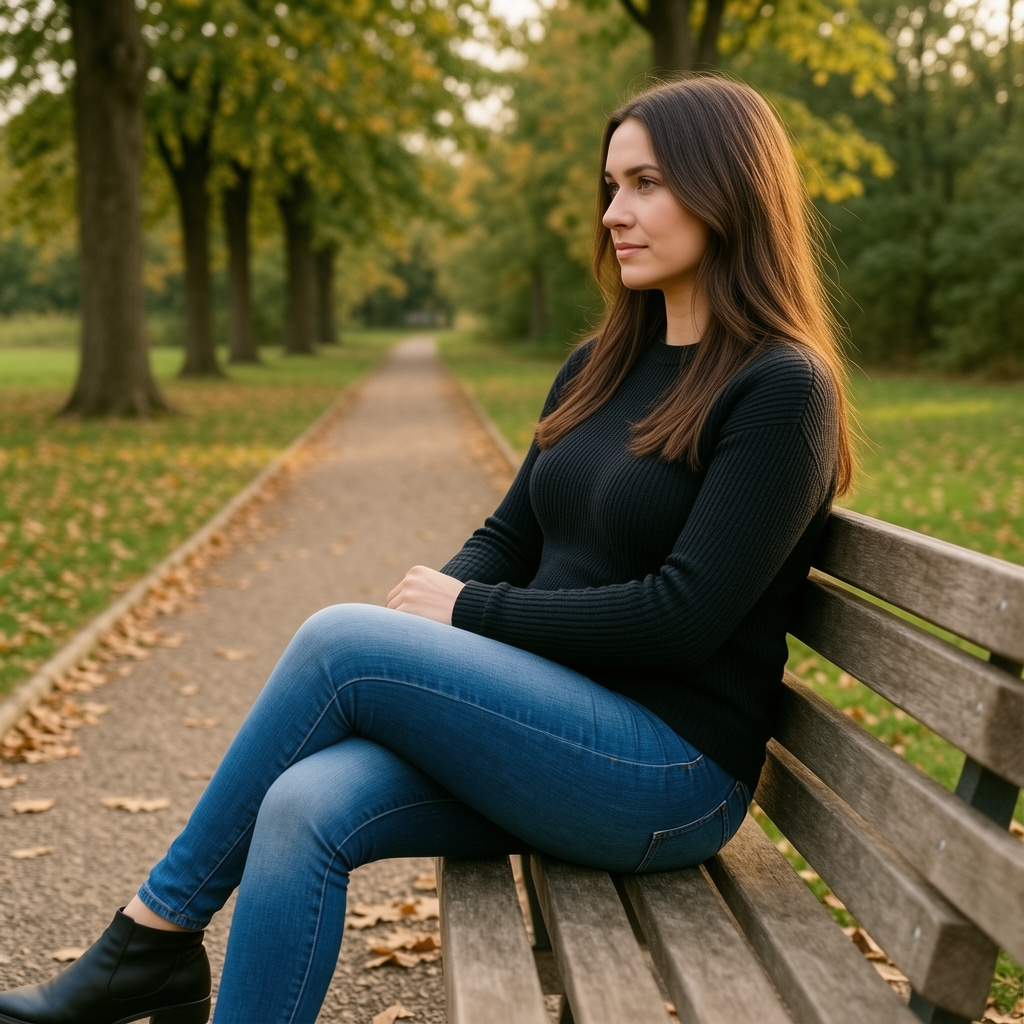}}
        \end{tabular}

        \vspace{4pt}

        \begin{tabular}{@{}p{0.47\linewidth}@{\hspace{0.04\linewidth}}p{0.47\linewidth}@{}}
            \centering\textbf{Seed A} & \centering\textbf{Seed B}
        \end{tabular}

        \vspace{3pt}

        \textit{What objects are visible?}

        \vspace{1pt}

        \begin{tabular}{@{}p{0.47\linewidth}@{\hspace{0.04\linewidth}}p{0.47\linewidth}@{}}
            \centering
            \answerframe{0.65cm}{Man, bench, street, buildings.}
            &
            \centering
            \answerframe{0.65cm}{Woman, bench, trees, grass.}
        \end{tabular}

        \vspace{4pt}

        \textit{What is the camera angle?}

        \vspace{1pt}

        \begin{tabular}{@{}p{0.47\linewidth}@{\hspace{0.04\linewidth}}p{0.47\linewidth}@{}}
            \centering
            \answerframe{0.35cm}{Eye level.}
            &
            \centering
            \answerframe{0.35cm}{Side view.}
        \end{tabular}

    \end{minipage}
    \hfill
    \begin{minipage}[t]{0.485\textwidth}
        \centering

        {\small\textbf{Prompt:} ``Young land surveyor in the workshop.''}

        \vspace{3pt}

        \setlength{\mirrorwidth}{0.235\linewidth}

        \begin{tabular}{@{}c@{\hspace{2pt}}c@{\hspace{2pt}}c@{\hspace{2pt}}c@{}}
            {\scriptsize\textbf{8\%}} &
            {\scriptsize\textbf{100\%}} &
            {\scriptsize\textbf{8\%}} &
            {\scriptsize\textbf{100\%}} \\[2pt]
            \fcolorbox{black}{white}{%
                \includegraphics[width=0.23\linewidth]{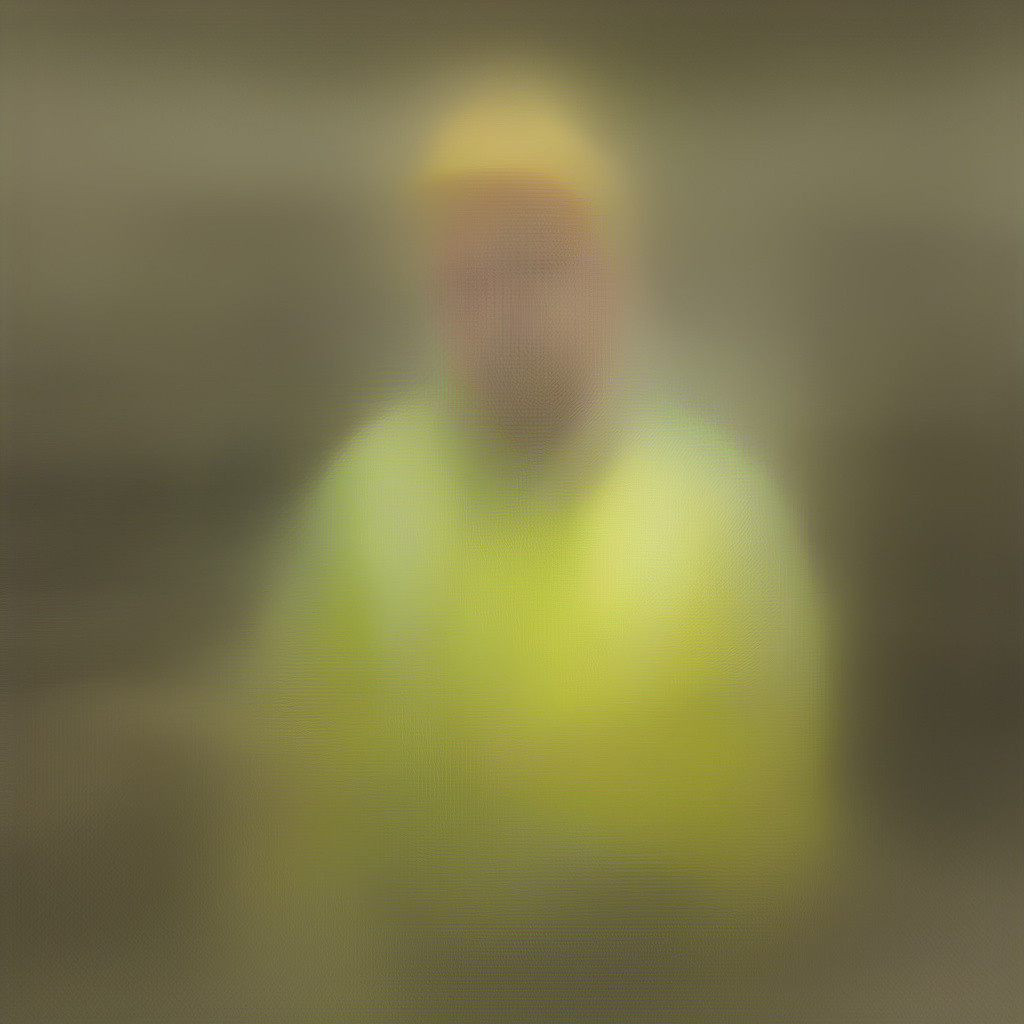}} &
            \fcolorbox{black}{white}{%
                \includegraphics[width=0.23\linewidth]{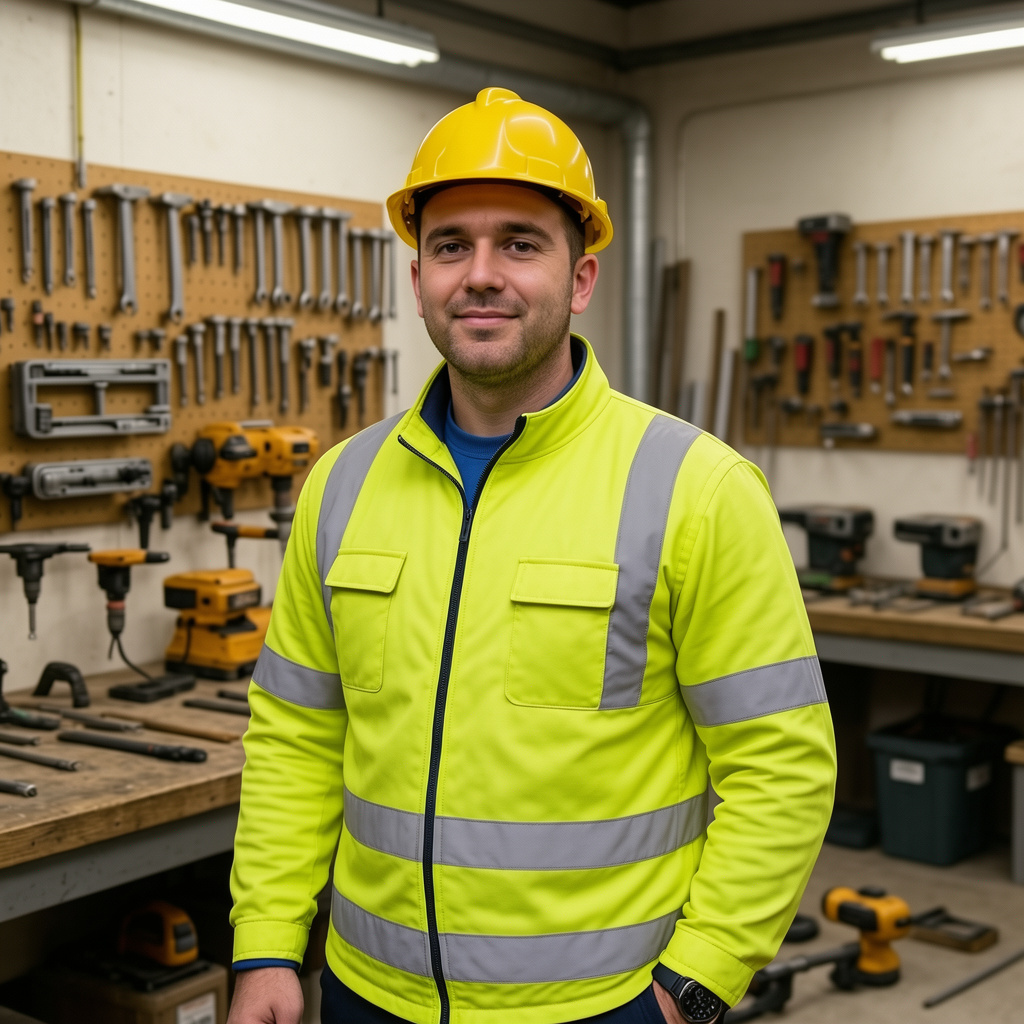}} &
            \fcolorbox{black}{white}{%
                \includegraphics[width=0.23\linewidth]{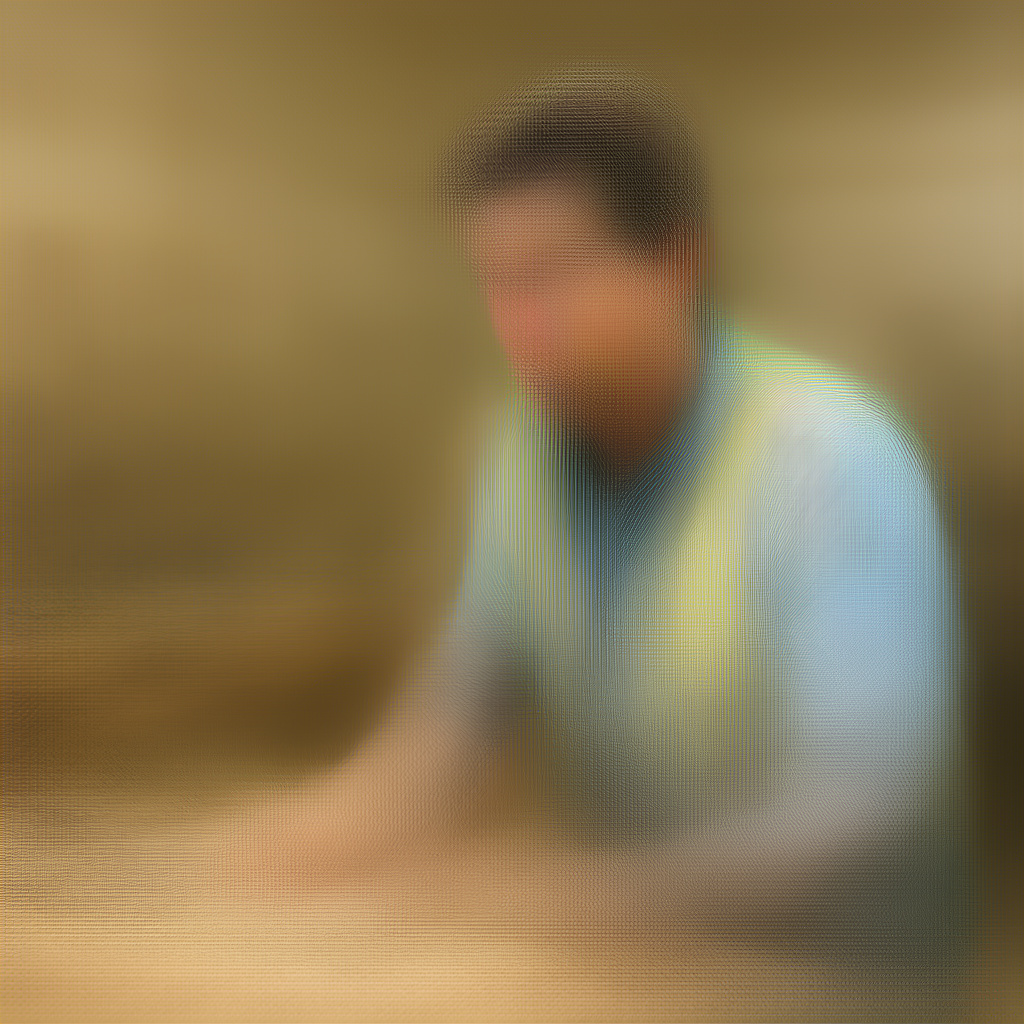}} &
            \fcolorbox{black}{white}{%
                \includegraphics[width=0.23\linewidth]{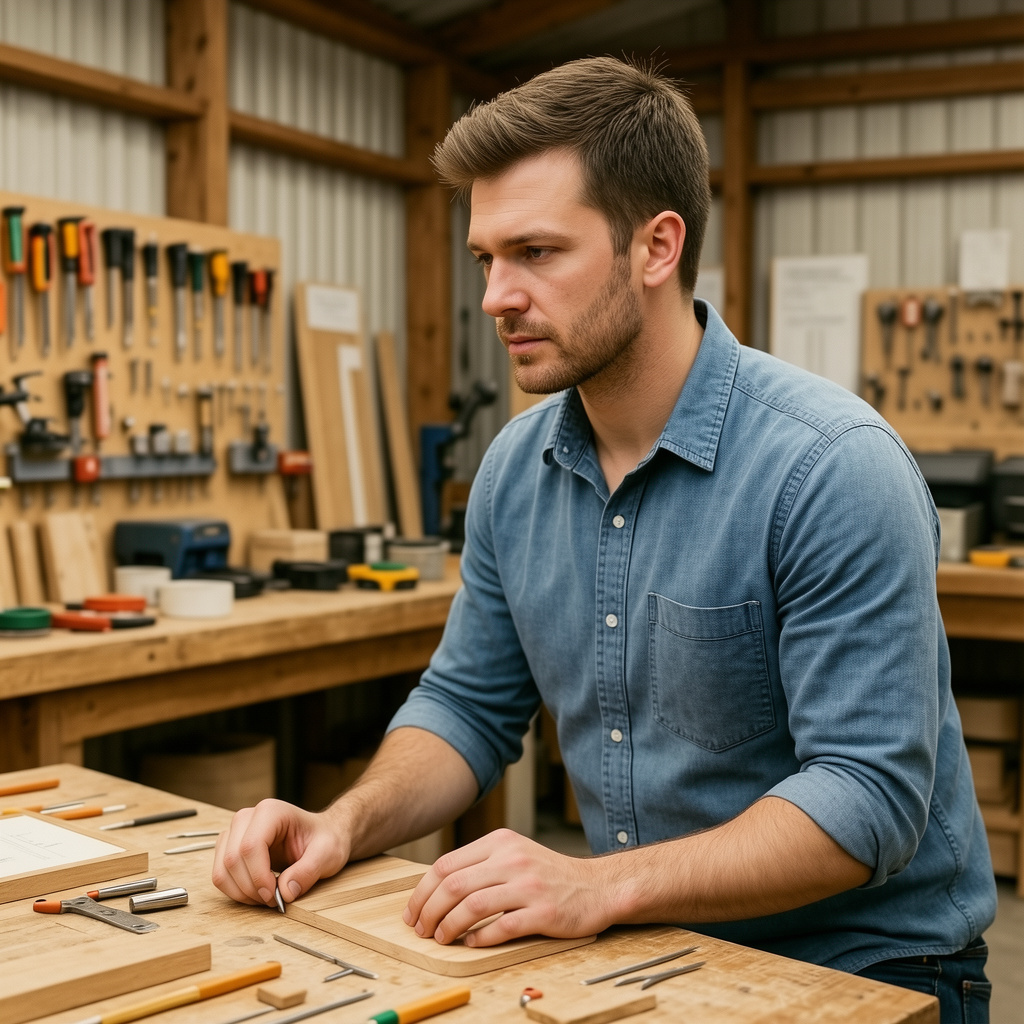}}
        \end{tabular}

        \vspace{4pt}

        \begin{tabular}{@{}p{0.47\linewidth}@{\hspace{0.04\linewidth}}p{0.47\linewidth}@{}}
            \centering\textbf{Seed A} & \centering\textbf{Seed B}
        \end{tabular}

        \vspace{3pt}

        \textit{What is the person's pose?}

        \vspace{1pt}

        \begin{tabular}{@{}p{0.47\linewidth}@{\hspace{0.04\linewidth}}p{0.47\linewidth}@{}}
            \centering
            \answerframe{0.65cm}{Standing, facing the camera.}
            &
            \centering
            \answerframe{0.65cm}{Sitting at a table, working.}
        \end{tabular}

        \vspace{4pt}

        \textit{What is the person wearing?}

        \vspace{1pt}

        \begin{tabular}{@{}p{0.47\linewidth}@{\hspace{0.04\linewidth}}p{0.47\linewidth}@{}}
            \centering
            \answerframe{0.35cm}{Safety vest and hard hat.}
            &
            \centering
            \answerframe{0.35cm}{Blue shirt and jeans.}
        \end{tabular}

    \end{minipage}

    \caption{
        \textbf{Additional plain-text predictions from our conditional Contextual Reader on FLUX.2.}
        For each prompt, we compare two random seeds that resolve underspecified
        attributes differently. We show the predicted clean images at $8\%$ of the
        denoising trajectory for both seeds, followed by their final images. The
        reader predictions are obtained from the contextual tokens at the same
        $8\%$ timestep.
    }
    \label{fig:conditional_sup}
\end{figure*}

\begin{figure*}[t]
    \centering
    \footnotesize

    \setlength{\fboxrule}{1pt}
    \setlength{\fboxsep}{0pt}

    \begin{minipage}[t]{0.485\textwidth}
        \centering

        {\small\textbf{Prompt:} ``A person carrying groceries.''}

        \vspace{3pt}

        \setlength{\mirrorwidth}{0.235\linewidth}

        \begin{tabular}{@{}c@{\hspace{2pt}}c@{\hspace{2pt}}c@{\hspace{2pt}}c@{}}
            {\scriptsize\textbf{18\%}} &
            {\scriptsize\textbf{100\%}} &
            {\scriptsize\textbf{18\%}} &
            {\scriptsize\textbf{100\%}} \\[2pt]
            \fcolorbox{black}{white}{%
                \includegraphics[width=0.23\linewidth]
                {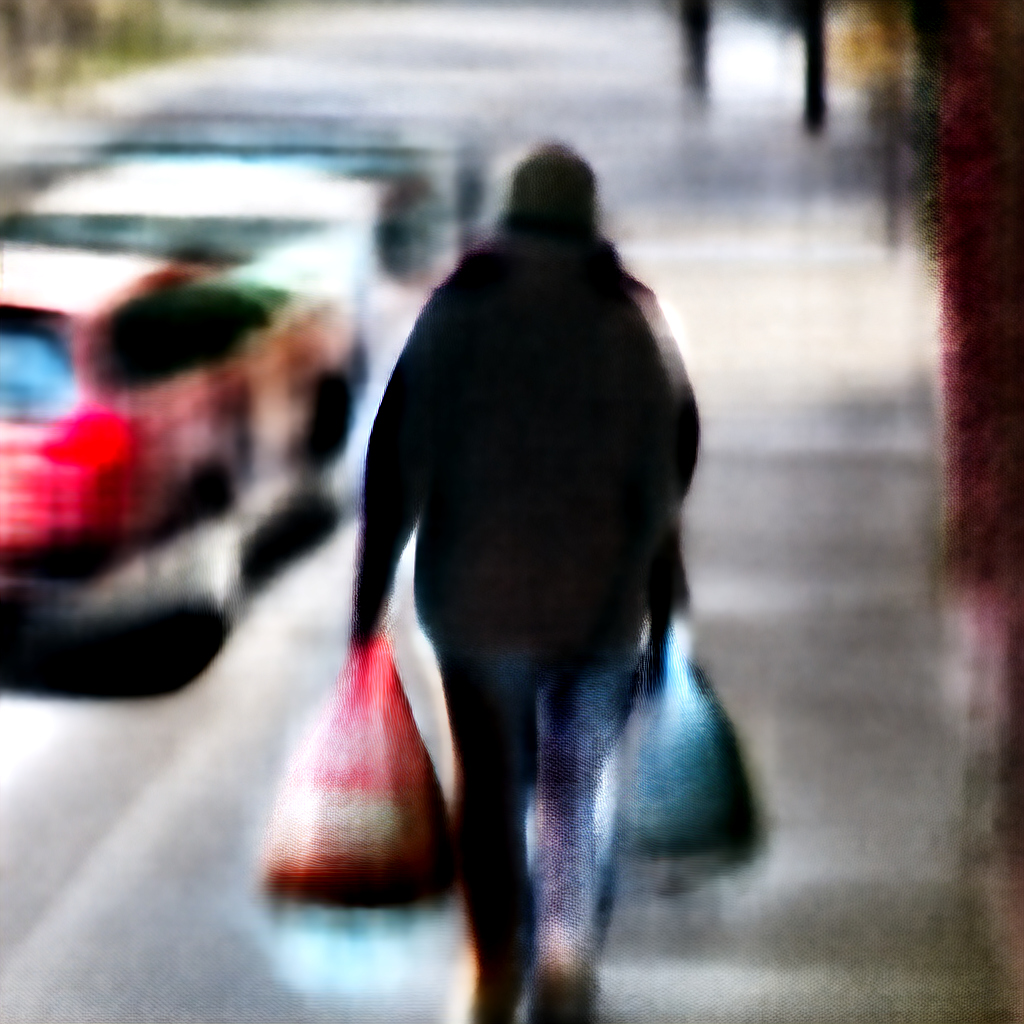}} &
            \fcolorbox{black}{white}{%
                \includegraphics[width=0.23\linewidth]
                {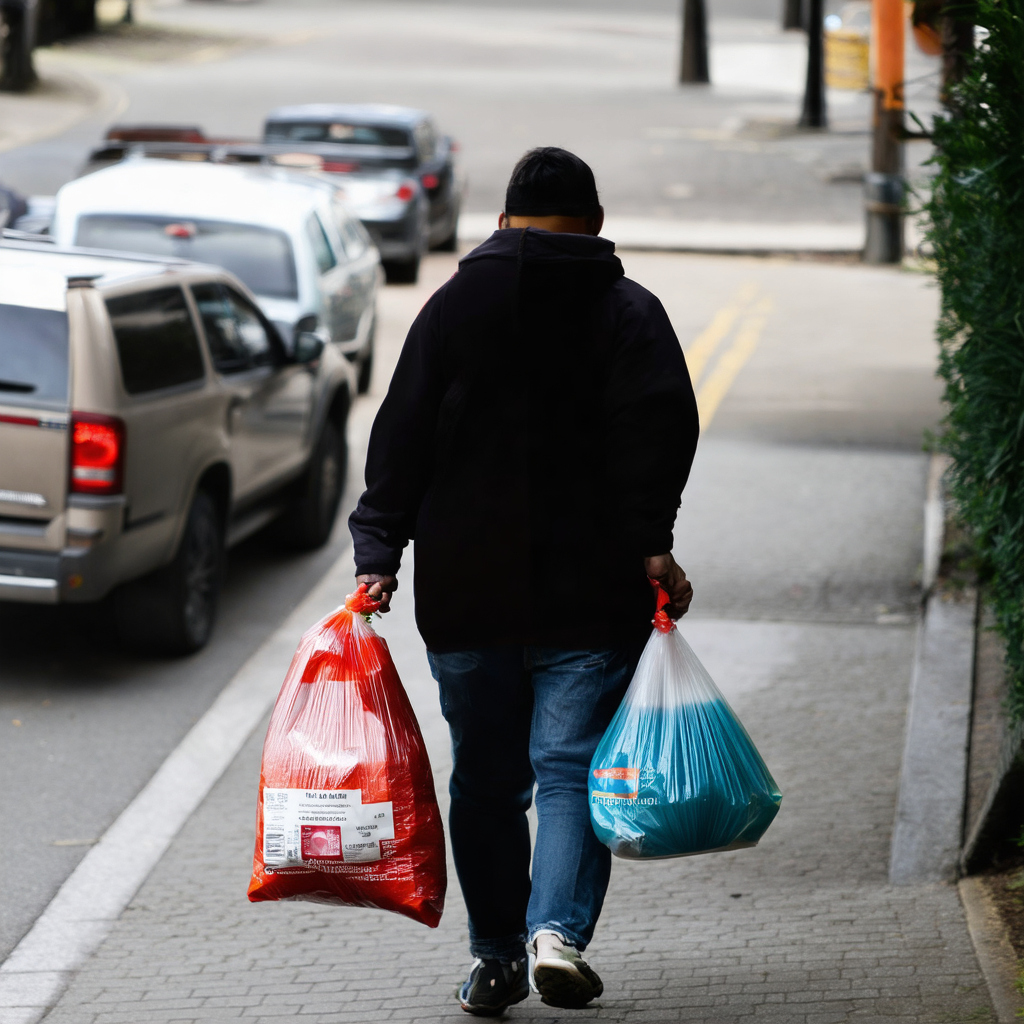}} &
            \fcolorbox{black}{white}{%
                \includegraphics[width=0.23\linewidth]
                {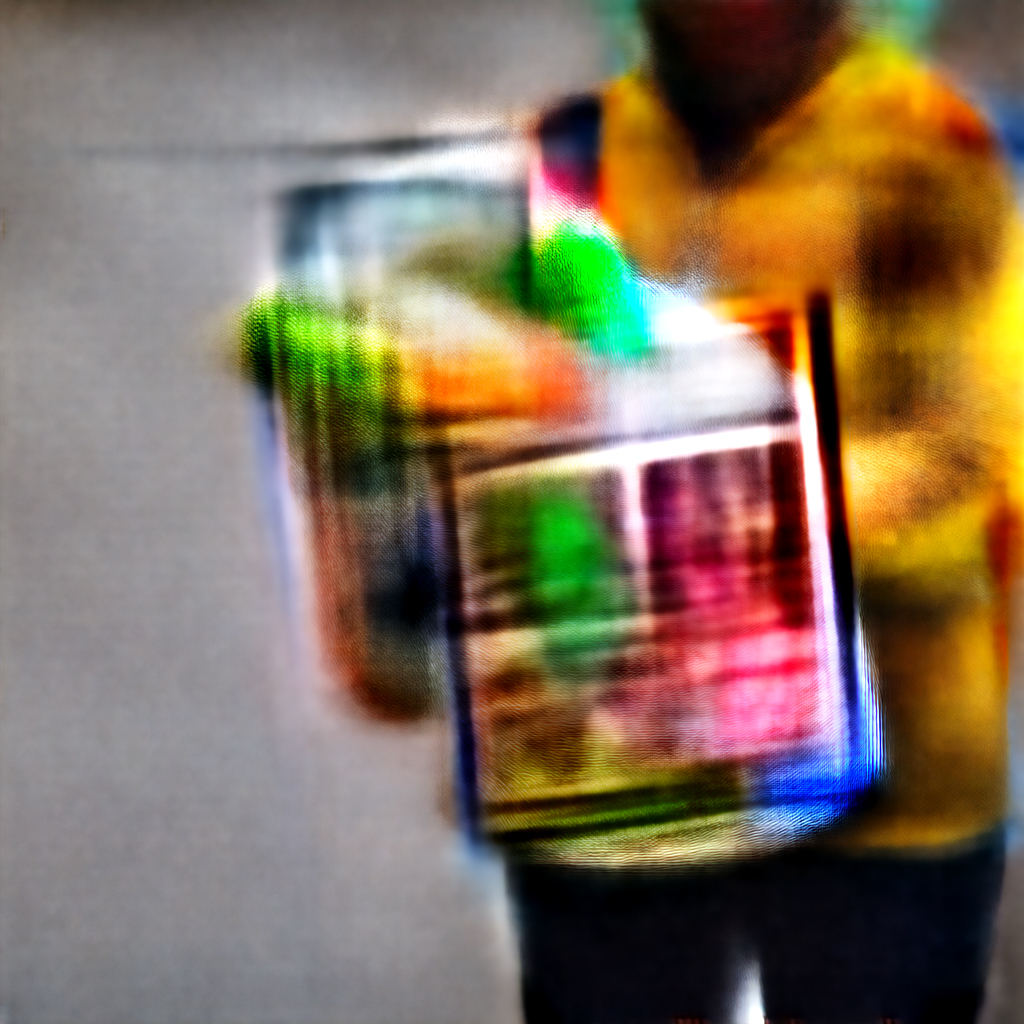}} &
            \fcolorbox{black}{white}{%
                \includegraphics[width=0.23\linewidth]
                {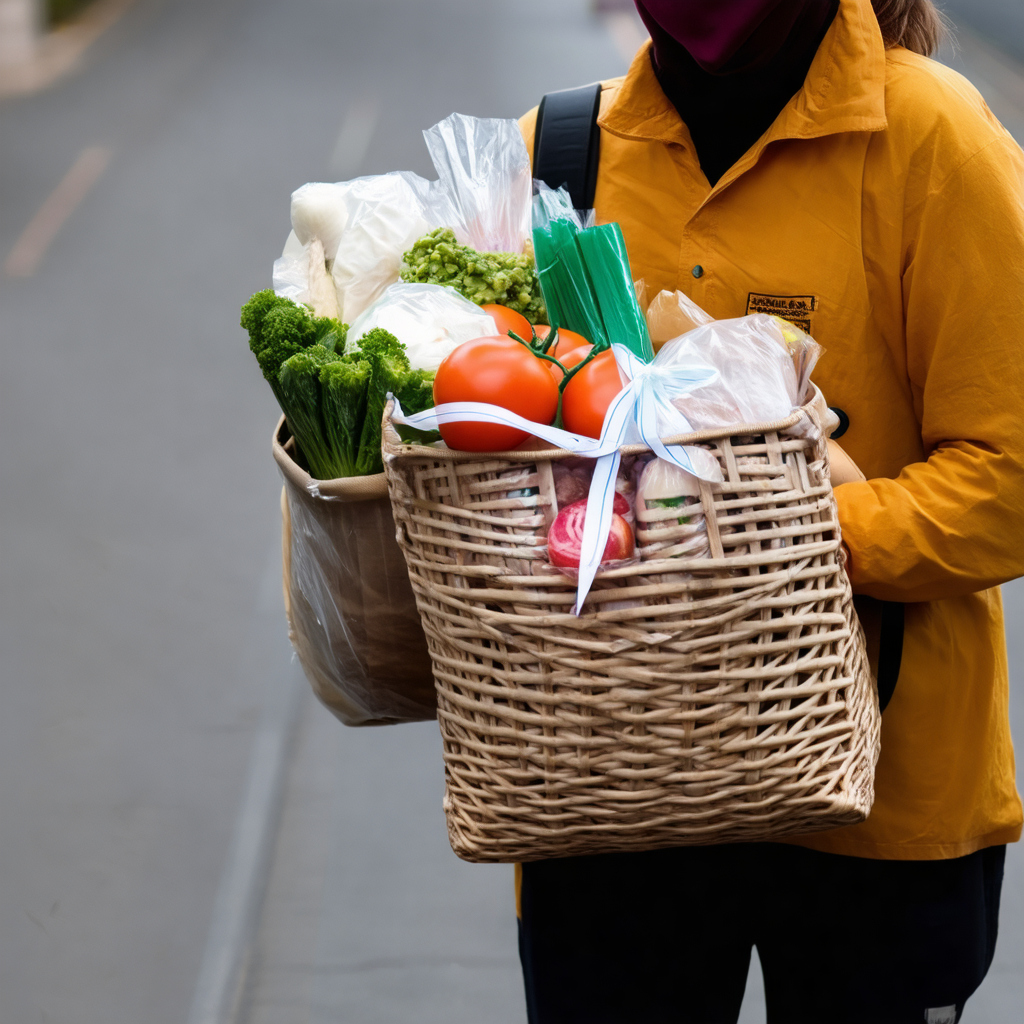}}
        \end{tabular}

        \vspace{4pt}

        \begin{tabular}{@{}p{0.47\linewidth}@{\hspace{0.04\linewidth}}p{0.47\linewidth}@{}}
            \centering\textbf{Seed A} &
            \centering\textbf{Seed B}
        \end{tabular}

        \vspace{3pt}

        \textit{What is the person's pose?}

        \vspace{1pt}

        \begin{tabular}{@{}p{0.47\linewidth}@{\hspace{0.04\linewidth}}p{0.47\linewidth}@{}}
            \centering
            \answerframe{1cm}{Walking with two large bags, holding them by handles.}
            &
            \centering
            \answerframe{1cm}{Carrying bags with both hands.}
        \end{tabular}

        \vspace{4pt}

        \textit{What objects are visible?}

        \vspace{1pt}

        \begin{tabular}{@{}p{0.47\linewidth}@{\hspace{0.04\linewidth}}p{0.47\linewidth}@{}}
            \centering
            \answerframe{0.655cm}{Person, trash bags, shoes, street, cars.}
            &
            \centering
            \answerframe{0.65cm}{Person, brown bags, fruits and vegetables.}
        \end{tabular}

    \end{minipage}
    \hfill
    \begin{minipage}[t]{0.485\textwidth}
        \centering

        {\small\textbf{Prompt:} ``A person looking at a laptop.''}

        \vspace{3pt}

        \setlength{\mirrorwidth}{0.235\linewidth}

        \begin{tabular}{@{}c@{\hspace{2pt}}c@{\hspace{2pt}}c@{\hspace{2pt}}c@{}}
            {\scriptsize\textbf{18\%}} &
            {\scriptsize\textbf{100\%}} &
            {\scriptsize\textbf{18\%}} &
            {\scriptsize\textbf{100\%}} \\[2pt]
            \fcolorbox{black}{white}{%
                \includegraphics[width=0.23\linewidth]
                {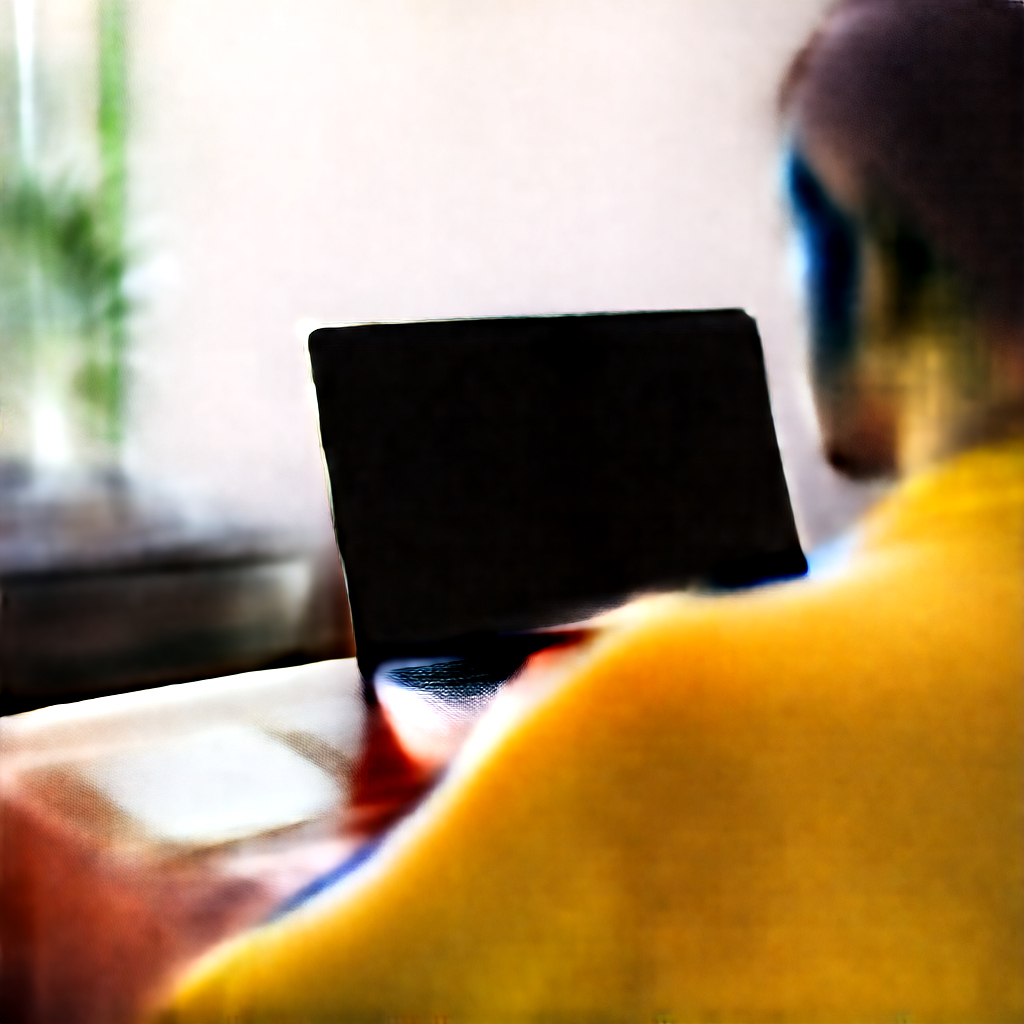}} &
            \fcolorbox{black}{white}{%
                \includegraphics[width=0.23\linewidth]
                {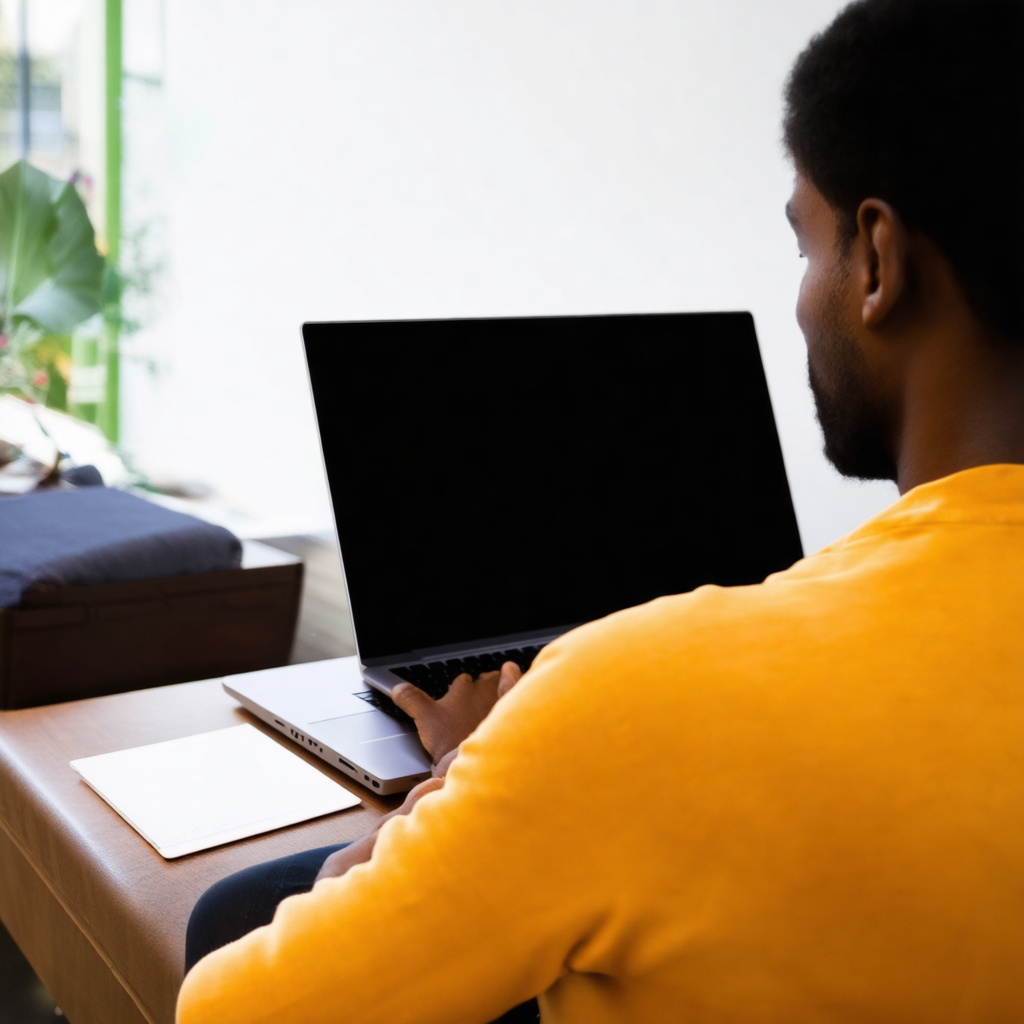}} &
            \fcolorbox{black}{white}{%
                \includegraphics[width=0.23\linewidth]
                {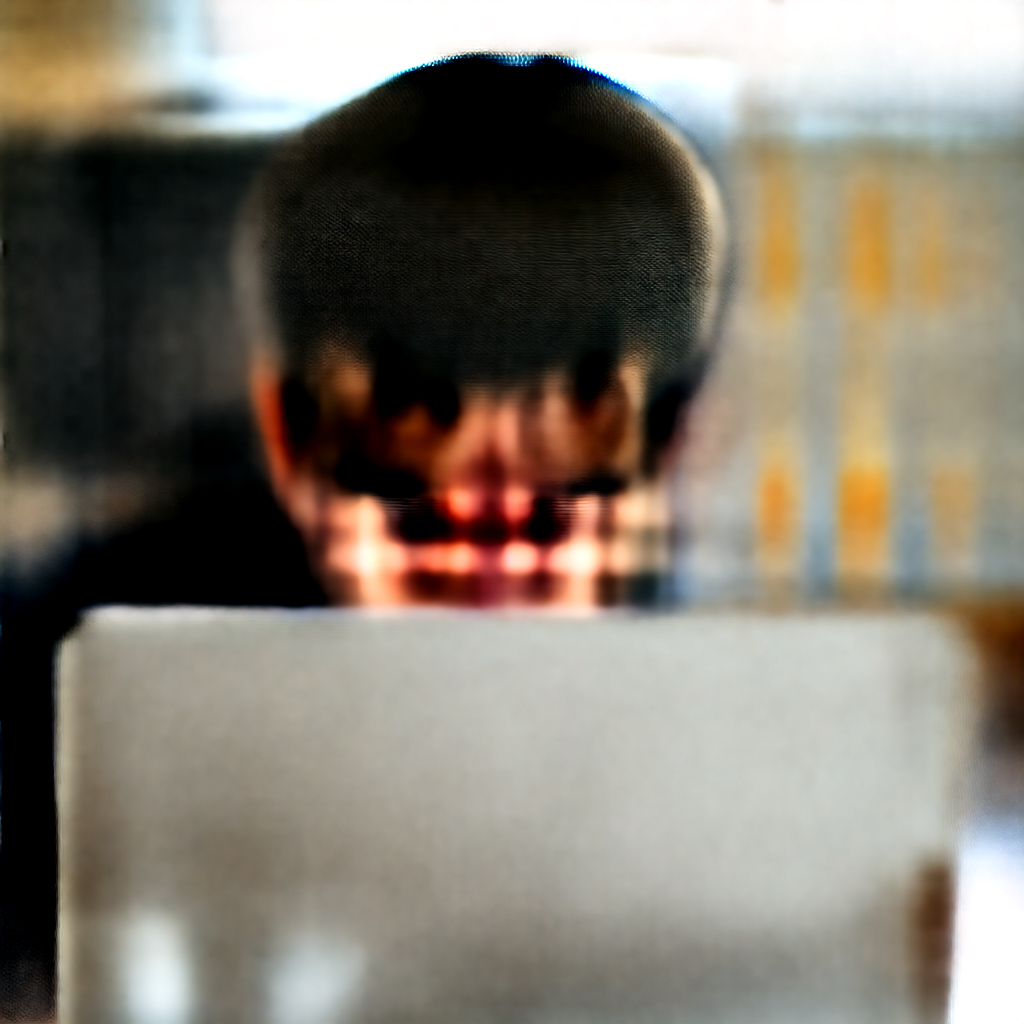}} &
            \fcolorbox{black}{white}{%
                \includegraphics[width=0.23\linewidth]
                {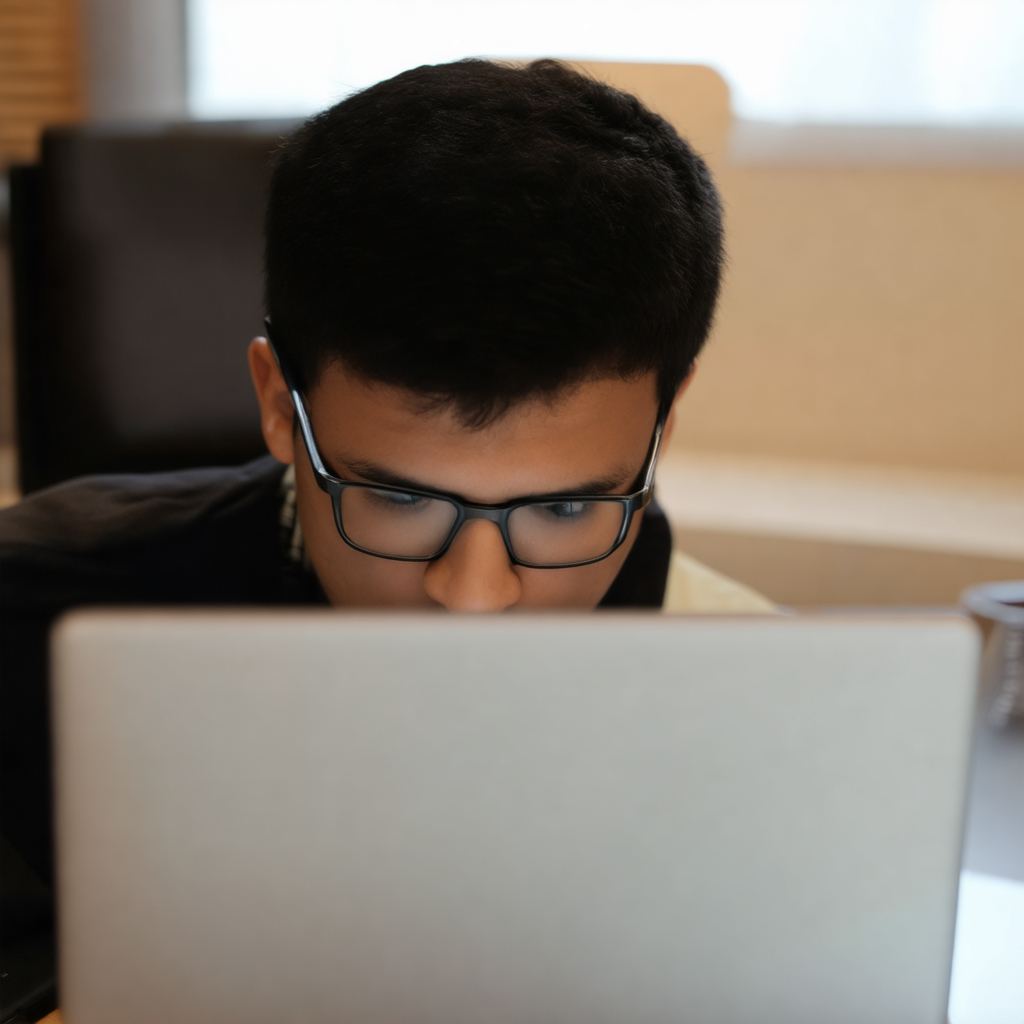}}
        \end{tabular}

        \vspace{4pt}

        \begin{tabular}{@{}p{0.47\linewidth}@{\hspace{0.04\linewidth}}p{0.47\linewidth}@{}}
            \centering\textbf{Seed A} &
            \centering\textbf{Seed B}
        \end{tabular}

        \vspace{3pt}

        \textit{What is the person's pose?}

        \vspace{1pt}

        \begin{tabular}{@{}p{0.47\linewidth}@{\hspace{0.04\linewidth}}p{0.47\linewidth}@{}}
            \centering
            \answerframe{1cm}{Sitting with head turned towards the screen, hands on keyboard.}
            &
            \centering
            \answerframe{1cm}{Looking at a screen with a focused expression.}
        \end{tabular}

        \vspace{4pt}

        \textit{What is the camera angle?}

        \vspace{1pt}

        \begin{tabular}{@{}p{0.47\linewidth}@{\hspace{0.04\linewidth}}p{0.47\linewidth}@{}}
            \centering
            \answerframe{0.65cm}{Back view.}
            &
            \centering
            \answerframe{0.65cm}{Close-up.}
        \end{tabular}

    \end{minipage}

    \vspace{10pt}

    \begin{minipage}[t]{0.485\textwidth}
        \centering

        {\small\textbf{Prompt:} ``A renowned politician.''}

        \vspace{3pt}

        \setlength{\mirrorwidth}{0.235\linewidth}

        \begin{tabular}{@{}c@{\hspace{2pt}}c@{\hspace{2pt}}c@{\hspace{2pt}}c@{}}
            {\scriptsize\textbf{18\%}} &
            {\scriptsize\textbf{100\%}} &
            {\scriptsize\textbf{18\%}} &
            {\scriptsize\textbf{100\%}} \\[2pt]
            \fcolorbox{black}{white}{%
                \includegraphics[width=0.23\linewidth]
                {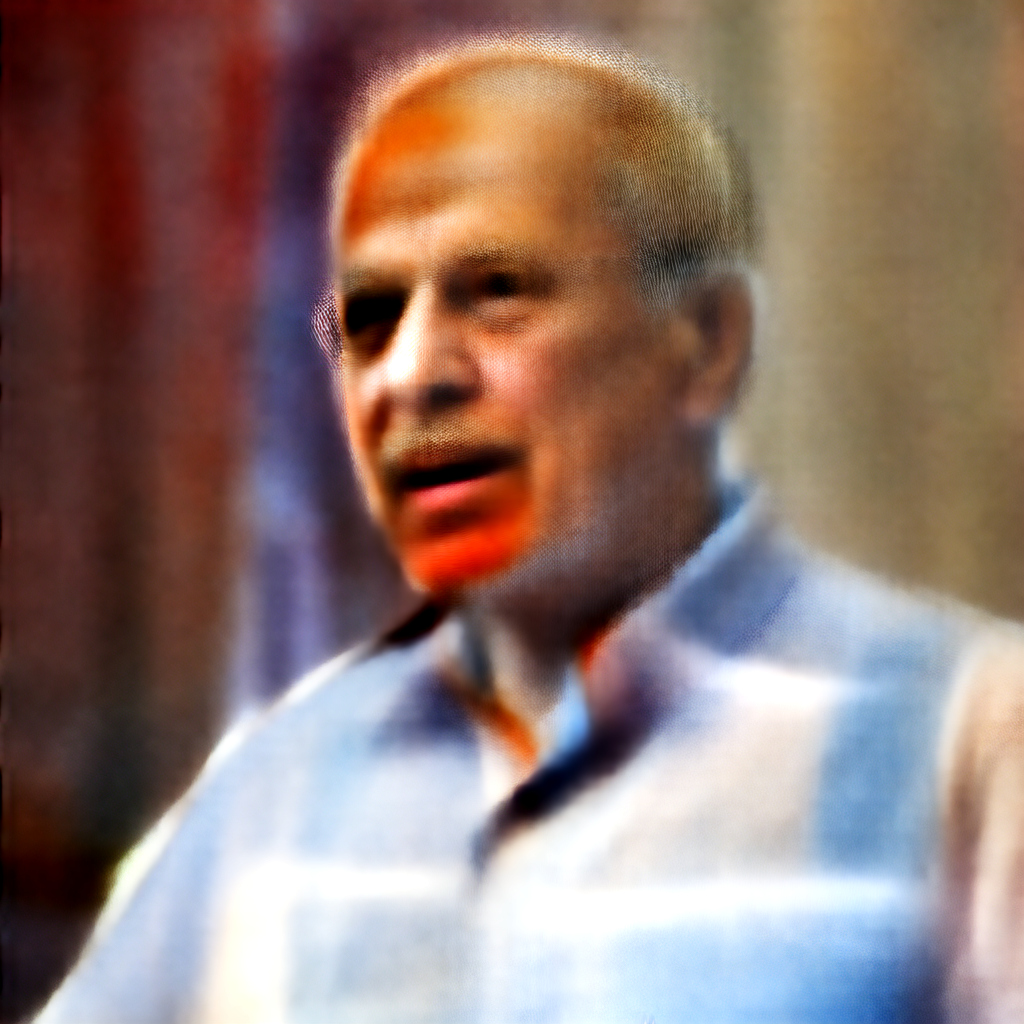}} &
            \fcolorbox{black}{white}{%
                \includegraphics[width=0.23\linewidth]
                {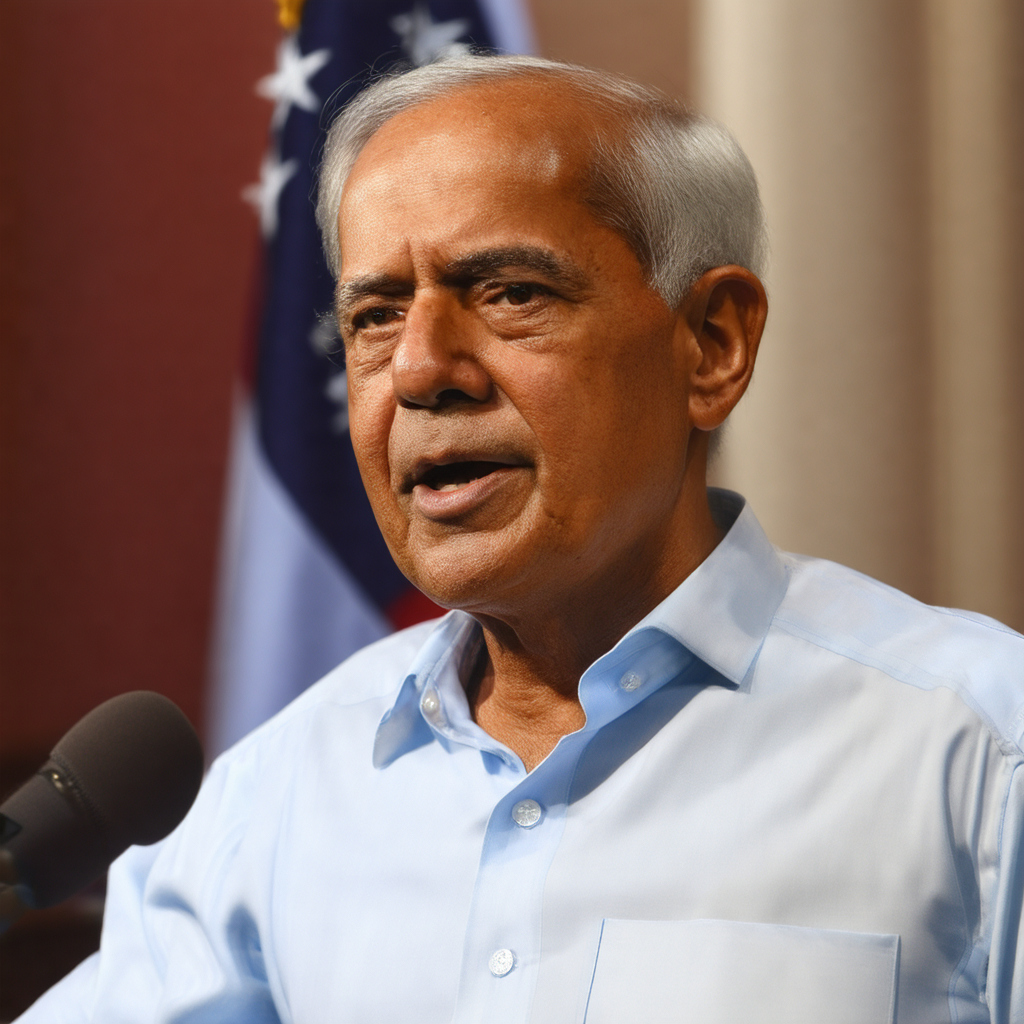}} &
            \fcolorbox{black}{white}{%
                \includegraphics[width=0.23\linewidth]
                {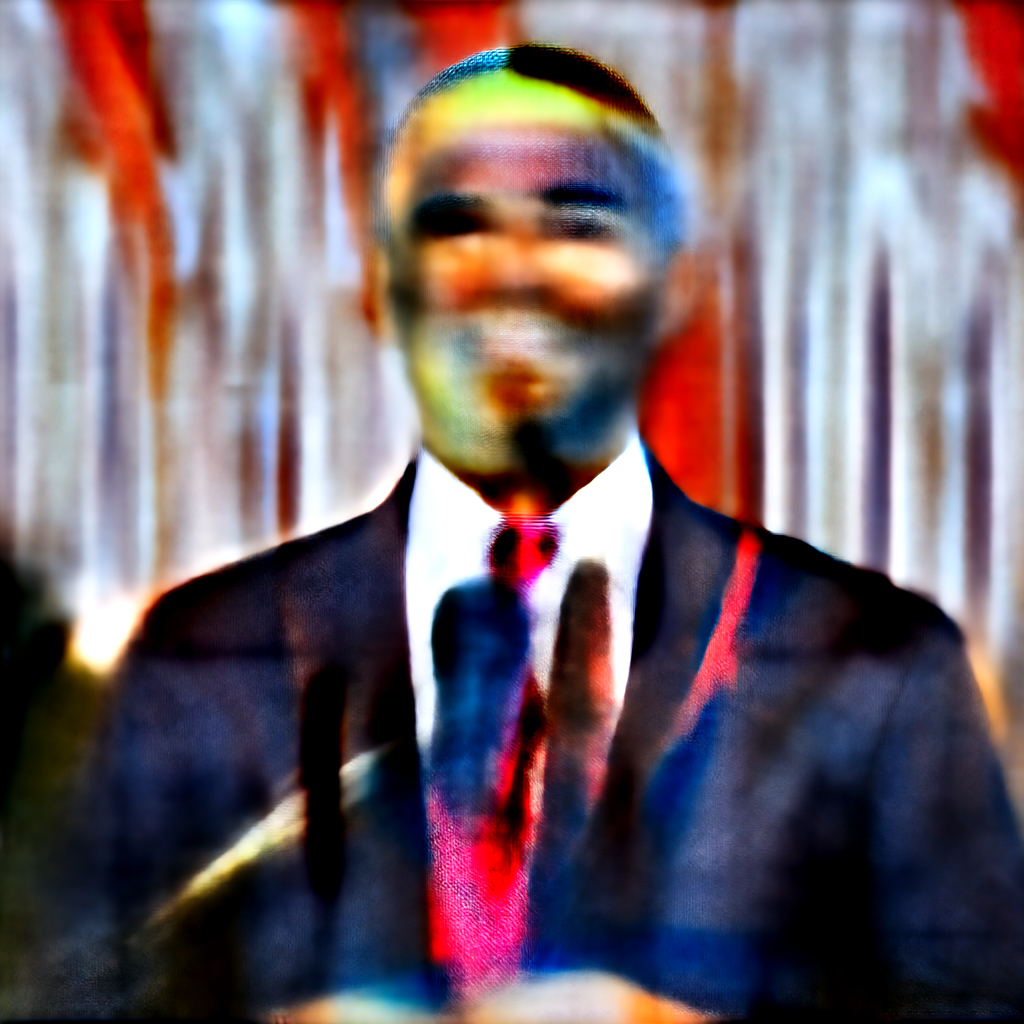}} &
            \fcolorbox{black}{white}{%
                \includegraphics[width=0.23\linewidth]
                {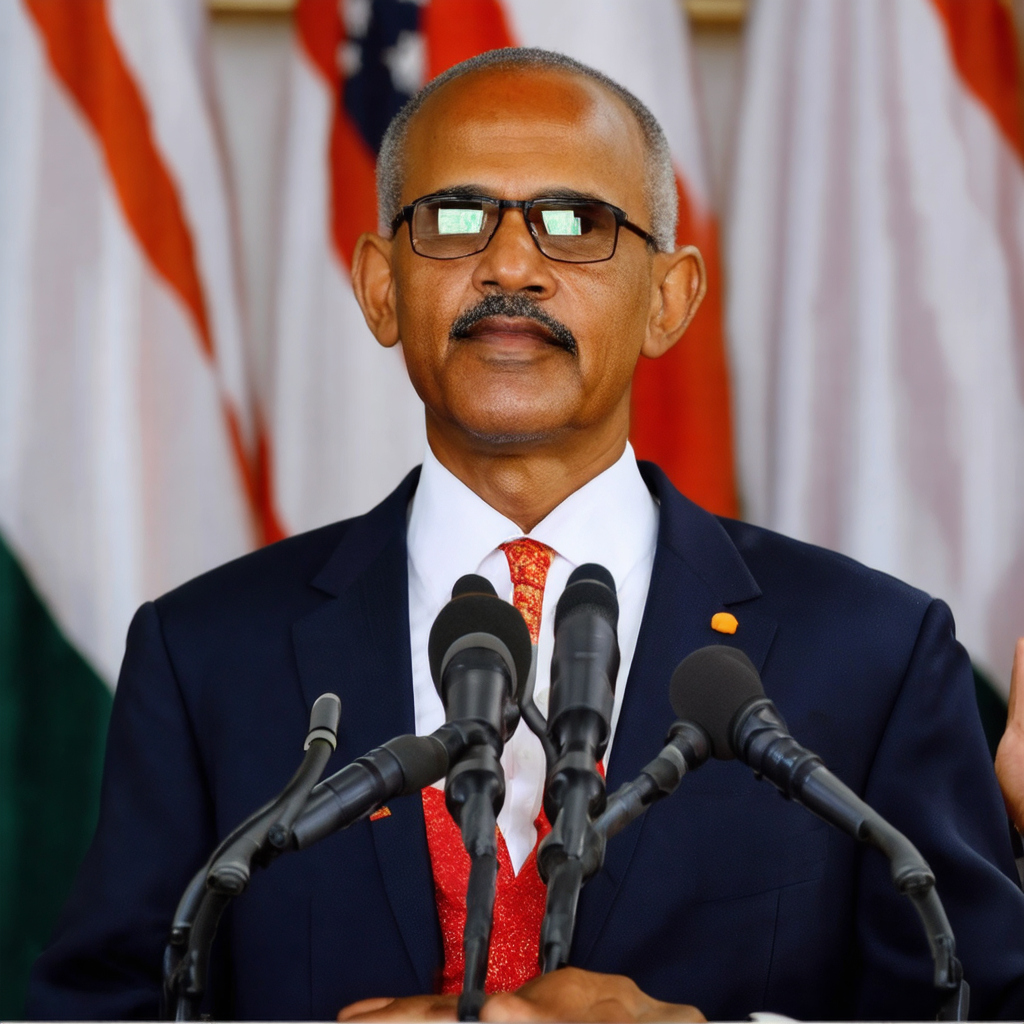}}
        \end{tabular}

        \vspace{4pt}

        \begin{tabular}{@{}p{0.47\linewidth}@{\hspace{0.04\linewidth}}p{0.47\linewidth}@{}}
            \centering\textbf{Seed A} &
            \centering\textbf{Seed B}
        \end{tabular}

        \vspace{3pt}

        \textit{What is the person wearing?}

        \vspace{1pt}

        \begin{tabular}{@{}p{0.47\linewidth}@{\hspace{0.04\linewidth}}p{0.47\linewidth}@{}}
            \centering
            \answerframe{1cm}{Blue shirt.}
            &
            \centering
            \answerframe{1cm}{Blue suit, white shirt, red tie.}
        \end{tabular}

        \vspace{4pt}

        \textit{Which direction is the person facing?}

        \vspace{1pt}

        \begin{tabular}{@{}p{0.47\linewidth}@{\hspace{0.04\linewidth}}p{0.47\linewidth}@{}}
            \centering
            \answerframe{0.65cm}{Turned slightly left.}
            &
            \centering
            \answerframe{0.65cm}{Front view.}
        \end{tabular}

    \end{minipage}
    \hfill
    \begin{minipage}[t]{0.485\textwidth}
        \centering

        {\small\textbf{Prompt:} ``A person standing near a train.''}

        \vspace{3pt}

        \setlength{\mirrorwidth}{0.235\linewidth}

        \begin{tabular}{@{}c@{\hspace{2pt}}c@{\hspace{2pt}}c@{\hspace{2pt}}c@{}}
            {\scriptsize\textbf{18\%}} &
            {\scriptsize\textbf{100\%}} &
            {\scriptsize\textbf{18\%}} &
            {\scriptsize\textbf{100\%}} \\[2pt]
            \fcolorbox{black}{white}{%
                \includegraphics[width=0.23\linewidth]
                {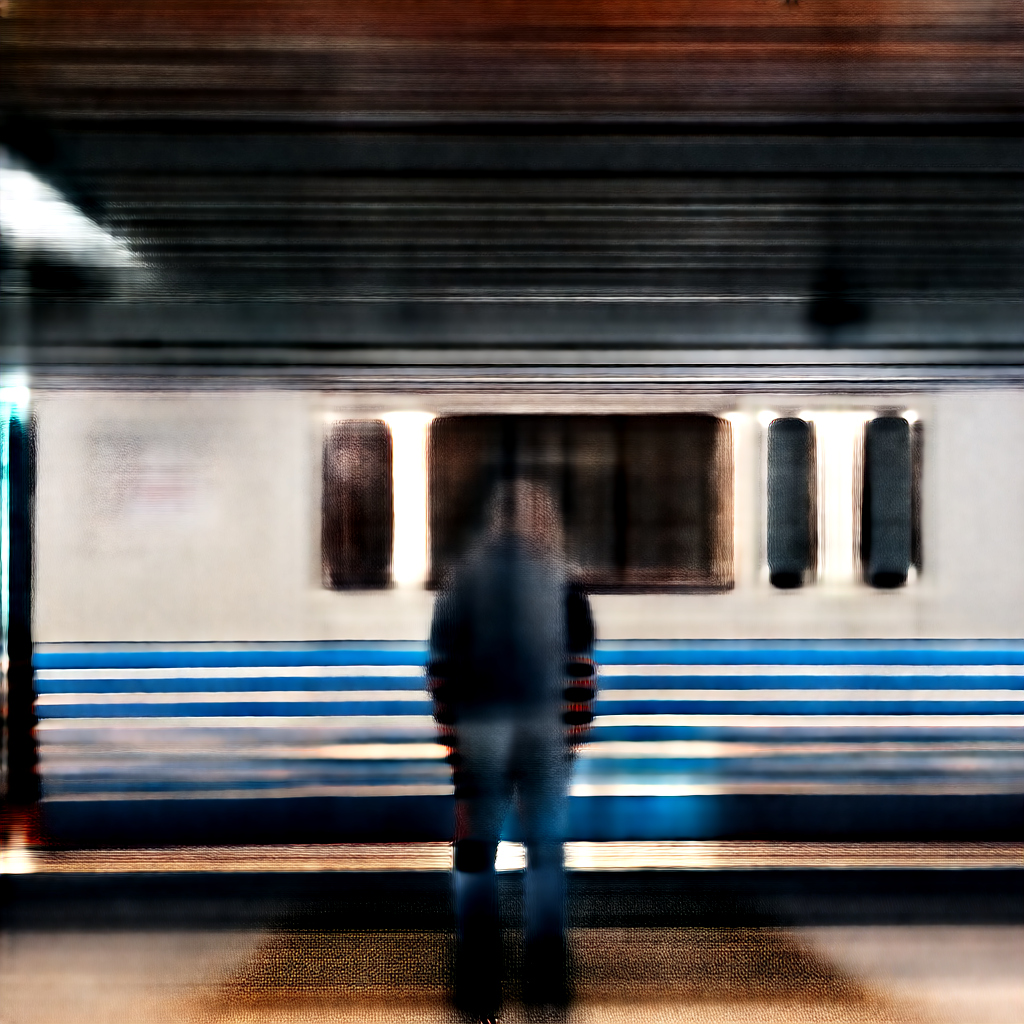}} &
            \fcolorbox{black}{white}{%
                \includegraphics[width=0.23\linewidth]
                {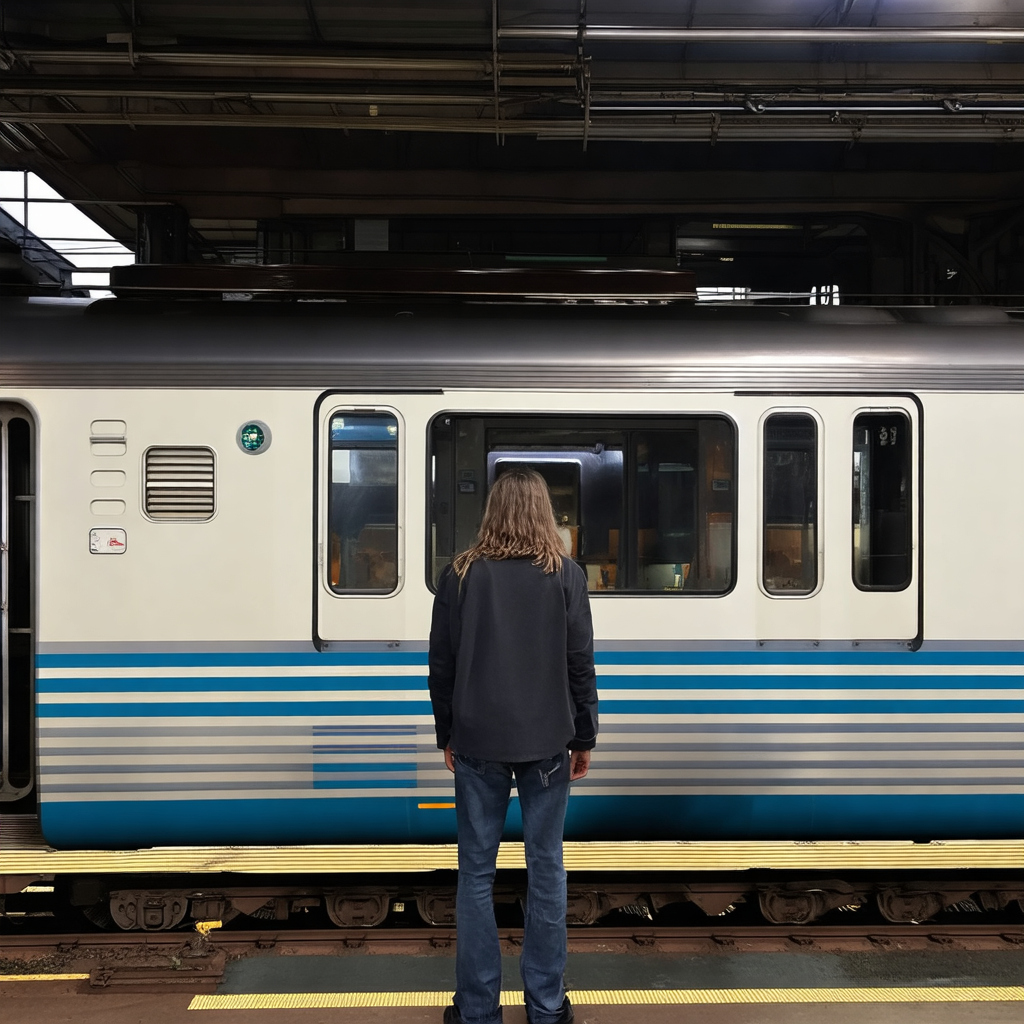}} &
            \fcolorbox{black}{white}{%
                \includegraphics[width=0.23\linewidth]
                {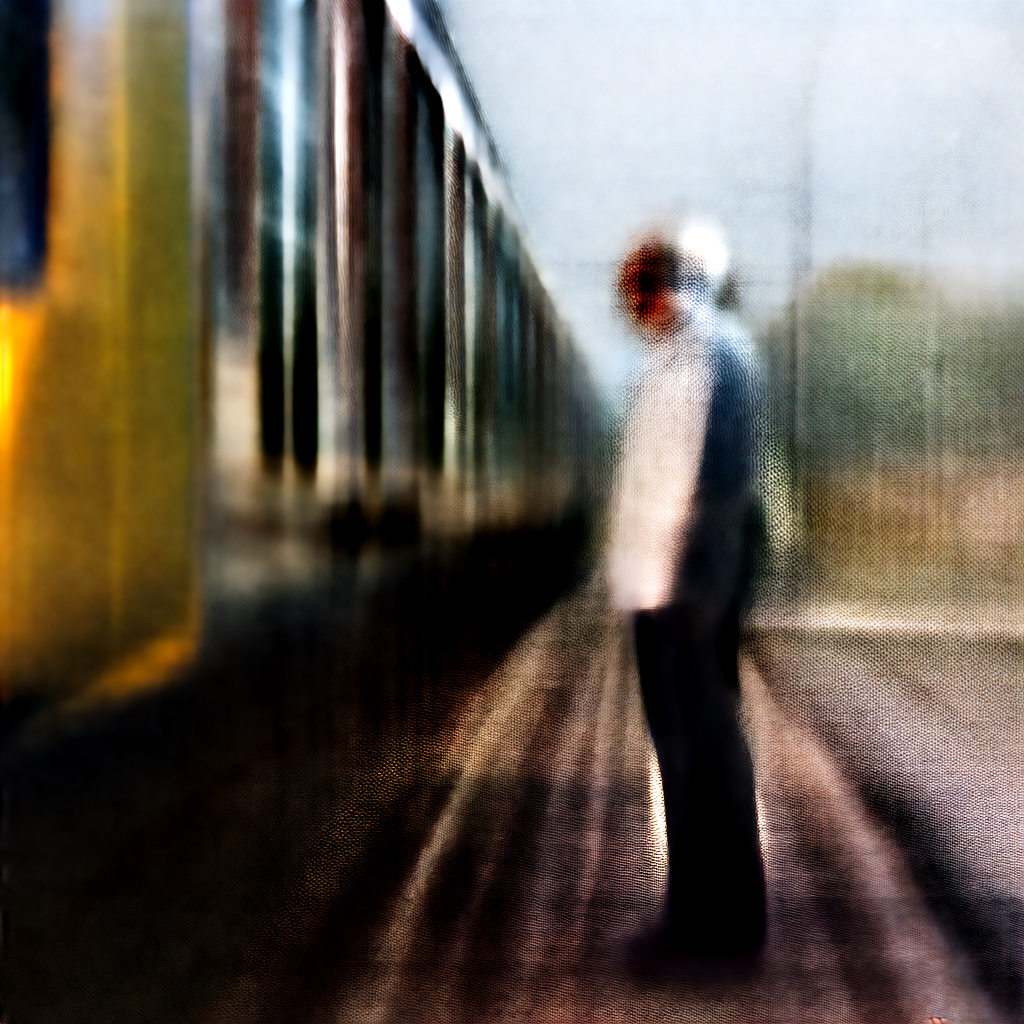}} &
            \fcolorbox{black}{white}{%
                \includegraphics[width=0.23\linewidth]
                {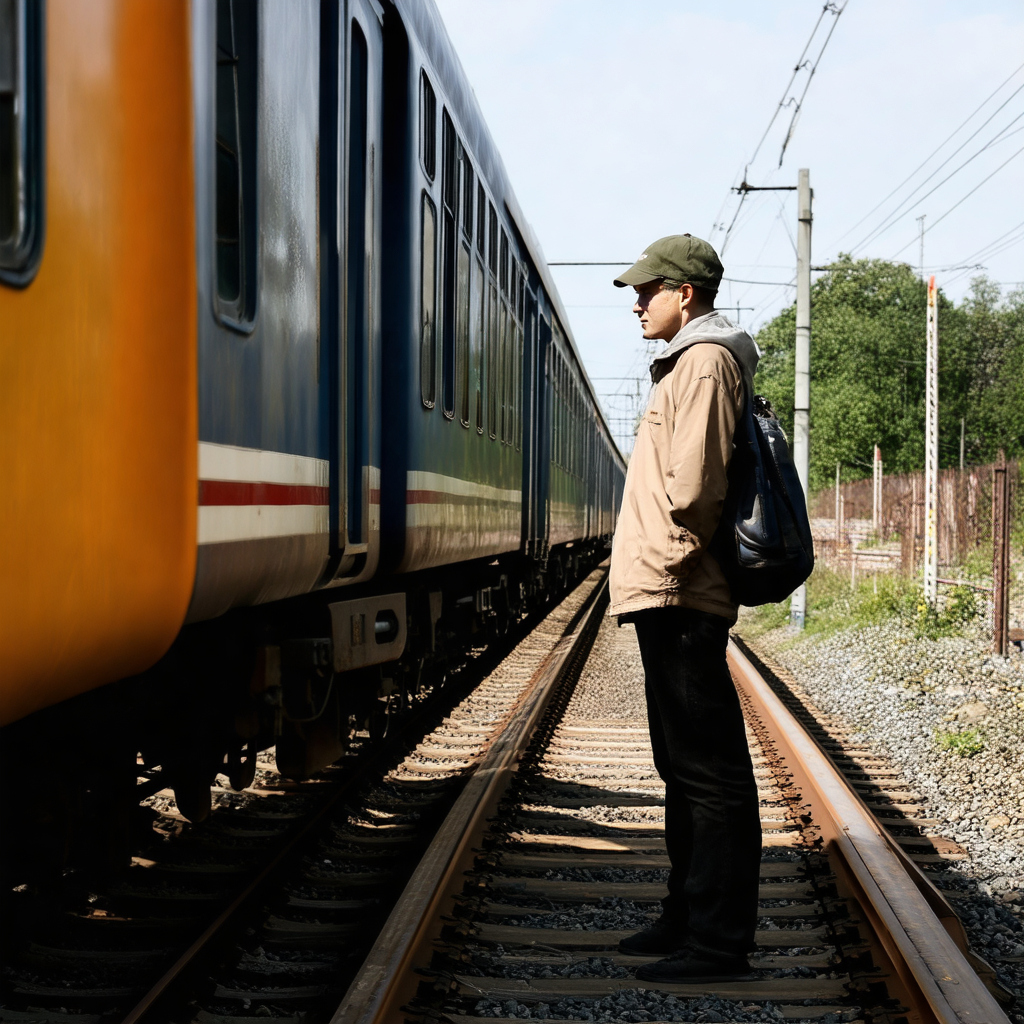}}
        \end{tabular}

        \vspace{4pt}

        \begin{tabular}{@{}p{0.47\linewidth}@{\hspace{0.04\linewidth}}p{0.47\linewidth}@{}}
            \centering\textbf{Seed A} &
            \centering\textbf{Seed B}
        \end{tabular}

        \vspace{3pt}

        \textit{What is the person's pose?}

        \vspace{1pt}

        \begin{tabular}{@{}p{0.47\linewidth}@{\hspace{0.04\linewidth}}p{0.47\linewidth}@{}}
            \centering
            \answerframe{1cm}{Standing with back to the camera, facing the train.}
            &
            \centering
            \answerframe{1cm}{Standing with hands in pocket, looking at the train.}
        \end{tabular}

        \vspace{4pt}

        \textit{What objects are visible?}

        \vspace{1pt}

        \begin{tabular}{@{}p{0.47\linewidth}@{\hspace{0.04\linewidth}}p{0.47\linewidth}@{}}
            \centering
            \answerframe{0.65cm}{Man, train, platform.}
            &
            \centering
            \answerframe{0.65cm}{Train, tracks, road, trees, man, backpack.}
        \end{tabular}

    \end{minipage}

    \caption{
        \textbf{Exemplary plain-text predictions from our conditional Contextual Reader on SD3.5.}
        For each prompt, we compare two random seeds that resolve underspecified
        attributes differently. We show the predicted clean images at $18\%$ of the
        denoising trajectory for both seeds, followed by their final images. The
        reader predictions are obtained from the contextual tokens at the same
        $18\%$ timestep.
    }
    \label{fig:sd35_conditional}
\end{figure*}

\begin{figure*}[t]
    \centering
    \footnotesize

    \setlength{\fboxrule}{1pt}
    \setlength{\fboxsep}{0pt}

    \begin{minipage}[t]{0.49\textwidth}
        \centering

        \setlength{\skiwidth}{0.23\linewidth}

        \begin{tabular}{@{}c@{\hspace{2pt}}c@{\hspace{2pt}}c@{\hspace{2pt}}c@{}}
            \textcolor{tzero}{\textbf{0\%}} &
            \textcolor{tfour}{\textbf{8\%}} &
            \textcolor{tten}{\textbf{20\%}} &
            \textbf{100\%}
            \\[2pt]

            \fcolorbox{tzero}{white}{%
                \includegraphics[width=\skiwidth]
                {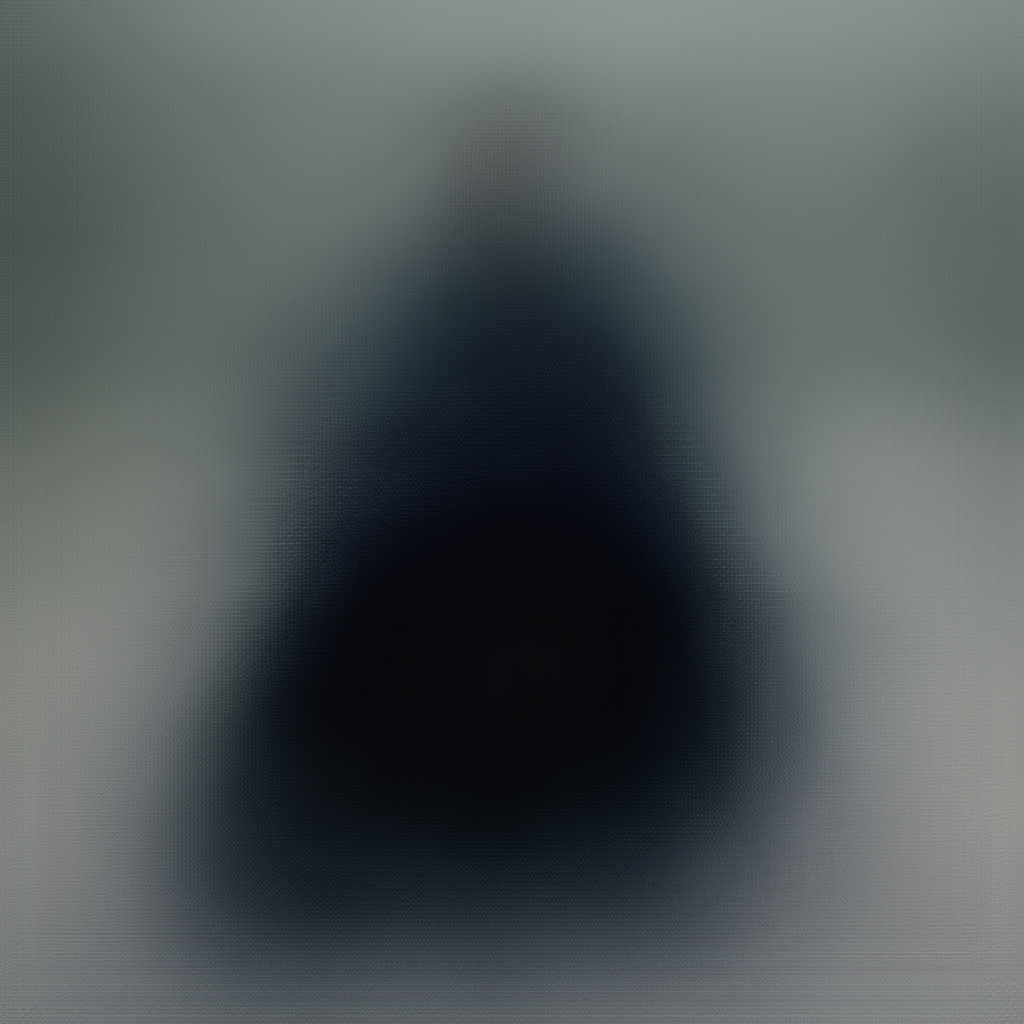}%
            }
            &
            \fcolorbox{tfour}{white}{%
                \includegraphics[width=\skiwidth]
                {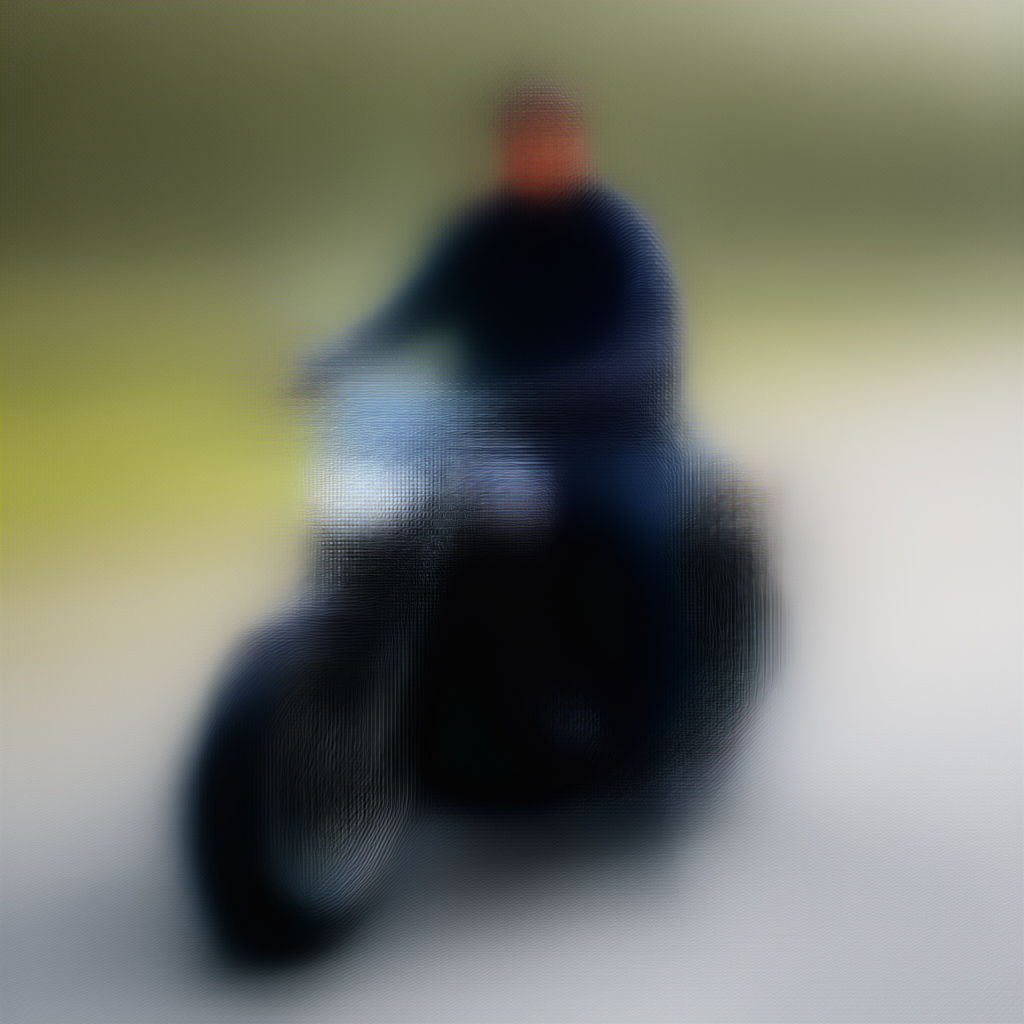}%
            }
            &
            \fcolorbox{tten}{white}{%
                \includegraphics[width=\skiwidth]
                {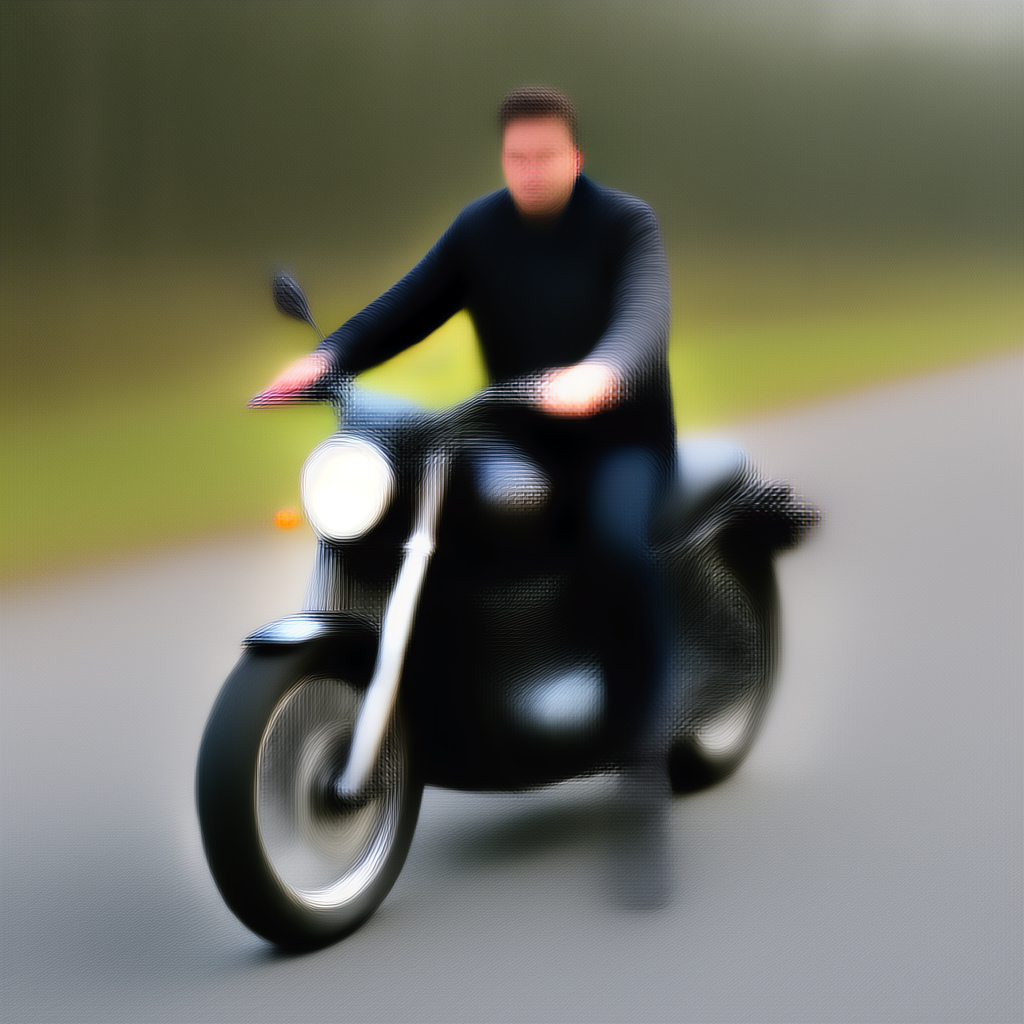}%
            }
            &
            \fcolorbox{black}{white}{%
                \includegraphics[width=\skiwidth]
                {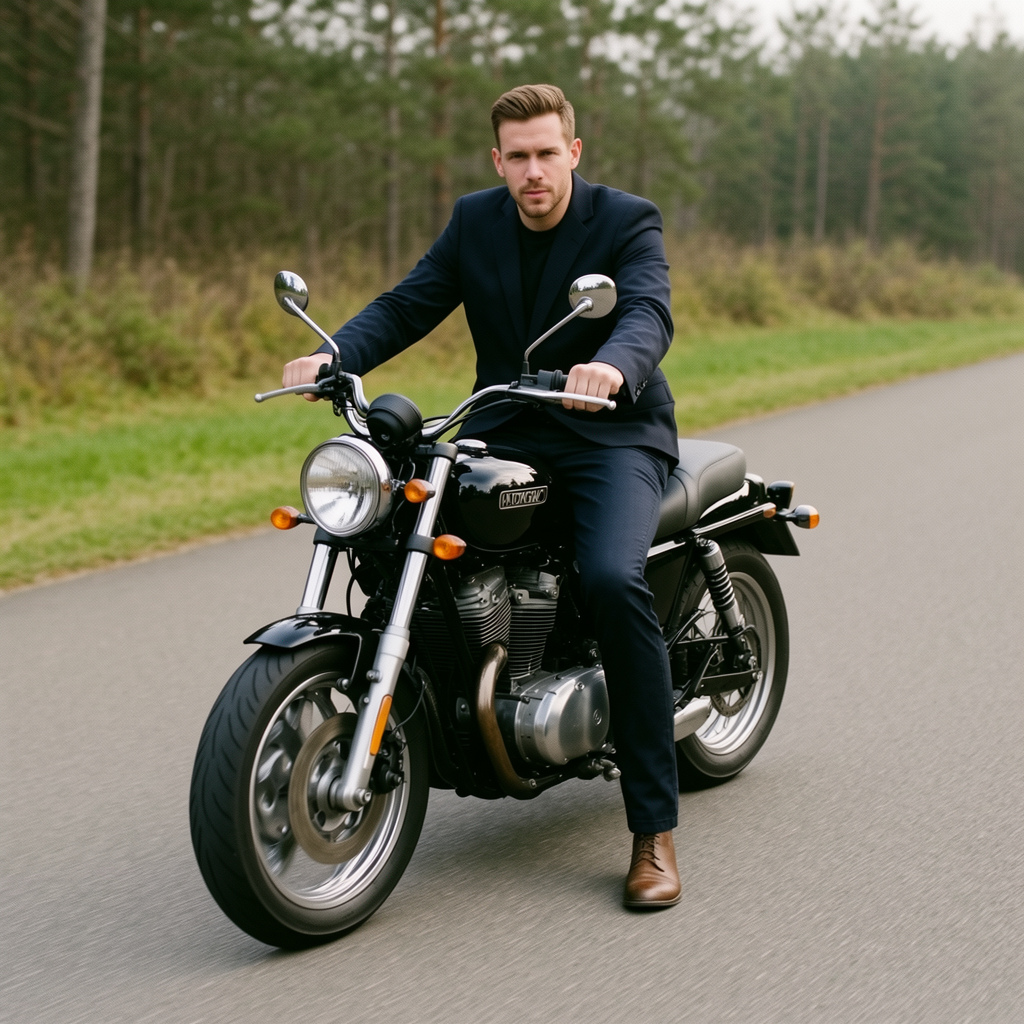}%
            }
        \end{tabular}

    \end{minipage}
    \hfill
    \begin{minipage}[t]{0.49\textwidth}
        \centering
        \setlength{\podiumwidth}{0.23\linewidth}

        \begin{tabular}{@{}c@{\hspace{2pt}}c@{\hspace{2pt}}c@{\hspace{2pt}}c@{}}
            \textcolor{tzero}{\textbf{0\%}} &
            \textcolor{tfour}{\textbf{8\%}} &
            \textcolor{tten}{\textbf{20\%}} &
            \textbf{100\%}
            \\[2pt]

            \fcolorbox{tzero}{white}{%
                \includegraphics[width=\podiumwidth]
                {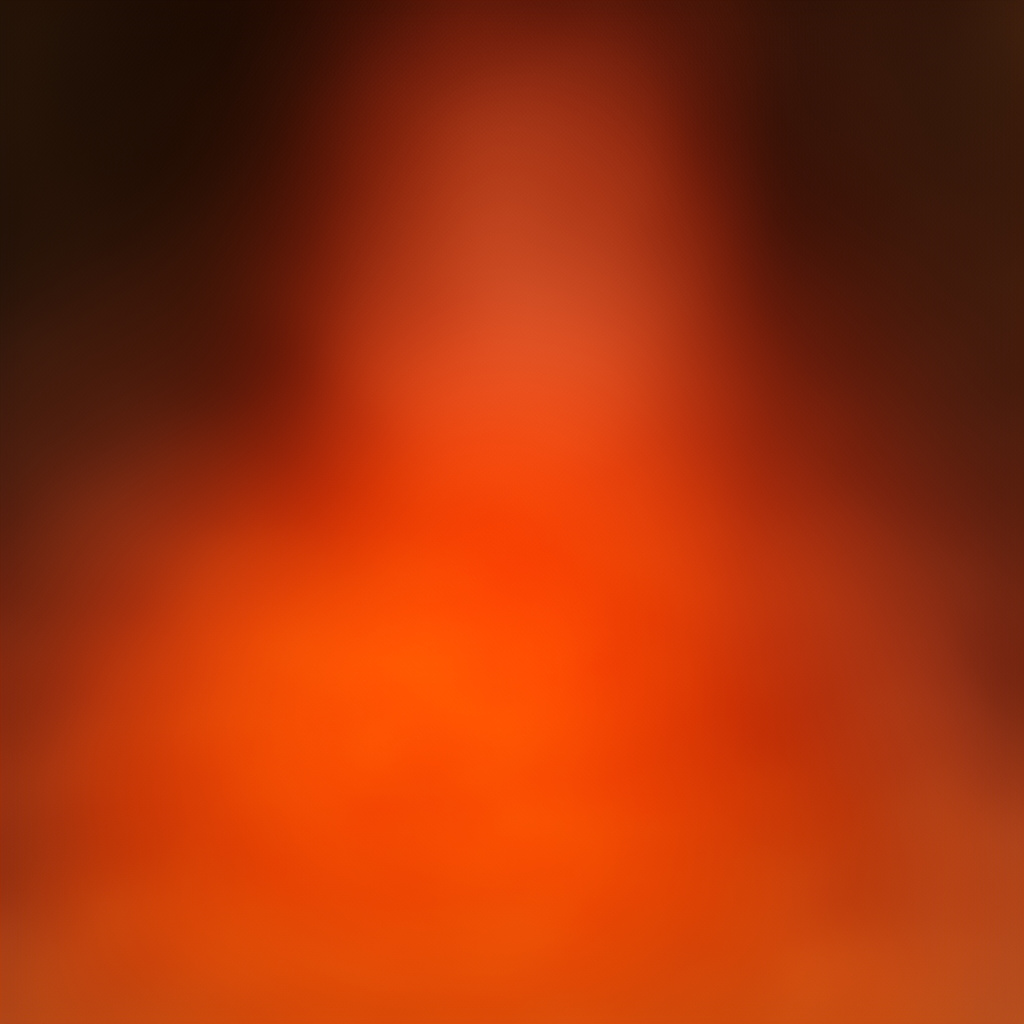}%
            }
            &
            \fcolorbox{tfour}{white}{%
                \includegraphics[width=\podiumwidth]
                {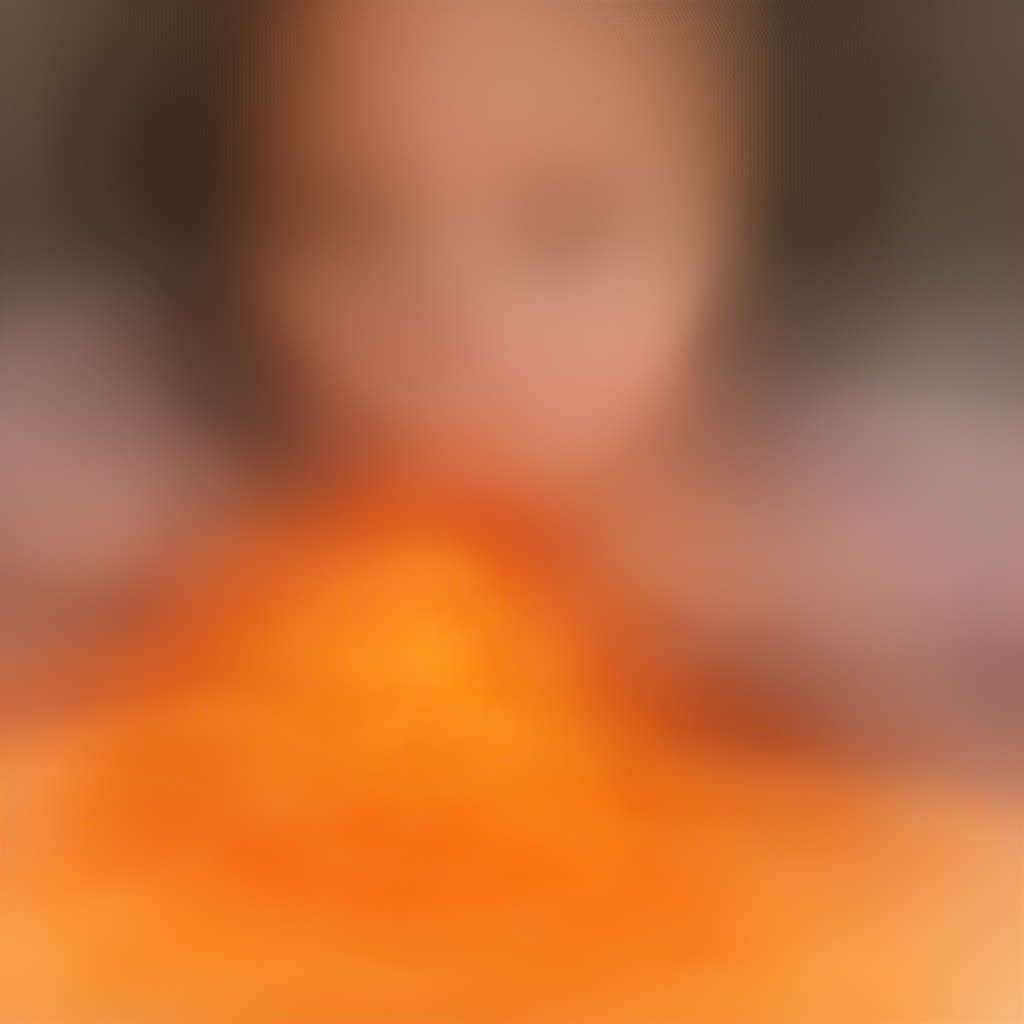}%
            }
            &
            \fcolorbox{tten}{white}{%
                \includegraphics[width=\podiumwidth]
                {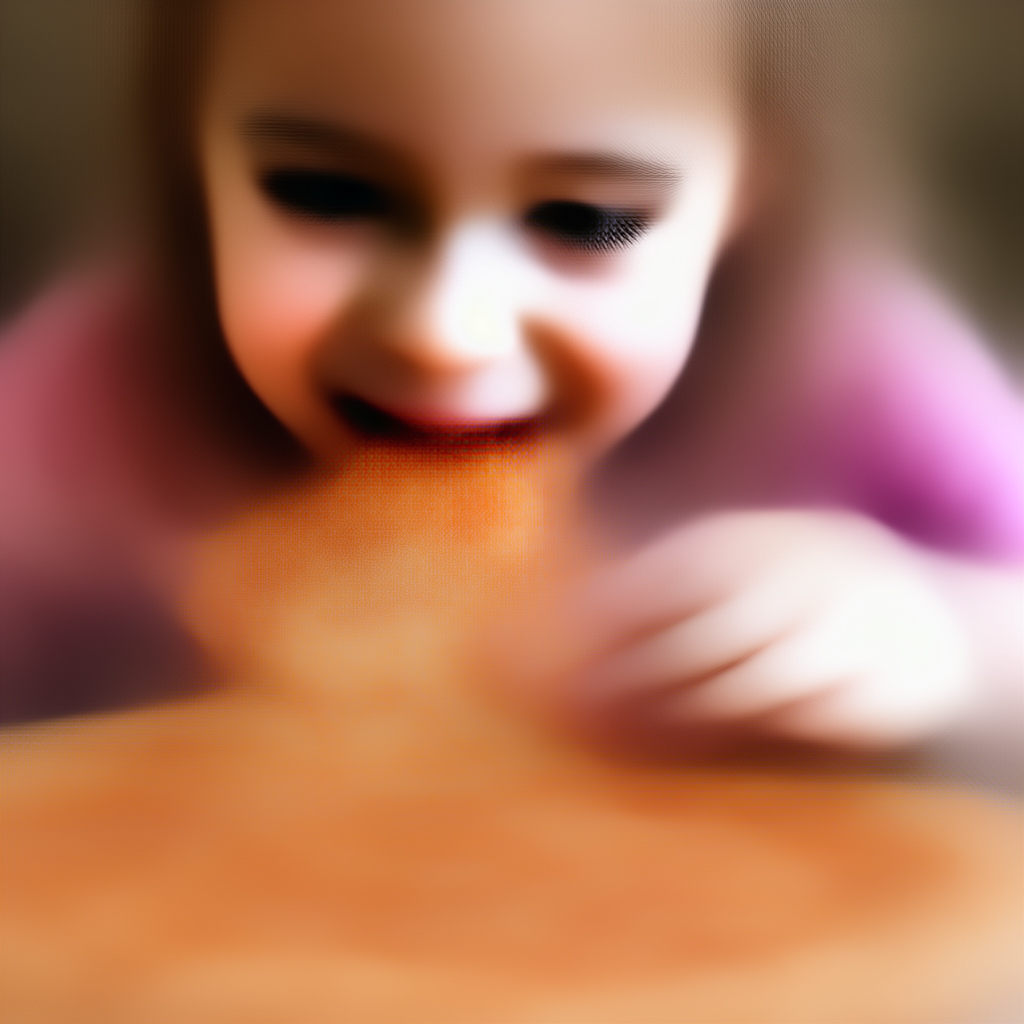}%
            }
            &
            \fcolorbox{black}{white}{%
                \includegraphics[width=\podiumwidth]
                {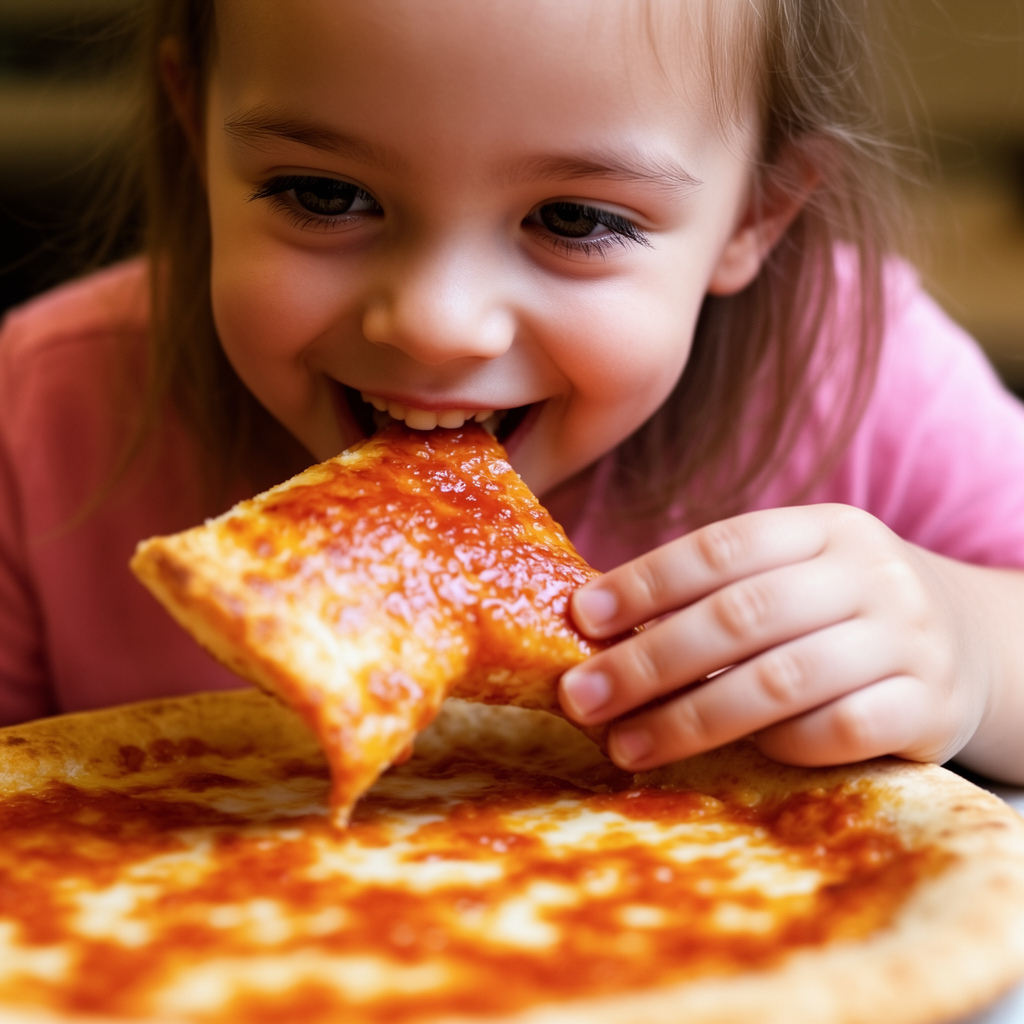}%
            }
        \end{tabular}

    \end{minipage}

    \vspace{3pt}

    \begin{minipage}[t]{0.49\textwidth}
        \centering

        \parbox[t]{\linewidth}{%
            \centering
            \textit{What is the person's pose?}

            \vspace{3pt}

            \answerframe{1.15cm}{%
                \answerbox{tzero}{Sitting on a couch.}
                \hspace{3pt}
                \answerbox{tfour}{Sitting on a motorcycle, holding handlebars with both hands.}
                \hspace{3pt}
                \answerbox{tten}{Sitting on a motorcycle, hands on handlebars, looking forward.}
            }%
        }

        \vspace{6pt}

        \parbox[t]{\linewidth}{%
            \centering
            \textit{What is the person wearing?}

            \vspace{3pt}

            \answerframe{0.95cm}{%
                \answerbox{tzero}{Black t-shirt and blue jeans.}
                \hspace{3pt}
                \answerbox{tfour}{Black jacket, jeans.}
                \hspace{3pt}
                \answerbox{tten}{Black jacket, jeans, and brown boots.}
            }%
        }

        \vspace{6pt}

        \parbox[t]{\linewidth}{%
            \centering
            \textit{What is in the background?}

            \vspace{3pt}

            \answerframe{0.95cm}{%
                \answerbox{tzero}{Trees and sky.}
                \hspace{3pt}
                \answerbox{tfour}{Trees and road.}
                \hspace{3pt}
                \answerbox{tten}{Trees, road, grassy field.}
            }%
        }

    \end{minipage}
    \hfill
    \begin{minipage}[t]{0.49\textwidth}
        \centering

        \parbox[t]{\linewidth}{%
            \centering
            \textit{What is the person doing?}

            \vspace{3pt}

            \answerframe{1.15cm}{%
                \answerbox{tzero}{Sitting on a bed.}
                \hspace{3pt}
                \answerbox{tfour}{Holding a slice of pizza.}
                \hspace{3pt}
                \answerbox{tten}{Smiling and eating pizza.}
            }%
        }

        \vspace{6pt}

        \parbox[t]{\linewidth}{%
            \centering
            \textit{What is the person's apparent age group?}

            \vspace{3pt}

            \answerframe{0.95cm}{%
                \answerbox{tzero}{Young adult.}
                \hspace{3pt}
                \answerbox{tfour}{Child.}
                \hspace{3pt}
                \answerbox{tten}{Child.}
            }%
        }

        \vspace{6pt}

        \parbox[t]{\linewidth}{%
            \centering
            \textit{What is in the camera angle?}

            \vspace{3pt}

            \answerframe{0.95cm}{%
                \answerbox{tzero}{Eye level.}
                \hspace{3pt}
                \answerbox{tfour}{Close-up.}
                \hspace{3pt}
                \answerbox{tten}{Close-up.}
            }%
        }

    \end{minipage}

    \caption{
        \textbf{Additional plain-text predictions from the unconditional
        Contextual Reader on FLUX.2.}
        We show predicted clean images at $0\%$, $8\%$, and
        $20\%$ of the denoising trajectory, together with the final
        generation. Colored highlights associate each intermediate
        timestep with the corresponding reader answer below. Despite
        receiving no text prompt, the reader recovers increasingly
        specific semantic information about the eventual image as
        denoising progresses.
    }

    \label{fig:unconditional_sup}

\end{figure*}

\begin{figure*}[t]
    \centering
    \footnotesize

    \setlength{\fboxrule}{1pt}
    \setlength{\fboxsep}{0pt}

    \begin{minipage}[t]{0.49\textwidth}
        \centering

        \setlength{\skiwidth}{0.23\linewidth}

        \begin{tabular}{@{}c@{\hspace{2pt}}c@{\hspace{2pt}}c@{\hspace{2pt}}c@{}}
            \textcolor{tzero}{\textbf{0\%}} &
            \textcolor{tfour}{\textbf{7\%}} &
            \textcolor{tten}{\textbf{18\%}} &
            \textbf{100\%}
            \\[2pt]

            \fcolorbox{tzero}{white}{%
                \includegraphics[width=\skiwidth]
                {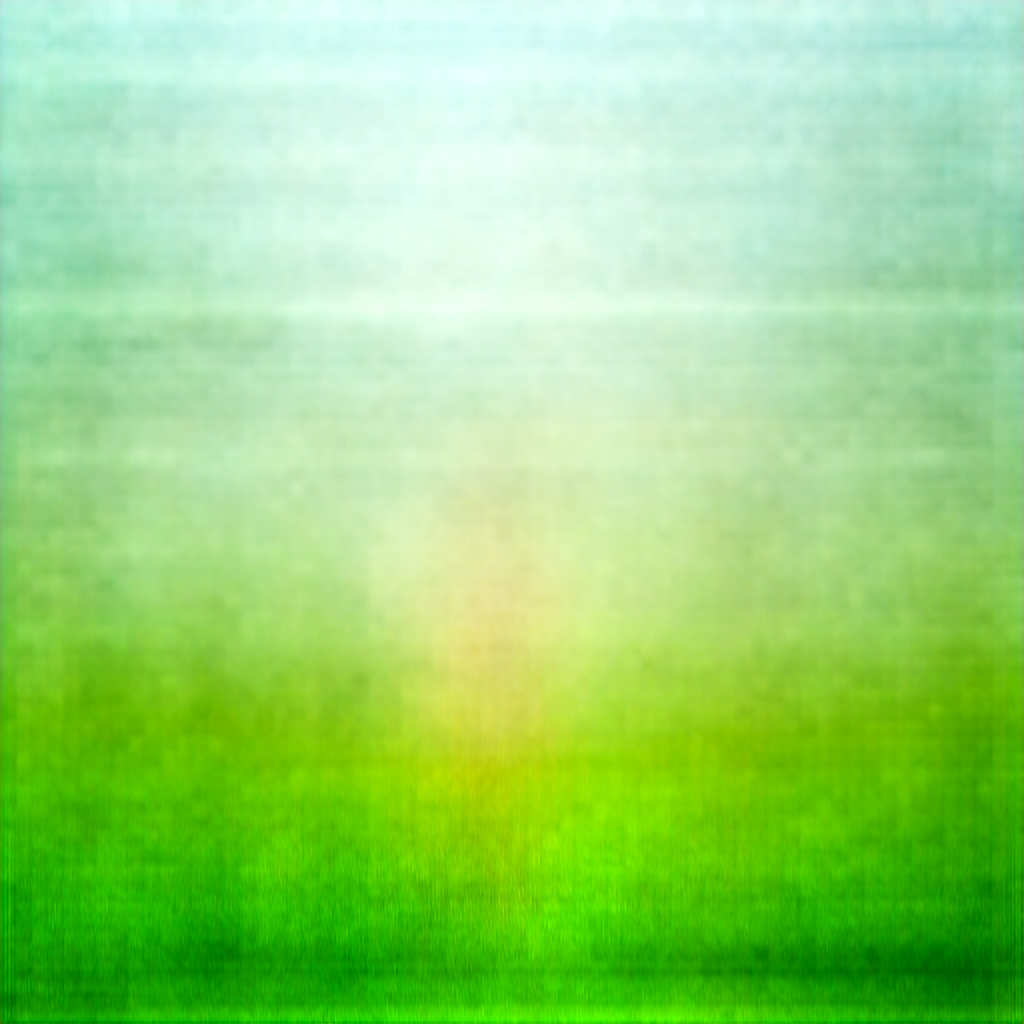}%
            }
            &
            \fcolorbox{tfour}{white}{%
                \includegraphics[width=\skiwidth]
                {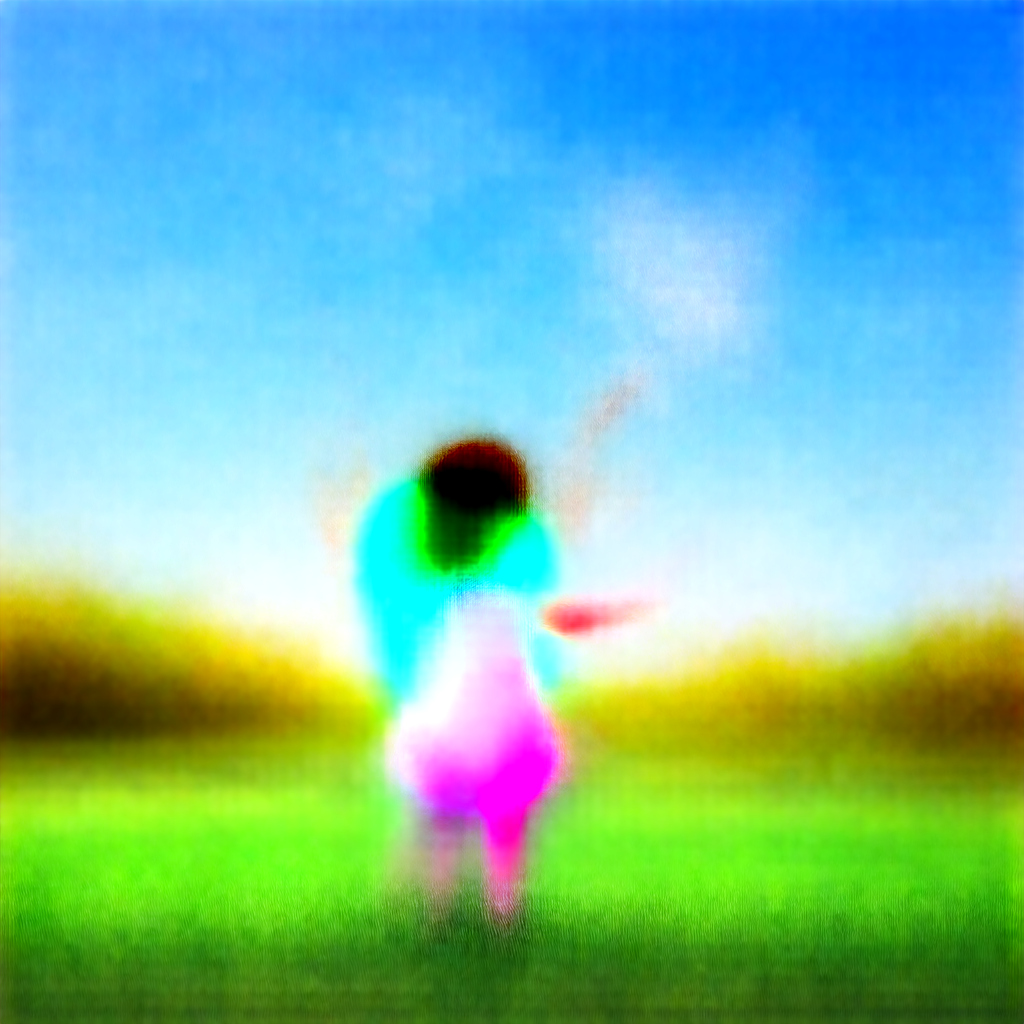}%
            }
            &
            \fcolorbox{tten}{white}{%
                \includegraphics[width=\skiwidth]
                {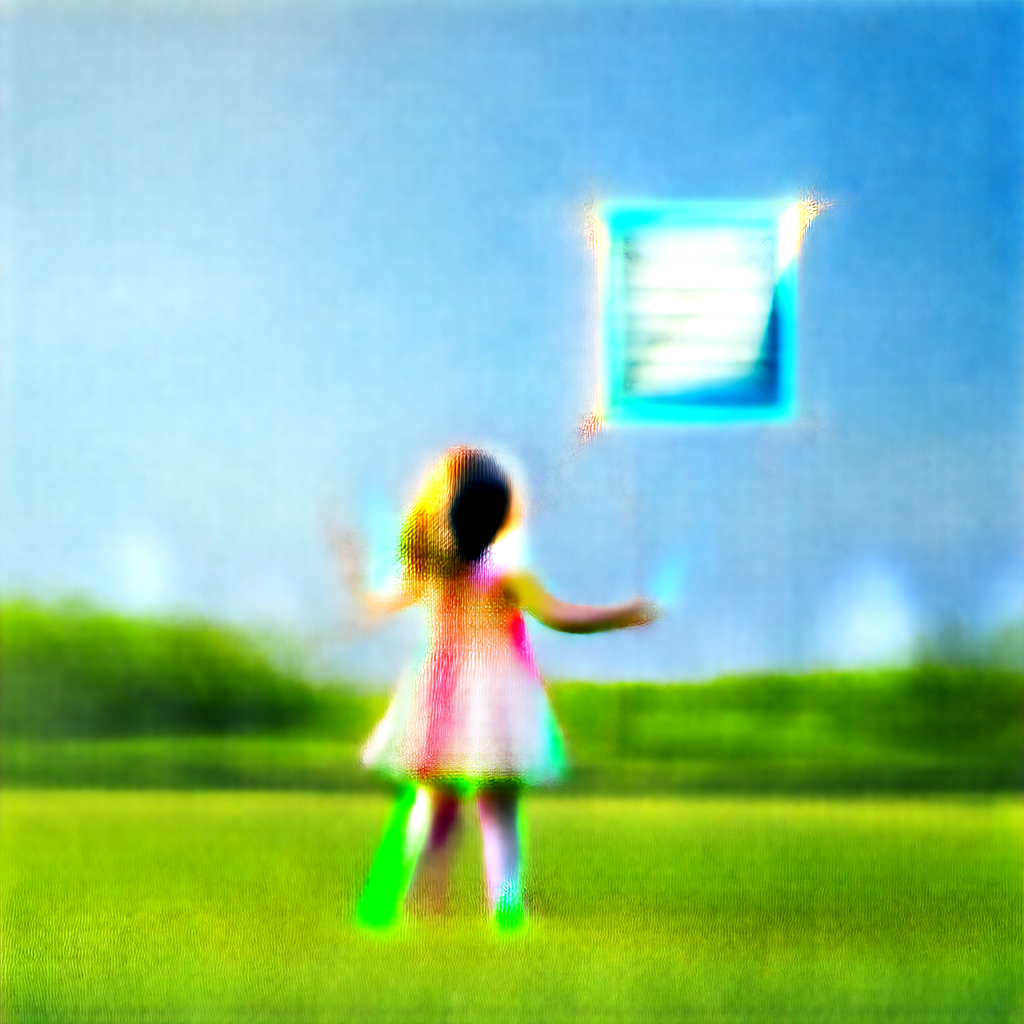}%
            }
            &
            \fcolorbox{black}{white}{%
                \includegraphics[width=\skiwidth]
                {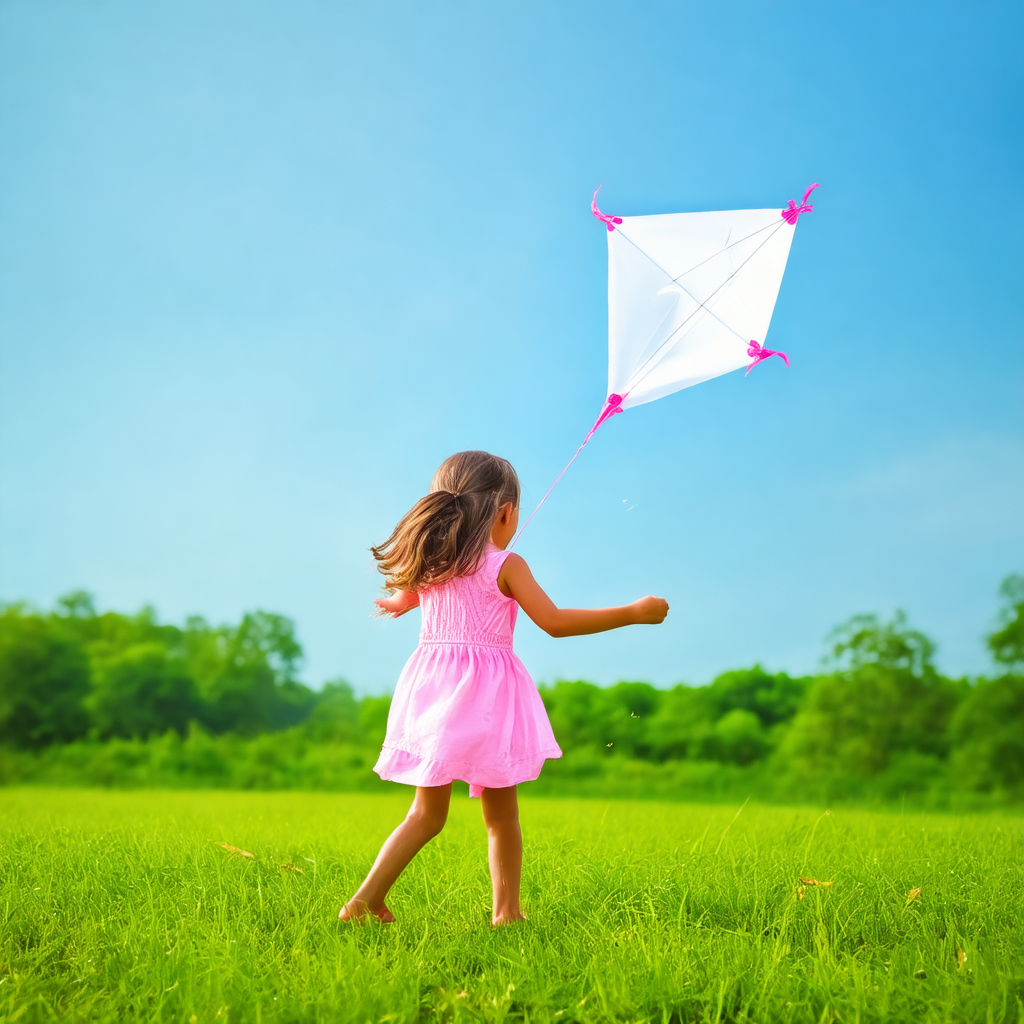}%
            }
        \end{tabular}

    \end{minipage}
    \hfill
    \begin{minipage}[t]{0.49\textwidth}
        \centering
        \setlength{\podiumwidth}{0.23\linewidth}

        \begin{tabular}{@{}c@{\hspace{2pt}}c@{\hspace{2pt}}c@{\hspace{2pt}}c@{}}
            \textcolor{tzero}{\textbf{0\%}} &
            \textcolor{tfour}{\textbf{7\%}} &
            \textcolor{tten}{\textbf{18\%}} &
            \textbf{100\%}
            \\[2pt]

            \fcolorbox{tzero}{white}{%
                \includegraphics[width=\podiumwidth]
                {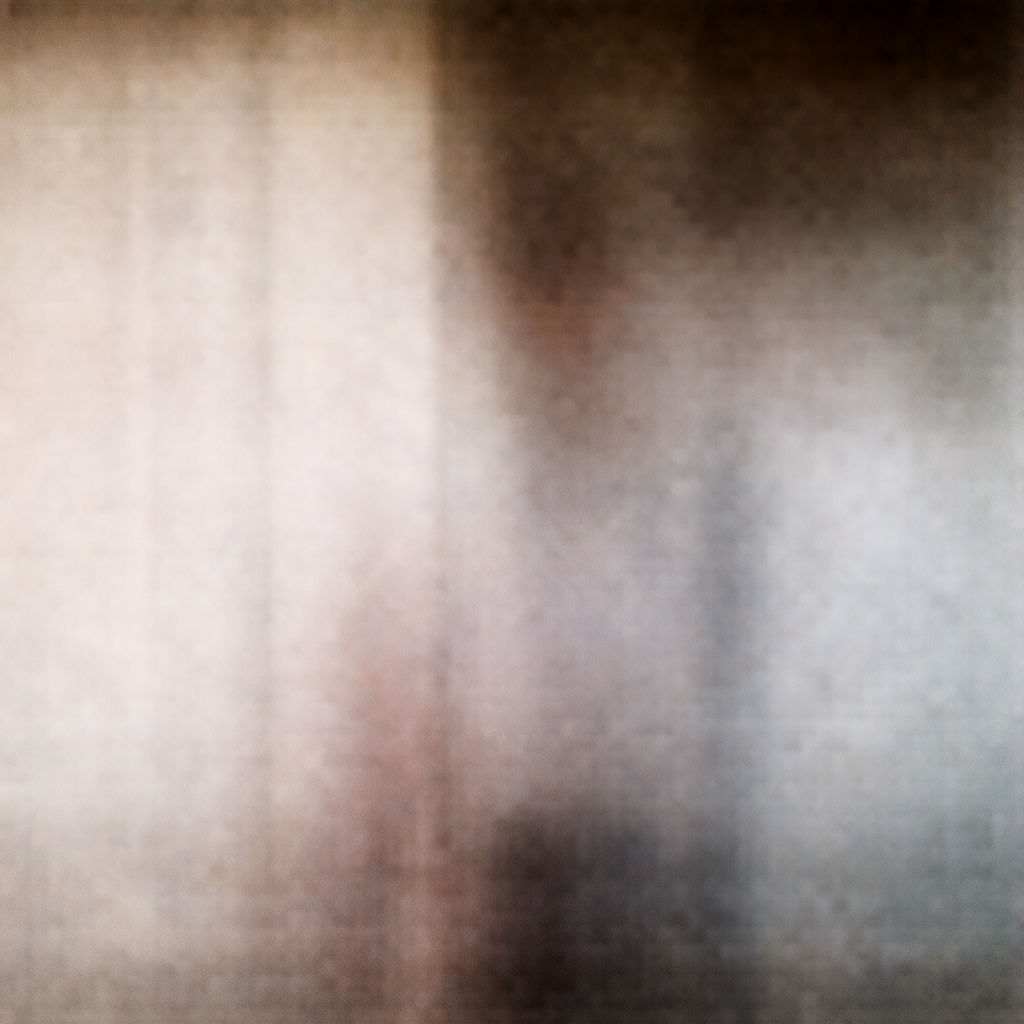}%
            }
            &
            \fcolorbox{tfour}{white}{%
                \includegraphics[width=\podiumwidth]
                {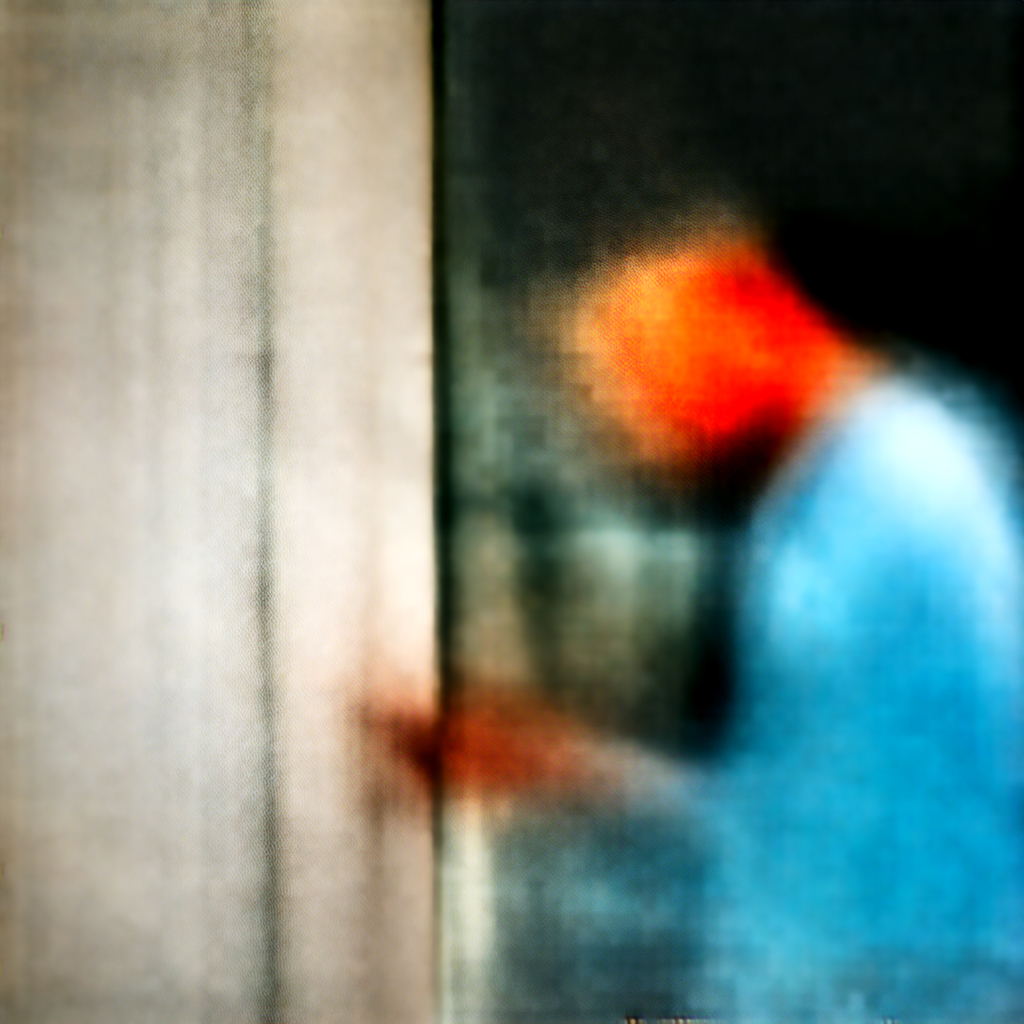}%
            }
            &
            \fcolorbox{tten}{white}{%
                \includegraphics[width=\podiumwidth]
                {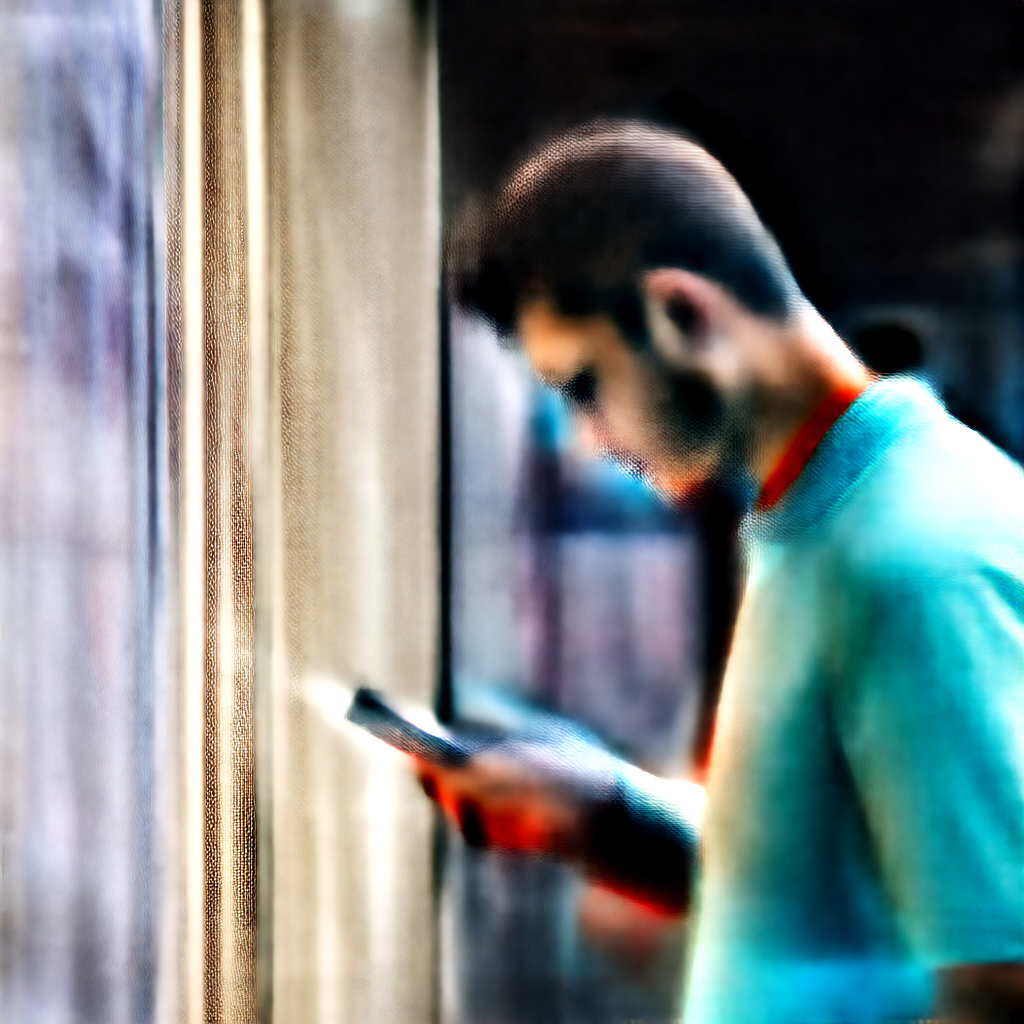}%
            }
            &
            \fcolorbox{black}{white}{%
                \includegraphics[width=\podiumwidth]
                {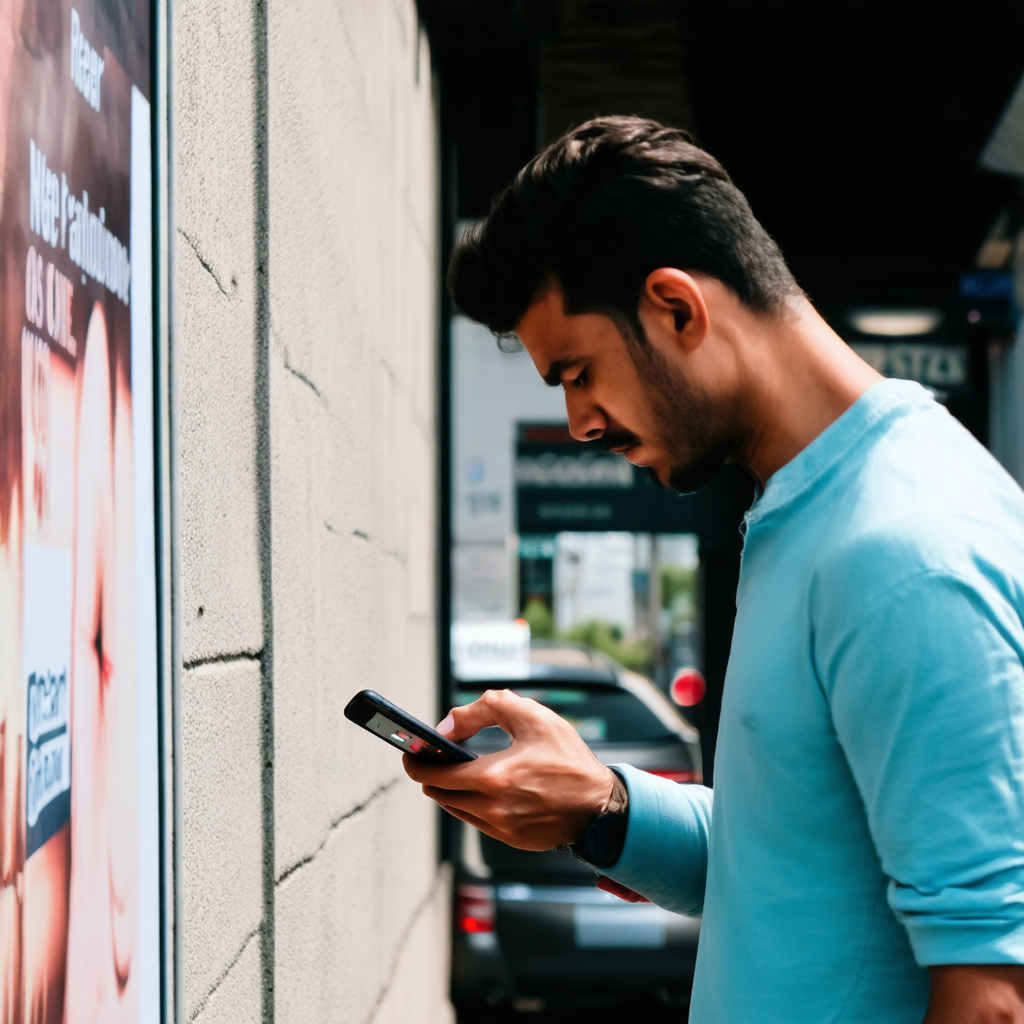}%
            }
        \end{tabular}

    \end{minipage}

    \vspace{3pt}

    \begin{minipage}[t]{0.49\textwidth}
        \centering

        \parbox[t]{\linewidth}{%
            \centering
            \textit{What is the person's hair like?}

            \vspace{3pt}

            \answerframe{0.95cm}{%
                \answerbox{tzero}{Short and dark.}
                \hspace{3pt}
                \answerbox{tfour}{Long and dark.}
                \hspace{3pt}
                \answerbox{tten}{Long and tied back.}
            }%
        }

        \vspace{6pt}

        \parbox[t]{\linewidth}{%
            \centering
            \textit{What is the person doing?}

            \vspace{3pt}

            \answerframe{0.95cm}{%
                \answerbox{tzero}{Holding a fishing rod.}
                \hspace{3pt}
                \answerbox{tfour}{Holding a kite string.}
                \hspace{3pt}
                \answerbox{tten}{Flying a kite.}
            }%
        }

        \vspace{6pt}

        \parbox[t]{\linewidth}{%
            \centering
            \textit{What is the person's apparent age group?}

            \vspace{3pt}

            \answerframe{0.35cm}{%
                \answerbox{tzero}{Young adult.}
                \hspace{3pt}
                \answerbox{tfour}{Child.}
                \hspace{3pt}
                \answerbox{tten}{Child.}
            }%
        }

    \end{minipage}
    \hfill
    \begin{minipage}[t]{0.49\textwidth}
        \centering

        \parbox[t]{\linewidth}{%
            \centering
            \textit{What is the person doing?}

            \vspace{3pt}

            \answerframe{0.95cm}{%
                \answerbox{tzero}{Sitting on a bench.}
                \hspace{3pt}
                \answerbox{tfour}{Holding a smartphone.}
                \hspace{3pt}
                \answerbox{tten}{Looking at a smartphone.}
            }%
        }

        \vspace{6pt}

        \parbox[t]{\linewidth}{%
            \centering
            \textit{Which parts of the person are visible?}

            \vspace{3pt}

            \answerframe{0.95cm}{%
                \answerbox{tzero}{Torso, arms.}
                \hspace{3pt}
                \answerbox{tfour}{Head, torso, arms.}
                \hspace{3pt}
                \answerbox{tten}{Head, torso, arms, hands.}
            }%
        }

        \vspace{6pt}

        \parbox[t]{\linewidth}{%
            \centering
            \textit{What is in the camera angle?}

            \vspace{3pt}

            \answerframe{0.35cm}{%
                \answerbox{tzero}{Eye level.}
                \hspace{3pt}
                \answerbox{tfour}{Side view.}
                \hspace{3pt}
                \answerbox{tten}{Side view.}
            }%
        }

    \end{minipage}

    \caption{
        \textbf{Exemplary plain-text predictions from the unconditional
        Contextual Reader on SD3.5.}
        We show predicted clean images at $0\%$, $7\%$, and
        $18\%$ of the denoising trajectory, together with the final
        generation. Colored highlights associate each intermediate
        timestep with the corresponding reader answer below. Despite
        receiving no text prompt, the reader recovers increasingly
        specific semantic information about the eventual image as
        denoising progresses.
    }

    \label{fig:sd35_unconditional}

\end{figure*}

\paragraph{Qualitative analysis.}
We provide additional qualitative results from our Contextual Reader in
Figures~\ref{fig:conditional_sup} and~\ref{fig:unconditional_sup} for FLUX.2,
and in Figures~\ref{fig:sd35_conditional} and~\ref{fig:sd35_unconditional}
for SD3.5.
For the conditional setting, we repeat the experiment in which we generate two
images from the same prompt, where underspecified attributes are realized
differently. As shown in Figures~\ref{fig:conditional_sup} and~\ref{fig:sd35_conditional},
the reader can recover these seed-specific differences from the contextual
tokens at early denoising timesteps. For the unconditional setting, we again
examine how the reader's predictions evolve as denoising progresses.
Figures~\ref{fig:unconditional_sup} and~\ref{fig:sd35_unconditional} show that
the reader first recovers coarse semantic properties and progressively provides
more fine-grained answers at later timesteps. Altogether, these results
demonstrate that contextual tokens can capture image-specific information
through their interaction with the visual tokens.

\begin{table}[t]
\centering
\caption{Ablation study conducted in both the full model training setting ($150k$ steps from random initialization on MS-COCO) and the fine-tuning setting ($30k$ steps on the Fine-T2I curated subset, starting from SD3). In full model training, Ours refers to REPA + \methodname{}; in fine-tuning, it refers to vanilla flow matching + \methodname{}. Bold/underline mark the best/second-best value per column within each setting.}
\label{tab:ablations}
\vspace{2mm}
\small
\setlength{\tabcolsep}{3.5pt}
\begin{tabular}{lccccccccc}
\toprule
 & \multicolumn{5}{c}{Full model training} & \multicolumn{3}{c}{Fine-tuning} \\
\cmidrule(lr){2-6} \cmidrule(lr){7-9}
Variant                       & FID$\downarrow$ & CLIP{\scriptsize$\times10^{2}$}$\uparrow$ & HPS{\scriptsize$\times10^{2}$}$\uparrow$ & Prec.$\uparrow$ & Rec.$\uparrow$ & FID$\downarrow$ & Prec.$\uparrow$ & Rec.$\uparrow$ \\
\midrule
Semantic Teacher: CLIP                 & 4.71 & 24.16 & 21.30 & 0.599 & 0.375 & 16.87 & \underline{0.994} & \underline{0.970} \\
Semantic Teacher: DINOv2                & 4.67 & 23.96 & 21.03 & 0.604 & \underline{0.392} & 16.92 & 0.993 & 0.964 \\
Semantic Teacher: Qwen3-VL-2B          & 5.03 & 24.09 & 21.23 & 0.588 & 0.352 & \underline{16.81} & 0.991 & 0.969 \\
\midrule
Contextual Layer: L3                     & 4.54 & 24.20 & 21.30 & 0.607 & 0.388 & 17.04 & 0.992 & 0.968 \\
Contextual Layer: L5                     & 4.78 & 24.18 & 21.23 & 0.597 & 0.364 & 16.97 & \underline{0.994} & 0.970 \\
Contextual Layer: L12                    & 5.30 & 23.99 & 21.05 & 0.575 & 0.318 & 17.27 & 0.991 & 0.968 \\
Contextual Layer: L20                    & 5.19 & 24.04 & 21.12 & 0.566 & 0.315 & 17.70 & 0.986 & 0.960 \\
\midrule
w/o Null-Condition Weighting & \underline{4.44} & \underline{24.31} & \underline{21.45} & \underline{0.617} & 0.389 & 17.60 & \textbf{0.994} & \textbf{0.970} \\
\midrule
Ours & \textbf{4.29} & \textbf{24.40} & \textbf{21.60} & \textbf{0.628} & \textbf{0.407} & \textbf{16.75} & 0.993 & 0.970 \\
\bottomrule
\end{tabular}
\end{table}

\subsection{Contextual Alignment}
\subsubsection{Ablation Studies}
\label{subsec:coal_ablations}

To validate our design choices, we ablate the semantic teacher, the contextual layer at which alignment is applied, and the null-condition weighting scheme, in both the full model training and fine-tuning settings (Table~\ref{tab:ablations}). Across both settings, \methodname{} is fairly robust to the choice of teacher, though SigLIP consistently edges out CLIP, DINOv2, and Qwen3-VL-2B. Aligning earlier contextual layers also consistently outperforms aligning later ones, likely because doing so gives more of the network the opportunity to make use of the well-aligned representation as it propagates forward. Finally, null-condition weighting proves especially important in the fine-tuning setting: since a pre-trained model already aligns reasonably well with the semantic teacher whenever the prompt is present, upweighting the loss contributes most on the text-dropped iterations, where that shortcut disappears.

\subsubsection{Extended Quantitative Comparisons}
\label{subsec:detailed_quantitative}
\begin{table}[t]
\centering
\caption{Extended quantitative comparison against baselines in the full model training setting ($150k$ steps from random initialization on MS-COCO 2014). Baseline rows: \textbf{bold}/\underline{underline} mark best/second-best among the baselines. \methodname{} rows: value colored \textcolor{ForestGreen}{green}/\textcolor{red}{red} if it improves/regresses on that metric relative to its own non-\methodname{} counterpart, with the percent change in parentheses.}
\label{tab:mscoco_supp_full}
\vspace{2mm}
\footnotesize
\setlength{\tabcolsep}{3pt}
\begin{tabular}{lccccccc}
\toprule
Method & FID$\downarrow$ & KID{\scriptsize$\times10^{3}$}$\downarrow$ & CLIP{\scriptsize$\times10^{2}$}$\uparrow$ & VQA$\uparrow$ & HPS{\scriptsize$\times10^{2}$}$\uparrow$ & Prec.$\uparrow$ & Rec.$\uparrow$ \\
\midrule
Vanilla & 6.52 & 1.53 & 23.27 & 0.630 & 20.10 & 0.496 & 0.206 \\
REG            & 7.33 & 2.10 & 23.79 & 0.613 & 20.81 & 0.486 & 0.268 \\
SRA            & 5.89 & 1.23 & 23.49 & 0.637 & 20.37 & 0.515 & 0.261 \\
HASTE          & 4.87 & 0.93 & 24.21 & 0.691 & 21.42 & 0.609 & 0.361 \\
REPA           & 4.95 & 0.94 & 24.20 & 0.691 & 21.38 & 0.591 & 0.338 \\
\midrule
Vanilla + Ours & \good{6.23}{-4.4\%} & \good{1.52}{-0.9\%} & \good{23.30}{+0.1\%} & \bad{0.629}{-0.2\%} & \good{20.17}{+0.3\%} & \good{0.515}{+3.7\%} & \good{0.274}{+33.2\%} \\
SRA + Ours     & \good{4.36}{-25.9\%} & \textbf{\good{0.76}{-38.7\%}} & \good{24.28}{+3.4\%} & \good{0.688}{+8.1\%} & \good{21.42}{+5.1\%} & \good{0.616}{+19.5\%} & \good{0.398}{+52.5\%} \\
HASTE + Ours   & \good{4.44}{-8.8\%} & \good{0.79}{-15.1\%} & \underline{\good{24.32}{+0.4\%}} & \textbf{\good{0.696}{+0.8\%}} & \underline{\good{21.56}{+0.7\%}} & \underline{\good{0.623}{+2.3\%}} & \underline{\good{0.400}{+10.6\%}} \\
REPA + Ours    & \textbf{\good{4.29}{-13.4\%}} & \underline{\good{0.76}{-19.4\%}} & \textbf{\good{24.40}{+0.8\%}} & \underline{\good{0.694}{+0.4\%}} & \textbf{\good{21.60}{+1.0\%}} & \textbf{\good{0.628}{+6.2\%}} & \textbf{\good{0.407}{+20.5\%}} \\
\bottomrule
\end{tabular}
\end{table}

\begin{table}[t]
\centering
\caption{Extended quantitative comparison against baselines on the Fine-T2I curated benchmark (val5k, 30k fine-tuning steps starting from SD3-Medium). Reference refers to the real Fine-T2I validation set images, and Base Model is the pretrained SD3-Medium checkpoint before fine-tuning. \textbf{Bold}/\underline{underline} mark the best/second-best value per column across all rows. \methodname{} rows are additionally colored \textcolor{ForestGreen}{green}/\textcolor{red}{red} if they improve/regress on that metric relative to their own non-\methodname{} counterpart, with the percent change in parentheses.}
\label{tab:finet2i_supp_full}
\vspace{2mm}
\footnotesize
\setlength{\tabcolsep}{3pt}
\begin{tabular}{lccccccc}
\toprule
Method & FID$\downarrow$ & KID{\scriptsize$\times10^{3}$}$\downarrow$ & CLIP{\scriptsize$\times10^{2}$}$\uparrow$ & VQA$\uparrow$ & HPS{\scriptsize$\times10^{2}$}$\uparrow$ & Prec.$\uparrow$ & Rec.$\uparrow$ \\
\midrule
Reference & -- & -- & 26.84 & 0.925 & 25.53 & -- & -- \\
Base Model   & 19.30 & 3.05 & 28.70 & \underline{0.940} & \textbf{29.04} & 0.993 & 0.939 \\
\midrule
Vanilla & 17.18 & 2.23 & \underline{28.87} & \textbf{0.941} & 27.78 & 0.993 & 0.966 \\
REG            & 21.13 & 3.68 & 28.43 & 0.923 & 24.18 & 0.976 & 0.961 \\
SRA            & 17.48 & 2.45 & \underline{28.87} & 0.940 & 27.71 & 0.991 & 0.964 \\
HASTE          & 17.53 & 2.34 & 28.80 & 0.938 & 27.71 & \underline{0.994} & 0.963 \\
REPA           & 17.45 & 2.13 & 28.83 & 0.939 & 27.77 & 0.993 & 0.966 \\
\midrule
Vanilla + Ours & \textbf{\good{16.75}{-2.5\%}} & \textbf{\good{1.82}{-18.4\%}} & \textbf{\good{28.92}{+0.2\%}} & \bad{0.938}{-0.3\%} & \underline{\good{27.92}{+0.5\%}} & \bad{0.993}{-0.0\%} & \textbf{\good{0.969}{+0.3\%}} \\
SRA + Ours     & \good{17.10}{-2.2\%} & \underline{\good{1.94}{-20.7\%}} & \bad{28.81}{-0.2\%} & \bad{0.937}{-0.2\%} & \bad{27.69}{-0.1\%} & \good{0.992}{+0.1\%} & \tie{0.964}{0.0\%} \\
HASTE + Ours   & \underline{\good{17.08}{-2.6\%}} & \good{2.09}{-10.5\%} & \good{28.81}{+0.0\%} & \bad{0.937}{-0.1\%} & \good{27.84}{+0.5\%} & \textbf{\good{0.994}{+0.0\%}} & \underline{\good{0.966}{+0.3\%}} \\
REPA + Ours    & \good{17.44}{-0.1\%} & \good{2.01}{-5.6\%} & \underline{\good{28.87}{+0.1\%}} & \bad{0.937}{-0.2\%} & \good{27.80}{+0.1\%} & \bad{0.993}{-0.0\%} & \bad{0.965}{-0.1\%} \\
\bottomrule
\end{tabular}
\end{table}

In addition to the results in the main paper, we provide an extended quantitative comparison for both the full model training and fine-tuning settings (Tables~\ref{tab:mscoco_supp_full} and \ref{tab:finet2i_supp_full}), reporting VQAScore~\citep{lin2024evaluating} and KID~\citep{binkowski2018demystifying} alongside the metrics used in the main paper. KID can be more informative than FID on smaller evaluation sets, which is relevant here since the Fine-T2I validation set contains only 5k images. We also add REG~\citep{wu2026representation} as an additional baseline in both settings. Similar to \methodname{}, it supervises a global semantic representation of the image rather than local, patch-level structure, which made it a natural point of comparison. However, it is designed for class-conditional architectures with no text branch, and it does not seem to translate well to MM-DiT, which already carries this kind of information through a fully functional contextual branch. Its FID and KID both come out worse than plain diffusion training. We therefore do not include it in the main manuscript or test it in combination with \methodname{}. For fine-tuning, we also report Reference and Base Model rows, where Reference denotes the real Fine-T2I validation images and Base Model is the pretrained SD3-Medium checkpoint before fine-tuning. For every configuration combined with \methodname{}, we report both how it compares to its own non-\methodname{} counterpart, colored green where it improves and red where it regresses, with the percent change alongside it, and how it ranks against every other configuration in the table, with bold/underline marking the best/second-best per column.

As already noted in the main paper for CLIP and HPS, Reference in fact scores below Base Model on all three preference and alignment style metrics in this extended comparison, including VQAScore. This reinforces that these axes should be read with caution in the fine-tuning setting: a real photograph scoring below a model's output does not mean the model produces more faithful images, only that these metrics reward image and caption conventions particular to their own training distributions.

In the full model training setting, the orthogonal comparison confirms the pattern from the main paper: every representation-alignment baseline improves on most metrics when combined with \methodname{}. Looking across the table as a whole rather than baseline by baseline, REPA + \methodname{} is not just the best combination relative to plain REPA, but the single best configuration overall, achieving the best value in every column we report. For fine-tuning, we additionally report REPA, HASTE, and SRA combined with \methodname{} for completeness. All three combinations improve KID substantially over their respective baselines (5.6 to 20.7\%), and where they do decrease Precision or Recall, the drop is marginal: REPA + \methodname{} is the only one of the three to regress on either metric, and only by 0.0\% and 0.1\%, respectively.

\subsubsection{Performance Across CFG Scales}
\label{subsec:coal_cfg_sweep}
\begin{figure}[t]
\centering
\begin{minipage}{0.495\linewidth}
    \centering
    \includegraphics[width=\linewidth]{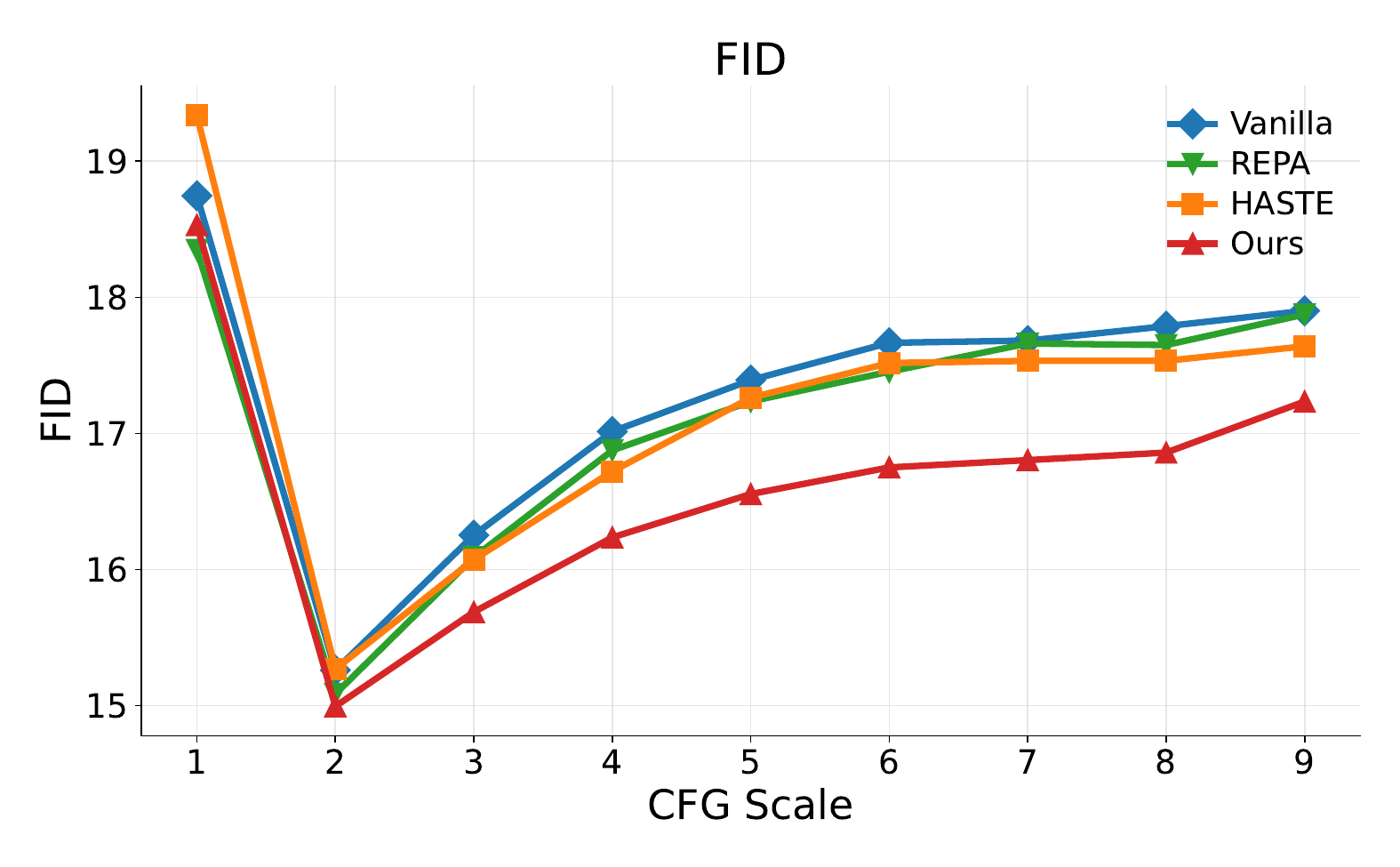}
\end{minipage}
\hfill
\begin{minipage}{0.495\linewidth}
    \centering
    \includegraphics[width=\linewidth]{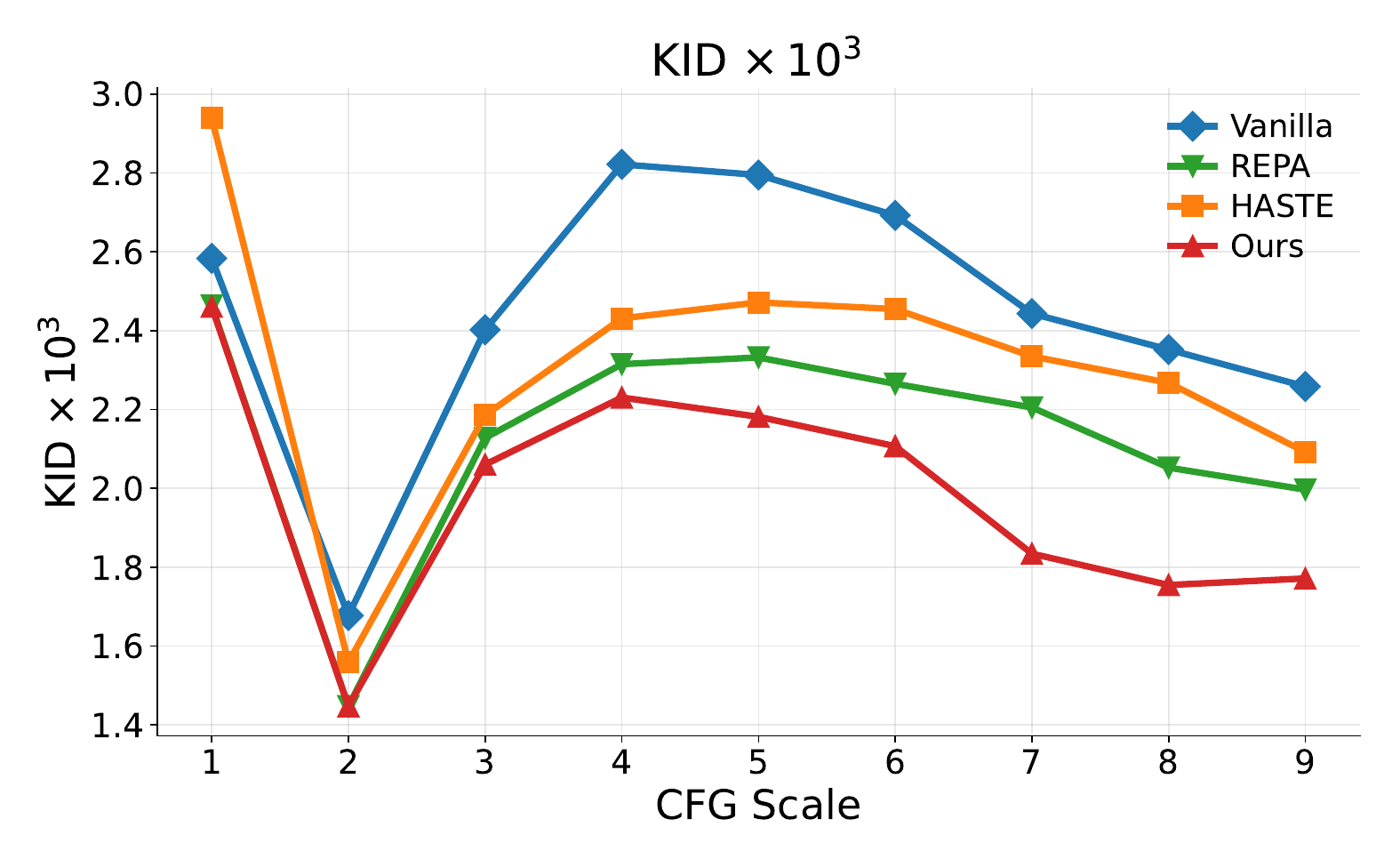}
\end{minipage}
\caption{Effect of classifier-free guidance (CFG) scale on FID and KID on the Fine-T2I curated benchmark (val5k), comparing Vanilla, REPA, HASTE, and \methodname{}.}
\label{fig:cfg_sweep}
\end{figure}

One might argue that our null-condition weighting scheme simply shifts the effective classifier-free guidance scale reached during training, and that evaluating at a different CFG value could reveal different behavior. To test this, we sweep CFG scale from 1 to 9 for Vanilla, REPA, HASTE, and Vanilla + \methodname{} on the Fine-T2I curated benchmark (Figure~\ref{fig:cfg_sweep}). \methodname{} is consistently the strongest of the four across the entire sweep. CFG$=2$ gives the best FID and KID for every method, ourselves included. We nonetheless use CFG$=7$ for all evaluations and generations elsewhere in the paper, since it offers a better balance between prompt adherence and distributional fidelity. Recent work ties image quality under classifier-free guidance to how closely the conditional and unconditional predictions agree with each other over the course of sampling: as the distance between them grows, the guidance direction takes large, sharp turns that can push generation off the data manifold~\citep{wang2026towards,yehezkel2025navigating}. \methodname{} handles large CFG values noticeably better than the other methods, which could suggest that its conditional and unconditional predictions agree with each other more closely. This is indeed what our null-condition weighting is designed to encourage: by encouraging the contextual branch to draw on information from the image tokens when the prompt is omitted, we in turn encourage similarity between the null and conditioned predictions, assuming the image is aligned with the prompt.

\subsubsection{Additional Qualitative Results}
\label{subsec:coal_additional_qual}
\begin{figure}[t]
\centering
\setlength{\qualvimgw}{0.16\linewidth}
\setlength{\tabcolsep}{1.5pt}
\renewcommand{\arraystretch}{0.9}
\begin{tabular}{cccccc}

\multicolumn{6}{p{0.95\linewidth}}{\centering\scriptsize\itshape ``A vibrant, high-angle shot capturing {\color{red}two} hands holding refreshing beverages--green and orange...''} \\[2pt]
\includegraphics[width=\qualvimgw]{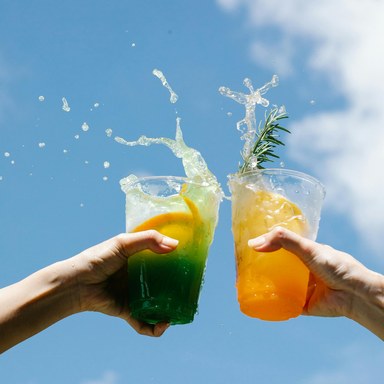} &
\includegraphics[width=\qualvimgw]{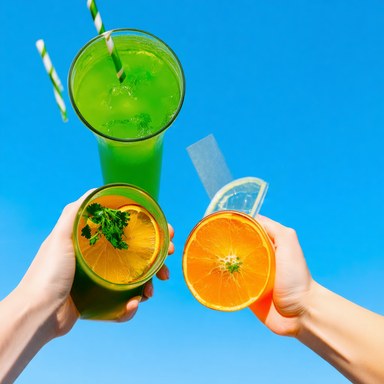} &
\includegraphics[width=\qualvimgw]{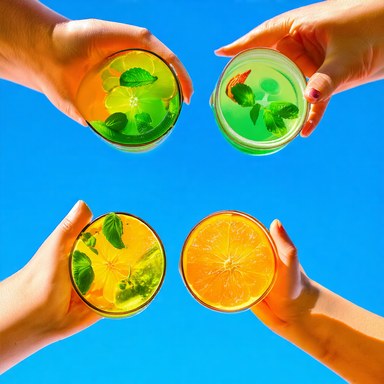} &
\includegraphics[width=\qualvimgw]{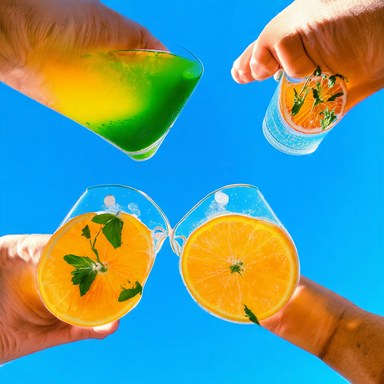} &
\includegraphics[width=\qualvimgw]{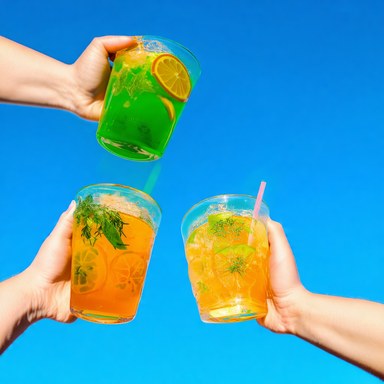} &
\includegraphics[width=\qualvimgw]{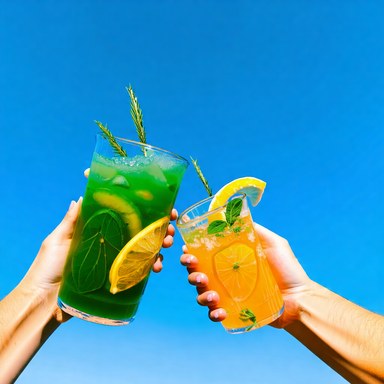} \\[3pt]

\multicolumn{6}{p{0.95\linewidth}}{\centering\scriptsize\itshape ``A black-and-white, high-contrast photograph featuring a young, shirtless man with short, curly hair...''} \\[2pt]
\includegraphics[width=\qualvimgw]{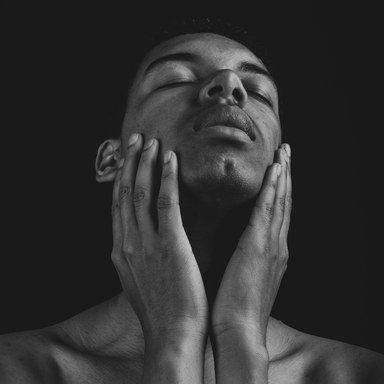} &
\includegraphics[width=\qualvimgw]{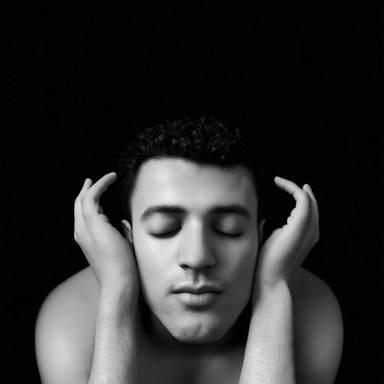} &
\includegraphics[width=\qualvimgw]{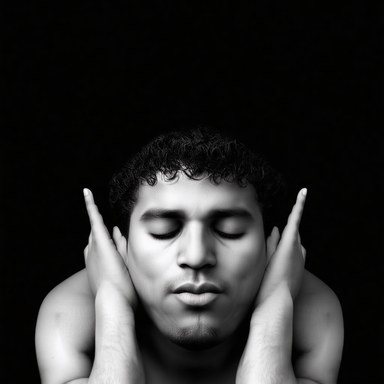} &
\includegraphics[width=\qualvimgw]{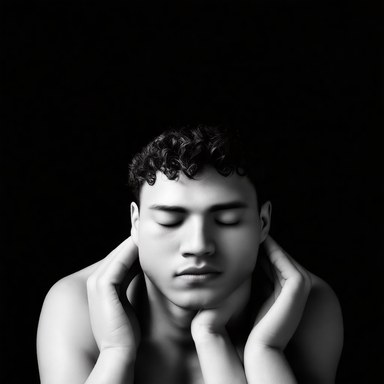} &
\includegraphics[width=\qualvimgw]{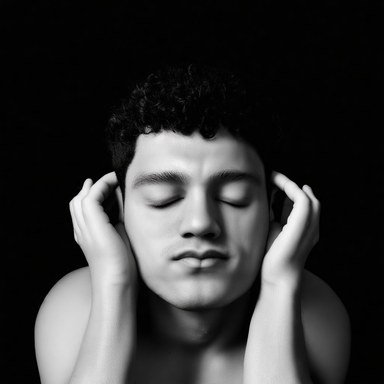} &
\includegraphics[width=\qualvimgw]{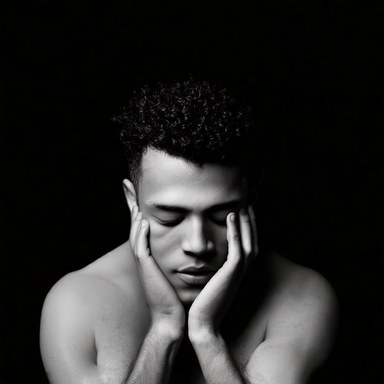} \\[3pt]

\multicolumn{6}{p{0.95\linewidth}}{\centering\scriptsize\itshape ``A dirt road curving into the distance, marked with the word {\color{red}"START"} in white lettering on a clear, blue sky day...''} \\[2pt]
\includegraphics[width=\qualvimgw]{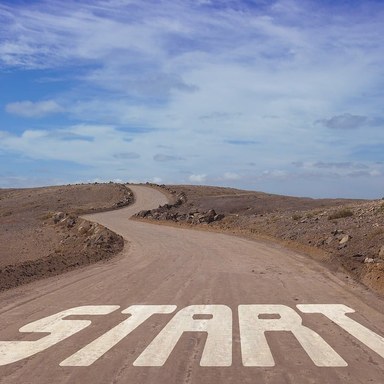} &
\includegraphics[width=\qualvimgw]{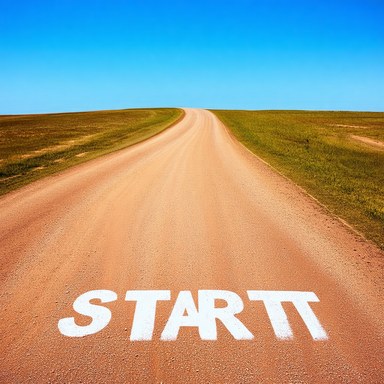} &
\includegraphics[width=\qualvimgw]{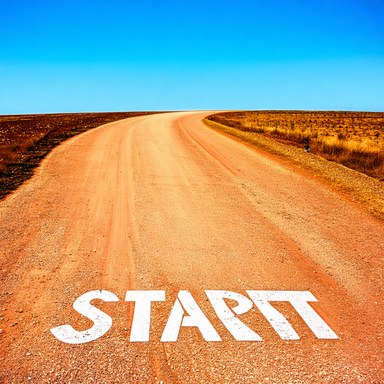} &
\includegraphics[width=\qualvimgw]{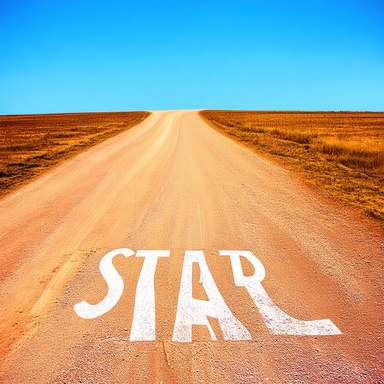} &
\includegraphics[width=\qualvimgw]{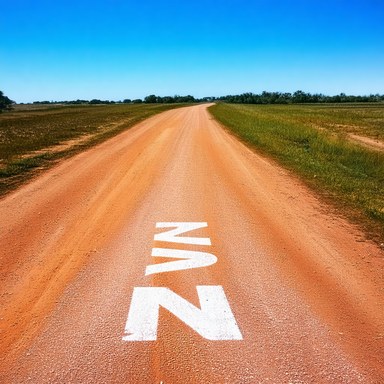} &
\includegraphics[width=\qualvimgw]{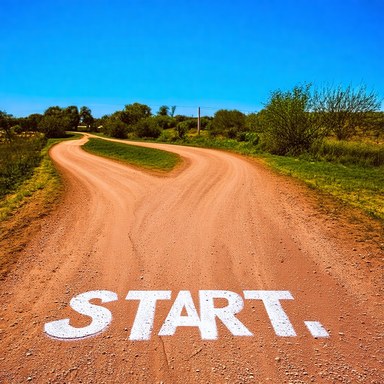} \\[3pt]

\multicolumn{6}{p{0.95\linewidth}}{\centering\scriptsize\itshape ``{\color{red}Three} women sit together in a bathtub surrounded by rows of rubber duckies on all sides, creating a tunnel-like effect...''} \\[2pt]
\includegraphics[width=\qualvimgw]{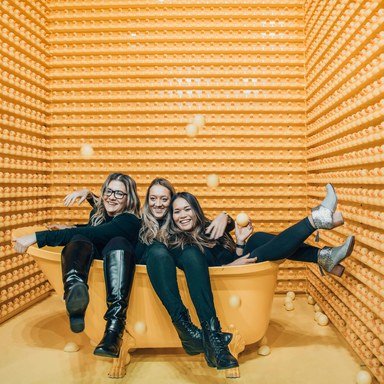} &
\includegraphics[width=\qualvimgw]{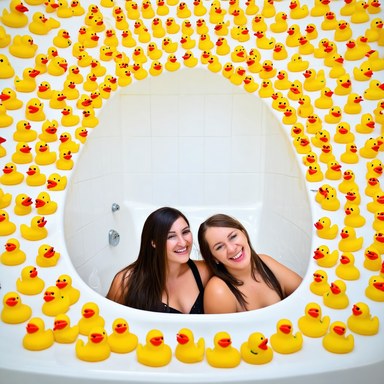} &
\includegraphics[width=\qualvimgw]{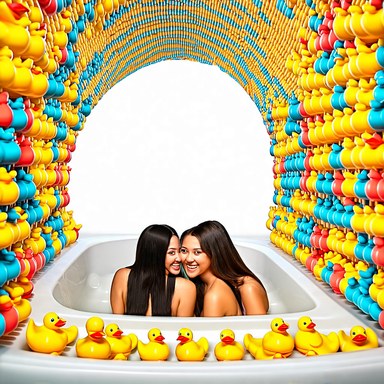} &
\includegraphics[width=\qualvimgw]{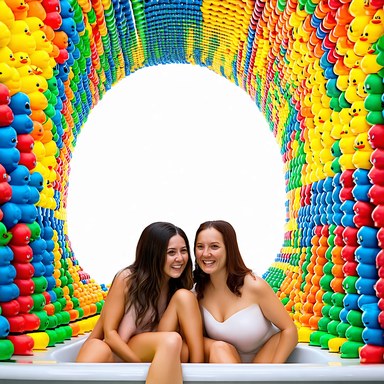} &
\includegraphics[width=\qualvimgw]{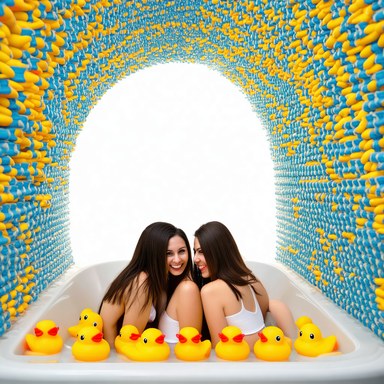} &
\includegraphics[width=\qualvimgw]{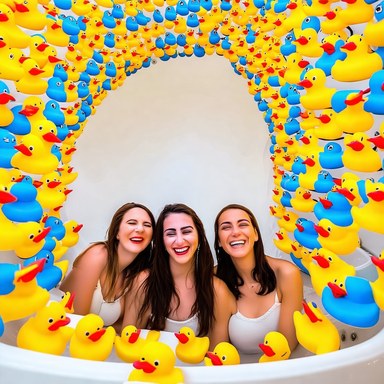} \\[3pt]

\multicolumn{6}{p{0.95\linewidth}}{\centering\scriptsize\itshape ``A young woman with long, dark hair, sitting on a concrete ledge at night, wearing a striped dress...''} \\[2pt]
\includegraphics[width=\qualvimgw]{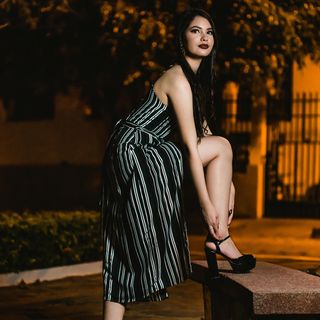} &
\includegraphics[width=\qualvimgw]{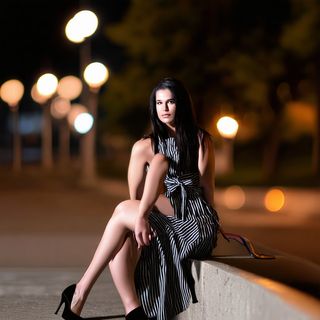} &
\includegraphics[width=\qualvimgw]{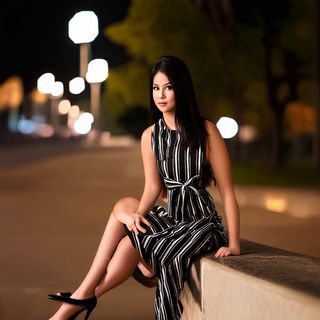} &
\includegraphics[width=\qualvimgw]{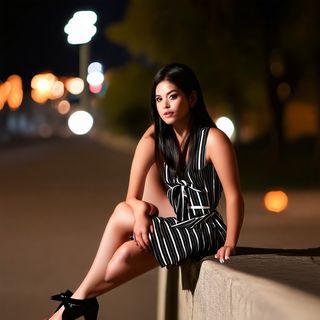} &
\includegraphics[width=\qualvimgw]{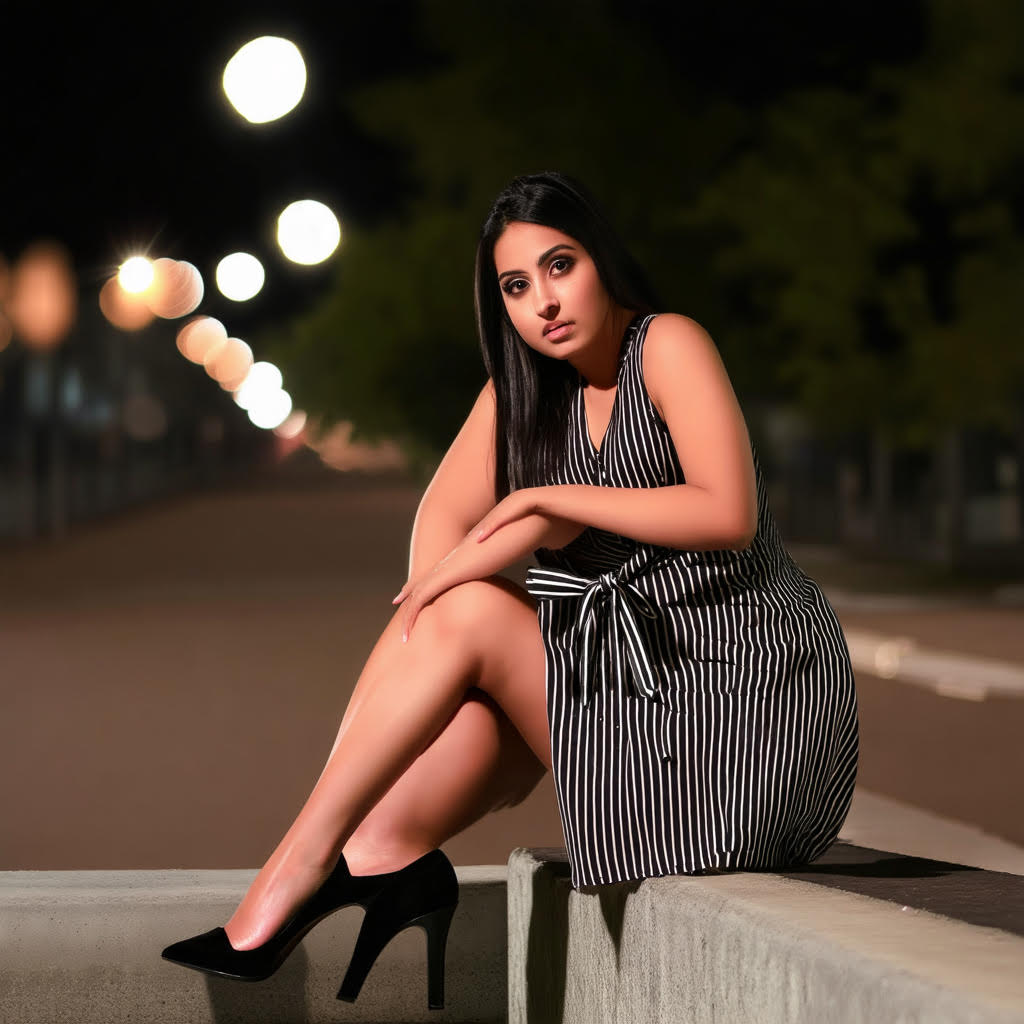} &
\includegraphics[width=\qualvimgw]{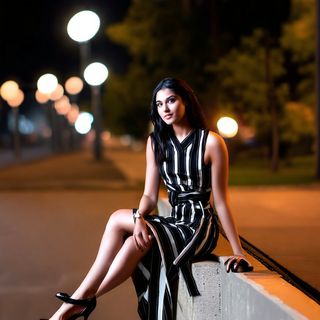} \\[3pt]

\multicolumn{6}{p{0.95\linewidth}}{\centering\scriptsize\itshape ``A vibrant geothermal pool with a gradient of colors, from deep blue to yellow, orange, and red...''} \\[2pt]
\includegraphics[width=\qualvimgw]{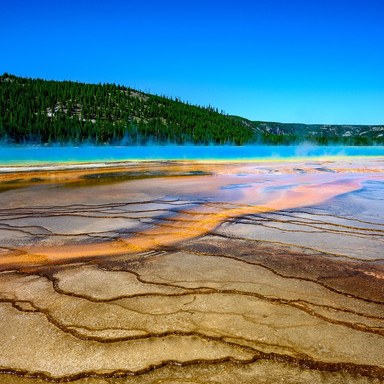} &
\includegraphics[width=\qualvimgw]{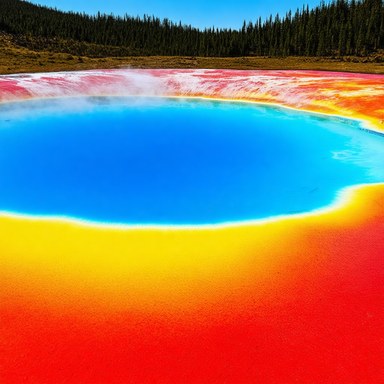} &
\includegraphics[width=\qualvimgw]{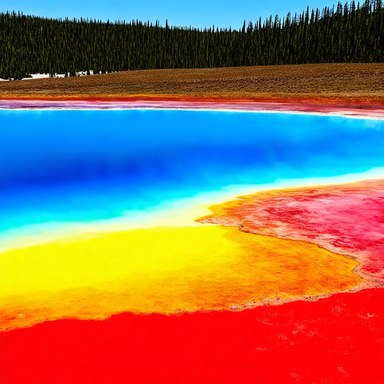} &
\includegraphics[width=\qualvimgw]{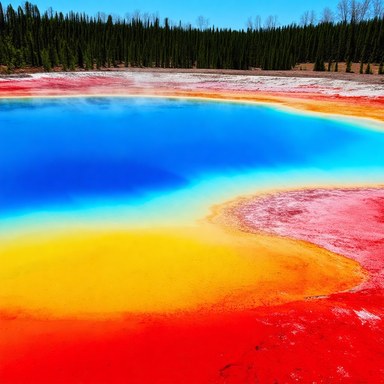} &
\includegraphics[width=\qualvimgw]{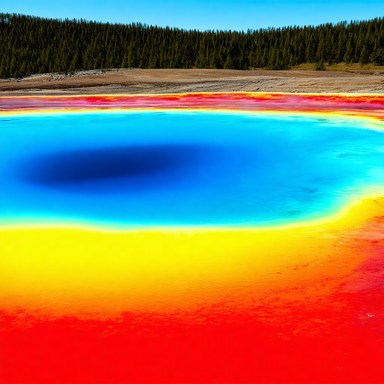} &
\includegraphics[width=\qualvimgw]{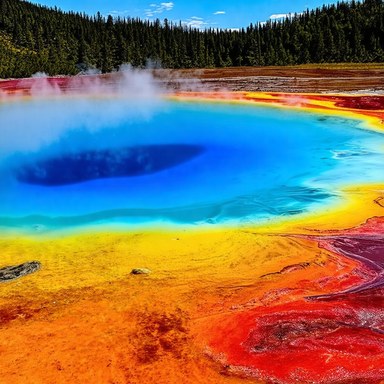} \\[3pt]

\multicolumn{6}{p{0.95\linewidth}}{\centering\scriptsize\itshape ``A minimalist photograph of the word {\color{red}`SALE'} spelled out in white tiles against a coral pink background...''} \\[2pt]
\includegraphics[width=\qualvimgw]{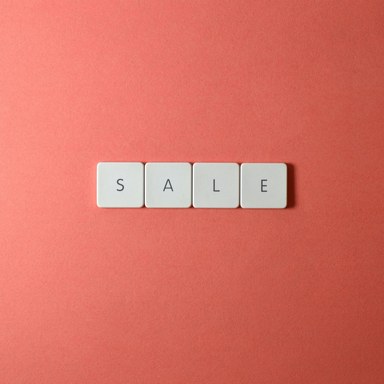} &
\includegraphics[width=\qualvimgw]{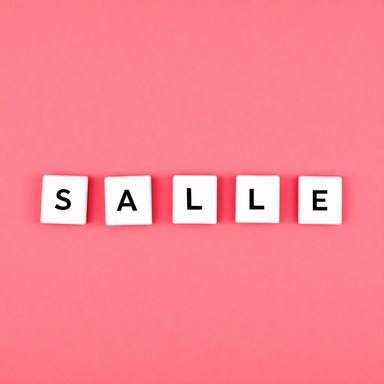} &
\includegraphics[width=\qualvimgw]{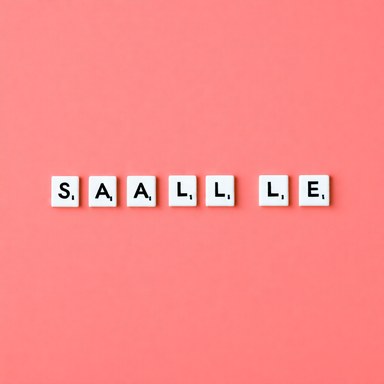} &
\includegraphics[width=\qualvimgw]{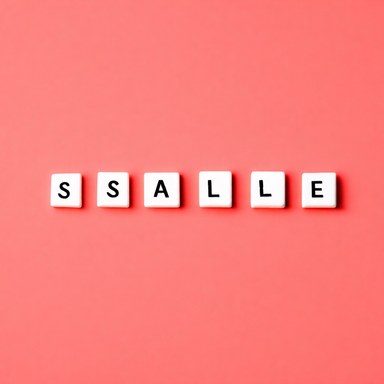} &
\includegraphics[width=\qualvimgw]{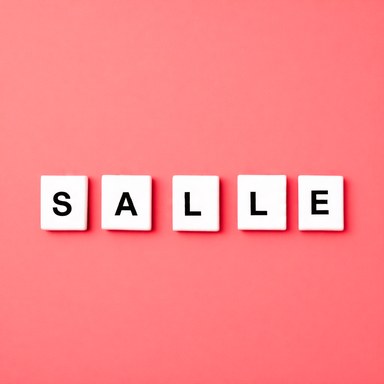} &
\includegraphics[width=\qualvimgw]{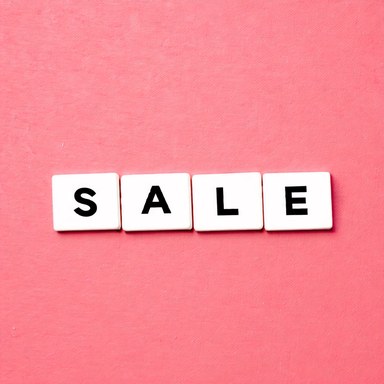} \\[3pt]

\small Reference & \small Vanilla & \small REPA & \small HASTE & \small SRA & \small \textbf{Ours} \\

\end{tabular}
\caption{Additional qualitative comparison against visual alignment methods over SD3-Medium fine-tuning on the Fine-T2I curated dataset.}
\label{fig:repa_qualitative_supp2}
\end{figure}

\begin{figure}[t]
\centering
\newlength{\qualvimgwsupp}
\setlength{\qualvimgwsupp}{0.16\linewidth}
\setlength{\tabcolsep}{1.5pt}
\renewcommand{\arraystretch}{0.9}
\begin{tabular}{cccccc}

\multicolumn{6}{p{0.95\linewidth}}{\centering\scriptsize\itshape ``A man stands in waist-deep, murky green water, holding a fishing rod with a black reel...''} \\[2pt]
\includegraphics[width=\qualvimgwsupp]{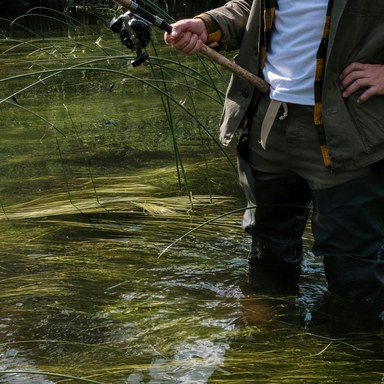} &
\includegraphics[width=\qualvimgwsupp]{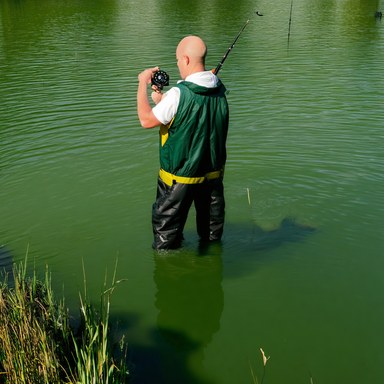} &
\includegraphics[width=\qualvimgwsupp]{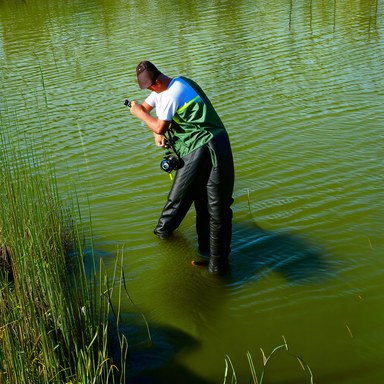} &
\includegraphics[width=\qualvimgwsupp]{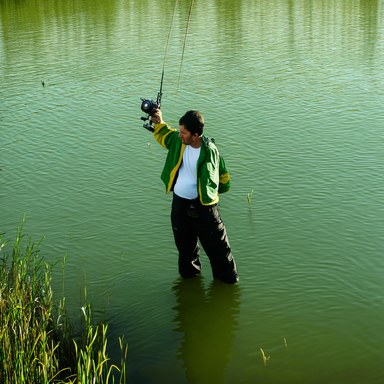} &
\includegraphics[width=\qualvimgwsupp]{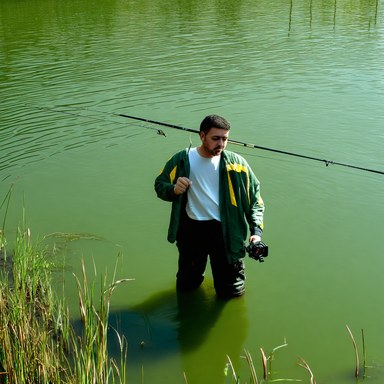} &
\includegraphics[width=\qualvimgwsupp]{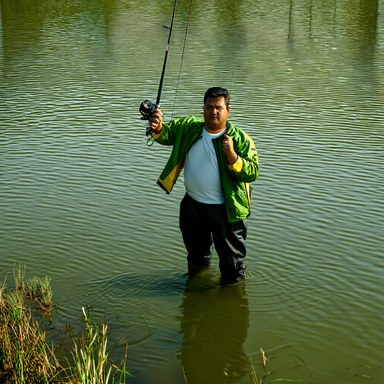} \\[3pt]

\multicolumn{6}{p{0.95\linewidth}}{\centering\scriptsize\itshape ``A close-up of a {\color{red}hand} wiping the words {\color{red}`Happy Birthday'} off a black chalkboard...''} \\[2pt]
\includegraphics[width=\qualvimgwsupp]{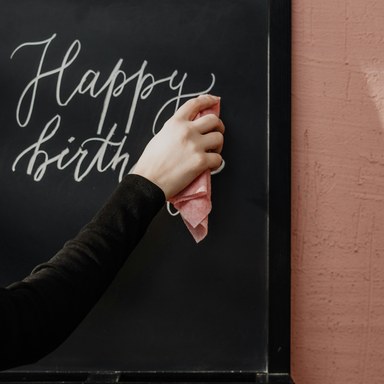} &
\includegraphics[width=\qualvimgwsupp]{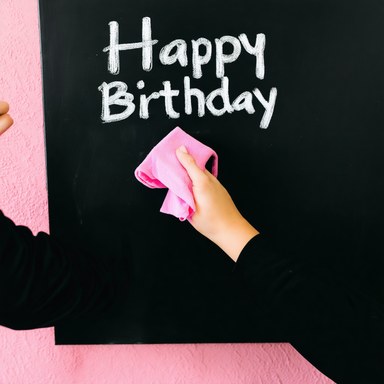} &
\includegraphics[width=\qualvimgwsupp]{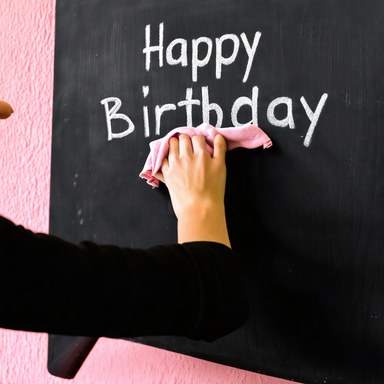} &
\includegraphics[width=\qualvimgwsupp]{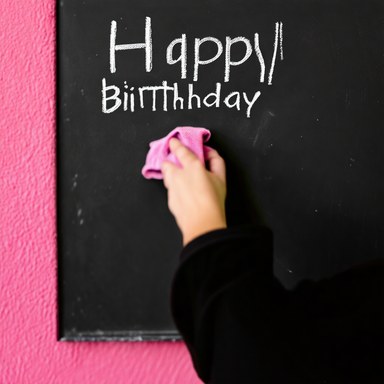} &
\includegraphics[width=\qualvimgwsupp]{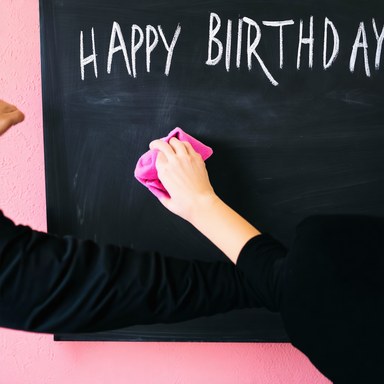} &
\includegraphics[width=\qualvimgwsupp]{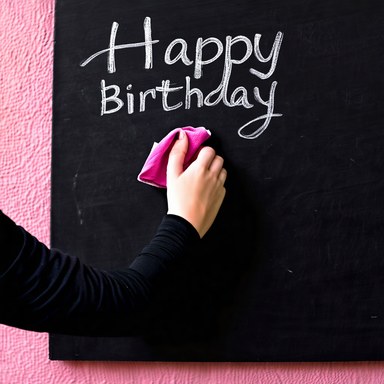} \\[3pt]

\multicolumn{6}{p{0.95\linewidth}}{\centering\scriptsize\itshape ``A double exposure photograph featuring a man playing the violin...''} \\[2pt]
\includegraphics[width=\qualvimgwsupp]{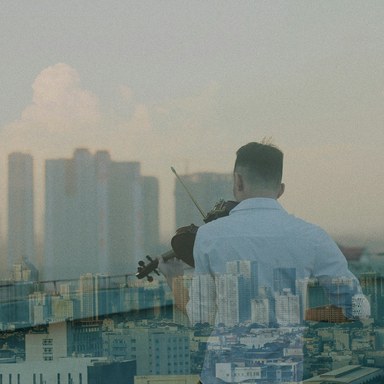} &
\includegraphics[width=\qualvimgwsupp]{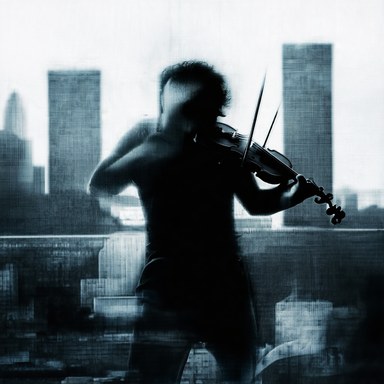} &
\includegraphics[width=\qualvimgwsupp]{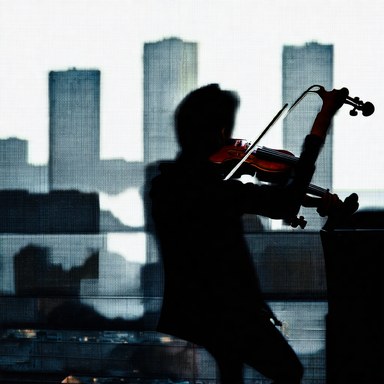} &
\includegraphics[width=\qualvimgwsupp]{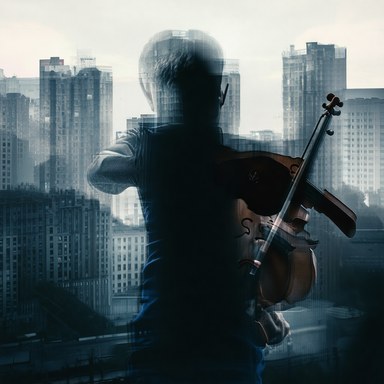} &
\includegraphics[width=\qualvimgwsupp]{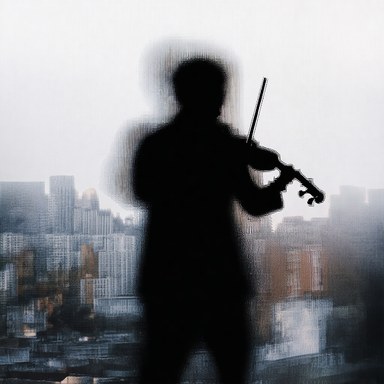} &
\includegraphics[width=\qualvimgwsupp]{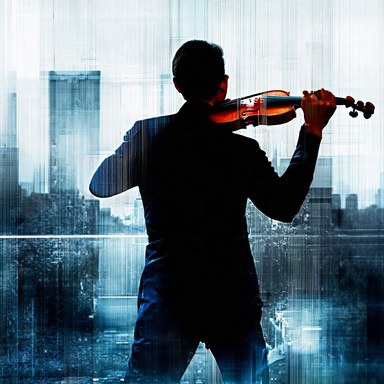} \\[3pt]

\multicolumn{6}{p{0.95\linewidth}}{\centering\scriptsize\itshape ``A close-up of a hand holding an old-fashioned Yashica 35mm camera...''} \\[2pt]
\includegraphics[width=\qualvimgwsupp]{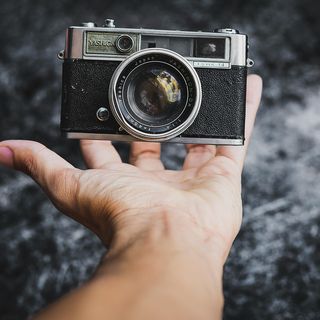} &
\includegraphics[width=\qualvimgwsupp]{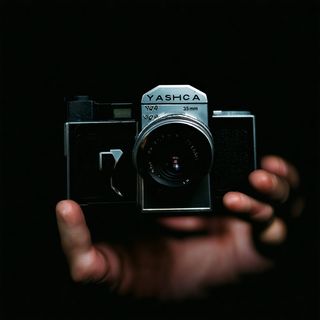} &
\includegraphics[width=\qualvimgwsupp]{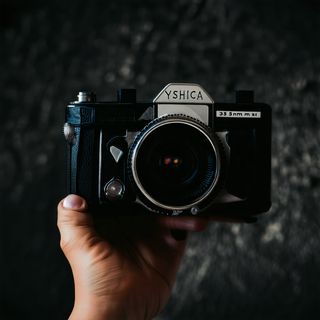} &
\includegraphics[width=\qualvimgwsupp]{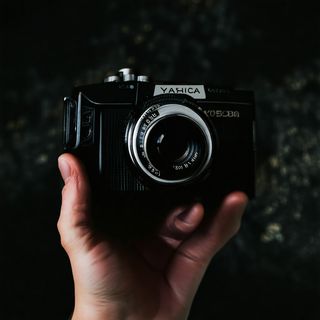} &
\includegraphics[width=\qualvimgwsupp]{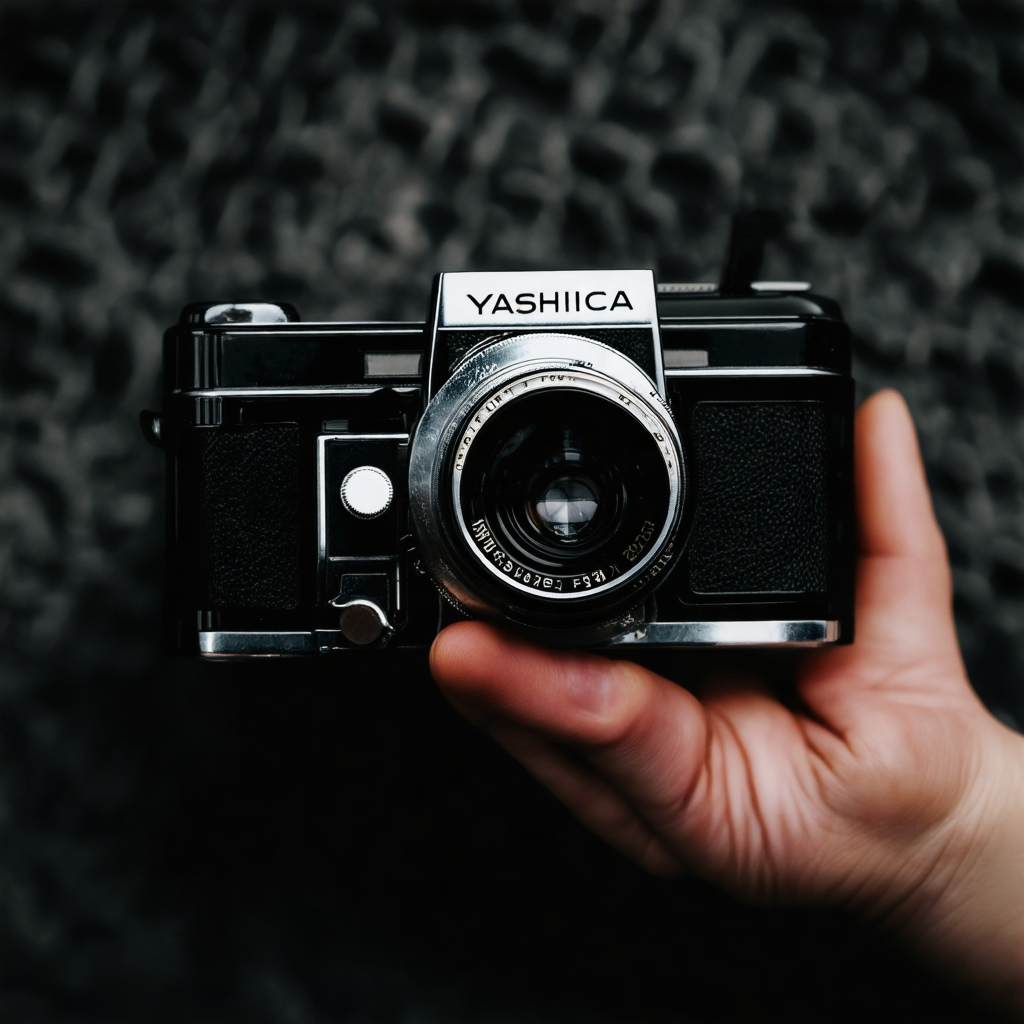} &
\includegraphics[width=\qualvimgwsupp]{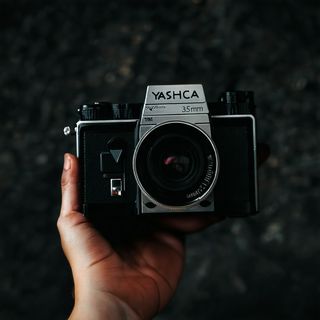} \\[3pt]

\multicolumn{6}{p{0.95\linewidth}}{\centering\scriptsize\itshape ``A {\color{red}couple} walking along a serene beach, dressed in white, under a vibrant blue sky...''} \\[2pt]
\includegraphics[width=\qualvimgwsupp]{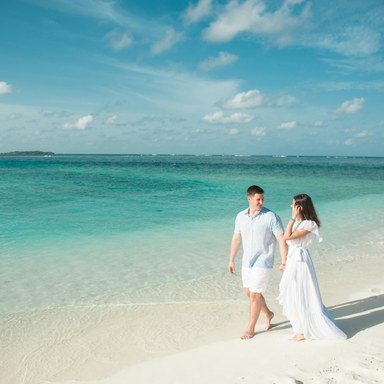} &
\includegraphics[width=\qualvimgwsupp]{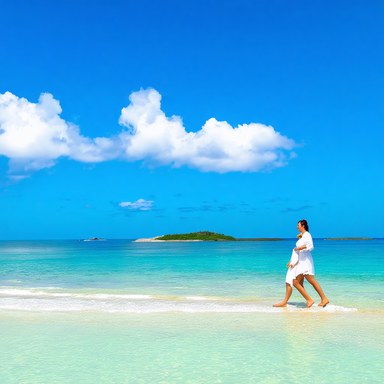} &
\includegraphics[width=\qualvimgwsupp]{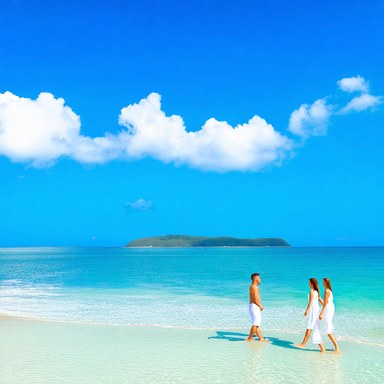} &
\includegraphics[width=\qualvimgwsupp]{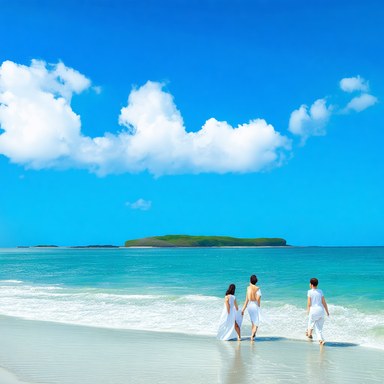} &
\includegraphics[width=\qualvimgwsupp]{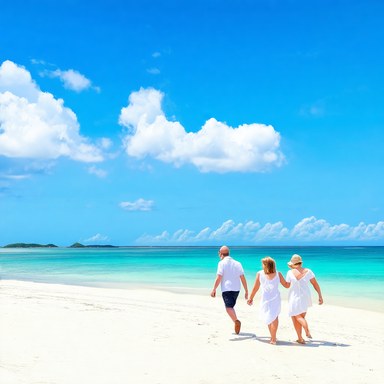} &
\includegraphics[width=\qualvimgwsupp]{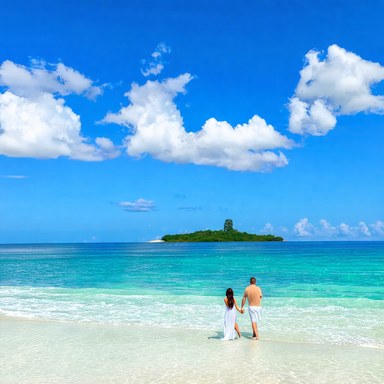} \\[3pt]

\multicolumn{6}{p{0.95\linewidth}}{\centering\scriptsize\itshape ``A neon sign in blue text reading {\color{red}`SWEET DREAMS ARE MADE OF THIS'} hangs on a wall...''} \\[2pt]
\includegraphics[width=\qualvimgwsupp]{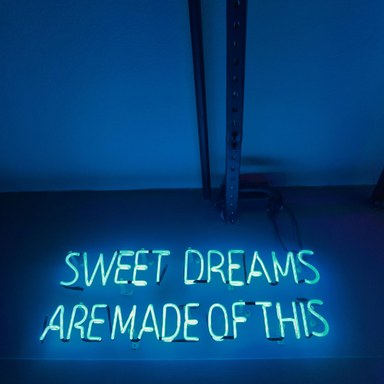} &
\includegraphics[width=\qualvimgwsupp]{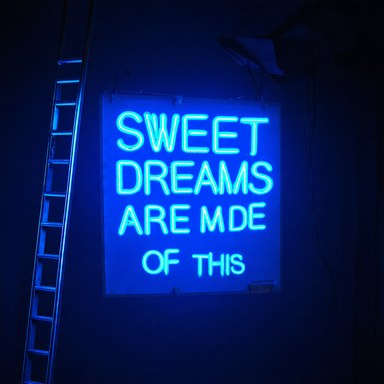} &
\includegraphics[width=\qualvimgwsupp]{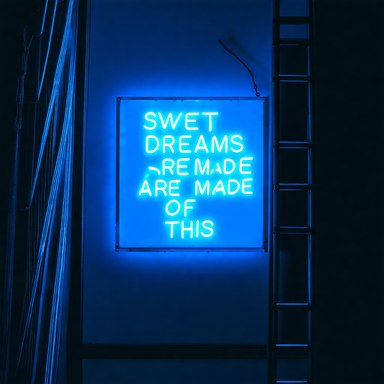} &
\includegraphics[width=\qualvimgwsupp]{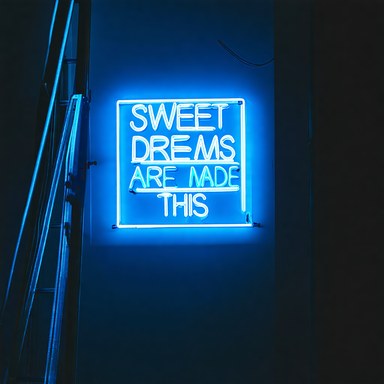} &
\includegraphics[width=\qualvimgwsupp]{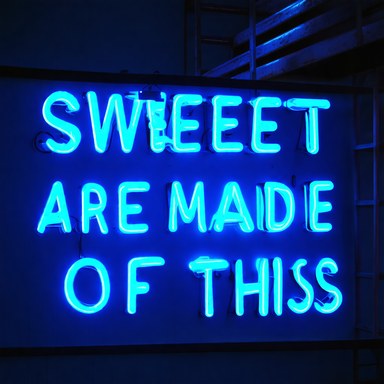} &
\includegraphics[width=\qualvimgwsupp]{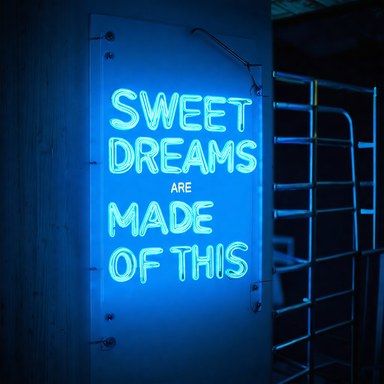} \\[3pt]

\multicolumn{6}{p{0.95\linewidth}}{\centering\scriptsize\itshape ``A close-up photograph through a circular lens, capturing two pelicans preening by a body of water...''} \\[2pt]
\includegraphics[width=\qualvimgwsupp]{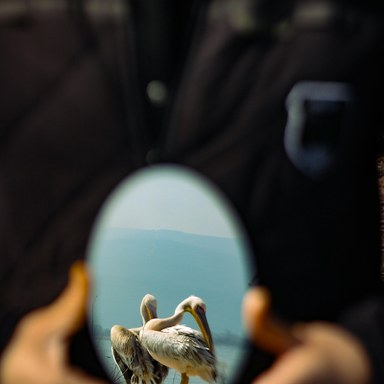} &
\includegraphics[width=\qualvimgwsupp]{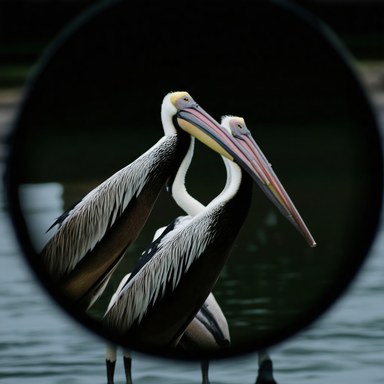} &
\includegraphics[width=\qualvimgwsupp]{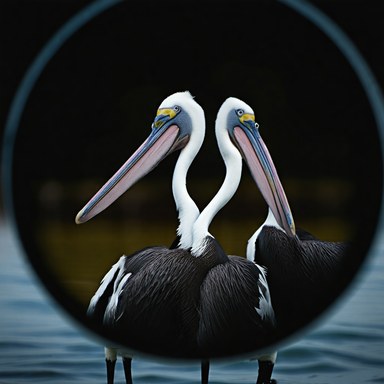} &
\includegraphics[width=\qualvimgwsupp]{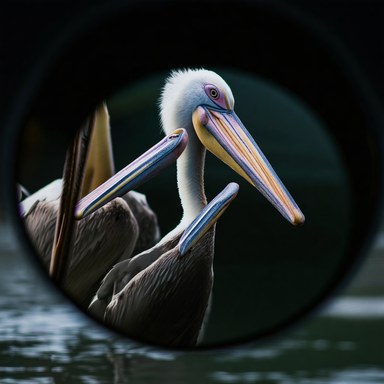} &
\includegraphics[width=\qualvimgwsupp]{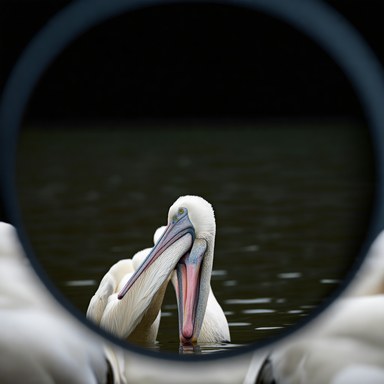} &
\includegraphics[width=\qualvimgwsupp]{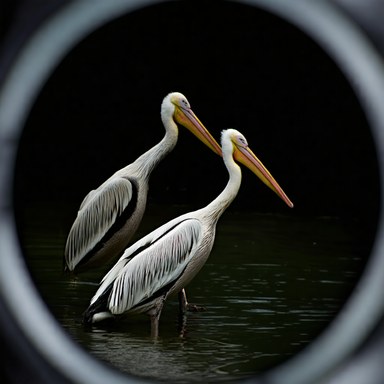} \\[3pt]

\small Ground Truth & \small Vanilla & \small REPA & \small HASTE & \small SRA & \small \textbf{Ours} \\

\end{tabular}
\caption{Additional qualitative comparison against visual alignment methods over SD3-Medium fine-tuning on the Fine-T2I curated dataset.}
\label{fig:repa_qualitative_supp}
\end{figure}

\begin{figure}[t]
\centering
\newlength{\qualimgw}
\setlength{\qualimgw}{0.15\linewidth}
\newlength{\qualcolw}
\setlength{\qualcolw}{0.305\linewidth}
\newcommand{\qualpair}[3]{%
  \begin{minipage}[t]{\qualcolw}
    \tiny\centering
    #1\hspace{1pt}%
    #2\\[1pt]
    #3
  \end{minipage}%
}
\setlength{\tabcolsep}{0pt}
\renewcommand{\arraystretch}{0.85}
\begin{tabular}{@{}c@{\hskip 6pt}c@{\hskip 6pt}c@{}}
\tiny

\qualpair{\includegraphics[width=\qualimgw]{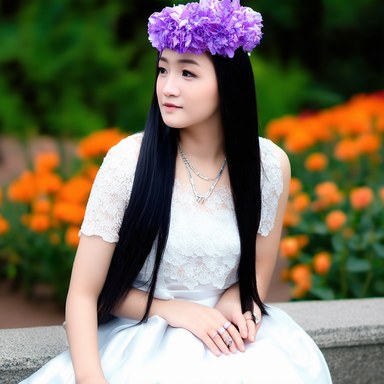}}{\includegraphics[width=\qualimgw]{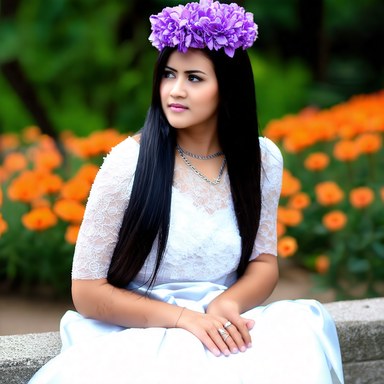}}{A young woman with long, straight black hair adorned with a purple flower crown, wearing a white lace top and white satin skirt. She is seated on a stone bench, looking off to the side with a soft, contemplative expression. Her outfit is complemented by a silver necklace and a ring, and she holds her knees with her hands. The background is a vibrant garden with orange flowers and greenery, blurred to focus on her.} &
\qualpair{\includegraphics[width=\qualimgw]{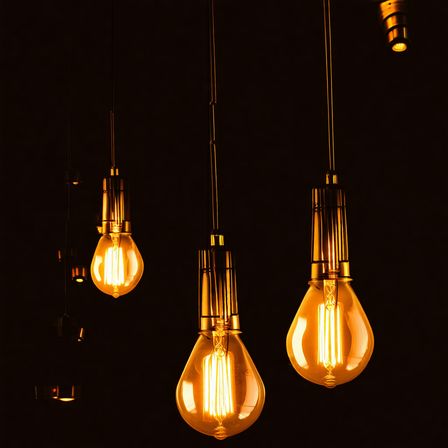}}{\includegraphics[width=\qualimgw]{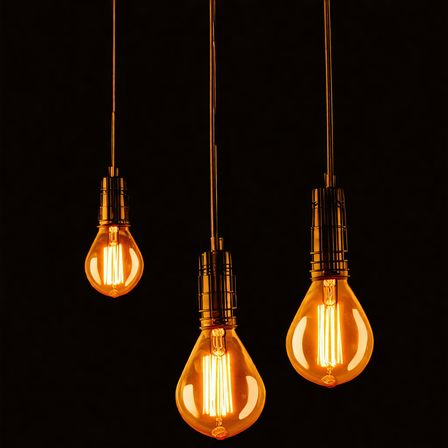}}{String of hanging Edison bulbs with black shades, illuminated in warm, amber glow against a dark background, creating a cozy, rustic ambiance.} &
\qualpair{\includegraphics[width=\qualimgw]{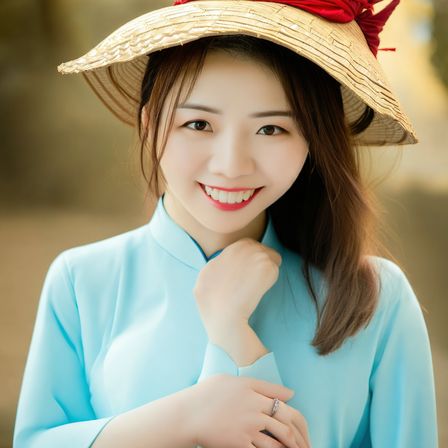}}{\includegraphics[width=\qualimgw]{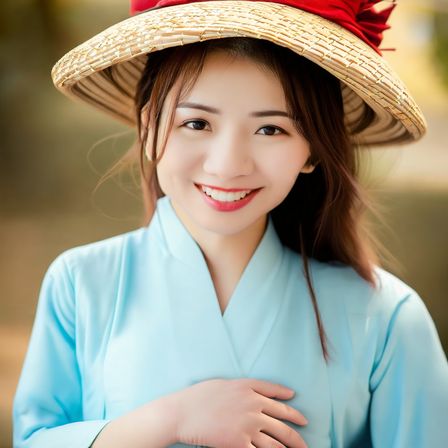}}{A young Asian woman with light brown hair, wearing a traditional Vietnamese ao dai in light blue and a wide-brimmed straw hat adorned with red fabric. She has a soft, gentle smile, and her hand rests gently on her chest. The background is blurred, suggesting an outdoor setting with warm, soft lighting.} \\[58pt]

\qualpair{\includegraphics[width=\qualimgw]{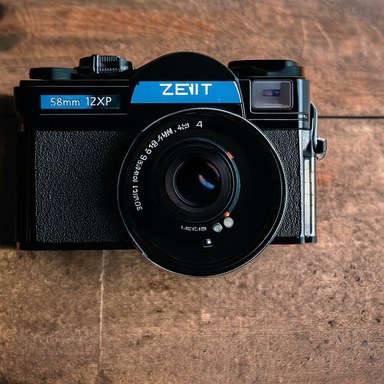}}{\includegraphics[width=\qualimgw]{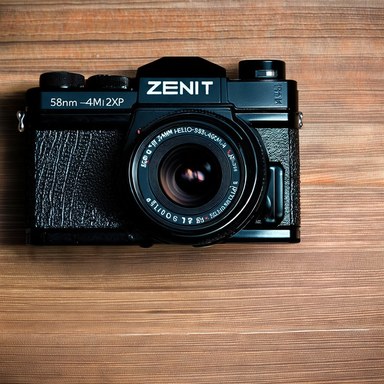}}{A close-up, high-resolution photograph of a Zenit 12XP camera, featuring a black body and a 58mm Helios-44M-4 lens, resting on a wooden surface with a textured, weathered appearance.} &
\qualpair{\includegraphics[width=\qualimgw]{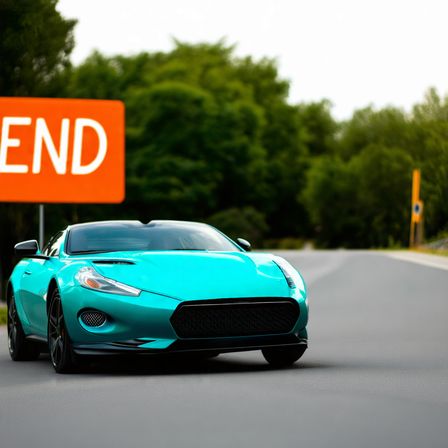}}{\includegraphics[width=\qualimgw]{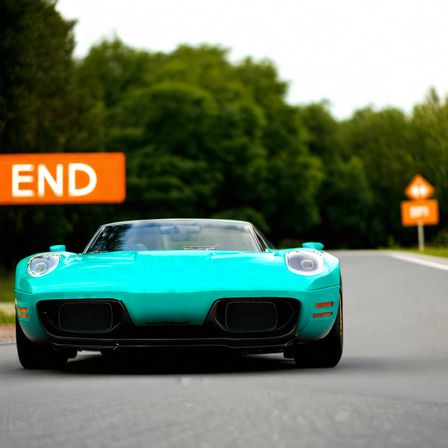}}{A teal-colored sports car parked at a road's end, with a bright orange "END" sign in the background, set against a softly blurred backdrop of trees and sky.} &
\qualpair{\includegraphics[width=\qualimgw]{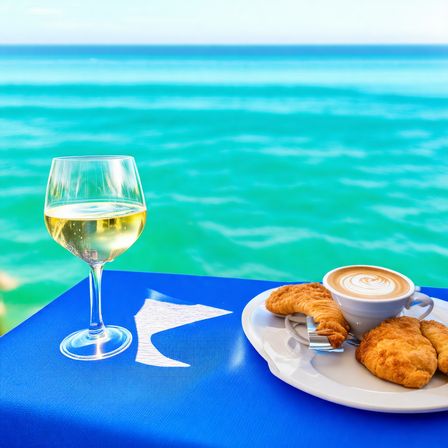}}{\includegraphics[width=\qualimgw]{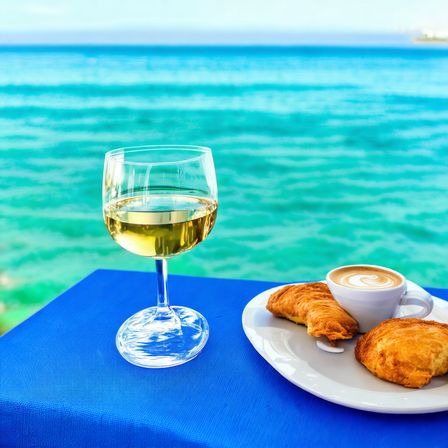}}{A bright, sunny day scene with a blue tablecloth and a glass of white wine, a cup of coffee, and a plate of fried pastries on a blue table overlooking a vibrant turquoise sea.} \\[40pt]

\qualpair{\includegraphics[width=\qualimgw]{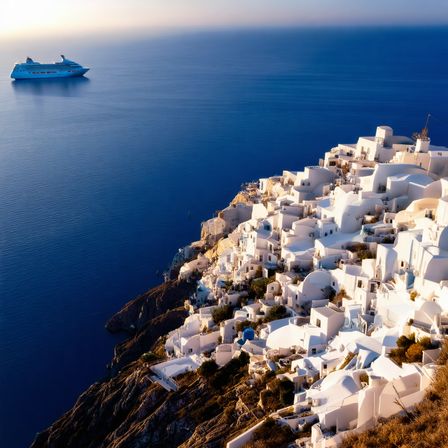}}{\includegraphics[width=\qualimgw]{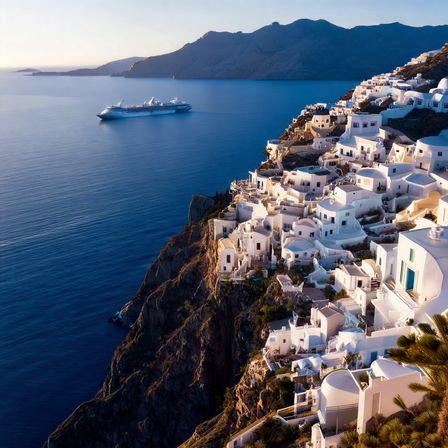}}{A stunning aerial view of Santorini, Greece, featuring a cluster of white, traditional Greek-style houses cascading down a steep cliff towards the deep blue Aegean Sea. The sun is setting, casting a warm glow over the landscape, enhancing the vivid colors. In the distance, a large cruise ship is anchored in the harbor.} &
\qualpair{\includegraphics[width=\qualimgw]{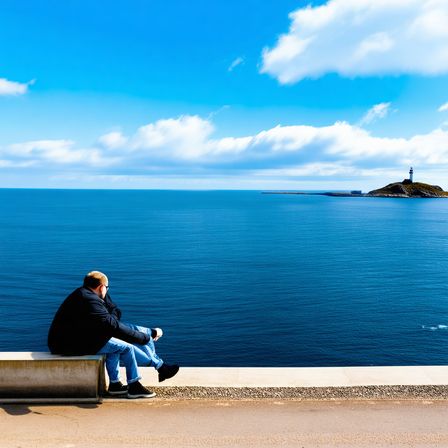}}{\includegraphics[width=\qualimgw]{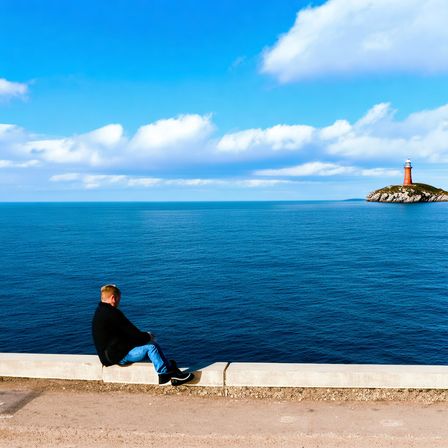}}{A man with light skin and short, light brown hair, sits alone on a concrete bench overlooking a vast, blue sea under a bright blue sky with scattered clouds. The scene is serene, with a distant rocky island and a small lighthouse on the horizon. The man, wearing a black jacket, blue jeans, and black sneakers, appears contemplative.} &
\qualpair{\includegraphics[width=\qualimgw]{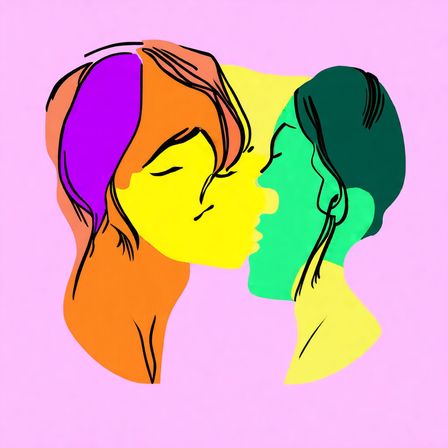}}{\includegraphics[width=\qualimgw]{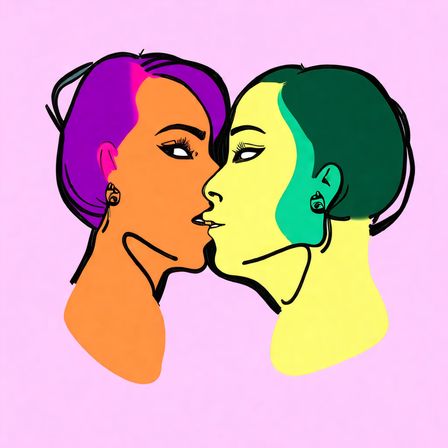}}{Abstract digital drawing depicting two women in profile, their faces merging in an intimate embrace. Faces are outlined in black, with skin tones in pastel yellow and orange, and vibrant purple, pink, and green hair. They are depicted in a minimalist, modern style with a soft pink background.} \\[55pt]

\qualpair{\includegraphics[width=\qualimgw]{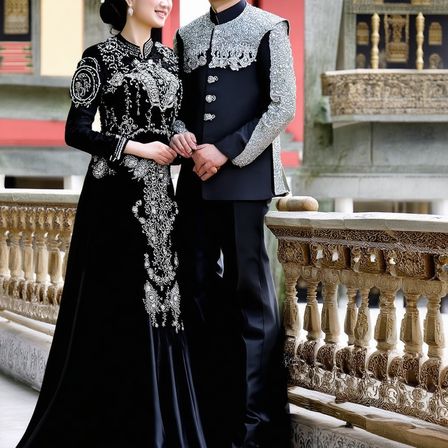}}{\includegraphics[width=\qualimgw]{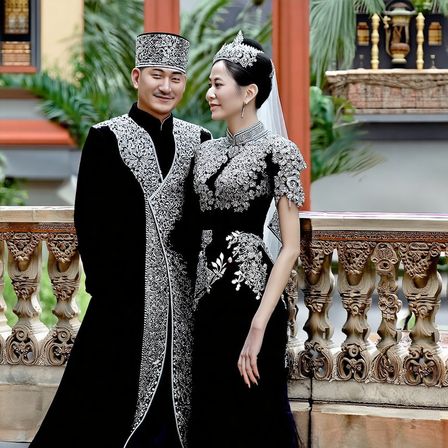}}{A couple dressed in traditional Indonesian wedding attire, the bride in a black, embroidered dress with a long train, and the groom in a silver and black outfit, stand beside an ornate stone railing adorned with intricate carvings.} &
\qualpair{\includegraphics[width=\qualimgw]{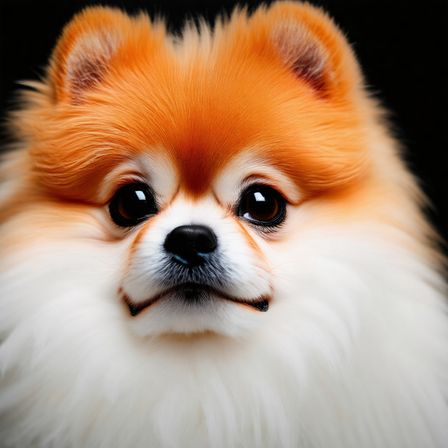}}{\includegraphics[width=\qualimgw]{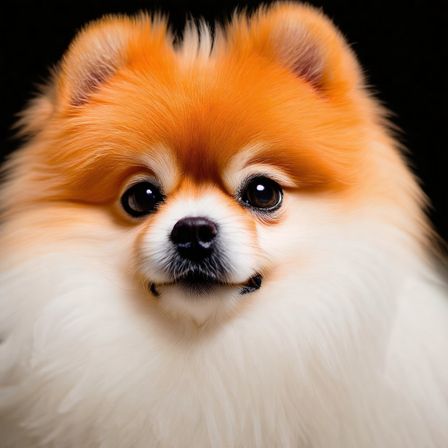}}{A close-up, high-resolution photograph of a Pomeranian dog with vibrant orange and white fur. The dog's expressive, dark eyes and small, pointed ears are prominently featured. The background is blurred, emphasizing the dog's soft, fluffy texture and compact, agile body.} &
\qualpair{\includegraphics[width=\qualimgw]{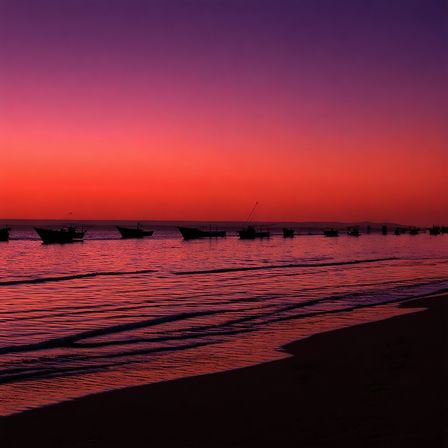}}{\includegraphics[width=\qualimgw]{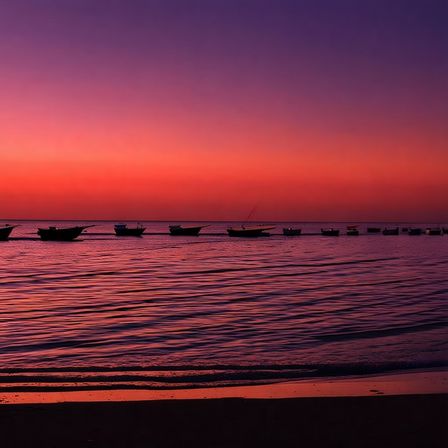}}{A tranquil sunset over a serene beach with a line of silhouetted boats floating on the calm, reflective water. The sky is a breathtaking gradient of deep reds, oranges, and purples, transitioning to darker hues towards the horizon. The sandy shore meets the water's edge, where gentle waves lap at the shore.} \\[55pt]

\qualpair{\includegraphics[width=\qualimgw]{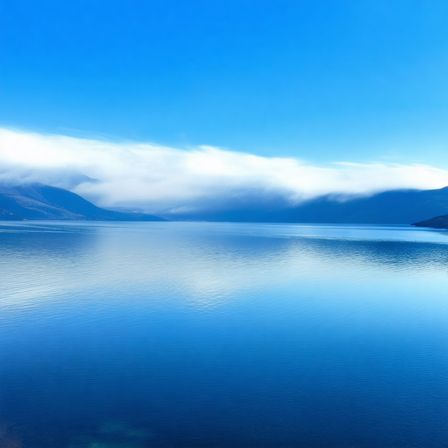}}{\includegraphics[width=\qualimgw]{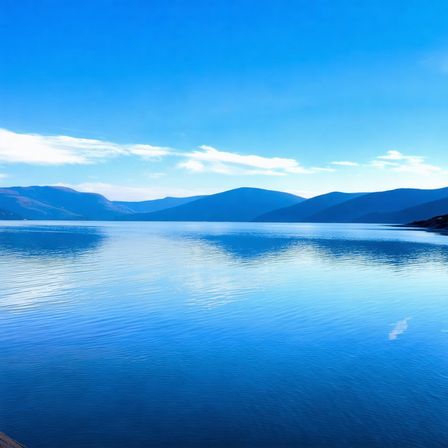}}{A serene coastal scene with calm, expansive blue water reflecting the sky. Rolling, misty mountains stretch into the distance under a partly cloudy, bright blue sky. The sun casts a soft glow, enhancing the tranquil mood.} &
\qualpair{\includegraphics[width=\qualimgw]{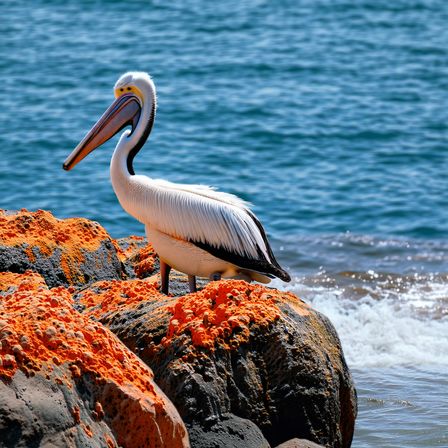}}{\includegraphics[width=\qualimgw]{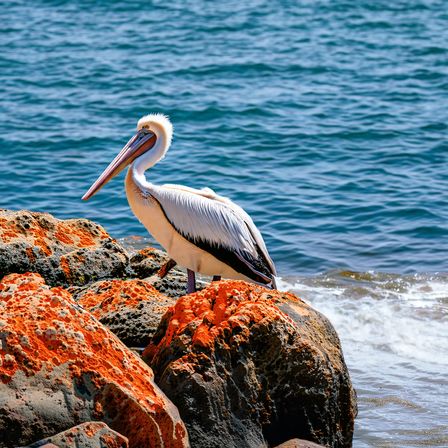}}{A pelican stands on a rocky shoreline with clear, blue water in the background. Large, weathered rocks are covered in orange lichen. Waves gently crash onto the shore.} &
\qualpair{\includegraphics[width=\qualimgw]{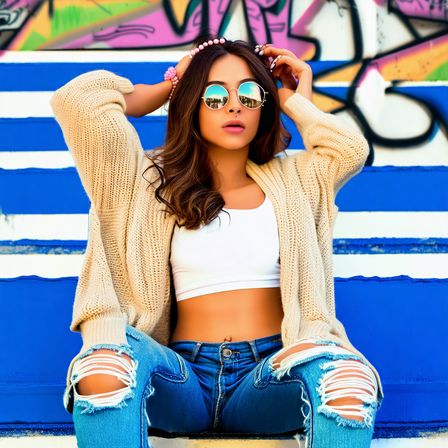}}{\includegraphics[width=\qualimgw]{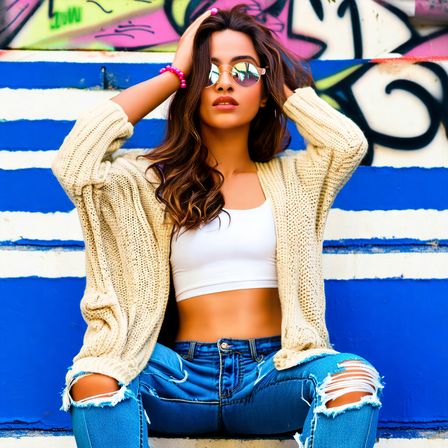}}{A young woman with long, wavy brown hair and light skin, wearing a white crop top, a beige knitted cardigan, and blue ripped jeans, sits confidently on blue and white striped steps against a graffiti-covered wall. She wears round, reflective sunglasses and a pink beaded bracelet.} \\[45pt]

\parbox[t]{\qualimgw}{\centering\small Diffusion}\hspace{1pt}\parbox[t]{\qualimgw}{\centering\small Diffusion\\+ Ours} &
\parbox[t]{\qualimgw}{\centering\small REPA}\hspace{1pt}\parbox[t]{\qualimgw}{\centering\small REPA\\+ Ours} &
\parbox[t]{\qualimgw}{\centering\small HASTE}\hspace{1pt}\parbox[t]{\qualimgw}{\centering\small HASTE\\+ Ours} \\

\end{tabular}
\caption{Qualitative comparisons on Fine-T2I val5k prompts: each pair shows a baseline (left) against the same baseline with our method added (right). Each pair uses its own set of example prompts.}
\label{fig:qualitative_comparison}
\end{figure}

We provide additional qualitative comparisons in Figures~\ref{fig:repa_qualitative_supp2},~\ref{fig:repa_qualitative_supp},and~\ref{fig:qualitative_comparison}.
Figures~\ref{fig:repa_qualitative_supp2} and~\ref{fig:repa_qualitative_supp} follow
the same format as Figure~\ref{fig:repa_qualitative}, comparing the reference held-out images from Fine-T2I, against Vanilla fine-tuning, REPA, HASTE, SRA, and \methodname{}, on
additional prompts. Figure~\ref{fig:qualitative_comparison} instead pairs each baseline directly against itself when combined with \methodname{}. We show results for Vanilla fine-tuning, REPA, and HASTE, with five examples per pair.
Across all figures, we observe improvements in image quality and sharpness, subject anatomy, text rendering, compositional coherence, and prompt following.

\end{document}